\documentclass{article}

\PassOptionsToPackage{numbers}{natbib}
\usepackage[preprint]{neurips_2026}

\usepackage[utf8]{inputenc} 
\usepackage[T1]{fontenc}    
\usepackage{hyperref}       
\usepackage{url}            
\usepackage{booktabs}       
\usepackage{amsfonts}       
\usepackage{nicefrac}       
\usepackage{microtype}      
\usepackage{xcolor}         

\usepackage{graphicx}
\usepackage{amsmath}
\usepackage{subcaption}
\usepackage[most]{tcolorbox}
\usepackage{wrapfig}

\title{Token-to-Token Alignment of Text Embeddings for Semantic Blending}

\author{
  Saar Huberman$^{1,2}$ \quad 
  Ron Mokady$^2$ \quad
  Or Patashnik$^1$ \quad  
  Daniel Cohen-Or$^1$ \\
  \vspace{1em}
  {\normalsize $^1$Tel Aviv University \quad $^2$BRIA AI}
  \vspace{-0.5em}
  \\
  {\small \url{https://saarhuberman.github.io/token-to-token/}}
}

\begin{document}

\maketitle



\newcommand{\imgwidtteaser}{0.122\linewidth}

\begin{figure*}[ht]
\vspace{-22pt}
    \centering
    \setlength{\tabcolsep}{0pt} 
    \setlength{\fboxsep}{0pt}
    \setlength{\fboxrule}{2pt}

    \makebox[\linewidth][l]{%
}

    \vspace{0.4pt}

    \begin{tabular}{cccccccc}
        \multicolumn{8}{c}{\textbf{Direct interpolation}}  \\
        \vspace{-2pt}
        \fcolorbox{red}{white}{\includegraphics[width=\imgwidtteaser]
        {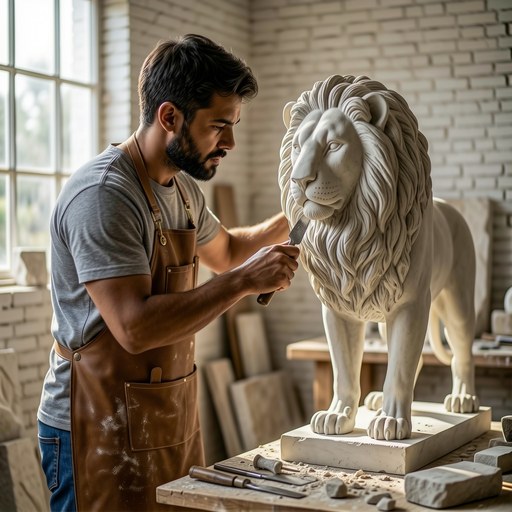}} &
        \includegraphics[width=\imgwidtteaser]{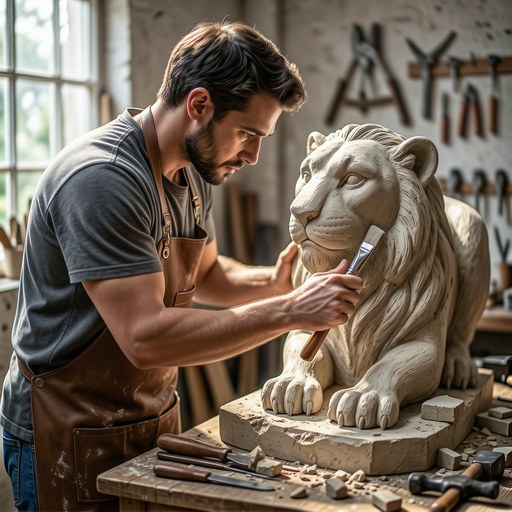} &
        \includegraphics[width=\imgwidtteaser]{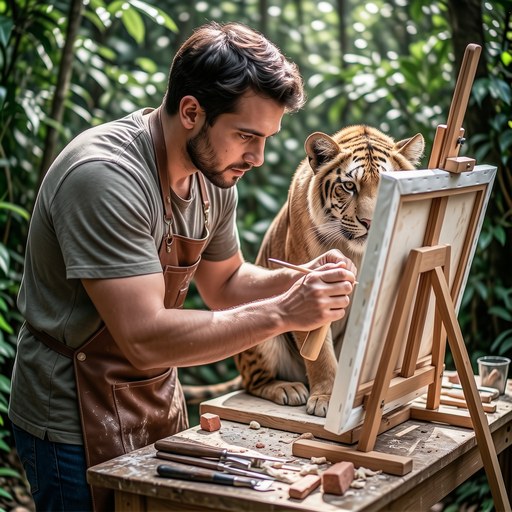} &
        \includegraphics[width=\imgwidtteaser]{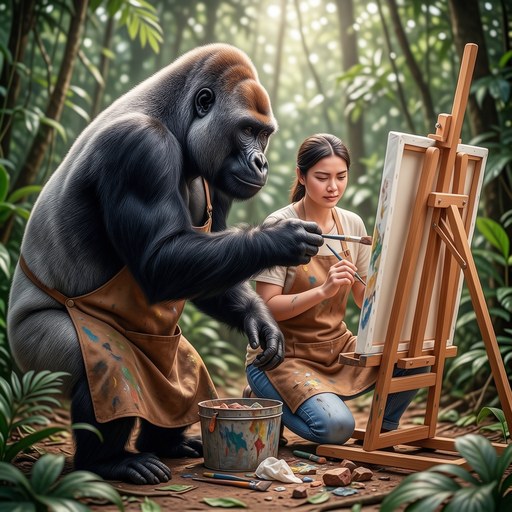} &
        \includegraphics[width=\imgwidtteaser]{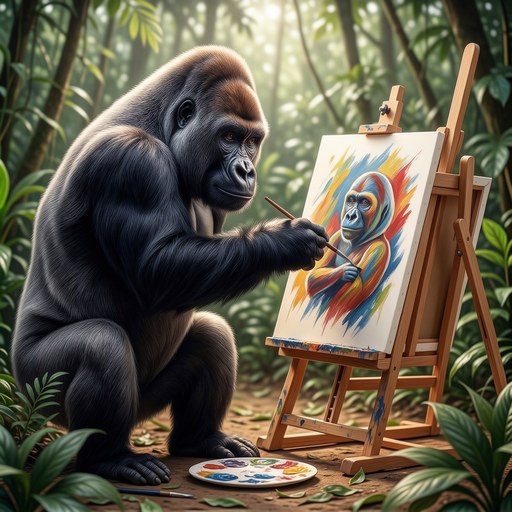} &
        \includegraphics[width=\imgwidtteaser]{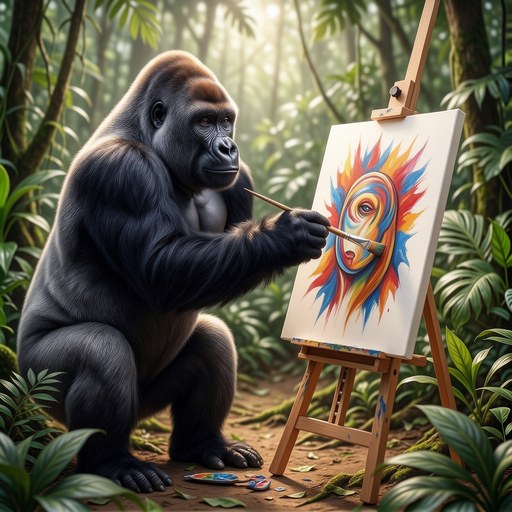} &
        \includegraphics[width=\imgwidtteaser]{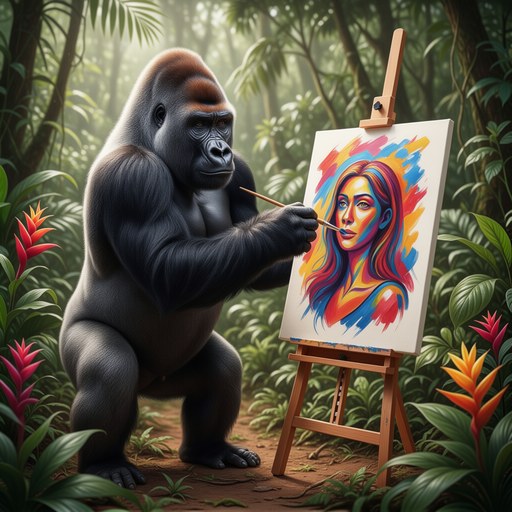} &
        \fcolorbox{red}{white}{\includegraphics[width=\imgwidtteaser]{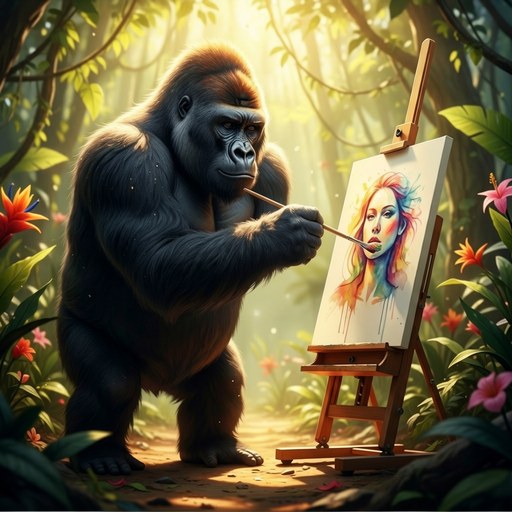}}
    \end{tabular}

    \begin{tabular}{cccccccc}
    \multicolumn{8}{c}{\textbf{Aligned interpolation}}  \\
    \vspace{-2pt}
        \fcolorbox{red}{white}{\includegraphics[width=\imgwidtteaser]
        {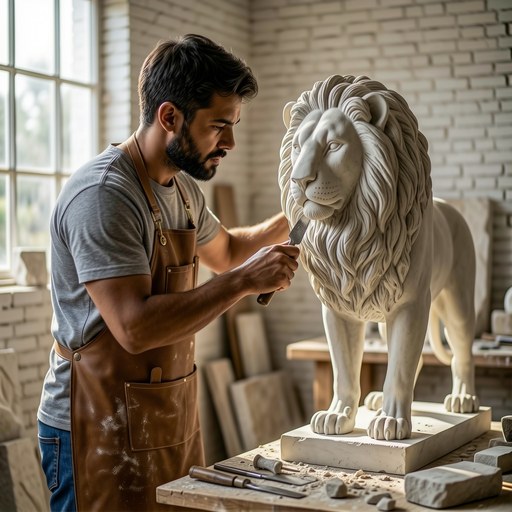}} &
        \includegraphics[width=\imgwidtteaser]{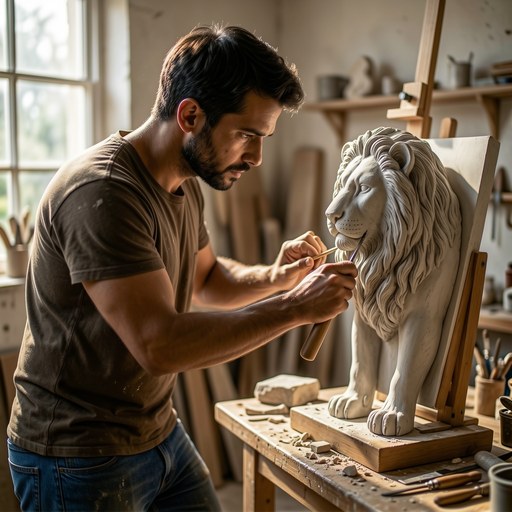} &
        \includegraphics[width=\imgwidtteaser]{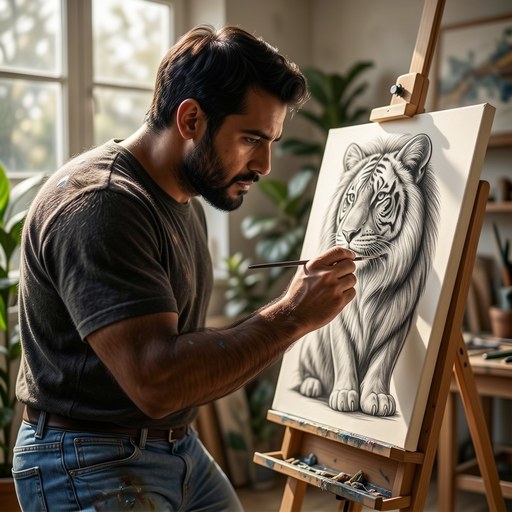} &
        \includegraphics[width=\imgwidtteaser]{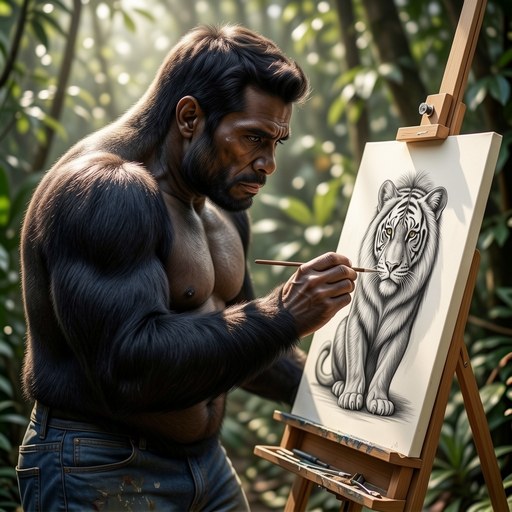} &
        \includegraphics[width=\imgwidtteaser]{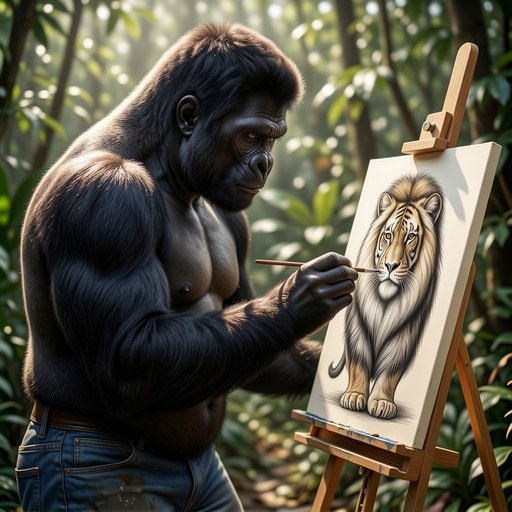} &
        \includegraphics[width=\imgwidtteaser]{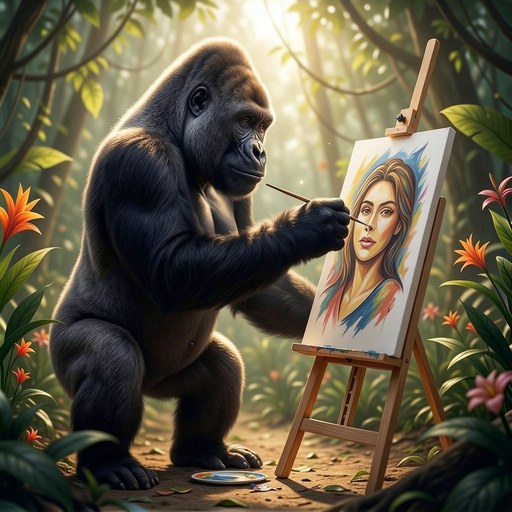} &
        \includegraphics[width=\imgwidtteaser]{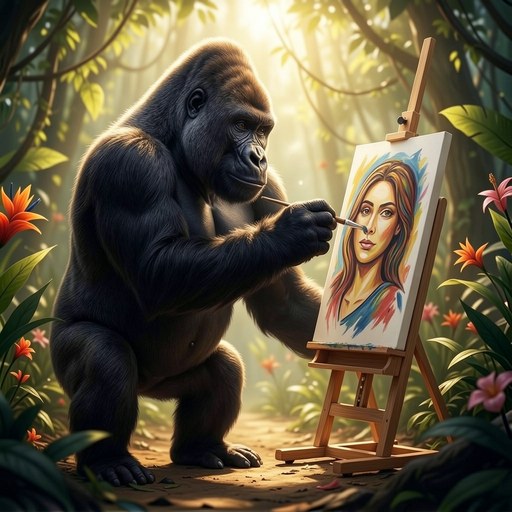} &
        \fcolorbox{red}{white}{\includegraphics[width=\imgwidtteaser]{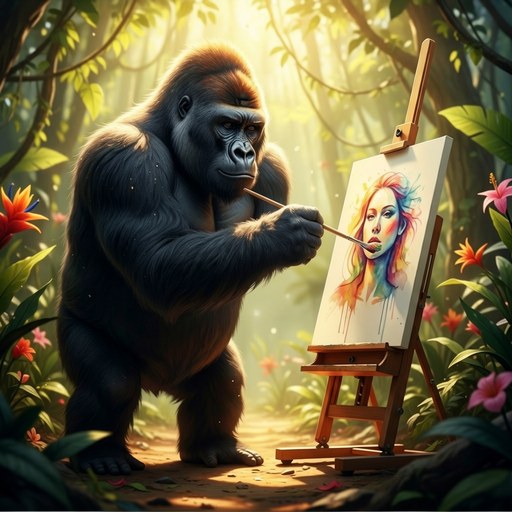}}
    \end{tabular}
    \vspace{-4pt}
    \caption{\textbf{Semantic interpolation through aligned text embeddings.}
Direct interpolation between prompt embeddings yields inconsistent and entangled transitions.
By aligning prompts at both the structural and embedding levels, we enforce token-wise semantic correspondence, enabling linear interpolation to follow a meaningful path in the text embedding space.
This reveals a continuous semantic manifold that diffusion models can traverse to generate smooth and consistent visual transformations.
    }
    \label{fig:teaser}
\end{figure*}
\vspace{-4pt}
\begin{abstract}
In modern generative models, images are specified and controlled through text prompts. In practice, images are generated from sequences of tokens derived from these prompts. 
However, the space of token sequences lacks a consistent accessible structure: semantically similar images may correspond to sequences that differ in wording, ordering, and placement of concepts, while similar token sequences may encode very different semantics. This apparent lack of structure makes it difficult to perform smooth transitions in this space, hindering applications such as image blending and continuous control of edits. We argue that this limitation stems not from the absence of semantic structure, but from misalignment between representations.
To address this misalignment, we introduce Token-to-Token alignment, a framework that establishes explicit semantic correspondence between tokens across prompts. Our approach transforms prompts into a structured representation in which semantically corresponding concepts are mapped to consistent positions across prompts, and then aligns their token embeddings based on semantic similarity. Concretely, the method consists of two stages: a structural alignment that rephrases prompts into a shared structured form, followed by an embedding-level alignment that matches token representations across prompts. With this alignment in place, simple linear interpolation becomes a meaningful operation, producing smooth and coherent semantic transitions and enabling applications such as blending and continuous editing. Our results show that text embedding spaces in text-to-image models implicitly encode a continuous semantic structure that becomes accessible once representations are properly aligned, suggesting that semantic control can be achieved by organizing existing representations rather than modifying the generative model.
\end{abstract}

\section{Introduction}

Text-to-image generative models~\cite{rombach2022high, ramesh2021zeroshottexttoimagegeneration, ho2020denoising} operate on sequences of text tokens, whose embeddings define the semantic conditioning of the generation process. While this representation is highly expressive, it lacks a consistent structure that supports continuous manipulation. In particular, linear interpolation between prompt embeddings -- an otherwise natural operation in latent spaces~\cite{shen2020interfacegan,harkonen2020ganspace} -- often produces incoherent intermediate states. This limits the ability to perform smooth semantic transitions, such as blending concepts or gradually applying edits.

This issue is not merely technical, but structural. For example, the prompts “A man holding a ginger cat” and “A ginger cat is being held by a man” describe essentially the same scene, yet their tokenizations and embeddings differ significantly. When interpolated with a third prompt, such as “A man holding a black cat”, these differences lead to inconsistent and entangled transitions. 
In this work, we argue that this limitation does not stem from the absence of semantic structure in the embedding space, but rather from misalignment between representations. Although text encoders implicitly organize semantics in a continuous manner, this structure is not directly accessible: corresponding concepts are not consistently represented across prompts. Without explicit correspondence, interpolation mixes unrelated attributes instead of following a meaningful semantic trajectory (see Figure~\ref{fig:teaser}).

To address this, we introduce \textit{Token-to-Token alignment}, a framework that establishes explicit semantic correspondence between tokens across prompts. Our approach operates in two stages. First, we perform \textit{structural alignment}, transforming prompts into a shared structured representation in which corresponding concepts occupy consistent semantic roles and textual positions. Second, we perform \textit{embedding-level alignment}, matching token representations based on semantic similarity to resolve discrepancies in the text encoder space.
Once representations are aligned, simple linear interpolation becomes a meaningful operation. Interpolating between aligned token embeddings produces coherent semantic trajectories, which the generative model renders as smooth and consistent visual transitions. This enables a unified framework for applications such as continuous image editing, and semantic blending, as demonstrated in Figure~\ref{fig:applications}.

More broadly, our results suggest that semantic control in text-to-image models is fundamentally a problem of representation rather than manipulation. The text embedding space already encodes a continuous semantic structure, but this structure is entangled and inaccessible under standard prompt formulations. By enforcing token-level correspondence, we effectively linearize this space, turning interpolation from an unreliable heuristic into a principled operation. In this view, the generative model plays a largely passive role, acting as a renderer of trajectories defined in an aligned semantic space.

We validate our approach through extensive qualitative and quantitative evaluations across both continuous editing and continuous semantic blending tasks. Our method produces semantically coherent intermediate states at every point along the trajectory, outperforming prior approaches in consistency while maintaining competitive smoothness and visual quality. Ablation studies confirm the complementary roles of structural and embedding-level alignment, and user studies further demonstrate a clear preference for the trajectories generated by our method.

\newcommand{\imgwidthapp}{0.145\linewidth}

\renewcommand{\imgwidthapp}{0.16\linewidth}

\begin{figure*}[t]
    \centering
    \setlength{\tabcolsep}{0pt} 
    \setlength{\fboxsep}{0pt}
    \setlength{\fboxrule}{2pt}
    \makebox[\linewidth][l]{%
    \textbf{Continuous Synthesis:}
    modify the prompt ``A $\langle$object$\rangle$ lying on the couch'' from cat$\rightarrow$lion
    }
    
    \vspace{0.2pt}

    \begin{tabular}{cccccc}
        \includegraphics[width=\imgwidthapp]{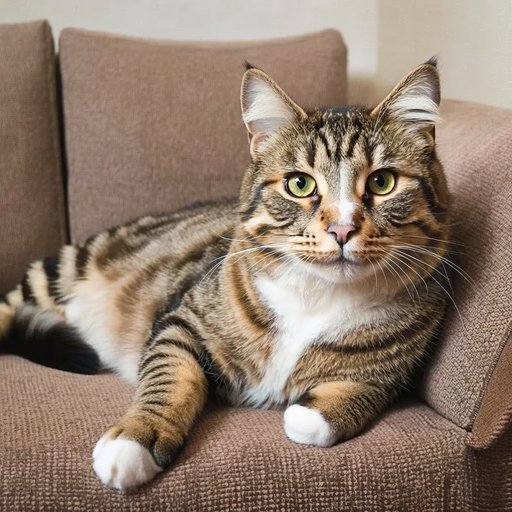} &
        \includegraphics[width=\imgwidthapp]{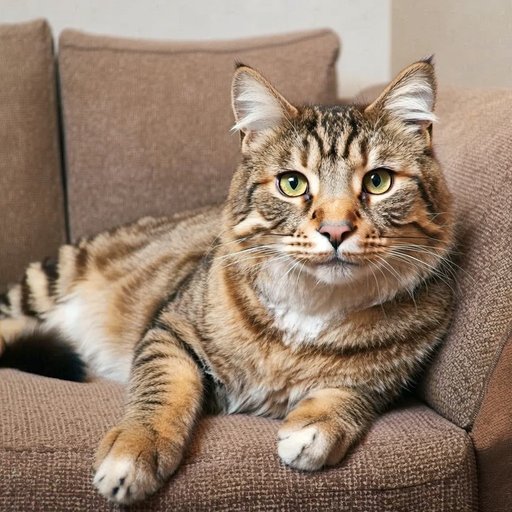} &
        \includegraphics[width=\imgwidthapp]{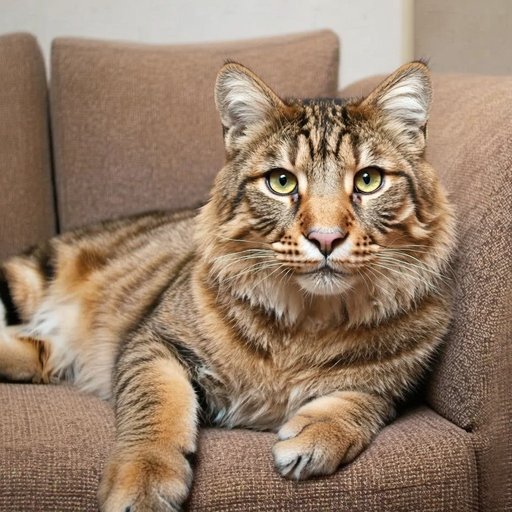} &
        \includegraphics[width=\imgwidthapp]{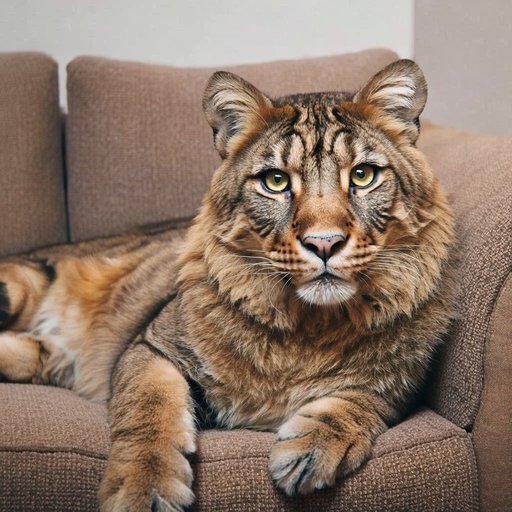} &
        \includegraphics[width=\imgwidthapp]{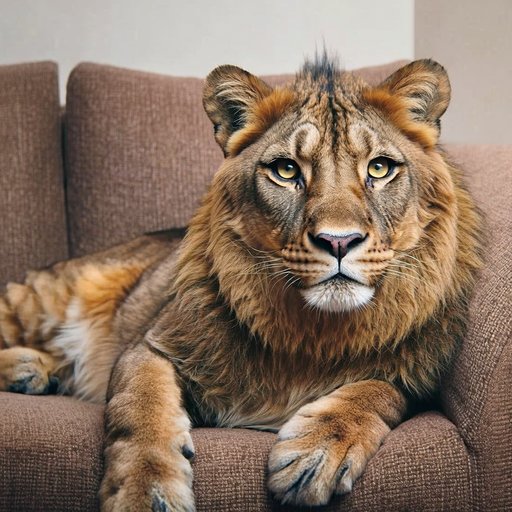} &
        \includegraphics[width=\imgwidthapp]{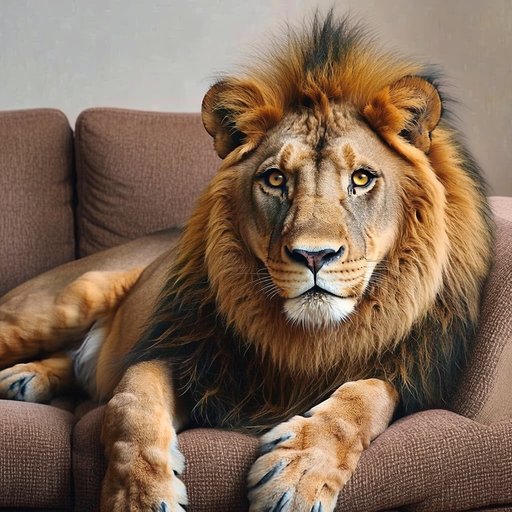}
    \end{tabular}

    \vspace{0.5pt}

    \makebox[\linewidth][l]{%
    \textbf{Continuous Editing:}
    Edit instruction: ``Turn the white origami crane into a green origami dragon''
    }

    \vspace{0.2pt}

    \begin{tabular}{cccccc}
        \fcolorbox{red}{white}{\includegraphics[width=\imgwidthapp]{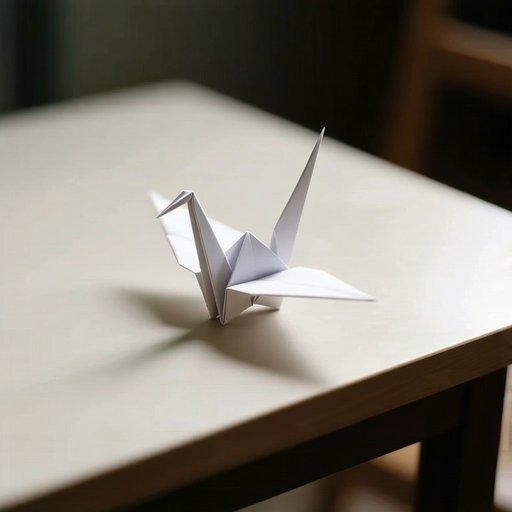}} &
        \includegraphics[width=\imgwidthapp]{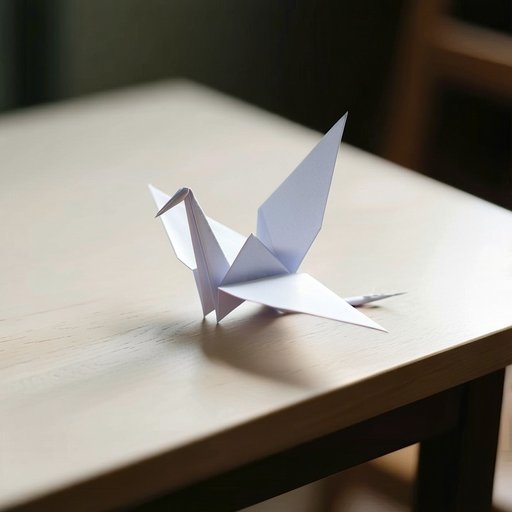} &
        \includegraphics[width=\imgwidthapp]{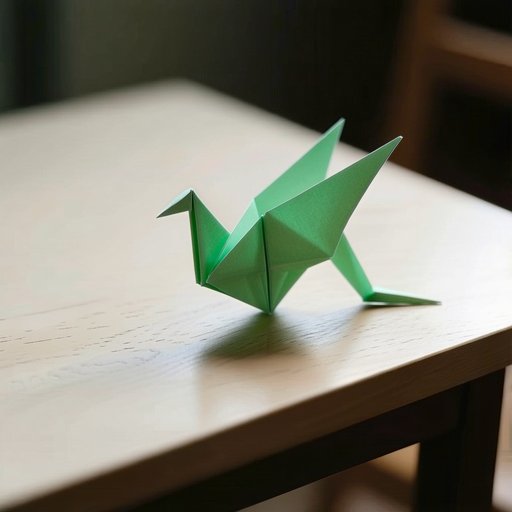} &
        \includegraphics[width=\imgwidthapp]{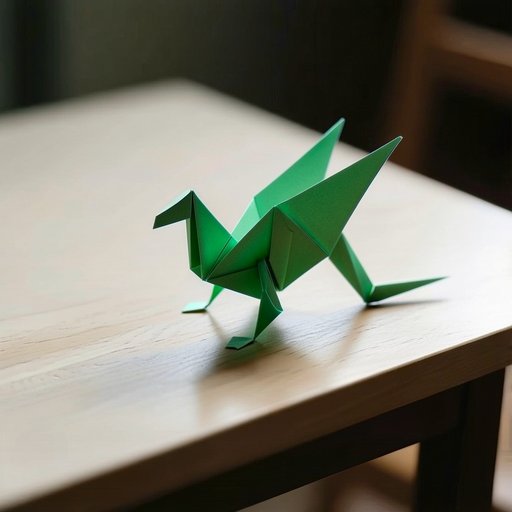} &
        \includegraphics[width=\imgwidthapp]{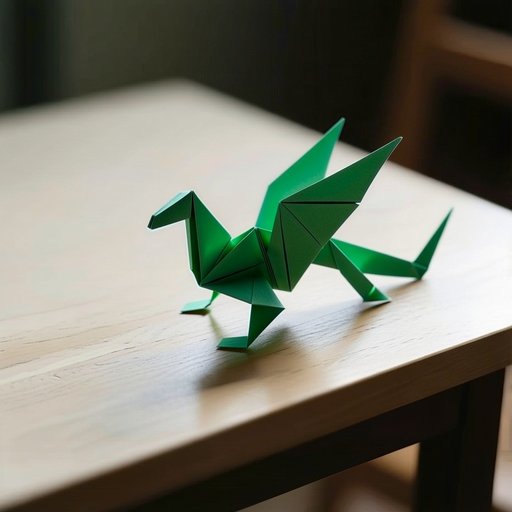} &
        \includegraphics[width=\imgwidthapp]{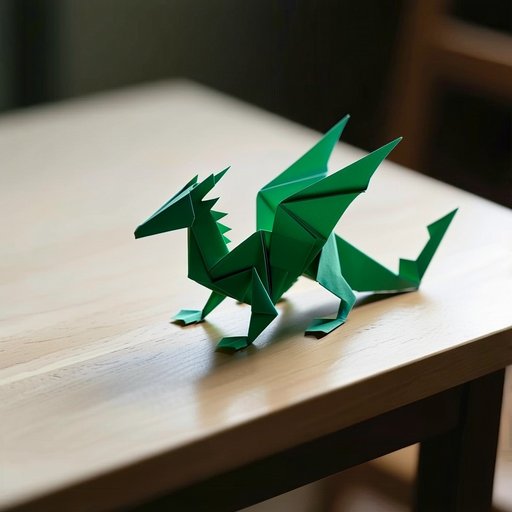}
    \end{tabular}

    \vspace{0.5pt}

    \makebox[\linewidth][l]{%
\textbf{Continuous Blending:} generate intermediate scenes
}

    \vspace{0.2pt}

    \begin{tabular}{cccccc}
        \fcolorbox{red}{white}{\includegraphics[width=\imgwidthapp]
        {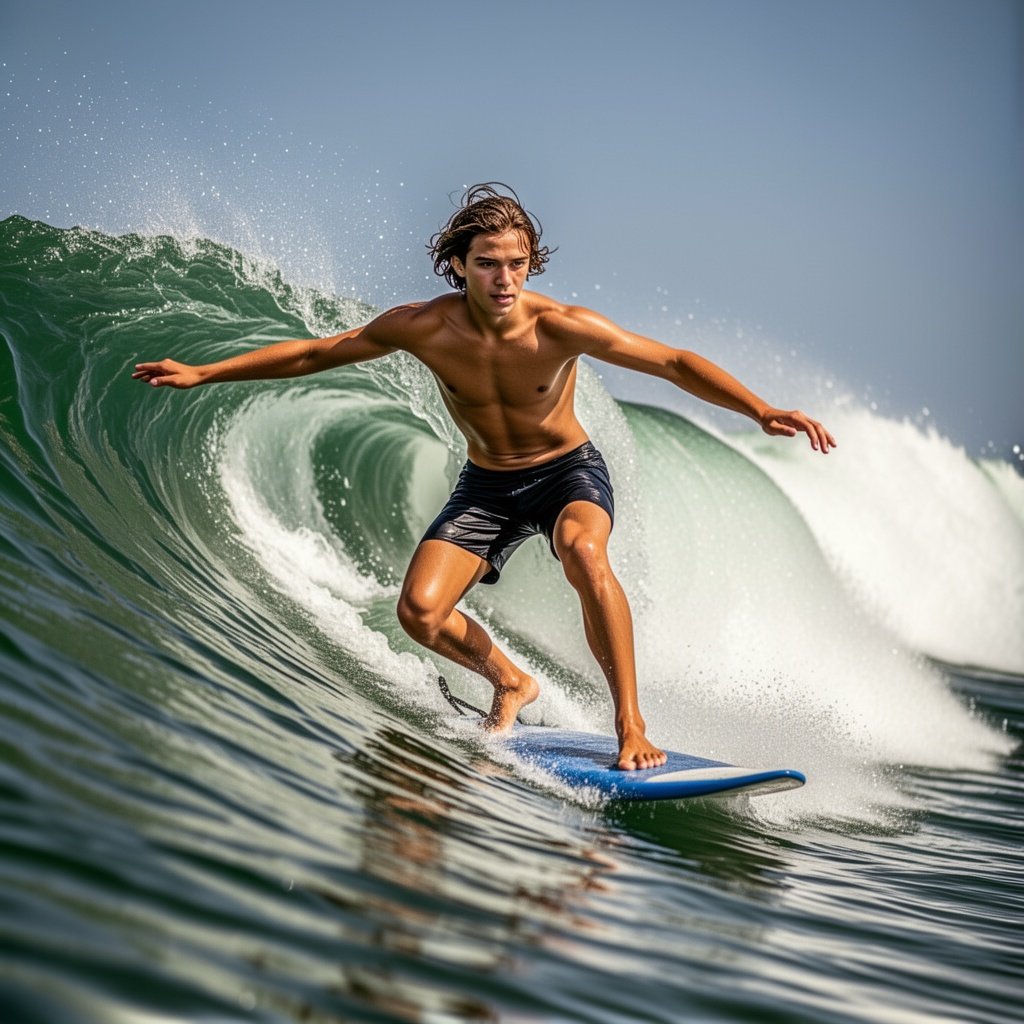}} &
        \includegraphics[width=\imgwidthapp]{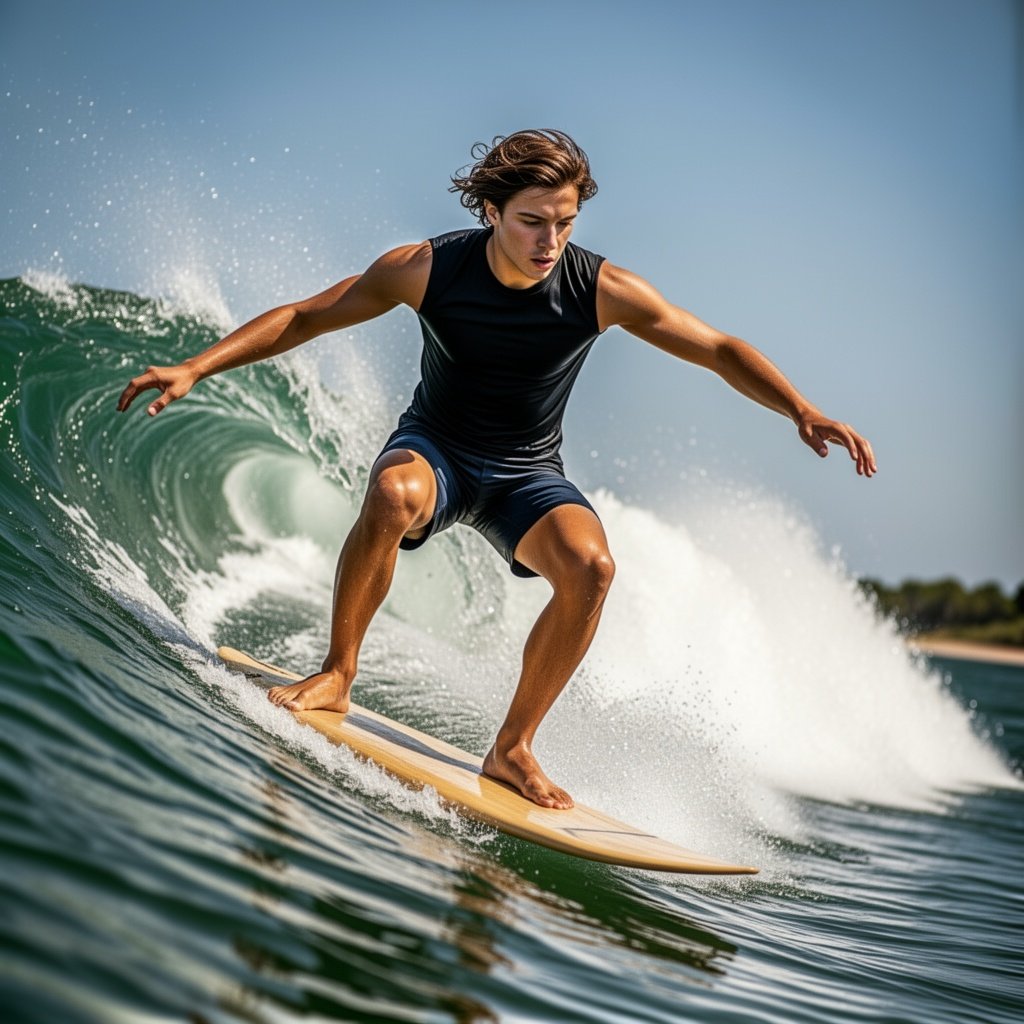} &
        \includegraphics[width=\imgwidthapp]{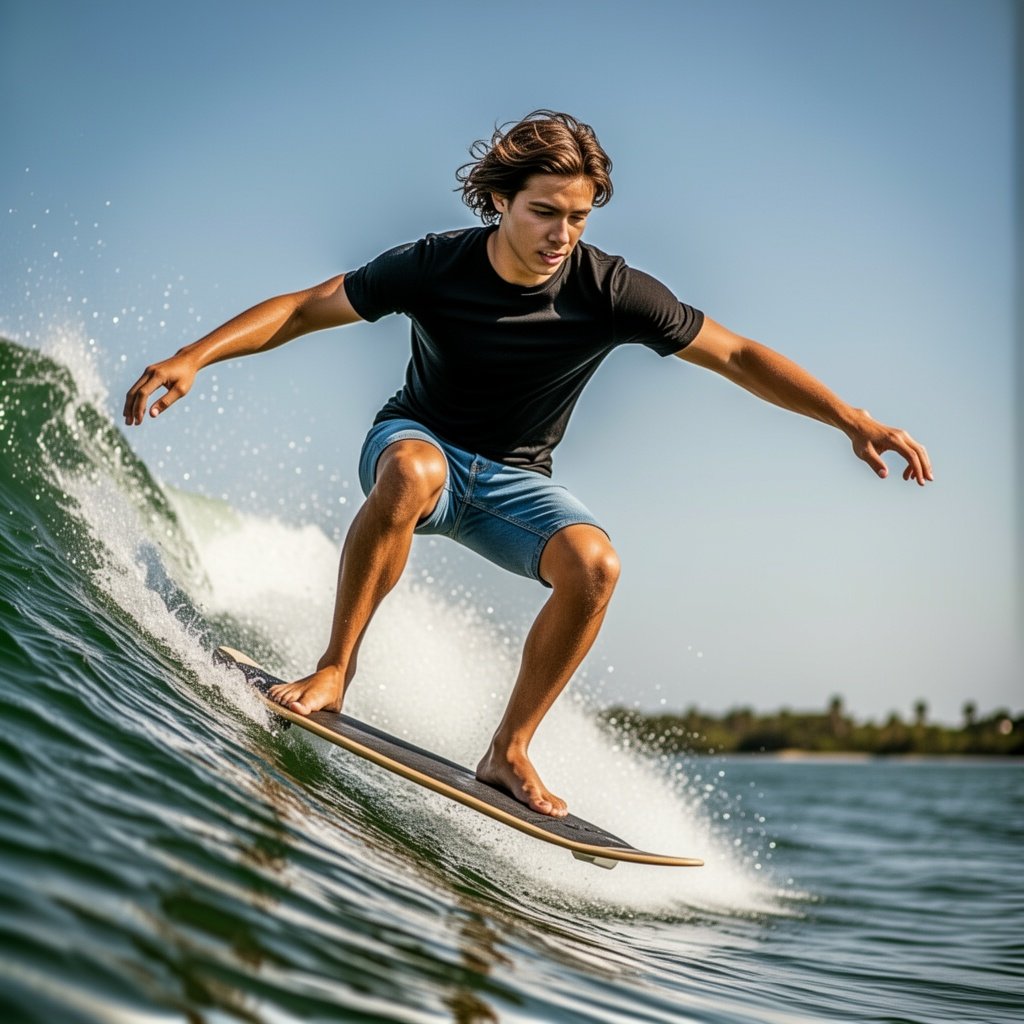} &
        \includegraphics[width=\imgwidthapp]{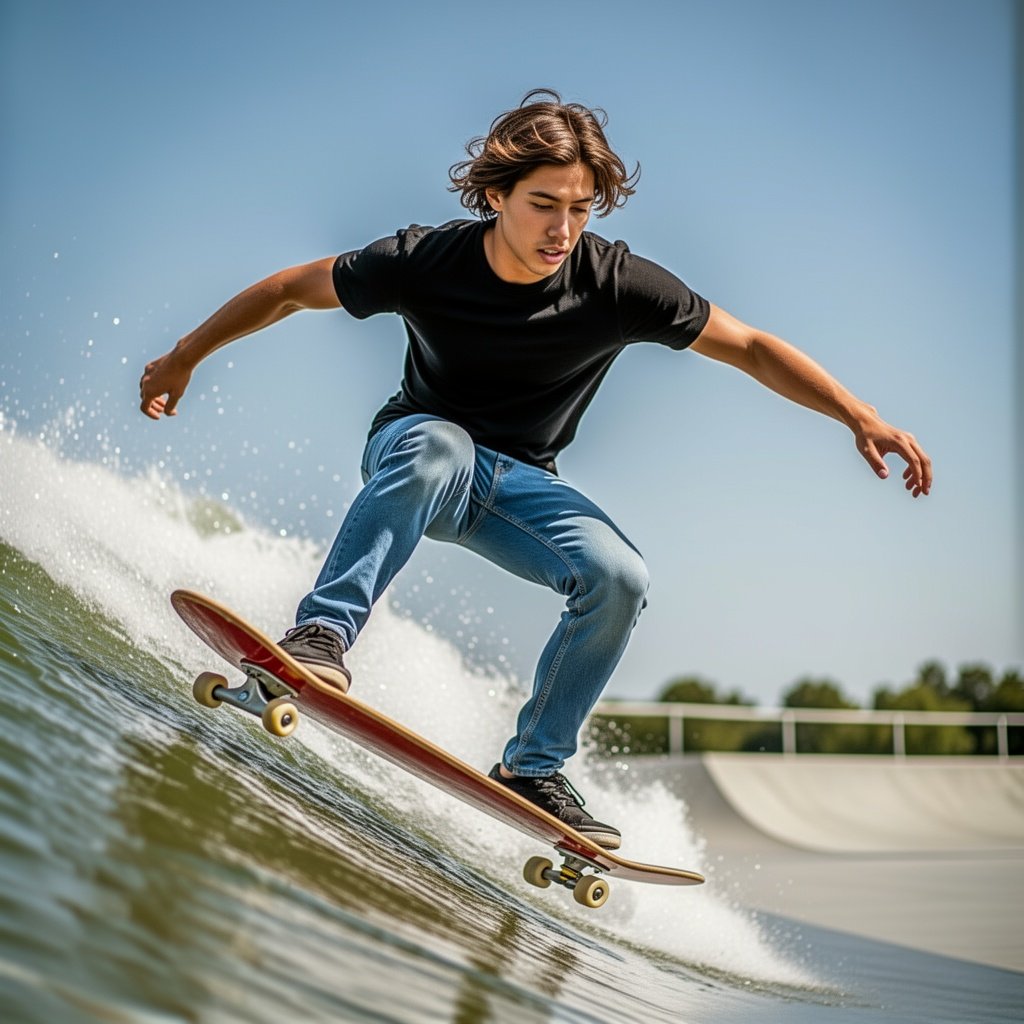} &
        \includegraphics[width=\imgwidthapp]{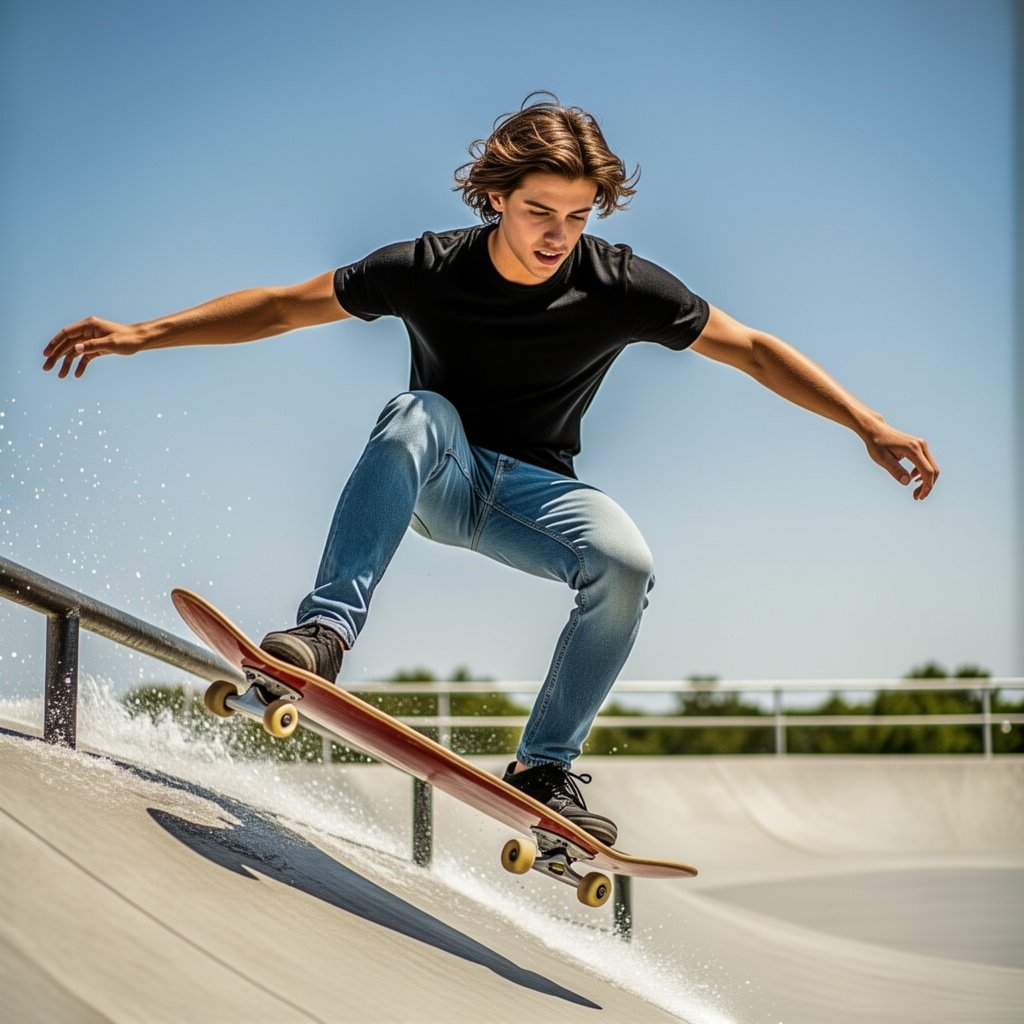} &
        \fcolorbox{red}{white}{\includegraphics[width=\imgwidthapp]{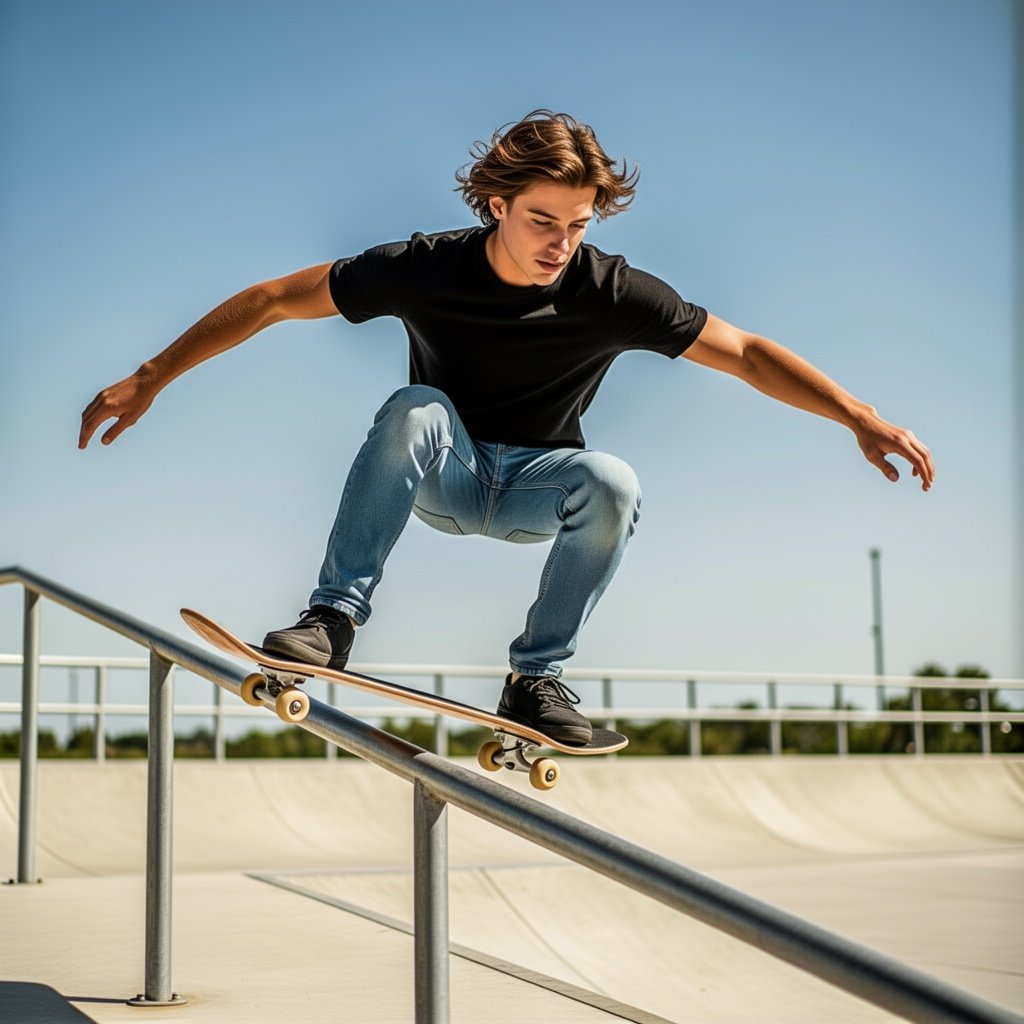}}
    \end{tabular}
    \vspace{-4pt}
    \caption{
    \textbf{Applications enabled by aligned text embedding interpolation.} Once representations are aligned, simple linear interpolation produces coherent semantic transitions. This unified framework supports continuous synthesis (top), image editing (middle), and image blending (bottom), generating meaningful intermediate states across tasks. Red boxes denote input images.
    }
    \vspace{-16pt}
    \label{fig:applications}
\end{figure*}

\vspace{-6pt}
\section{Related Work}

\vspace{-6pt}

\paragraph{\textbf{Structured Text Representations}}
Recent advances in text-to-image diffusion models have shifted from short, underspecified prompts toward richer, more detailed descriptions~\citep{betker2023improving, hu2024ella, liu2025llm4gen, gutflaish2025generating, liu2024playground, kachlon2026bbq}. While alignment between short prompts is often trivial, underspecified prompts can still induce entangled priors~\citep{huberman2026image}, and the increased complexity and variability of richer descriptions introduce challenges in establishing correspondence across prompts.
In natural language processing, canonicalization and structure-aware paraphrasing reduce such variability~\citep{berant2014semantic, huang2021disentangling}, but operate on individual sentences rather than enforcing shared structure across different prompts. In contrast, we jointly transform prompt pairs into a shared structured form, enabling explicit token-level alignment.

More broadly, alignment of representation spaces has been studied across domains, including sequence matching and embedding alignment, using methods such as dynamic time warping~\cite{sakoe1978dynamic}, optimal transport~\citep{peyre2019computational}, and cross-lingual embedding alignment~\citep{mikolov2013exploiting, conneau2017word, artetxe2018robust}. Our setting involves embeddings from a shared text encoder with similar structure, where differences arise from both intended semantic changes (e.g., “cat” vs. “lion”) and local variations in phrasing, tokenization, and contextualization, rather than global modality gaps.

\vspace{-10pt}
\paragraph{\textbf{Latent Space Interpolation}}
Latent spaces in generative models often exhibit meaningful structure, enabling simple operations such as linear interpolation to produce smooth and semantically coherent transitions between images~\citep{shen2020interfacegan, harkonen2020ganspace, patashnik2021styleclip}.
In text-conditioned diffusion models, semantic manipulation has been explored across multiple representation spaces, including noisy latent trajectories~\cite{meng2021sdedit, kulikov2025flowedit, dalva2023noiseclr}, intermediate features such as attention maps~\cite{hertz2022prompt, tumanyan2023plug, liu2026tokendialcontinuousattributecontrol}, and parameter-space modifications such as LoRA-based methods~\cite{gandikota2024concept, dravidinterpreting, zarei2025slideredit, chiu2026text}.
Closest to our work are methods operating in text embedding and contextual representation spaces, including approaches that decompose text embeddings into interpretable components~\citep{kamenetsky2025saedit, baumann2025continuous, yu2024uncovering}, perform per-edit optimization~\citep{kawar2023imagic, wu2023uncovering}, operate in global embedding spaces~\citep{dalva2024fluxspace, garibi2025tokenverseversatilemulticonceptpersonalization}, or directly modify textual representations~\citep{ekin2026unreasonable, dahary2026fly}.
In contrast, we align prompt pairs at the token level, establishing explicit semantic correspondence across their representations.


\vspace{-10pt}
\paragraph{\textbf{Image Editing, Blending and Continuous Control}}
Early diffusion-based editing methods rely on text prompts as the primary control mechanism, but this interface mainly supports discrete changes and provides limited control over transformation strength~\citep{meng2021sdedit, hertz2022prompt, batifol2025flux, brooks2023instructpix2pix}.
A growing body of recent work focuses on continuous control, introducing mechanisms to modulate edit strength.
These mechanisms include guidance-based control~\citep{wolf2026continuous, zhang2025group}, LoRA-based editing directions~\citep{zarei2025slideredit, gandikota2024concept}, and fine-tuned instruction-based editors with explicit strength scalars~\citep{parihar2025kontinuous, xu2025numerikontrol, liu2026tokendialcontinuousattributecontrol}.
Related work has also leveraged video generation for image editing~\citep{rotstein2025pathways} and analyzed and smoothed semantic progression in generated sequences~\citep{metzer2026video}.

Continuous transitions have also been explored through pixel or latent-space interpolation~\citep{cao2025freemorph, zhang2024diffmorpher}, but often produce visual blends with artifacts such as ghosting or inconsistent object composition, rather than semantically meaningful intermediate states. More recent work blends distinct concepts by identifying shared semantic attributes~\citep{yang2025vibe}. 
However, these approaches do not explicitly enforce structured correspondence between concepts, which can lead to intermediate states where semantic attributes are not consistently preserved.

In contrast, our approach enables structured, continuous semantic transitions directly in the text conditioning space, without task-specific training or modifications to the generative model.
Moreover, it generalizes to both spatially similar and spatially distinct images.

\begin{figure}[t]
\vspace{-4pt}
\begin{center}
	\includegraphics[width=1\linewidth]{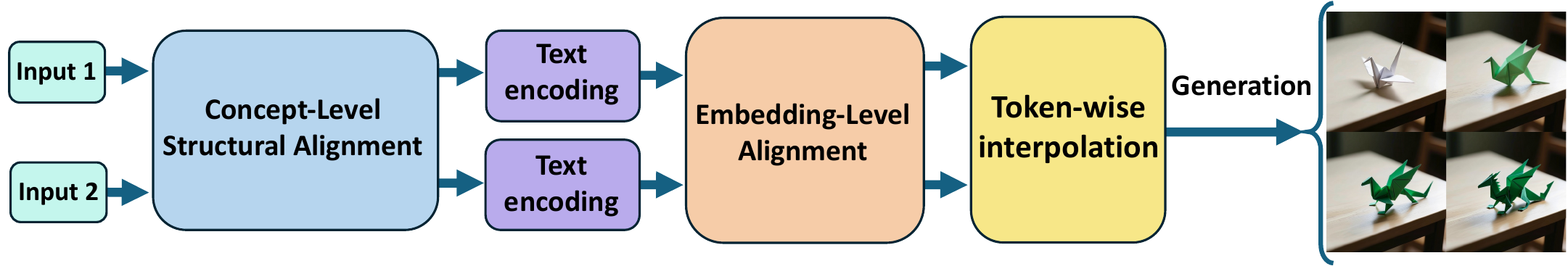}
\end{center}
\vspace{-5pt}

\caption{
\textbf{Method overview.}
Inputs are transformed into structured descriptions under a shared schema with aligned textual phrasing, establishing correspondence between scene components and how they are expressed. Their token embeddings are then aligned in the text encoder space to enforce token-wise semantic correspondence. Interpolation between aligned embeddings produces intermediate representations, which the generative model renders as smooth semantic transitions.}

\label{fig:method_overview}
\end{figure}
\begin{figure}[t]
\vspace{-4pt}
\begin{center}
	\includegraphics[width=1\linewidth]{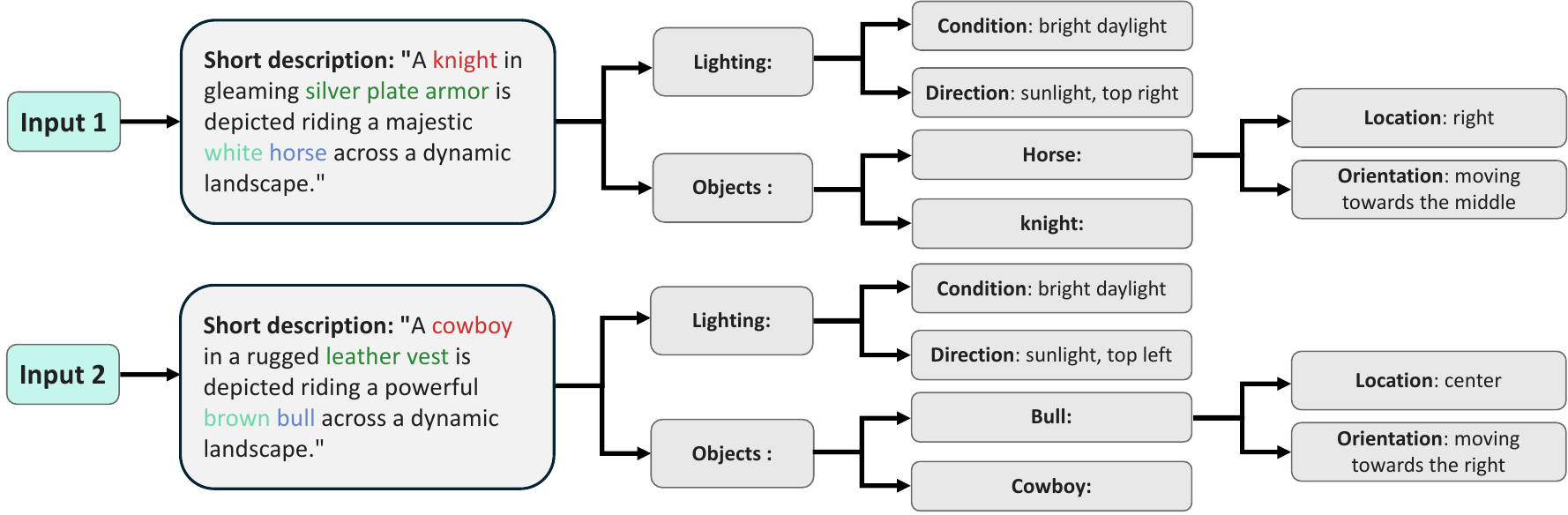}
\end{center}
\vspace{-4pt}

\caption{
\textbf{Concept level structural alignment.}
Inputs are transformed into structured descriptions under a shared schema, where scene components are decomposed into semantic fields (e.g., lighting, objects) with consistent ordering. Corresponding elements are expressed using aligned textual phrasing within each field, establishing coarse correspondence at the textual level.
}

\vspace{-4pt}
\label{fig:structural_alignment}
\end{figure}
\begin{figure}[t]
\vspace{-4pt}
\begin{center}
	\includegraphics[width=1\linewidth]{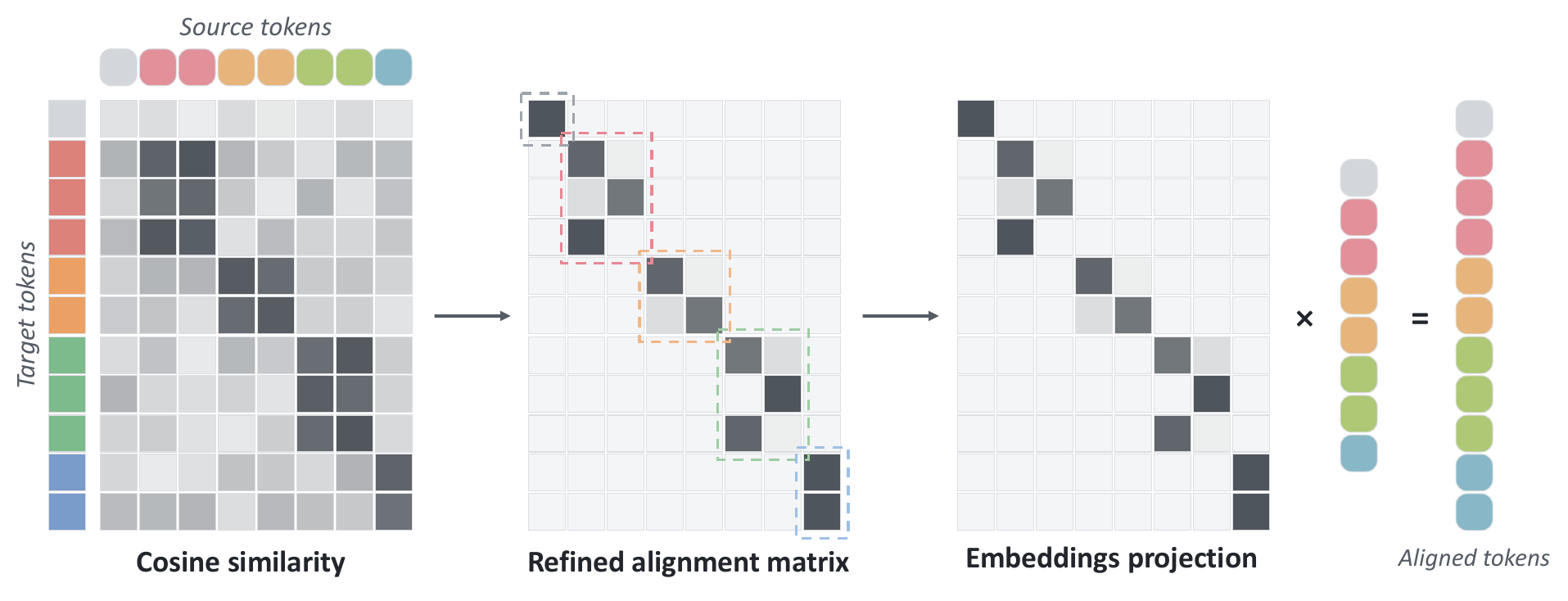}
\end{center}
\vspace{-10pt}
\caption{
\textbf{Embedding-level alignment.}
We compute a pairwise cosine similarity matrix between token embeddings, then refine it using field-based masking and a locality bias, and apply a temperature-scaled softmax to obtain alignment weights. These weights define a projection that maps source embeddings onto the target token structure, re-expressing their semantic content in the target layout and establishing token-wise semantic correspondence.
}

\vspace{-10pt}
\label{fig:embedding_alignment}
\end{figure}
\section{Method}

In this section, we present our framework for continuous semantic control in image generation and editing.
At the core of our framework is a Token-to-Token alignment method that receives as input a pair of natural language text prompts $p^A, p^B$, and produces a corresponding pair of aligned token embedding sequences $E^A, E^B$. These aligned sequences establish semantic correspondences across tokens, enabling meaningful interpolation between the prompts.

Given the aligned embedding sequences, $E^A$ and $E^B$, we interpolate between them to obtain intermediate representations that capture gradual semantic transitions. These interpolated embeddings are then used to condition a generative model $G$, producing a sequence of images along the transition.
The generative model is a diffusion model, which takes Gaussian noise as input and progressively denoises it while being conditioned on a text embedding. In our case, this conditioning is provided by the interpolated embeddings produced by our alignment method. For applications that require an input image, we use diffusion models that additionally support image conditioning alongside the text embedding. 
Specifically, we demonstrate our framework using FIBO~\cite{gutflaish2025generating} and Flux2~\cite{flux-2-2025}. FIBO has two variants: one conditioned solely on a text prompt, and another that additionally supports image conditioning. In Flux2, the same model can be used with or without image conditioning.

Next, we present our Token-to-Token method, and then in Section~\ref{sec:applications} we describe how it is applied to different generative applications. 
An overview of our method is shown in Figure~\ref{fig:method_overview}.

\subsection{Token-to-Token}

\paragraph{Concept-Level Structural Alignment}

The first stage of Token-to-Token establishes concept-level structural alignment using an LLM. Given a pair of prompts $p^A, p^B$, we instruct the LLM to jointly construct aligned scene descriptions $J^A, J^B$ under explicit structural constraints. The goal of this stage is to reduce linguistic variability and enforce consistent semantic structure prior to tokenization.

We illustrate a pair of aligned scene descriptions in Figure~\ref{fig:structural_alignment}. Each description is represented as a JSON string with the same fixed semantic fields (e.g., objects, attributes, lighting) and consistent ordering. Corresponding elements use parallel phrasing within each field, such that semantic differences appear as local substitutions within a shared context rather than independent rephrasing (e.g., \textit{knight} vs. \textit{cowboy}, \textit{horse} vs. \textit{bull}), highlighted by matching colors in the figure.

While these descriptions establish concept-level alignment, they operate prior to tokenization and do not guarantee correspondence in the text encoder space. Semantically corresponding words (e.g., \textit{knight} vs. \textit{cowboy}) may tokenize into different numbers of tokens, shifting token positions and misaligning the resulting embedding sequences.

\vspace{-6pt}
\paragraph{Embedding-Level Alignment}
\label{sec:embedding_alignemnt}

Let $V^A = \{v^A_i\}_{i=1}^{n}$ and $V^B = \{v^B_j\}_{j=1}^{m}$ denote the token embeddings produced by the text encoder for $J^A$ and $J^B$, respectively. Due to differences in phrasing and tokenization, these sequences may differ in length and token boundaries, even after structural alignment.

To address this, we perform embedding-level alignment to establish token correspondences across the two sequences, as illustrated in Figure~\ref{fig:embedding_alignment}. We first compute a pairwise similarity matrix $S \in \mathbb{R}^{n \times m}$ using cosine similarity:

\vspace{-8pt}

\[
S_{ij} = \frac{\langle v^A_i, v^B_j \rangle}{\|v^A_i\| \|v^B_j\|}.
\]

\vspace{-4pt}

Token embeddings from transformer-based text encoders are contextual and reflect semantic roles within the description, enabling similarity-based alignment rather than exact token matching, even when corresponding elements differ in value (e.g., colors, clothing, or expressions).

Alignment is constrained by the semantic structure obtained in the previous stage. Specifically, we restrict matching to tokens within corresponding semantic fields by masking $S$, ensuring that correspondence respects the scene decomposition. To encourage stable correspondences, we incorporate a locality bias over token positions and apply temperature scaling:
$\tilde{S}_{ij} = {S_{ij}}/{\tau} + \lambda \cdot \text{pos}(i,j)$,
where $\text{pos}(i,j)$ is a Gaussian bias over relative token positions within each semantic field. Following structural alignment, corresponding tokens are likely to appear at similar relative positions, motivating the locality bias. Temperature scaling amplifies similarity differences before normalization, preventing correspondence weights from spreading across many unrelated tokens.

We obtain a soft alignment matrix $A$ by normalizing similarities over tokens in $V^A$,
$A_{ij} = \exp(\tilde{S}_{ij})/{\sum_{i'} \exp(\tilde{S}_{i'j})}$.
For each token in $V^B$, this defines a soft correspondence over tokens in $V^A$, allowing sequences of different lengths to be matched. We then project the semantic content of $V^A$ onto the structure of $V^B$, yielding $E^A = A^\top V^A$ and $E^B = V^B$, where $\top$ denotes transpose.

\vspace{-6pt}
\paragraph{Interpolation in Aligned Embedding Space}
\label{sec:interpolation}

Given the aligned token embedding sequences $E^A = \{e_j^A\}_{j=1}^{m}$ and $E^B = \{e_j^B\}_{j=1}^{m}$, we interpolate between them by linearly combining corresponding embeddings:
\[
E^\alpha = \{ (1-\alpha)e_j^A + \alpha e_j^B \}_{j=1}^{m}, \quad \alpha \in [0,1].
\]

\vspace{-6pt}
Crucially, interpolation is performed between semantically corresponding tokens, as established by the alignment stages, ensuring that intermediate representations reflect gradual and coherent semantic changes rather than mixtures of unrelated attributes. These interpolated embeddings can be directly used as conditioning for the generative model, which renders them as smooth semantic transitions in the image space.

\subsection{Applications}
\label{sec:applications}

Natural language can specify the direction of a semantic change, but offers limited control over its magnitude. The interpolation coefficient $\alpha$ provides continuous control over the transformation strength. We next describe how Token-to-Token enables continuous synthesis, editing, and blending.

\vspace{-8pt}
\paragraph{Continuous Synthesis}
In this application, shown in the top row of Figure~\ref{fig:applications}, the input consists of a pair of prompts $p^A, p^B$ that differ in a specific attribute (e.g., the object on the couch). We apply Token-to-Token to construct a sequence of interpolated embeddings $\{E^\alpha\}_{\alpha \in [0,1]}$, and generate the corresponding images $\{G(E^\alpha)\}_{\alpha \in [0,1]}$ using text-only conditioning.

\vspace{-8pt}
\paragraph{Continuous Editing}
In this application, shown in the middle row of Figure~\ref{fig:applications}, the input consists of an image $I$ and a text prompt $p^B$ describing the desired edit. We first caption the input image to obtain a source prompt $p^A$, and then apply Token-to-Token to align $p^A$ with the target prompt $p^B$ and construct a sequence of interpolated embeddings $\{E^\alpha\}_{\alpha \in [0,1]}$. We generate the corresponding images $\{G(I, E^\alpha)\}_{\alpha \in [0,1]}$ by conditioning the model on both the input image and each interpolated embedding. In this setting, we use Flux2 and the image-conditioned variant of FIBO. This produces a continuous transition from the input image toward the edited result, enabling gradual control over the edit strength.

\vspace{-8pt}
\paragraph{Continuous Blending}
In this application, shown in the bottom row of Figure~\ref{fig:applications}, the input consists of a pair of images $I^A, I^B$. We caption the images to obtain a pair of prompts $p^A, p^B$, and apply Token-to-Token to construct a sequence of interpolated embeddings $\{E^\alpha\}_{\alpha \in [0,1]}$. We generate the corresponding images $\{G(I^A, I^B, E^\alpha)\}_{\alpha \in [0,1]}$ by conditioning the model on both input images and each interpolated embedding.
To incorporate both input images, their embeddings are concatenated in the model input and distinguished using separate temporal positional encodings, allowing the model to identify their respective roles. The interaction between the generated image, the input images, and the text embeddings is handled through the attention layers of the diffusion transformer, where tokens attend to each other to exchange information.

A key challenge in this application is balancing the contribution of the two input images throughout the interpolation. Naively conditioning on both images can lead to dominance of visual features, reducing the process to appearance-based blending. To address this, we modulate the attention weights using an $\alpha$-dependent scaling, weighting the contributions of $I^A$ and $I^B$ by $(1-\alpha)^2$ and $\alpha^2$, respectively. This emphasizes image conditioning near the endpoints to preserve fidelity, while reducing it in intermediate regions to better follow the interpolated semantic trajectory. This results in coherent blending, where semantic attributes evolve smoothly while remaining visually grounded. In this setting, we use Flux2 and the image-conditioned variant of FIBO.
\section{Experiments and Results}
\label{sec:experiments}

We evaluate our framework on both applications through qualitative (Section~\ref{sec:qualitative}) and quantitative (Section~\ref{sec:quantitative}) comparisons, and conduct ablation studies (Section~\ref{sec:ablation}) to assess the contribution of key components.


\renewcommand{\arraystretch}{0}

\begin{figure*}[t]
    \centering
    \small
    \newcommand{\gw}{0.098\linewidth} 
    \setlength{\tabcolsep}{0pt} 
    
    \begin{tabular}{c @{\hspace{1pt}} ccccc @{\hspace{2pt}} ccccc}
        & \includegraphics[width=\gw]{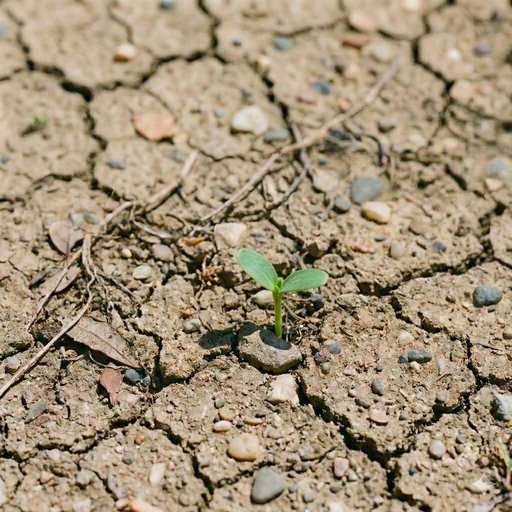} & \multicolumn{4}{c}{\textit{\shortstack{``Grow the sprout into a vibrant orange \\ California poppy flower.''}}}  & \includegraphics[width=\gw]{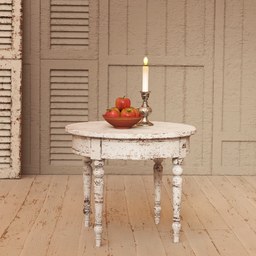} & \multicolumn{4}{c}{\textit{``Change the room to a forest''}} \\

        \raisebox{16pt}{\rotatebox[origin=t]{90}{\scriptsize{Kontinuous K.}}} &
        \includegraphics[width=\gw]{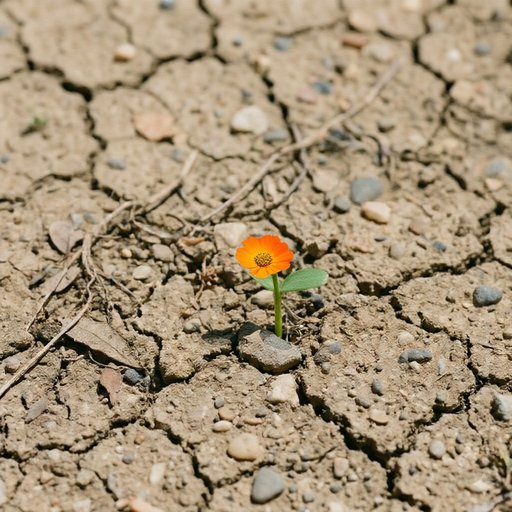} &
        \includegraphics[width=\gw]{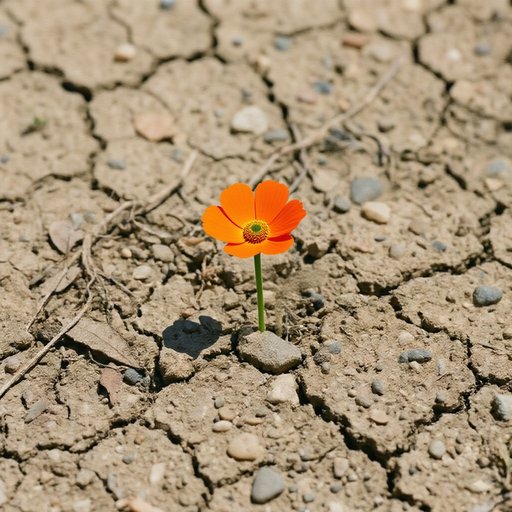} &
        \includegraphics[width=\gw]{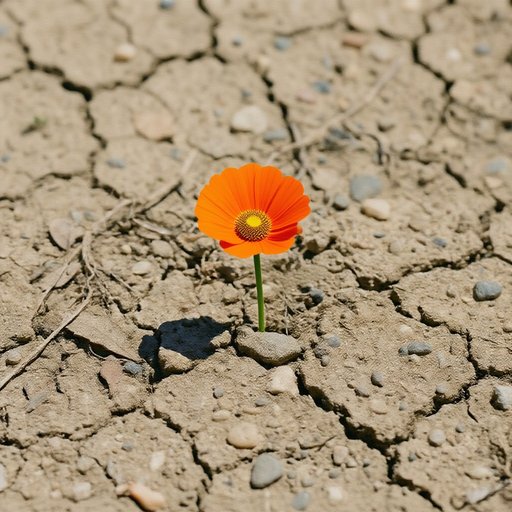} &
        \includegraphics[width=\gw]{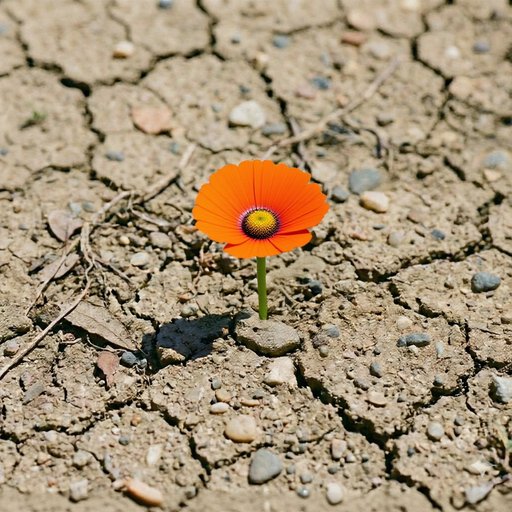} &
        \includegraphics[width=\gw]{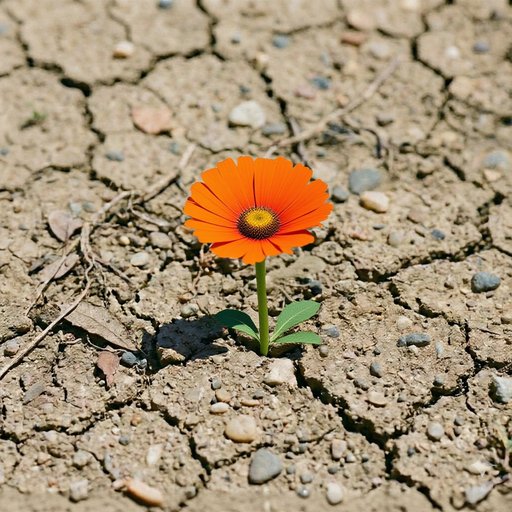} &
        \includegraphics[width=\gw]{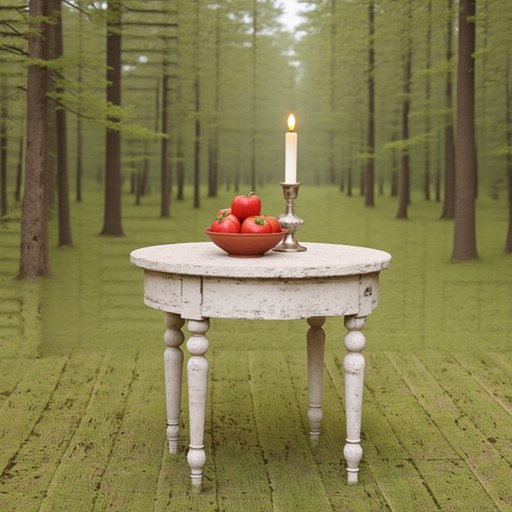} &
        \includegraphics[width=\gw]{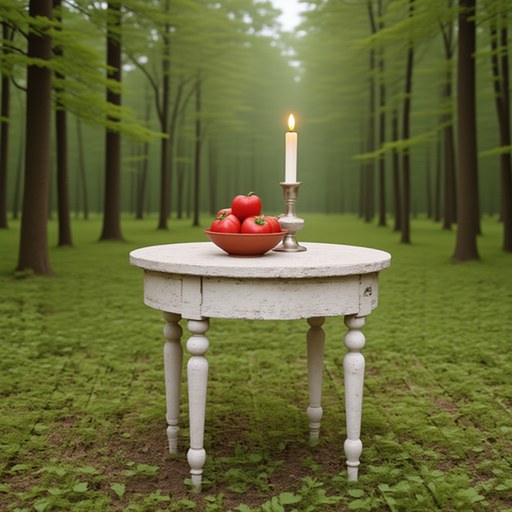} &
        \includegraphics[width=\gw]{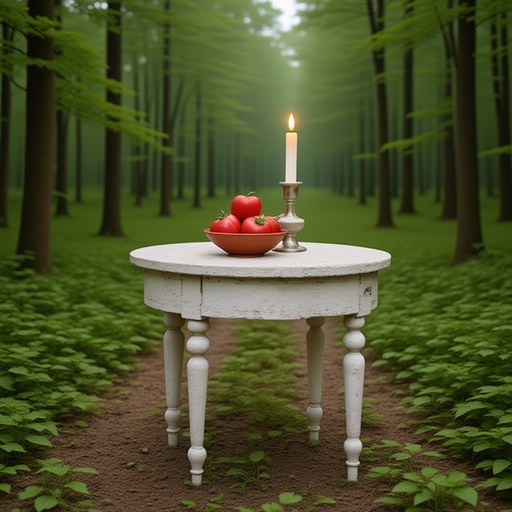} &
        \includegraphics[width=\gw]{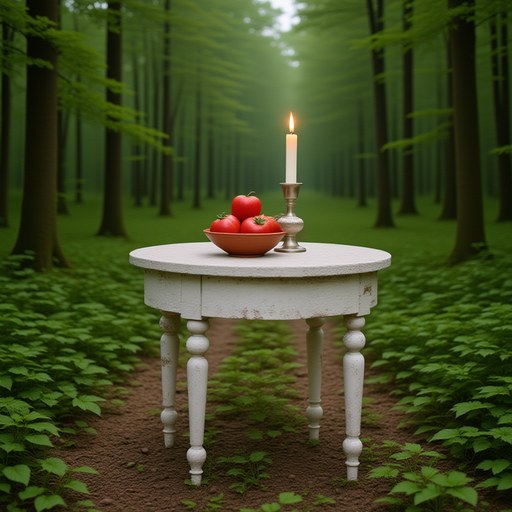} &
        \includegraphics[width=\gw]{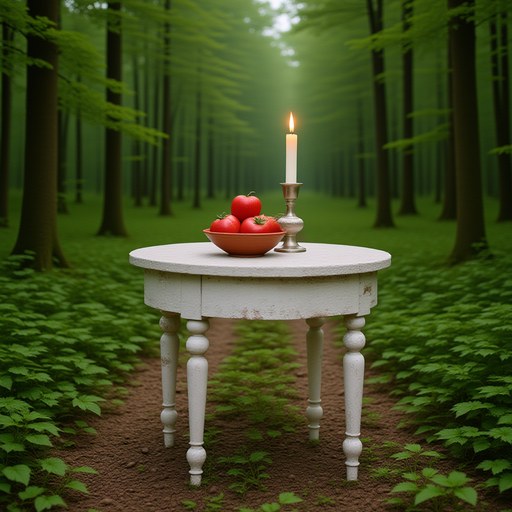} \\

        \raisebox{15pt}{\rotatebox[origin=t]{90}{\scriptsize{SliderEdit}}} &
        \includegraphics[width=\gw]{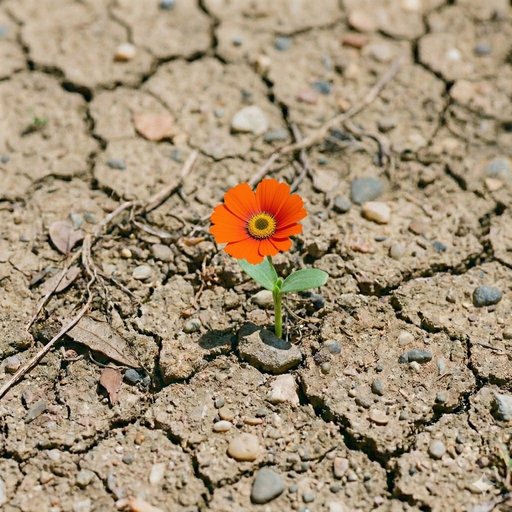} &
        \includegraphics[width=\gw]{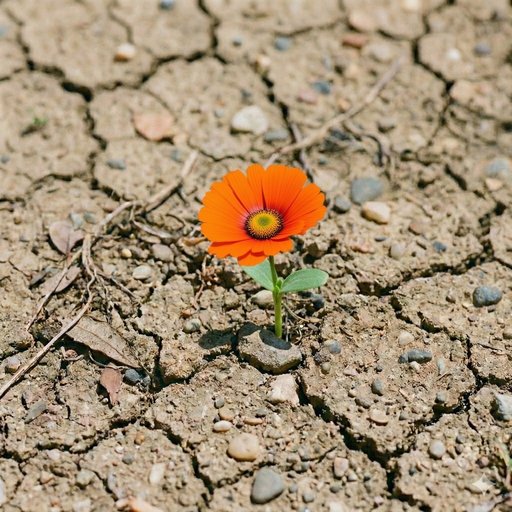} &
        \includegraphics[width=\gw]{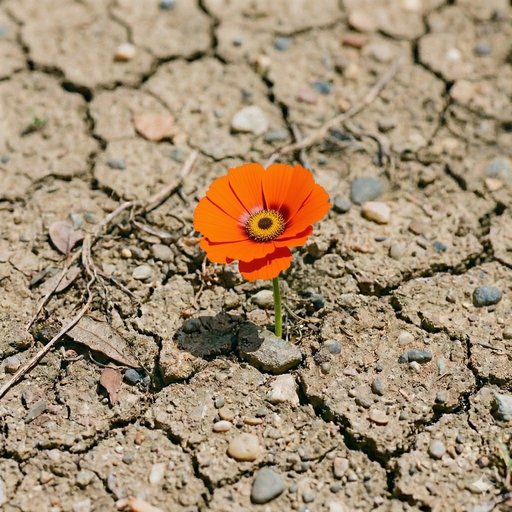} &
        \includegraphics[width=\gw]{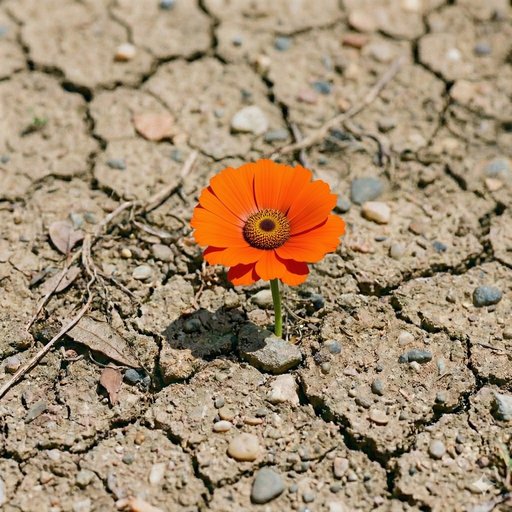} &
        \includegraphics[width=\gw]{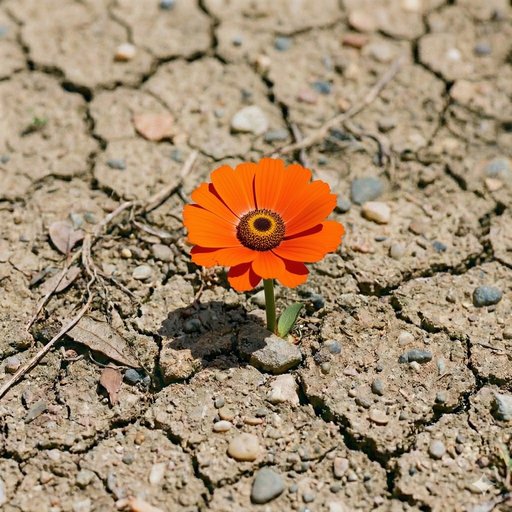} &
        \includegraphics[width=\gw]{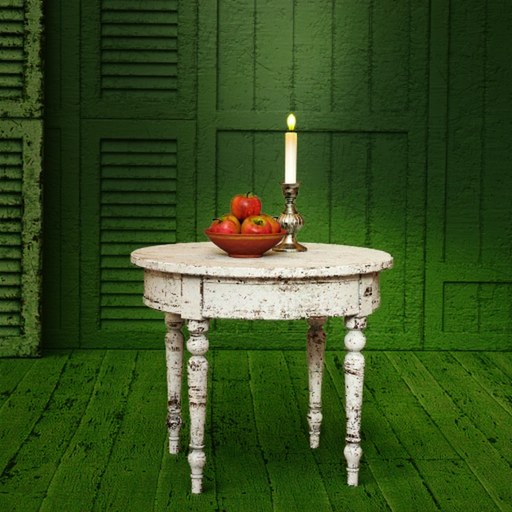} &
        \includegraphics[width=\gw]{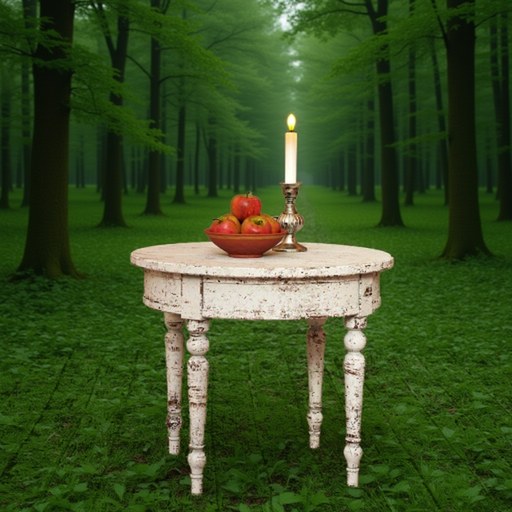} &
        \includegraphics[width=\gw]{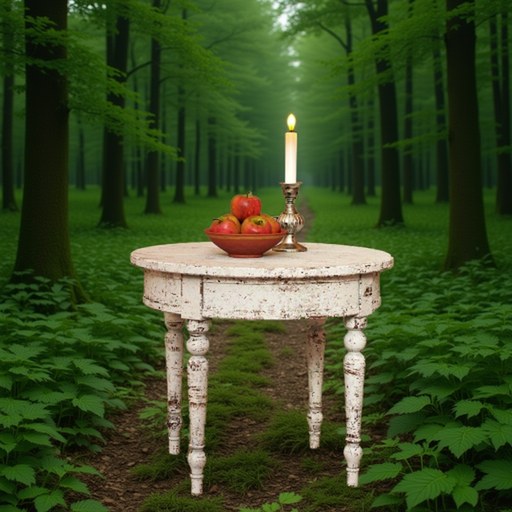} &
        \includegraphics[width=\gw]{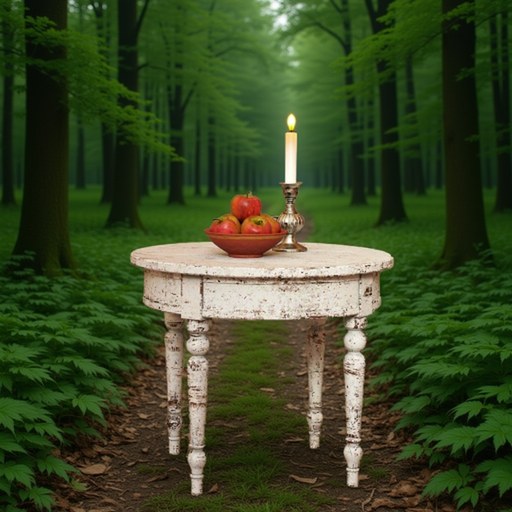} &
        \includegraphics[width=\gw]{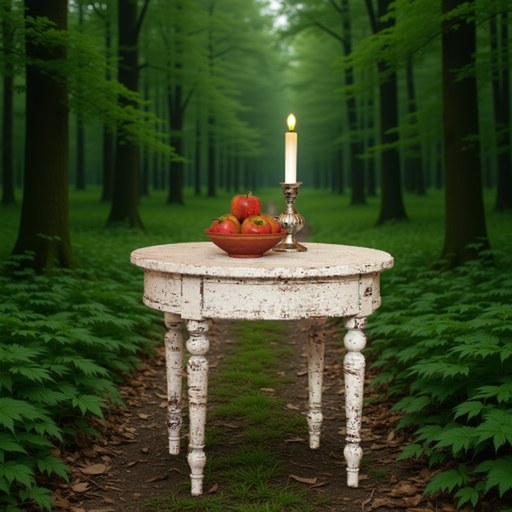} \\

        \raisebox{15pt}{\rotatebox[origin=t]{90}{\scriptsize{GRAG}}} &
        \includegraphics[width=\gw]{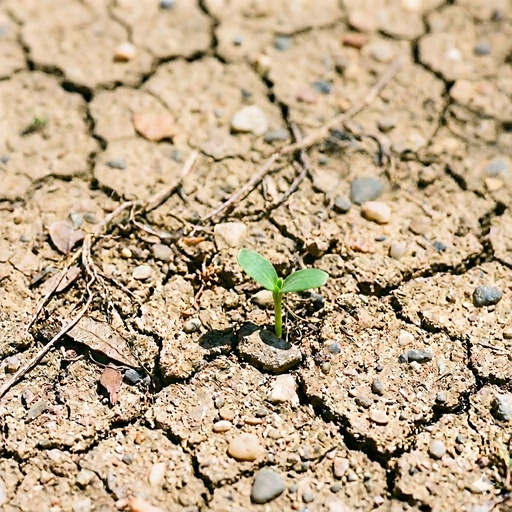} &
        \includegraphics[width=\gw]{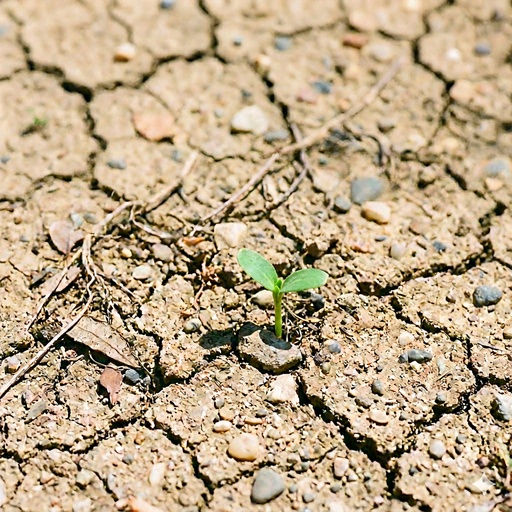} &
        \includegraphics[width=\gw]{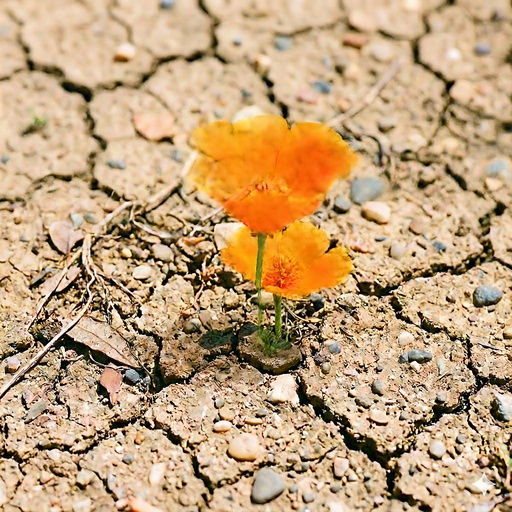} &
        \includegraphics[width=\gw]{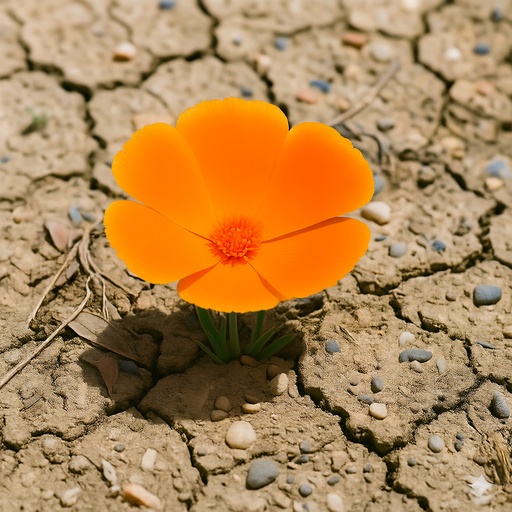} &
        \includegraphics[width=\gw]{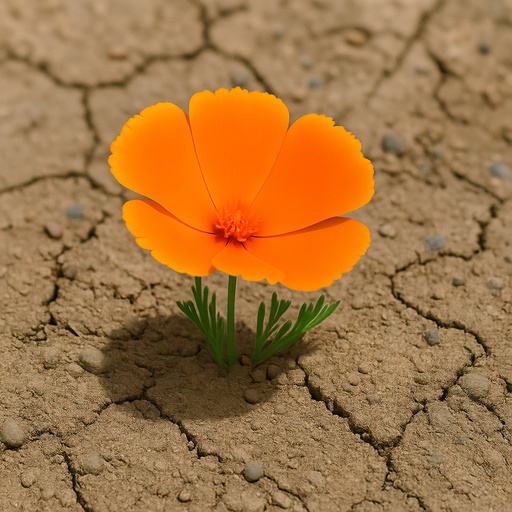} &
        \includegraphics[width=\gw]{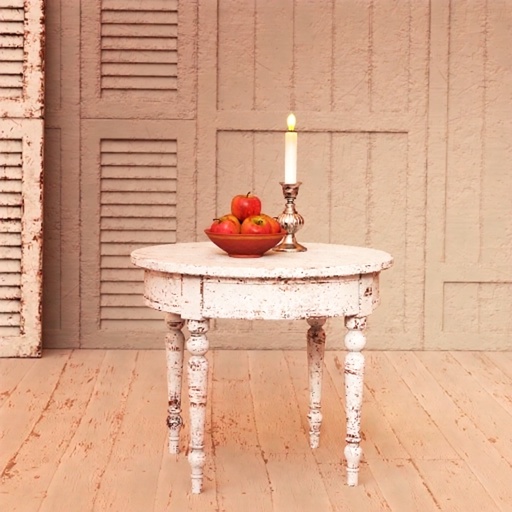} &
        \includegraphics[width=\gw]{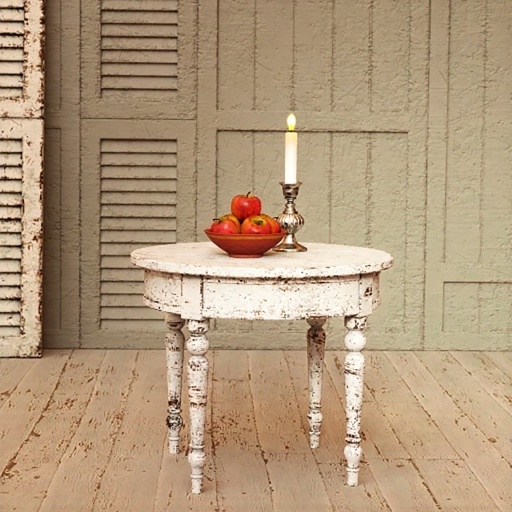} &
        \includegraphics[width=\gw]{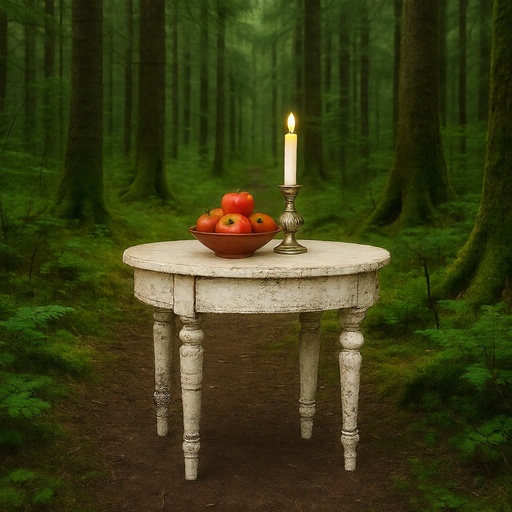} &
        \includegraphics[width=\gw]{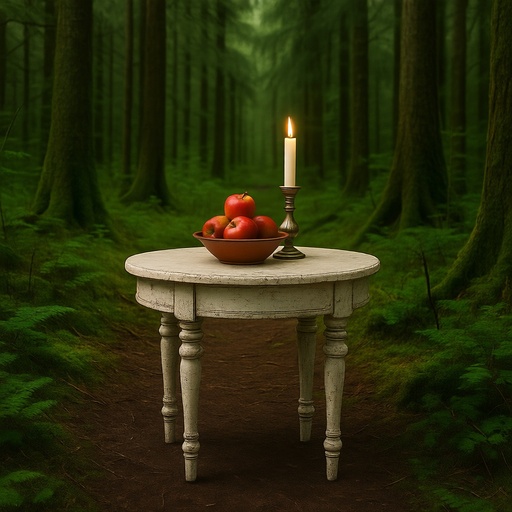} &
        \includegraphics[width=\gw]{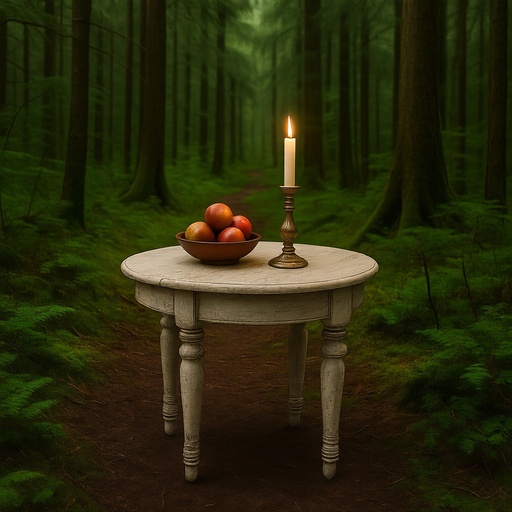} \\

        \raisebox{15pt}{\rotatebox[origin=t]{90}{\scriptsize{T2T (Ours)}}} &
        \includegraphics[width=\gw]{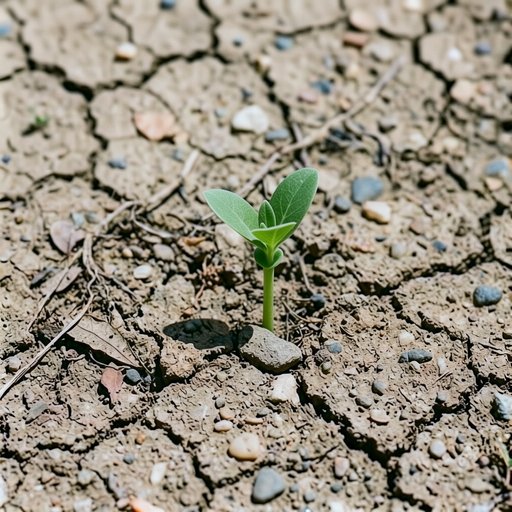} &
        \includegraphics[width=\gw]{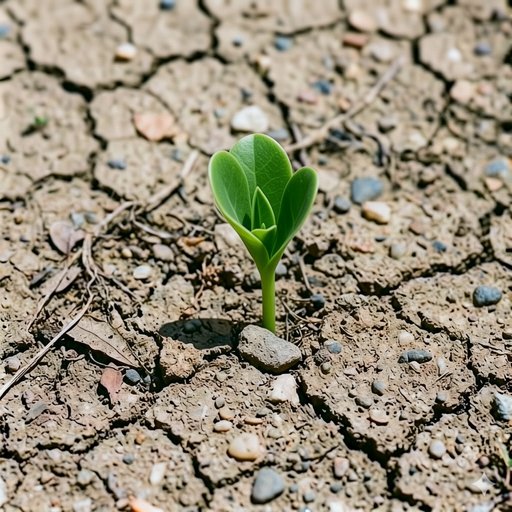} &
        \includegraphics[width=\gw]{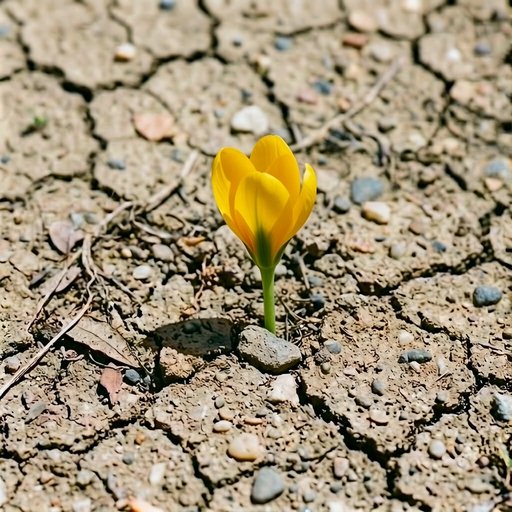} &
        \includegraphics[width=\gw]{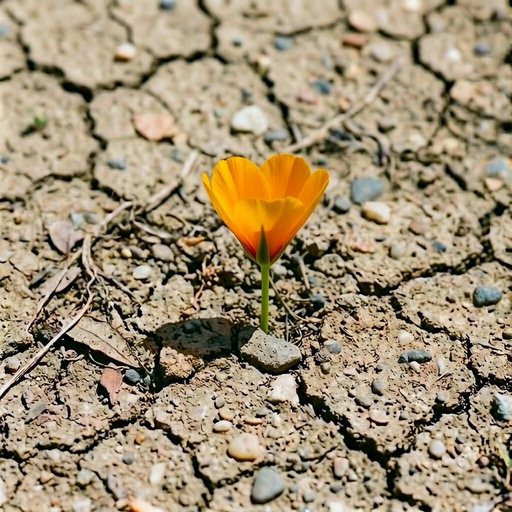} &
        \includegraphics[width=\gw]{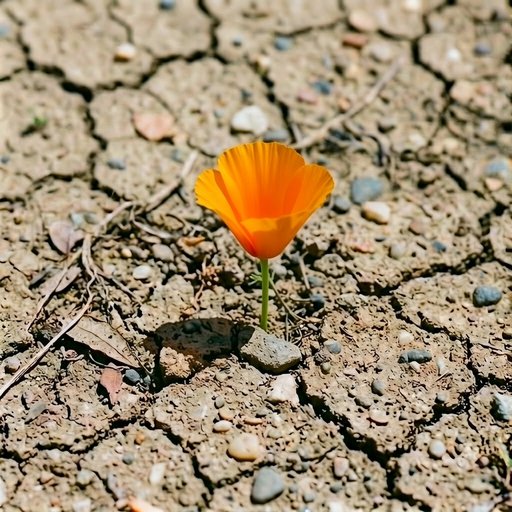} &
        \includegraphics[width=\gw]{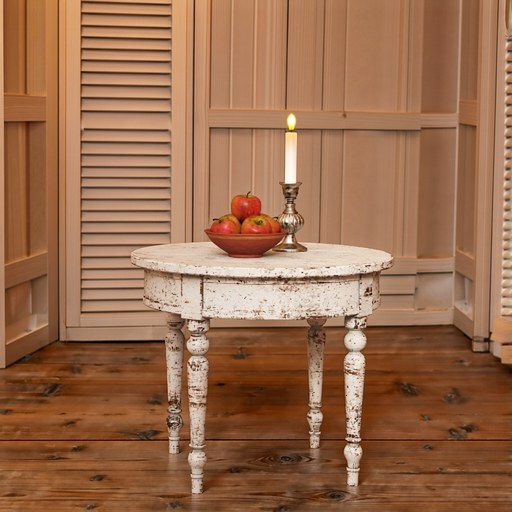} &
        \includegraphics[width=\gw]{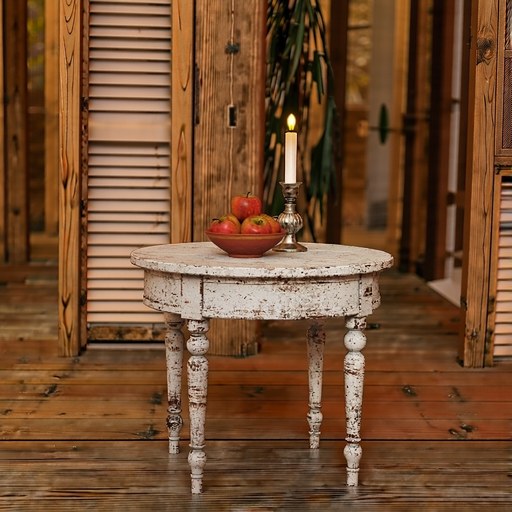} &
        \includegraphics[width=\gw]{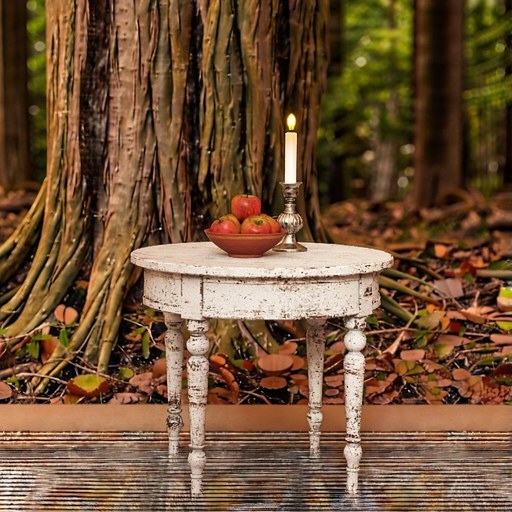} &
        \includegraphics[width=\gw]{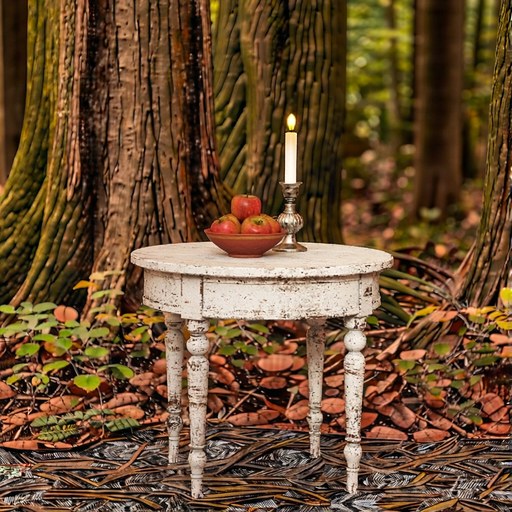} &
        \includegraphics[width=\gw]{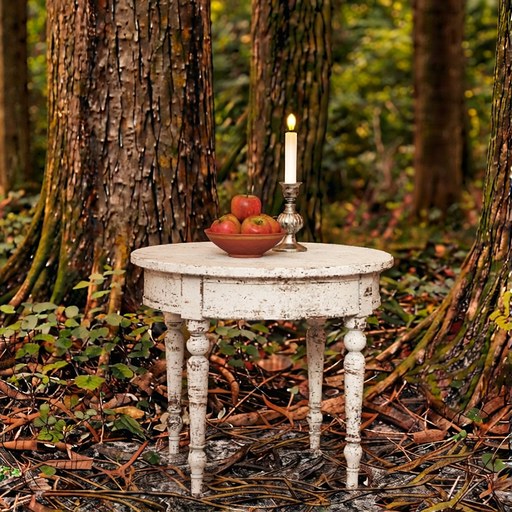} \\

    \noalign{\vspace{3pt}} 


    & \scriptsize{0.2} & \scriptsize{0.4} & \scriptsize{0.6} & \scriptsize{0.8} & \scriptsize{1.0} & \scriptsize{0.2} & \scriptsize{0.4} & \scriptsize{0.6} & \scriptsize{0.8} & \scriptsize{1.0} \\

    \end{tabular}
    \vspace{-4pt}
    
    \caption{\textbf{Qualitative comparison with continuous editing methods.} 
    Existing methods often rely on appearance-based interpolation or introduce abrupt semantic changes, whereas our method produces smooth and semantically coherent transitions with meaningful intermediate states.}

    \vspace{-12pt}
    \label{fig:edit_comparison}
\end{figure*}
\vspace{-4pt}
\renewcommand{\arraystretch}{1}

\vspace{-8pt}
\renewcommand{\arraystretch}{0}

\begin{figure*}[t]
    \centering
    \small
    \newcommand{\gw}{0.098\linewidth} 
    \setlength{\tabcolsep}{0pt} 
    
    \begin{tabular}{c @{\hspace{1pt}} ccccc @{\hspace{2pt}} ccccc}
        & \scriptsize{Input A} & & & & \scriptsize{Input B} & \scriptsize{Input A} & & & & \scriptsize{Input B} \\
        & \includegraphics[width=\gw]{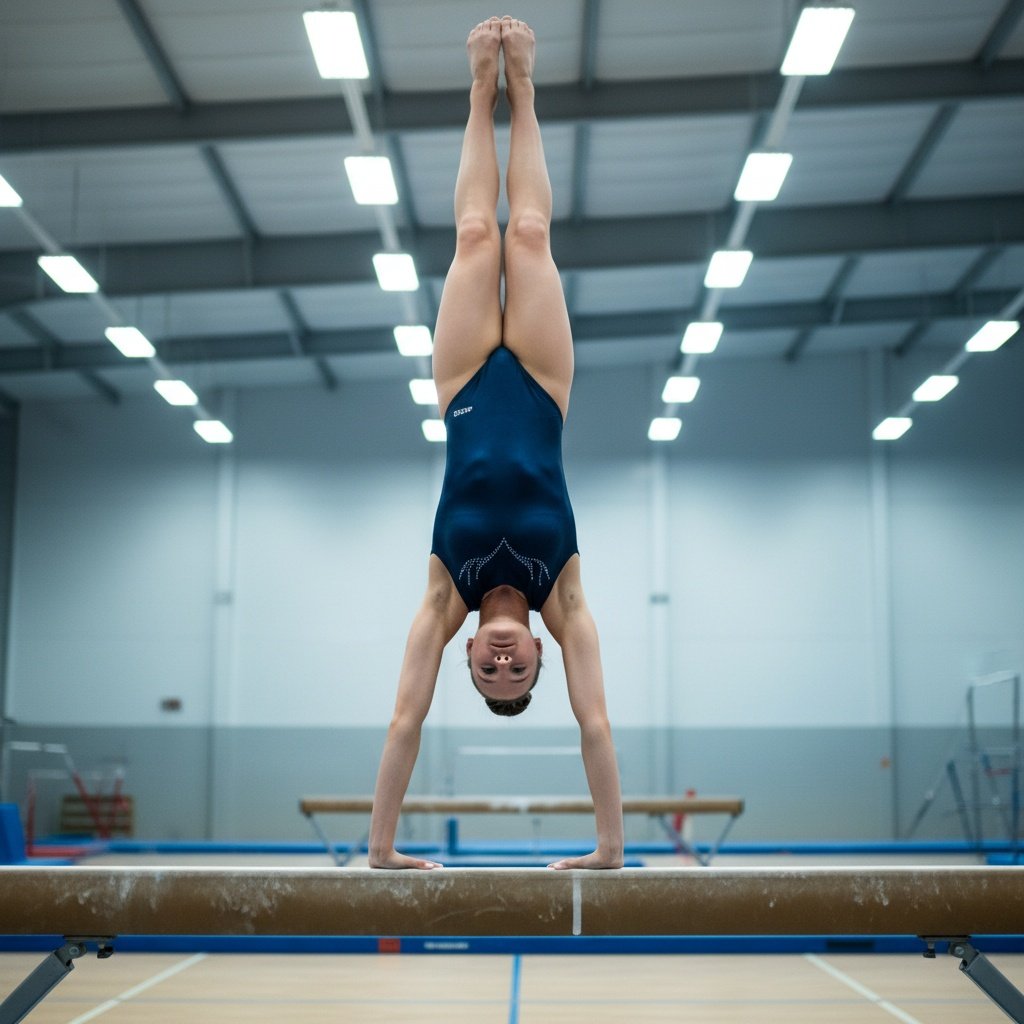} & & & & \includegraphics[width=\gw]{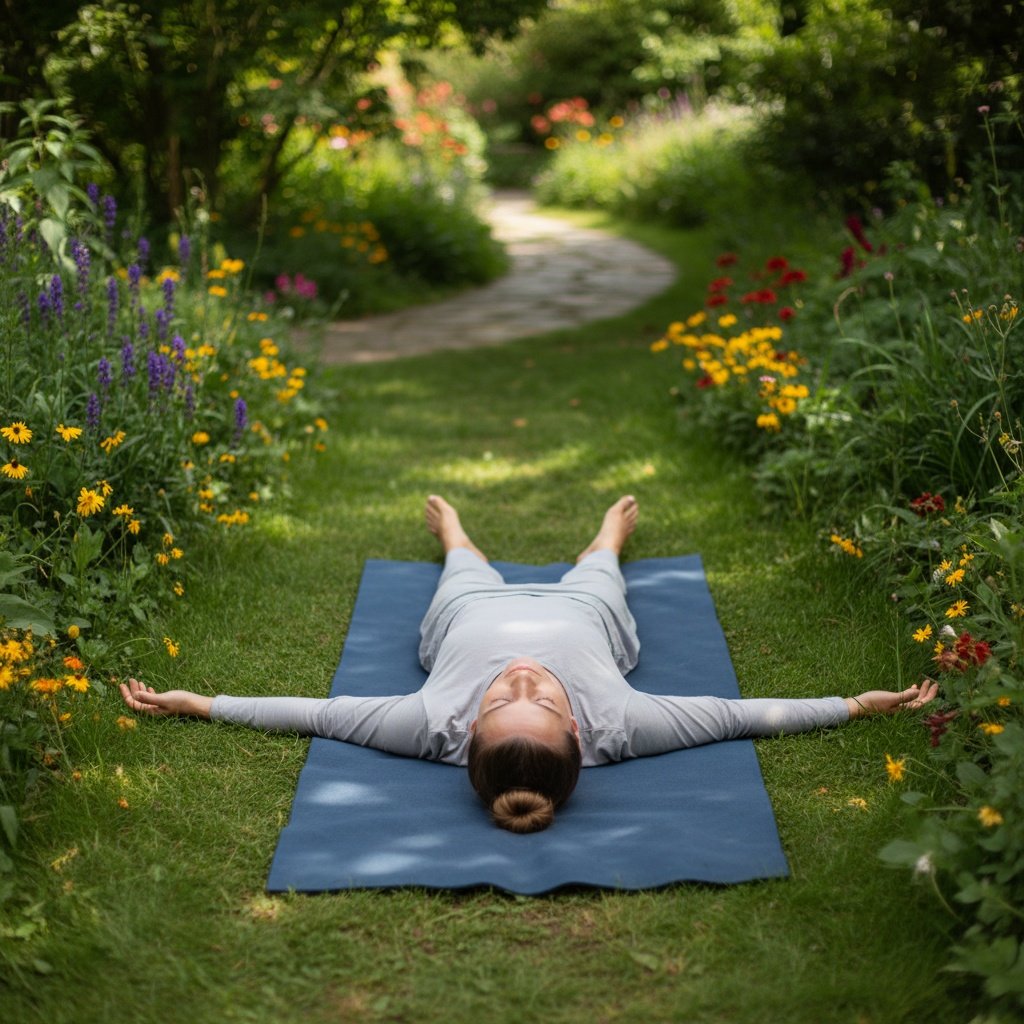} & \includegraphics[width=\gw]{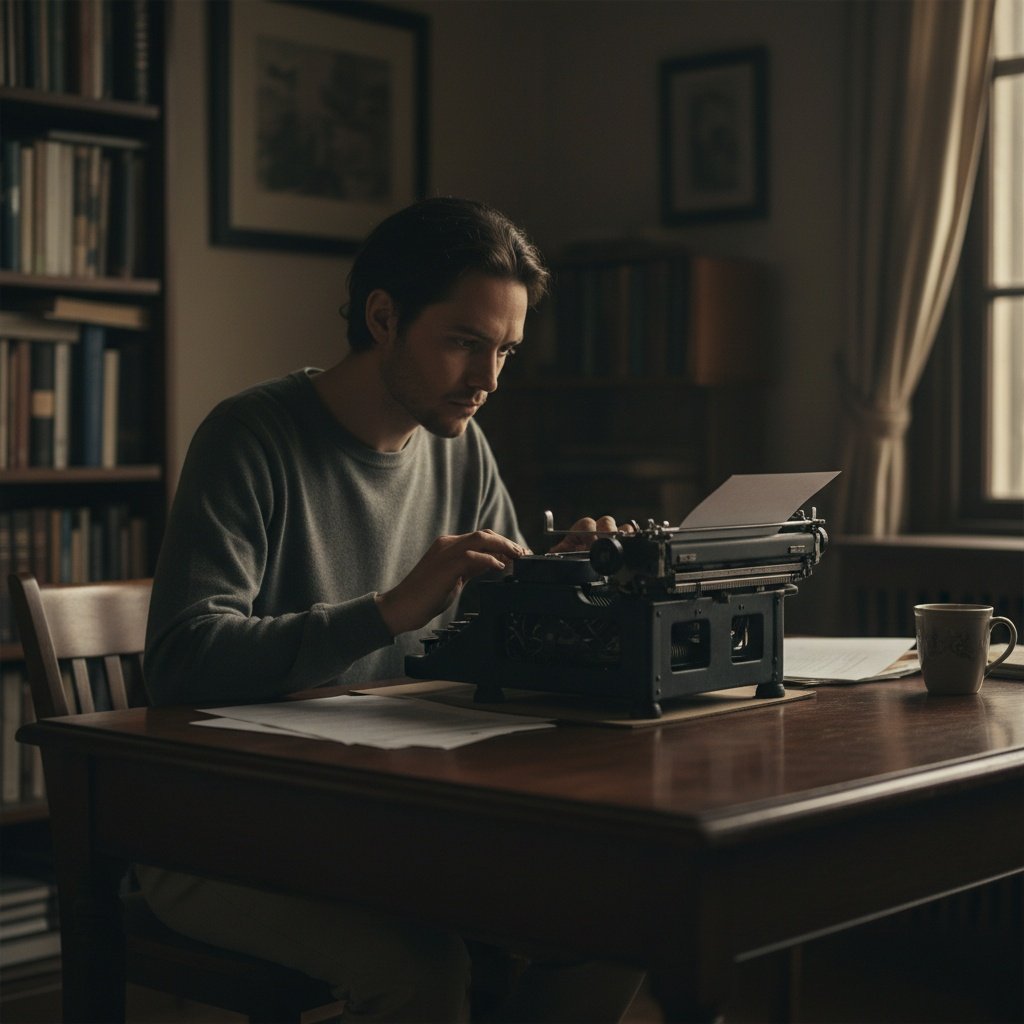} & & & & \includegraphics[width=\gw]{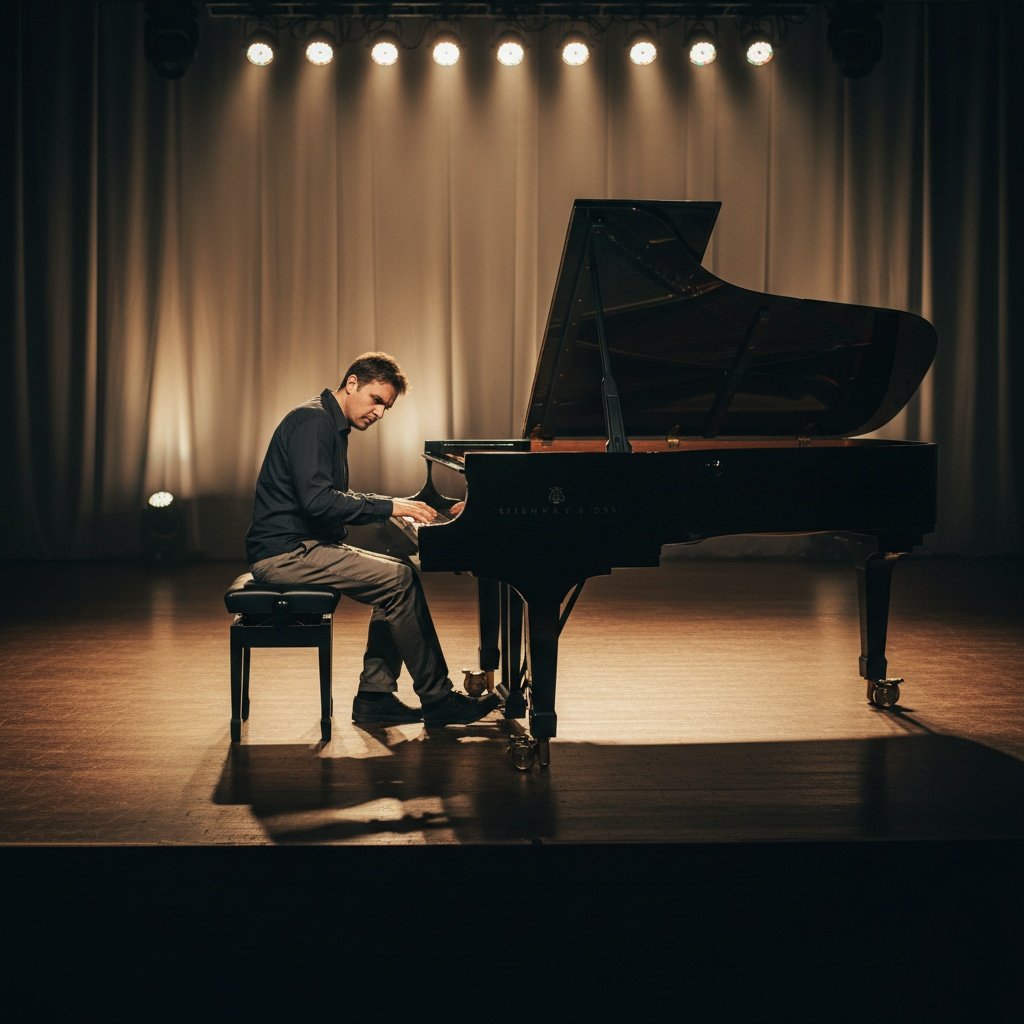} \\

        \raisebox{15pt}{\rotatebox[origin=t]{90}{\scriptsize{DiffMorpher}}} &
        \includegraphics[width=\gw]{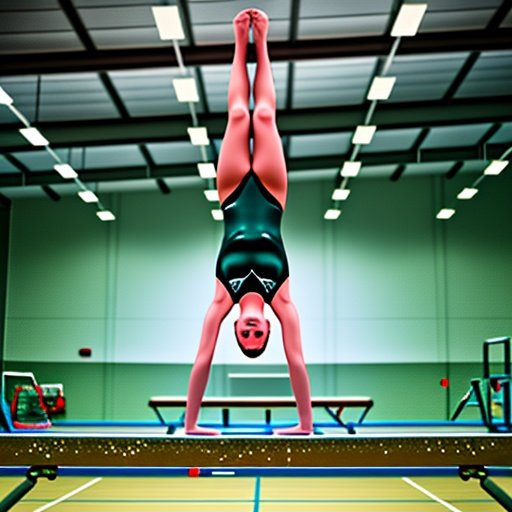} &
        \includegraphics[width=\gw]{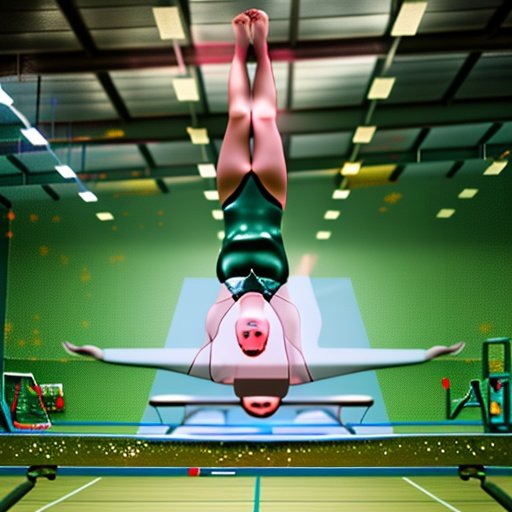} &
        \includegraphics[width=\gw]{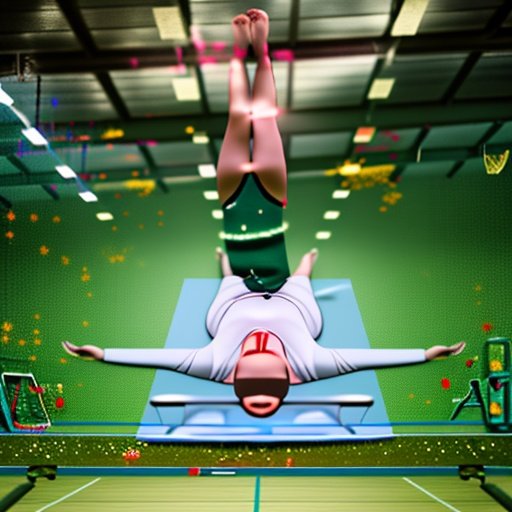} &
        \includegraphics[width=\gw]{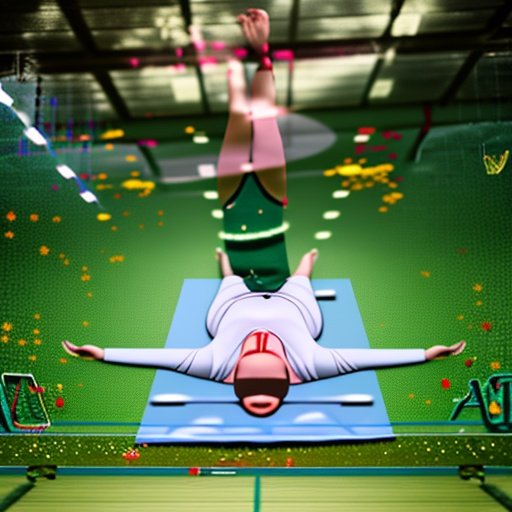} &
        \includegraphics[width=\gw]{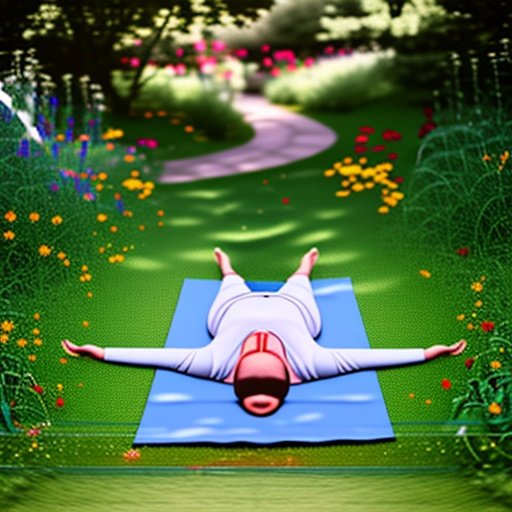} &
        \includegraphics[width=\gw]{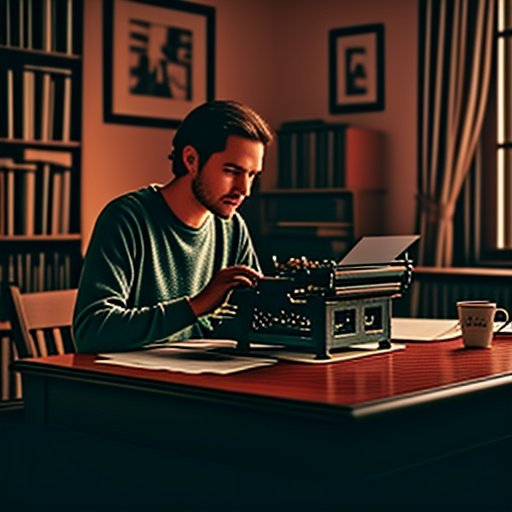} &
        \includegraphics[width=\gw]{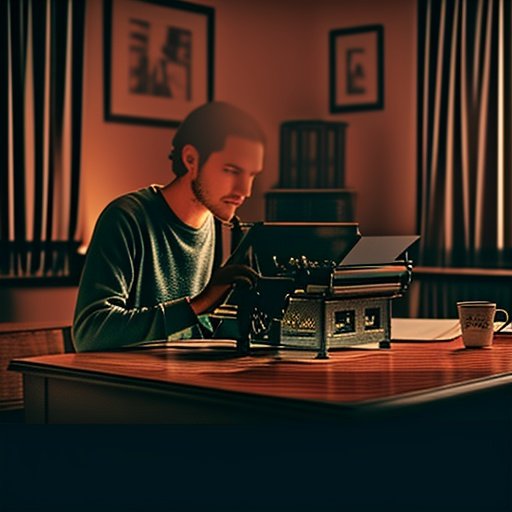} &
        \includegraphics[width=\gw]{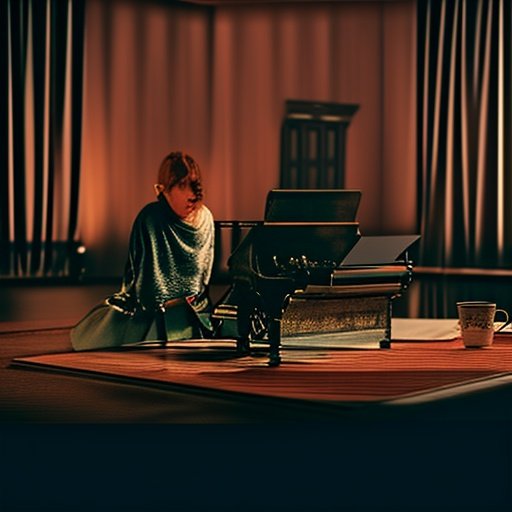} &
        \includegraphics[width=\gw]{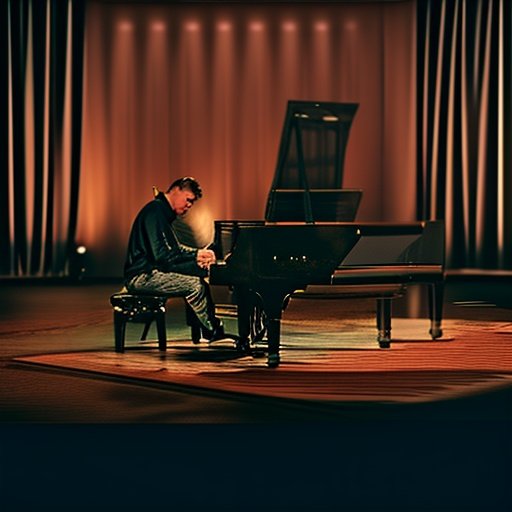} &
        \includegraphics[width=\gw]{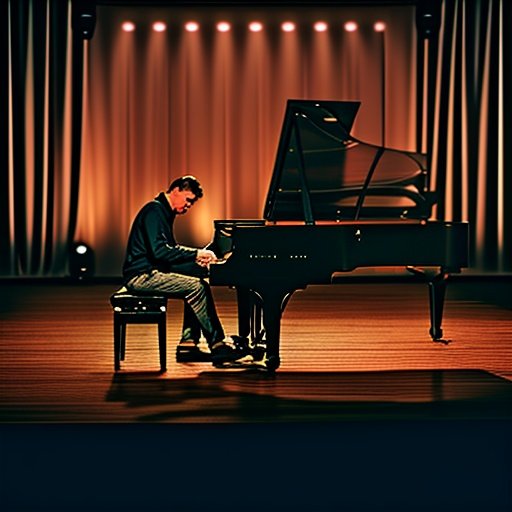} \\

        \raisebox{15pt}{\rotatebox[origin=t]{90}{\scriptsize{FreeMorph}}} &
        \includegraphics[width=\gw]{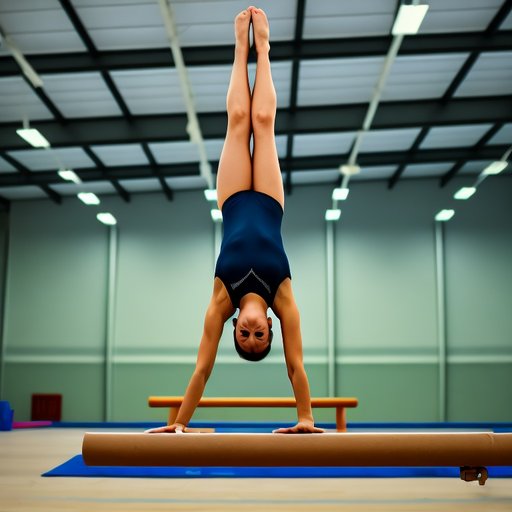} &
        \includegraphics[width=\gw]{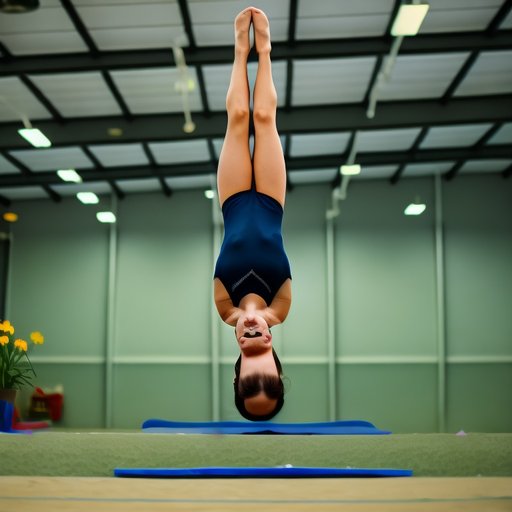} &
        \includegraphics[width=\gw]{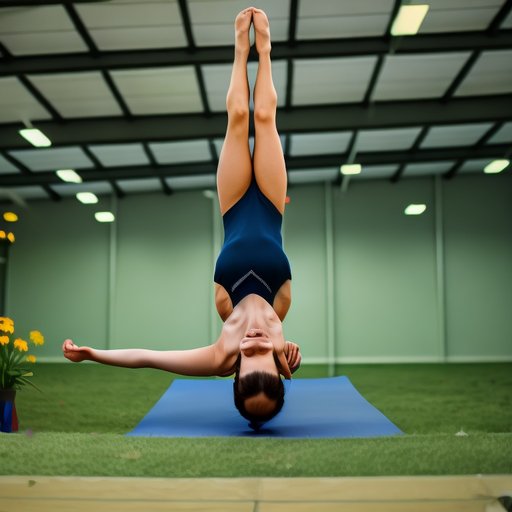} &
        \includegraphics[width=\gw]{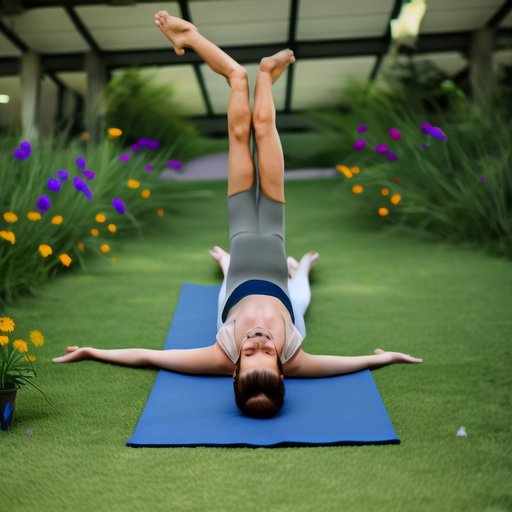} &
        \includegraphics[width=\gw]{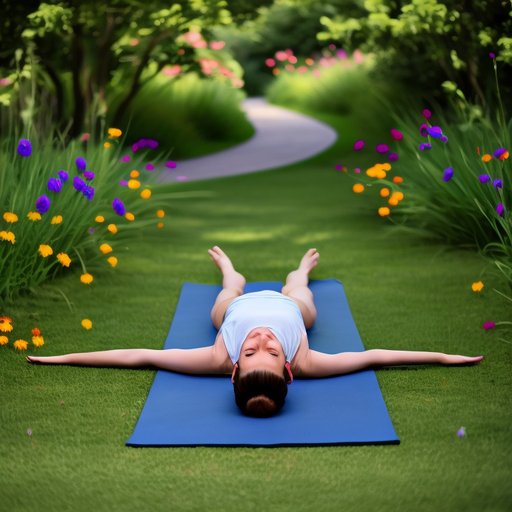} &
        \includegraphics[width=\gw]{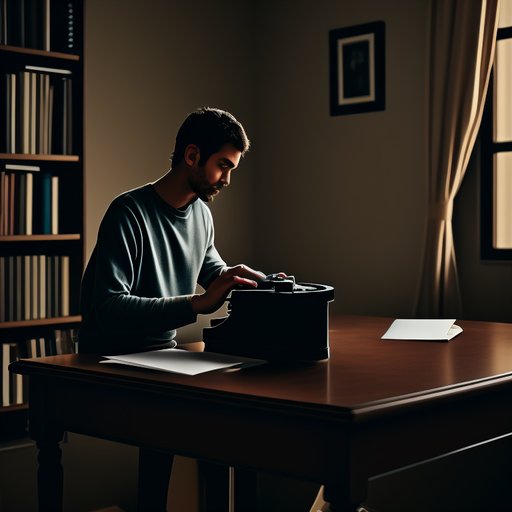} &
        \includegraphics[width=\gw]{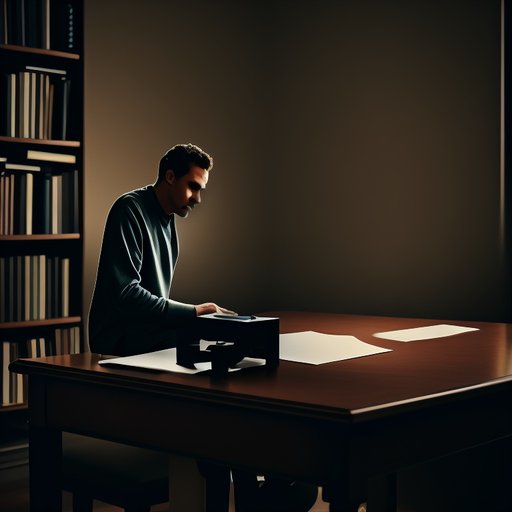} &
        \includegraphics[width=\gw]{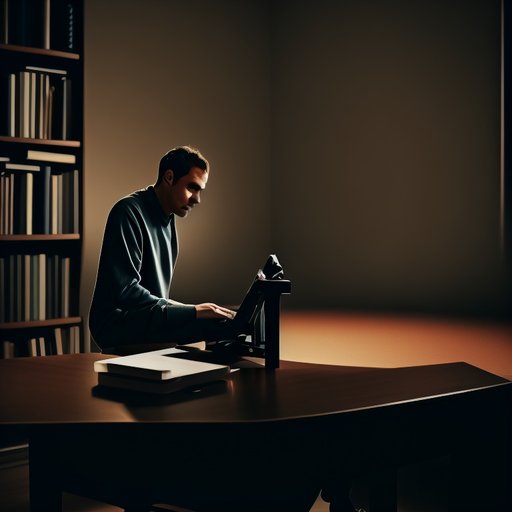} &
        \includegraphics[width=\gw]{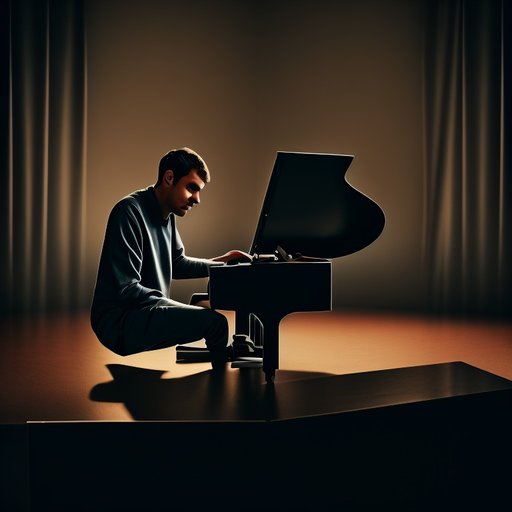} &
        \includegraphics[width=\gw]{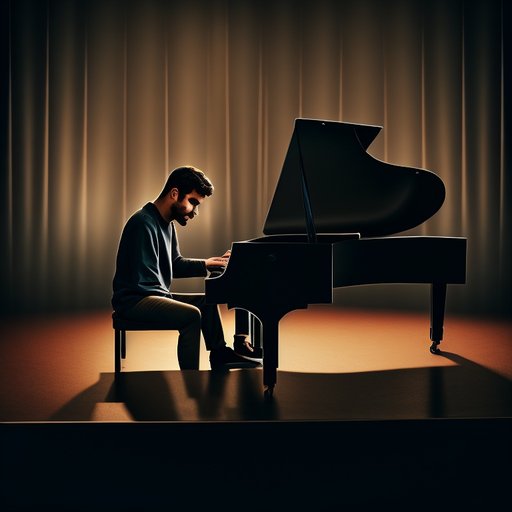} \\

        \raisebox{15pt}{\rotatebox[origin=t]{90}{\scriptsize{Vibe Space}}} &
        \includegraphics[width=\gw]{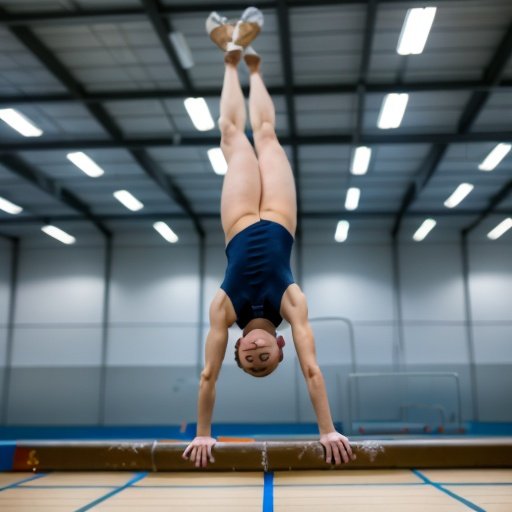} &
        \includegraphics[width=\gw]{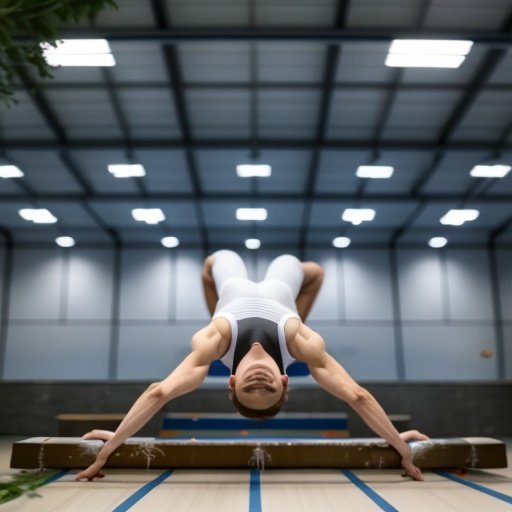} &
        \includegraphics[width=\gw]{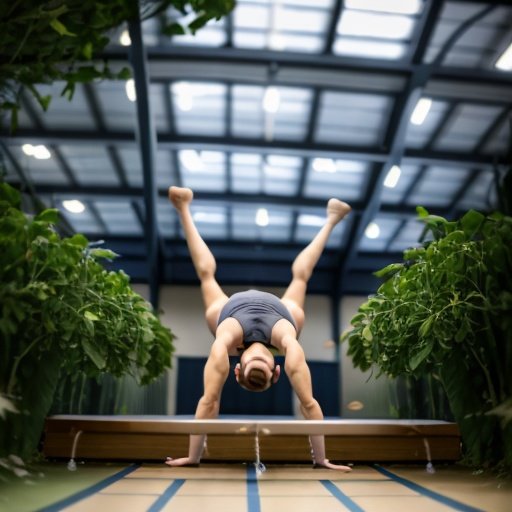} &
        \includegraphics[width=\gw]{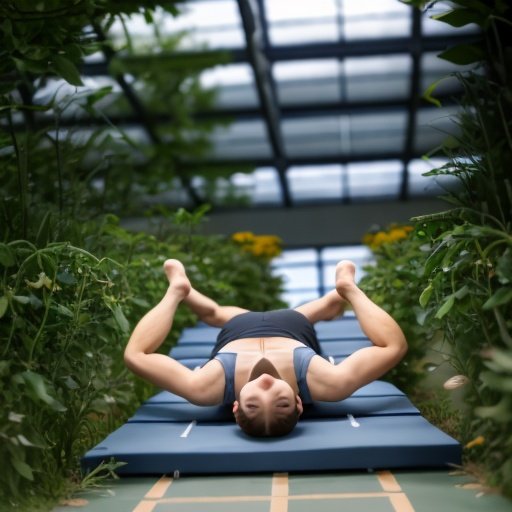} &
        \includegraphics[width=\gw]{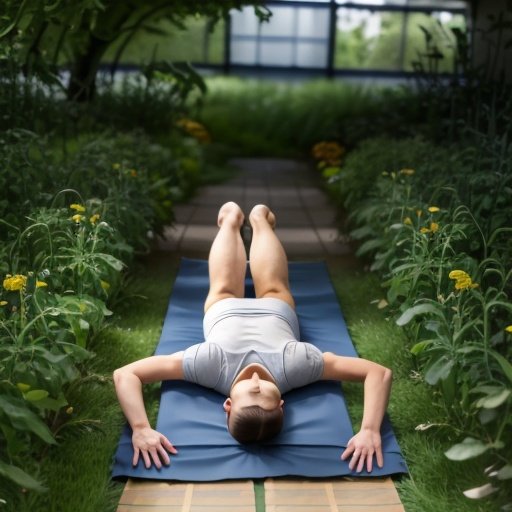} &
        \includegraphics[width=\gw]{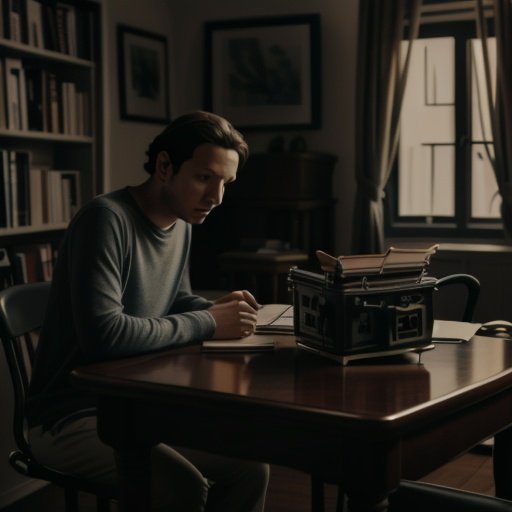} &
        \includegraphics[width=\gw]{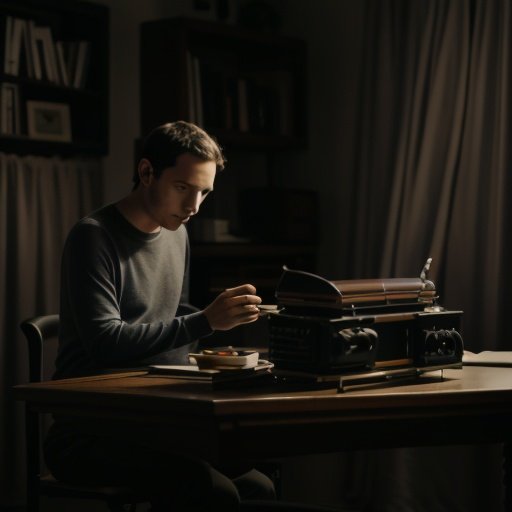} &
        \includegraphics[width=\gw]{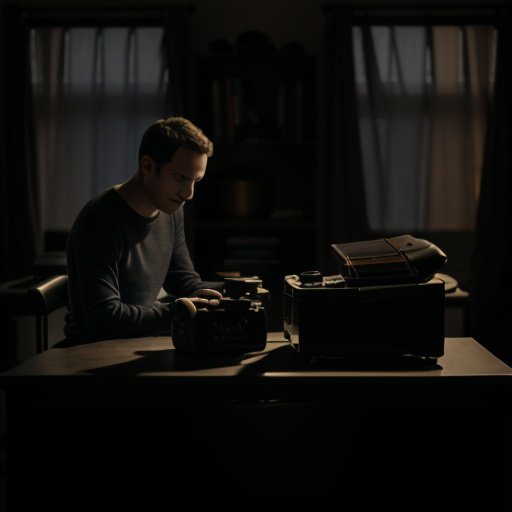} &
        \includegraphics[width=\gw]{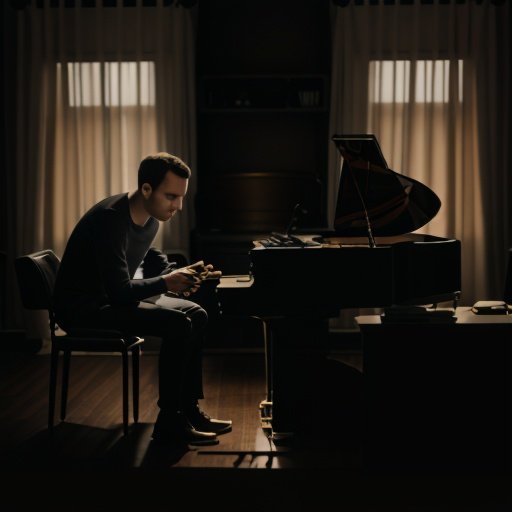} &
        \includegraphics[width=\gw]{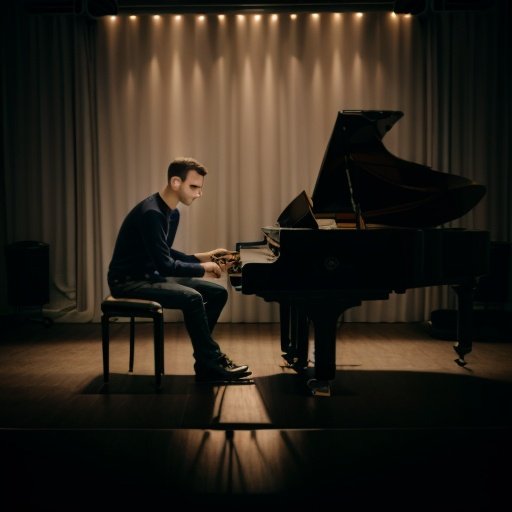} \\

        \raisebox{15pt}{\rotatebox[origin=t]{90}{\scriptsize{T2T (Ours)}}} &
        \includegraphics[width=\gw]{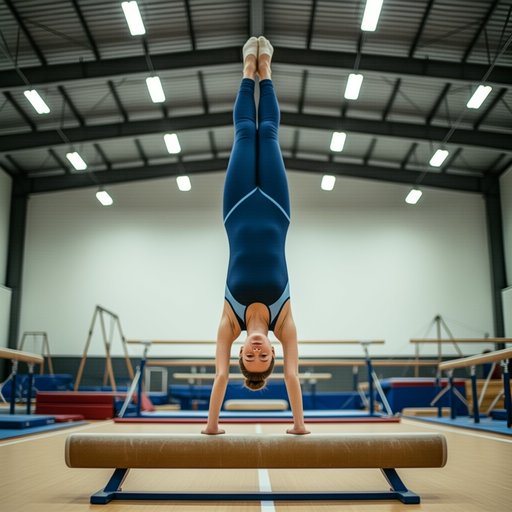} &
        \includegraphics[width=\gw]{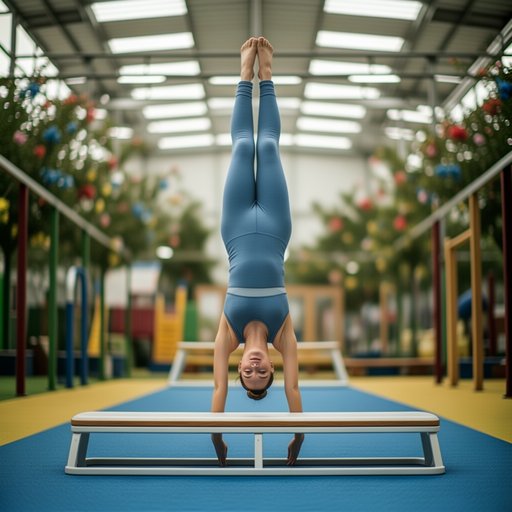} &
        \includegraphics[width=\gw]{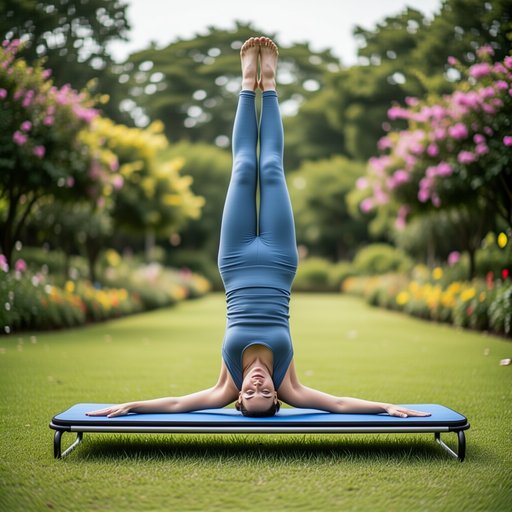} &
        \includegraphics[width=\gw]{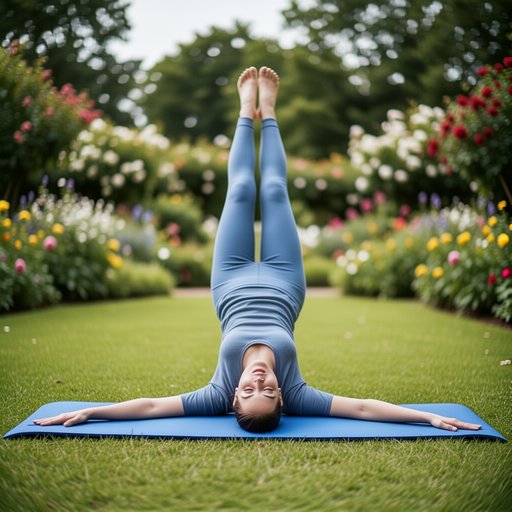} &
        \includegraphics[width=\gw]{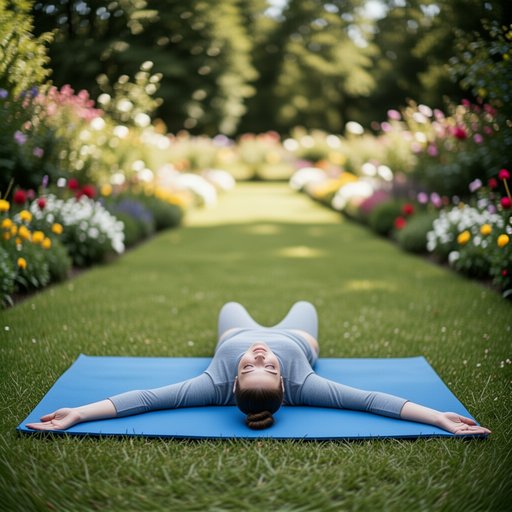} &
        \includegraphics[width=\gw]{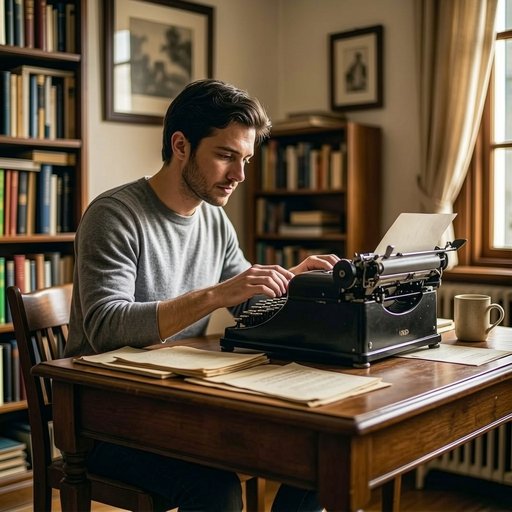} &
        \includegraphics[width=\gw]{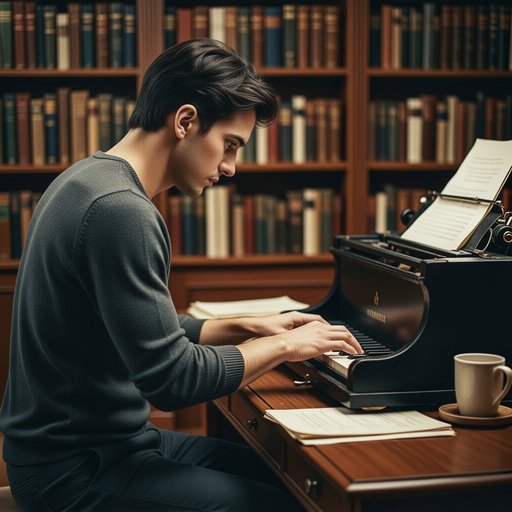} &
        \includegraphics[width=\gw]{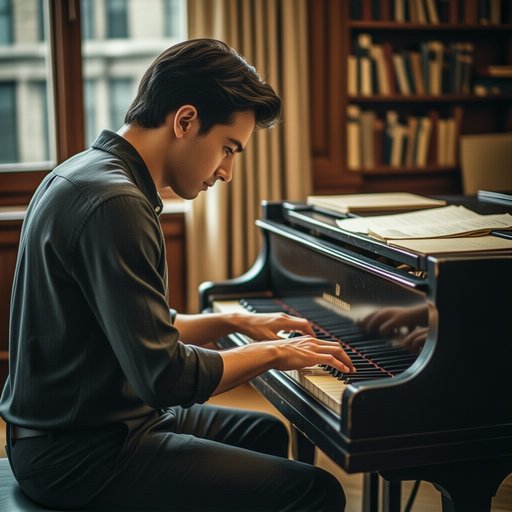} &
        \includegraphics[width=\gw]{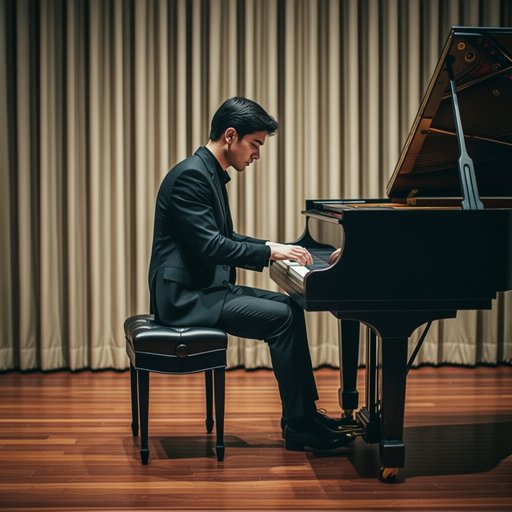} &
        \includegraphics[width=\gw]{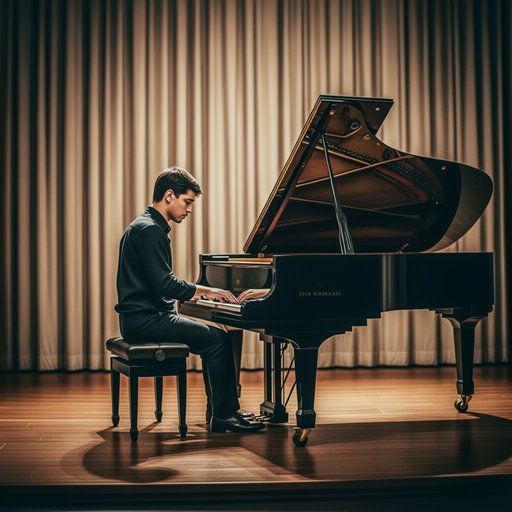} \\

    \noalign{\vspace{1pt}} 


    \multicolumn{6}{c}{\scriptsize{ $\xleftarrow{\hspace{35pt}}$ {Intermediate transitions ($0 < \alpha < 1$)} $\xrightarrow{\hspace{35pt}}$}} & \multicolumn{5}{c}{\scriptsize{ $\xleftarrow{\hspace{35pt}}$ {Intermediate transitions ($0 < \alpha < 1$)} $\xrightarrow{\hspace{35pt}}$}} \\

    \end{tabular}
    \vspace{-4pt}
    \caption{\textbf{Qualitative comparison with continuous blending methods.} While prior approaches may produce mixed or drifting attributes along the transition, our method interpolates in aligned text embedding space, yielding coherent and semantically valid intermediate images at every step. 
    }

    \label{fig:blend_comparison}
    \vspace{-14pt}
\end{figure*}

\renewcommand{\arraystretch}{1}

\paragraph{\textbf{Implementation Details.}}
We evaluate our method using FLUX-2 (Klein 9B)~\citep{flux-2-2025} and FIBO-edit~\citep{gutflaish2025generating}, with Qwen3 (8B)~\citep{yang2025qwen3} and SmolLM3-3B~\citep{bakouch2025smollm3} as their corresponding text encoders. Structured scene descriptions are generated using Gemini 2.5~\citep{comanici2025gemini} under a shared JSON schema. For embedding alignment, we use intermediate text representations from multiple encoder layers and compute layer-wise similarity matrices with temperature scaling (\(\tau = 0.05\)). For continuous blending, image contributions are modulated using \(\alpha\)-dependent quadratic weighting. Unless otherwise specified, inference settings are kept consistent across methods for fair comparison.

\vspace{-4pt}
\paragraph{\textbf{Baselines.}}
We compare our method against prior work across two settings. For continuous image blending, we consider DiffMorpher~\citep{zhang2024diffmorpher}, a diffusion-based method that morphs between images by interpolating LoRA representations; FreeMorph~\citep{cao2025freemorph}, a tuning-free approach that blends attention features; and VibeSpace~\citep{yang2025vibe}, which learns a joint embedding space and performs latent interpolation. For continuous image editing, we consider Kontinuous Kontext~\citep{parihar2025kontinuous}, a supervised method that learns a scalar control for edit strength; SliderEdit~\citep{zarei2025slideredit}, which enables continuous control via token-selective LoRA adapters; and GRAG~\citep{zhang2025group}, a training-free method that modulates attention representations. All baselines are evaluated using their official implementations and recommended settings.

\vspace{-4pt}
\subsection{Qualitative Results}
\label{sec:qualitative}

We present qualitative evaluations for continuous editing in Figure~\ref{fig:edit_comparison} and continuous blending in Figure~\ref{fig:blend_comparison}. As shown in Figure~\ref{fig:edit_comparison}, prior methods often rely on appearance-based transitions or introduce abrupt semantic changes during the edit trajectory. For example, in the left example, existing methods either abruptly generate the flower or mainly modify its scale without modeling a meaningful semantic growth process. In contrast, our method produces smooth semantic transitions with coherent intermediate states that progressively reflect the intended edit. As shown in Figure~\ref{fig:blend_comparison}, methods that operate in pixel or latent spaces (e.g., DiffMorpher, FreeMorph) often generate distorted intermediate images, particularly under large geometric or viewpoint changes. DiffMorpher exhibits ghosting artifacts with overlapping structures, while FreeMorph may produce implausible body configurations with structural discontinuities. Although VibeSpace operates in a semantic space, it still exhibits attribute drift and less consistent transitions across intermediate results. In contrast, our method produces coherent intermediate images by interpolating between semantically aligned text embeddings, preserving structure and ensuring that each step corresponds to a valid scene. Additional editing examples and blending transitions are provided in the supplementary material.

\vspace{-4pt}
\subsection{Quantitative Results}
\label{sec:quantitative}
\vspace{-4pt}

\begin{table*}[t]
\centering

\scriptsize
\setlength{\tabcolsep}{2.5pt}

\begin{minipage}[t]{0.5\textwidth}
\vspace{0pt}
\centering

\caption{Quantitative evaluation with existing continuous blending methods.}
\label{tab:blend}
\vspace{-6pt}
\begin{tabular}{l ccc ccc}
\toprule
& \multicolumn{3}{c}{\textbf{Morph4Data}} 
& \multicolumn{3}{c}{\textbf{BlendBench}} \\
\cmidrule(lr){2-4} \cmidrule(lr){5-7}

\textbf{Method} 
& {$\text{PPL} \downarrow$} 
& {$\text{MUSIQ} \uparrow$} 
& {$\text{FID} \downarrow$}
& {$\text{PPL} \downarrow$} 
& {$\text{MUSIQ} \uparrow$} 
& {$\text{FID} \downarrow$} \\

\midrule

DiffMorpher                  
& \textbf{12.40} 
& 62.27 
& 113.68 
& \textbf{21.54} 
& 65.95 
& 146.80 \\

FreeMorph                    
& \underline{27.34} 
& 67.51 
& 105.85 
& \underline{63.06} 
& 71.17 
& 129.53 \\

Vibe Space                   
& 61.96 
& 71.04 
& \underline{101.76} 
& 252.47 
& 71.53 
& 139.60 \\

\midrule

$\text{T2T}_\text{Flux2-klein}$ 
& 45.38 
& \textbf{74.61} 
& 105.97 
& 147.63 
& \textbf{73.27} 
& \textbf{85.06} \\

$\text{T2T}_\text{FIBO-edit}$    
& 53.49 
& \underline{73.06} 
& \textbf{85.71} 
& 164.71 
& \underline{73.08} 
& \underline{103.64} \\

\bottomrule
\end{tabular}
\end{minipage}
\hfill
\begin{minipage}[t]{0.46\textwidth}
\vspace{0pt}
\centering

\caption{Quantitative evaluation with existing continuous editing methods.}
\label{tab:edit}
\vspace{-6pt}
\begin{tabular}{lccc}
\toprule
& \multicolumn{3}{c}{\textbf{PIE-Bench}} \\
\cmidrule(lr){2-4}
\textbf{Method} & 
$\boldsymbol{\delta_{\text{smooth}}}\downarrow$ & 
\textbf{Continuity}$\uparrow$ & 
\textbf{Norm CLIP-Dir}$\uparrow$ \\

\midrule

Kontinuous K.     
& \textbf{0.246} 
& \underline{1.33} 
& 0.259 \\

Slider Edit       
& \underline{0.395} 
& 1.08 
& 0.353 \\

GRAG              
& 0.515 
& 0.97 
& 0.343 \\

\midrule

$\text{T2T}_\text{Flux2-klein}$  
& 0.562 
& 1.20 
& \underline{0.355} \\

$\text{T2T}_\text{FIBO-edit}$    
& 0.530 
& \textbf{1.46} 
& \textbf{0.393} \\

\bottomrule
\end{tabular}

\end{minipage}
\vspace{-4pt}
\end{table*}
\begin{figure}[t]
\vspace{-4pt}
\begin{center}
	\includegraphics[width=1\linewidth]{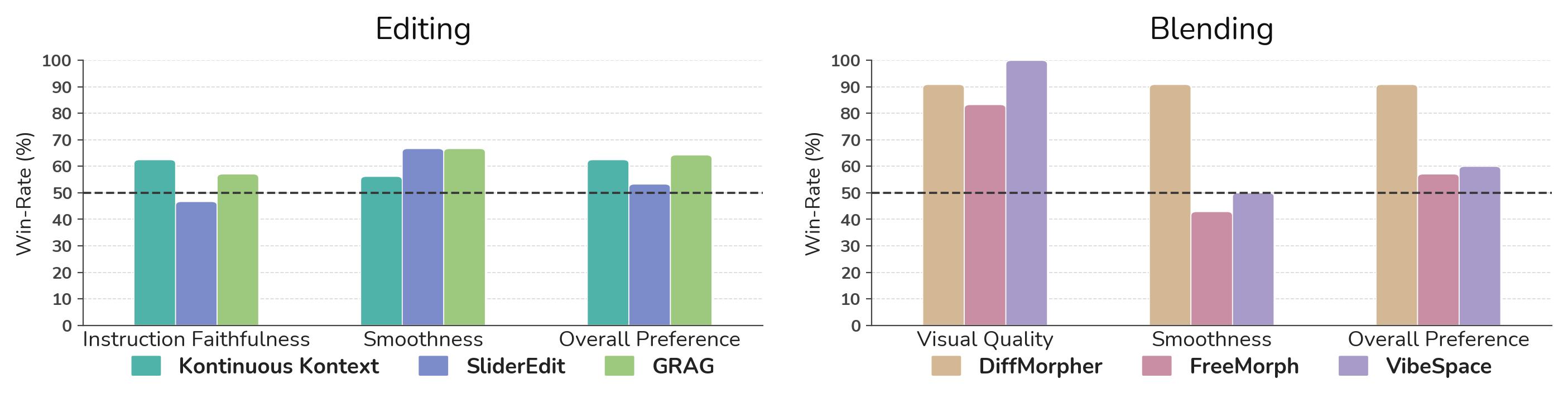}
\end{center}
\vspace{-7pt}
\caption{\textbf{User study results.} Win rates (\%) of our method in pairwise comparisons against editing (left) and blending (right) baselines across evaluation criteria.}

\vspace{-14pt}
\label{fig:user_study}
\end{figure}

\paragraph{Continuous Blending.}
We evaluate continuous blending on \textbf{Morph4Data}~\cite{cao2025freemorph} and our proposed \textbf{BlendBench}. Morph4Data contains 76 image pairs with partial spatial alignment, enabling blending through local correspondence. To evaluate semantic blending in more challenging settings, we introduce BlendBench, a dataset of 100 complex pairs with variations in objects, poses, and viewpoints, requiring simultaneous semantic changes. These scenarios lack reliable spatial correspondence and highlight limitations of morphing-based methods, such as ghosting and structural artifacts, making blending more challenging. We further quantify this distinction in Appendix~\ref{sec:benchmarks}. We follow~\cite{cao2025freemorph} using 5 intermediate steps for Morph4Data, and use 9 steps for BlendBench to capture more complex transitions. We evaluate blending in terms of \textbf{trajectory smoothness}, measured using PPL~\cite{karras2020analyzing} 
and \textbf{visual quality}, measured using MUSIQ~\cite{ke2021musiq} and FID computed against the input image distribution~\cite{heusel2017gans} (see supplementary material for details).

As shown in Table~\ref{tab:blend}, morphing-based methods (DiffMorpher, FreeMorph) achieve the lowest PPL, indicating smooth perceptual trajectories. However, PPL captures local perceptual smoothness and does not distinguish a semantically meaningful transition from a visually smooth blend; such trajectories may therefore achieve low PPL while exhibiting ghosting or implausible intermediate states. The lower MUSIQ and higher FID further reflect this degradation in visual quality. VibeSpace improves semantic consistency but exhibits less stable trajectories and weaker adherence to the input images. In contrast, our method produces more grounded and structurally consistent transitions while maintaining competitive smoothness, with higher image quality and better alignment with the input distribution.

\vspace{-8pt}
\paragraph{Continuous Editing.}
\begin{figure*}[t]
\centering

\begin{minipage}[t]{0.49\textwidth}
\centering
\setlength{\tabcolsep}{0pt}
\newcommand{\imgwidthsyn}{0.18\linewidth}
{\small
\begin{tabular}{@{}ccccc@{}}
    \multicolumn{5}{c}{\textit{``Apple''$\rightarrow$``Pumice stone''}} \\

    \includegraphics[width=\imgwidthsyn]{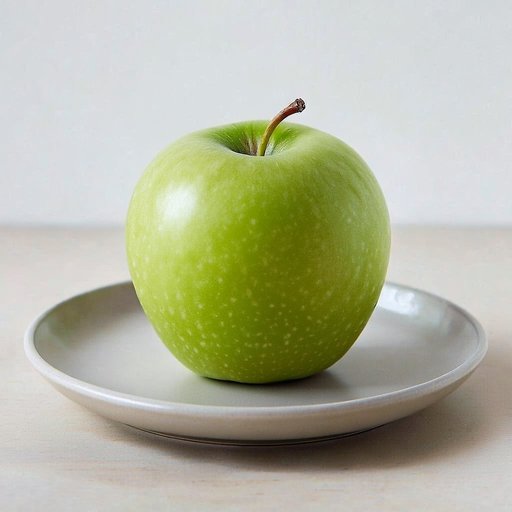} &
    \includegraphics[width=\imgwidthsyn]{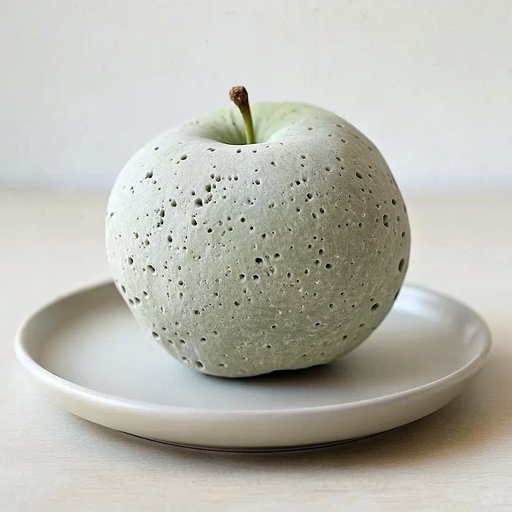} &
    \includegraphics[width=\imgwidthsyn]{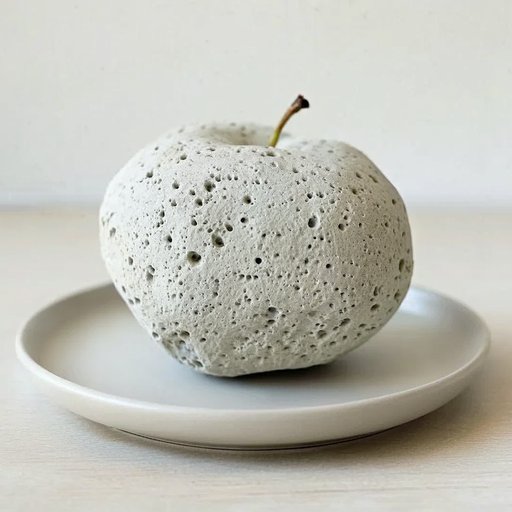} &
    \includegraphics[width=\imgwidthsyn]{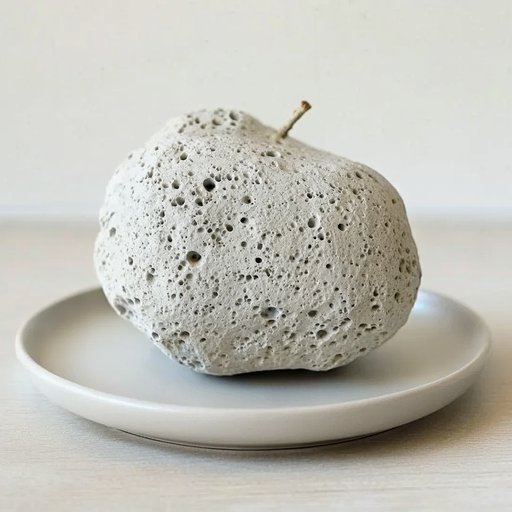} &
    \includegraphics[width=\imgwidthsyn]{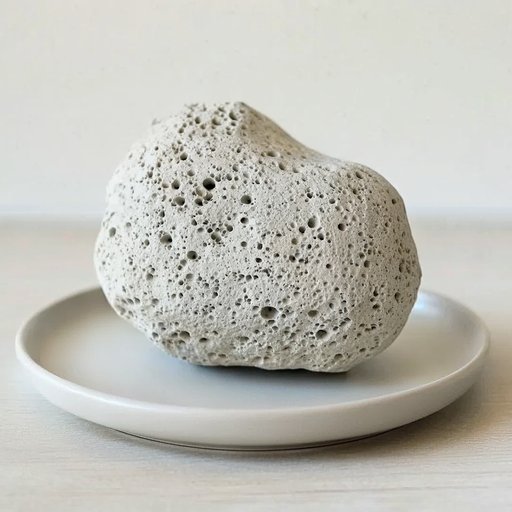} \\

    \multicolumn{5}{c}{\textit{``Teenager''$\rightarrow$``Elderly man''}} \\
    \includegraphics[width=\imgwidthsyn]{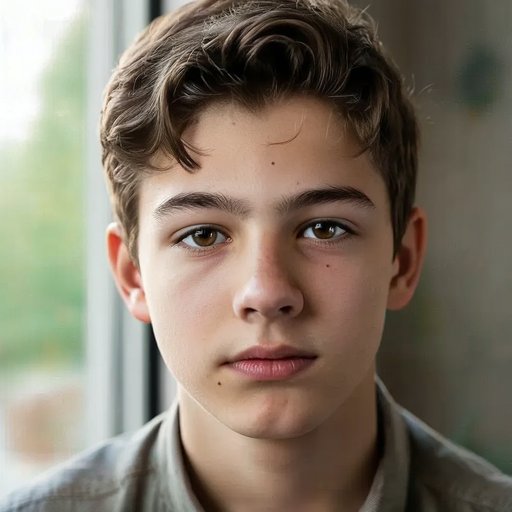} &
    \includegraphics[width=\imgwidthsyn]{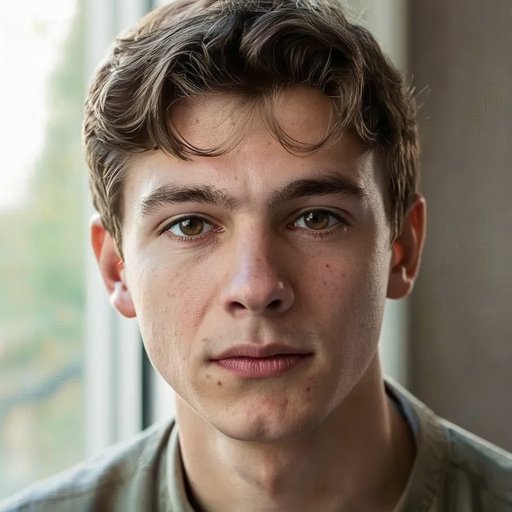} &
    \includegraphics[width=\imgwidthsyn]{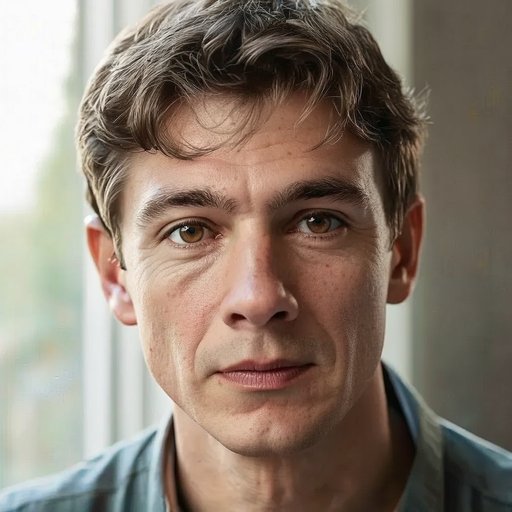} &
    \includegraphics[width=\imgwidthsyn]{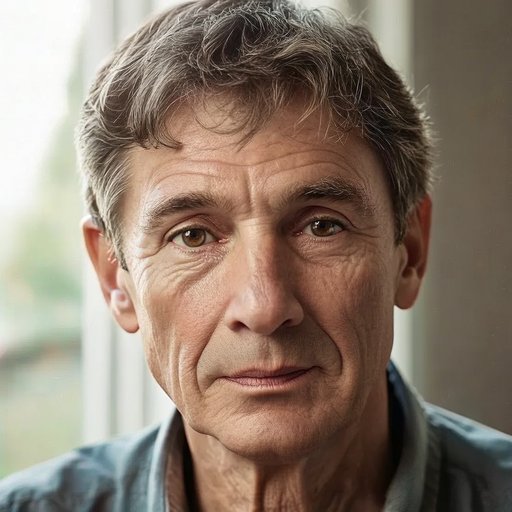} &
    \includegraphics[width=\imgwidthsyn]{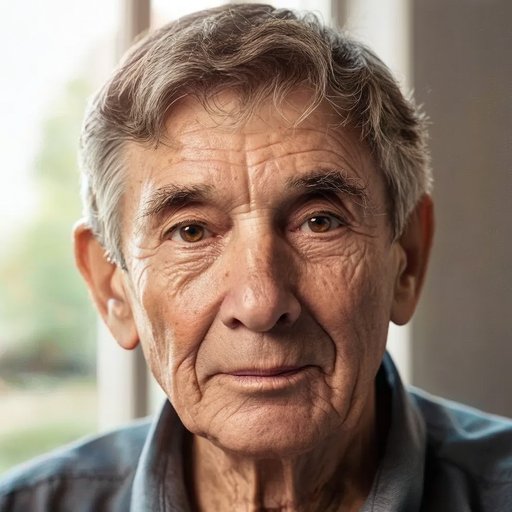} \\

    \multicolumn{5}{c}{\textit{``Alley''$\rightarrow$``Canal with boats''}} \\
    \includegraphics[width=\imgwidthsyn]{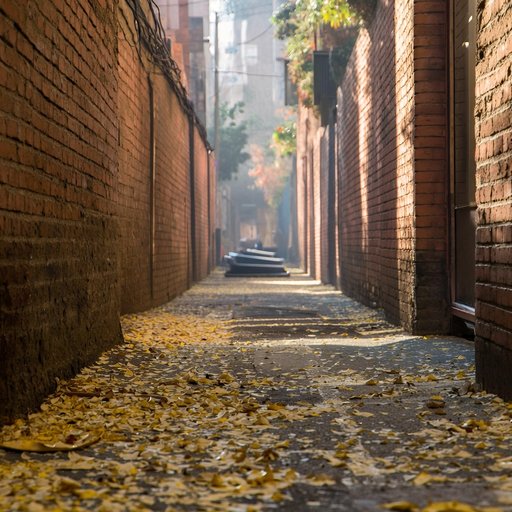} &
    \includegraphics[width=\imgwidthsyn]{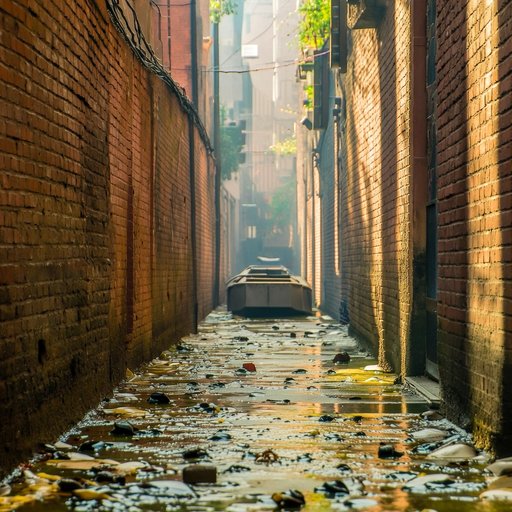} &
    \includegraphics[width=\imgwidthsyn]{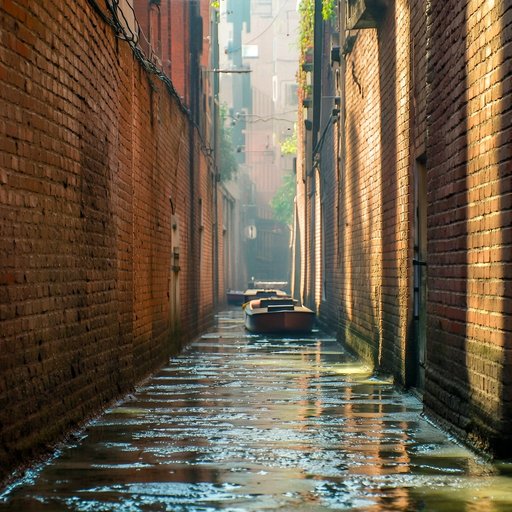} &
    \includegraphics[width=\imgwidthsyn]{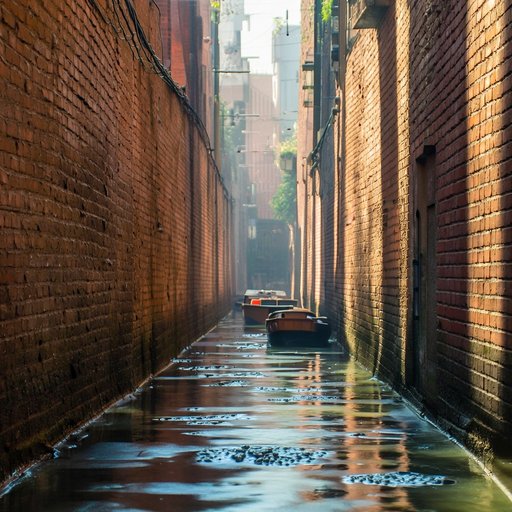} &
    \includegraphics[width=\imgwidthsyn]{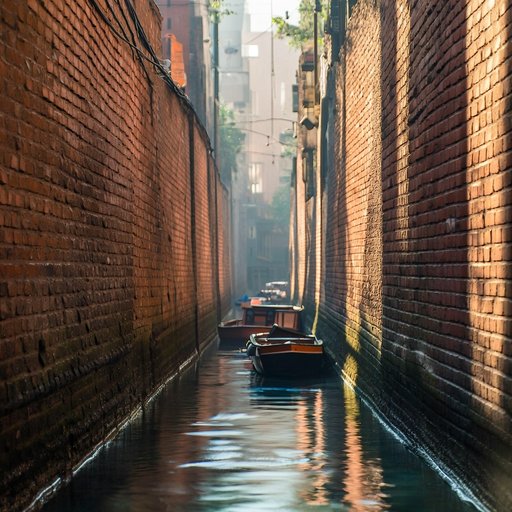} \\

\end{tabular}
}

\vspace{-2pt}

\caption{
\footnotesize{\textbf{Continuous synthesis using Token-to-Token alignment.} Interpolating between aligned text embeddings produces smooth semantic transitions while preserving structural consistency throughout the trajectory.
}
}

\vspace{-6pt}
\label{fig:aditional_results}

\end{minipage}
\hfill
\begin{minipage}[t]{0.49\textwidth}
\centering
\setlength{\tabcolsep}{0pt}
\newcommand{\imgwidthablate}{0.15\linewidth}
{\small
\begin{tabular}{@{}c@{\hspace{2pt}}ccccccc@{}}
    \multicolumn{7}{c}{\textit{``Erase the graffiti from the wall''}} \\

    \raisebox{10pt}{\rotatebox[origin=t]{90}{\scriptsize{Direct}}} &
    \includegraphics[width=\imgwidthablate]{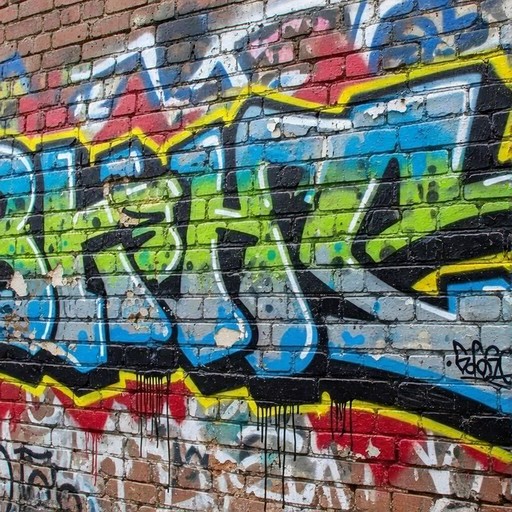} &
    \includegraphics[width=\imgwidthablate]{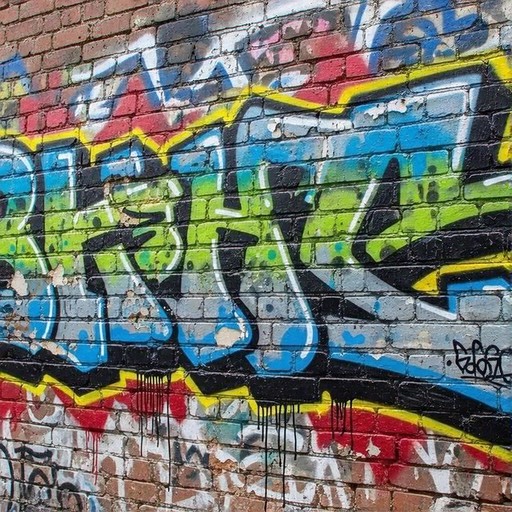} &
    \includegraphics[width=\imgwidthablate]{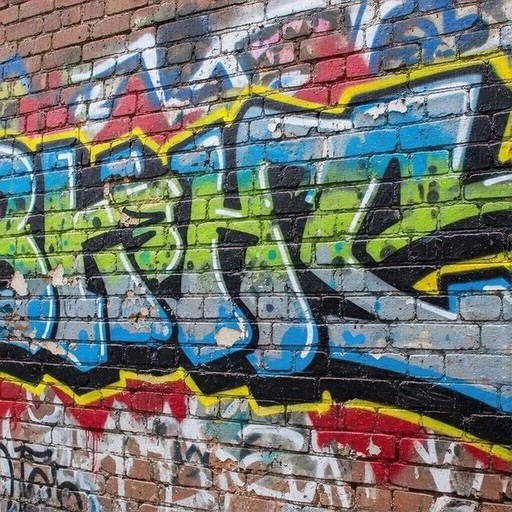} &
        \includegraphics[width=\imgwidthablate]{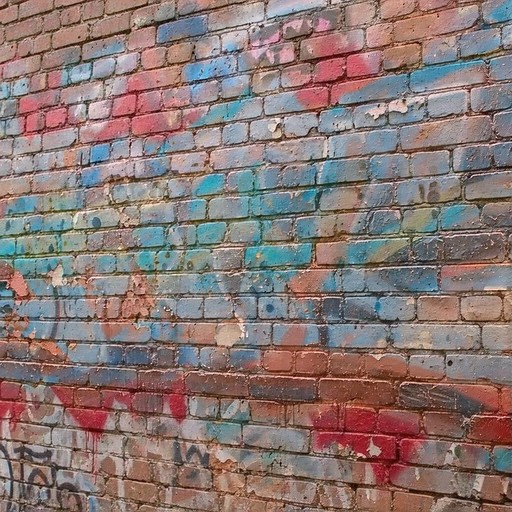} &
    \includegraphics[width=\imgwidthablate]{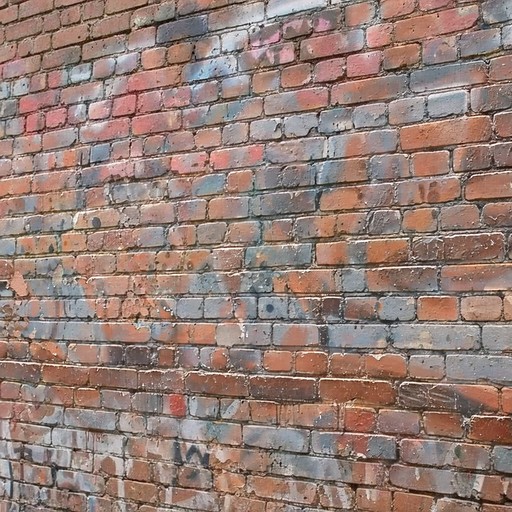} &
    \includegraphics[width=\imgwidthablate]{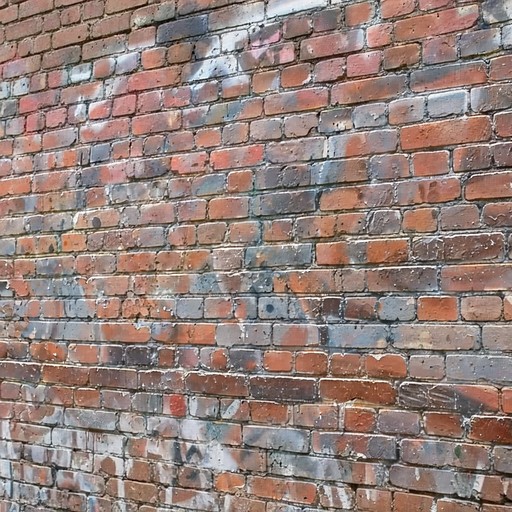} \\

    \raisebox{10pt}{\rotatebox[origin=t]{90}{\scriptsize{Text Al.}}} &
    \includegraphics[width=\imgwidthablate]{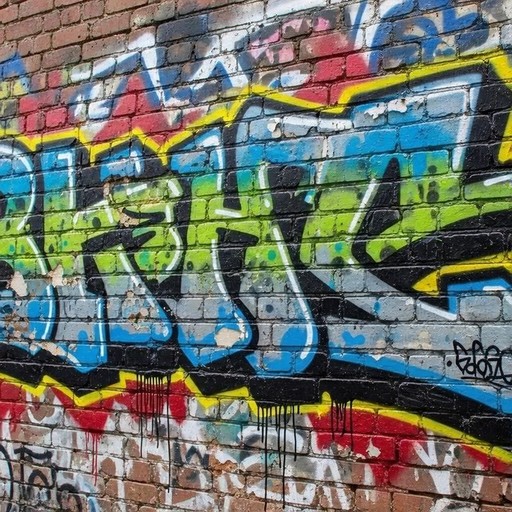} &
    \includegraphics[width=\imgwidthablate]{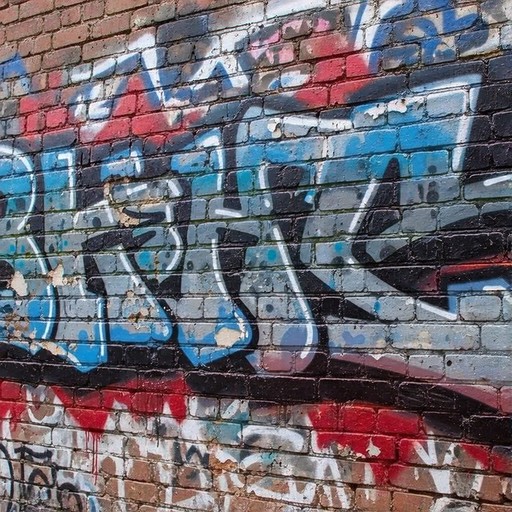} &
    \includegraphics[width=\imgwidthablate]{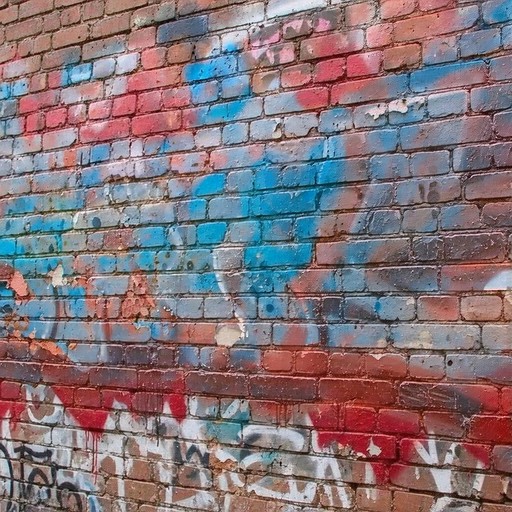} &
    \includegraphics[width=\imgwidthablate]{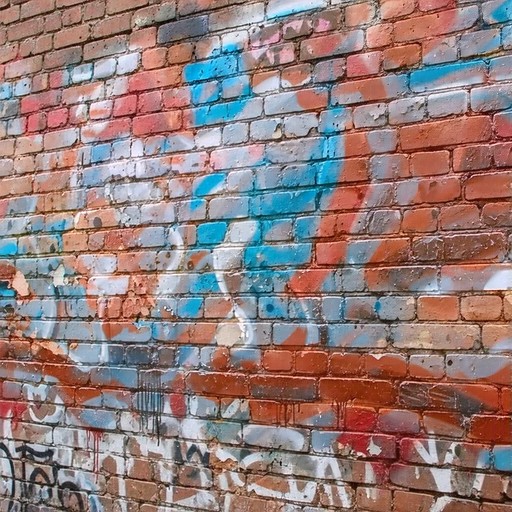} &
    \includegraphics[width=\imgwidthablate]{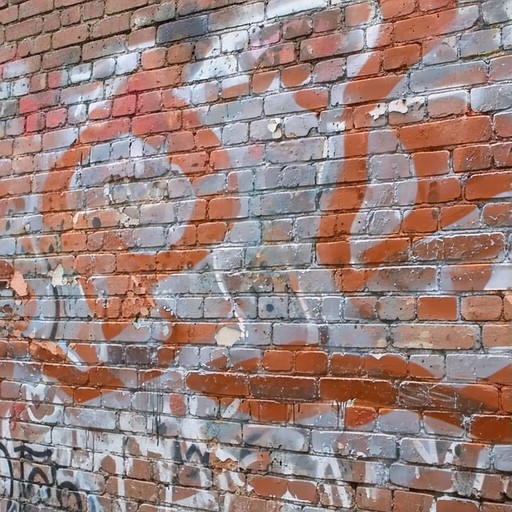} &
    \includegraphics[width=\imgwidthablate]{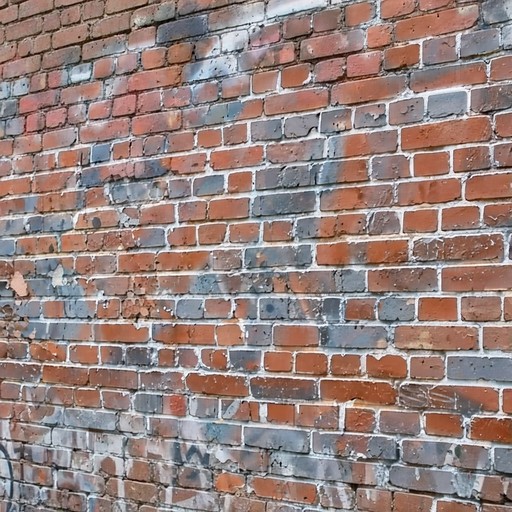} \\

    \raisebox{10pt}{\rotatebox[origin=t]{90}{\scriptsize{Embed Al.}}} &
    \includegraphics[width=\imgwidthablate]{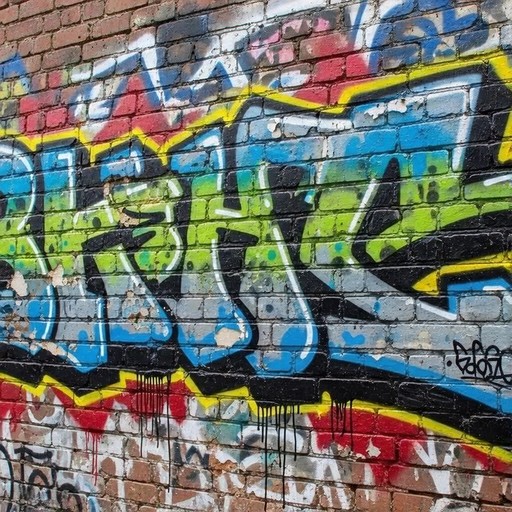} &
    \includegraphics[width=\imgwidthablate]{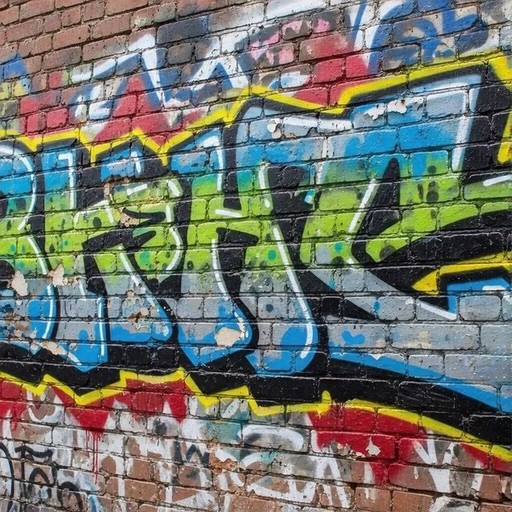} &
    \includegraphics[width=\imgwidthablate]{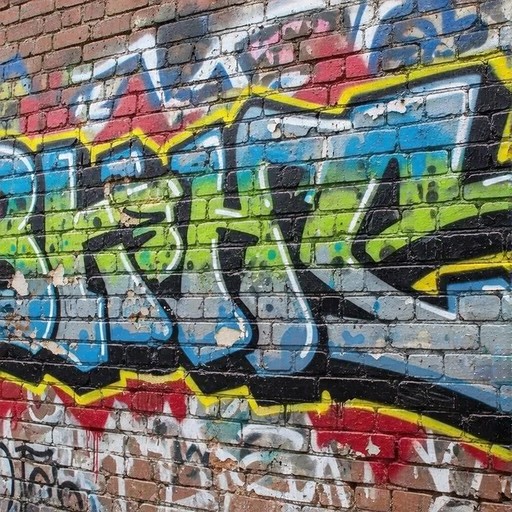} &
    \includegraphics[width=\imgwidthablate]{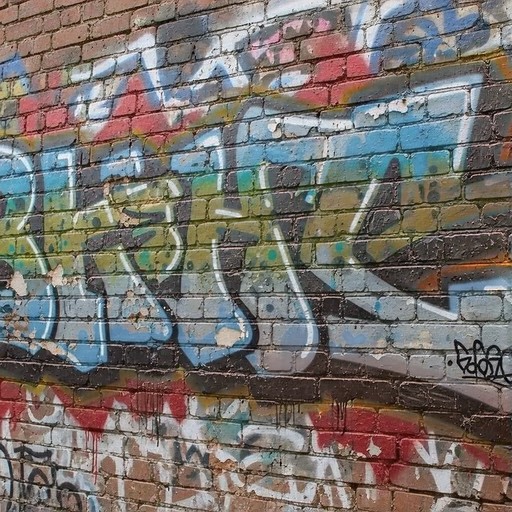} &
    \includegraphics[width=\imgwidthablate]{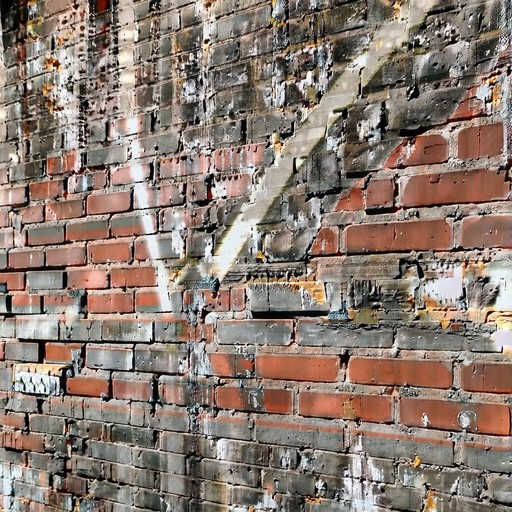} &
    \includegraphics[width=\imgwidthablate]{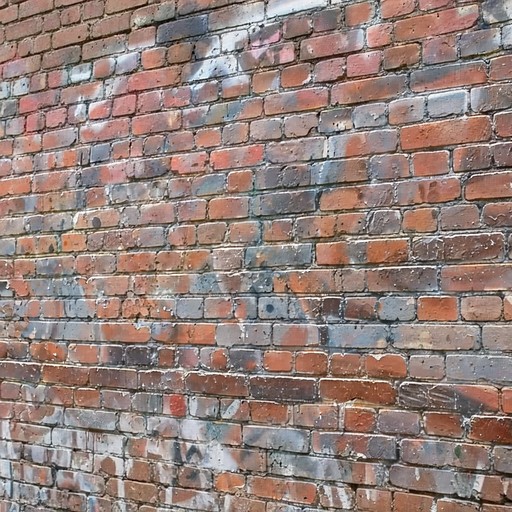} \\

    \raisebox{10pt}{\rotatebox[origin=t]{90}{\scriptsize{Full}}} &
    \includegraphics[width=\imgwidthablate]{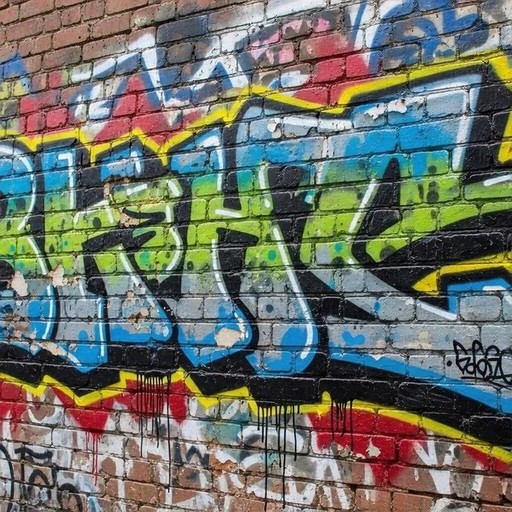} &
    \includegraphics[width=\imgwidthablate]{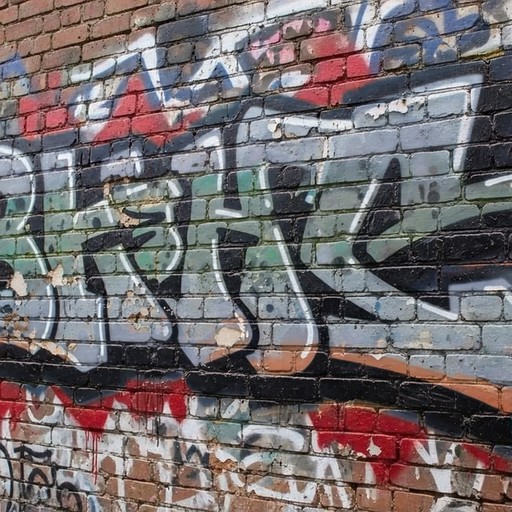} &
    \includegraphics[width=\imgwidthablate]{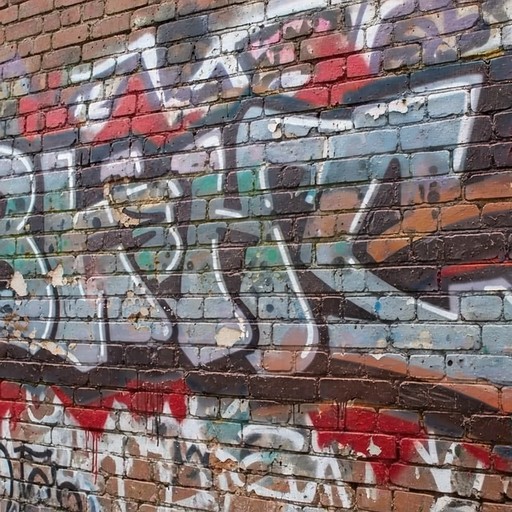} &
    \includegraphics[width=\imgwidthablate]{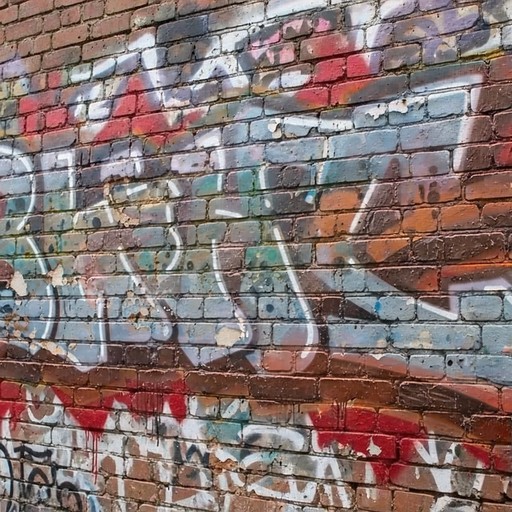} &
    \includegraphics[width=\imgwidthablate]{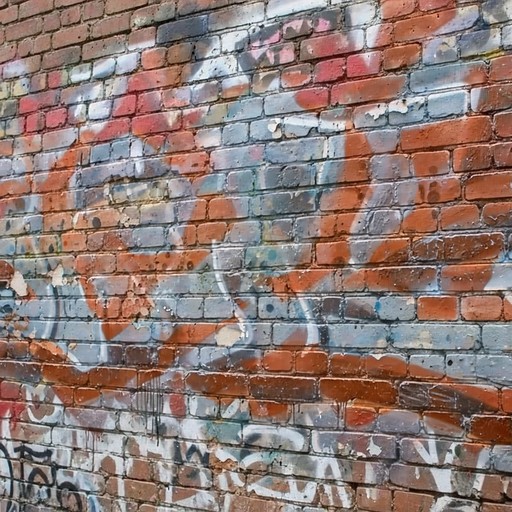} &
    \includegraphics[width=\imgwidthablate]{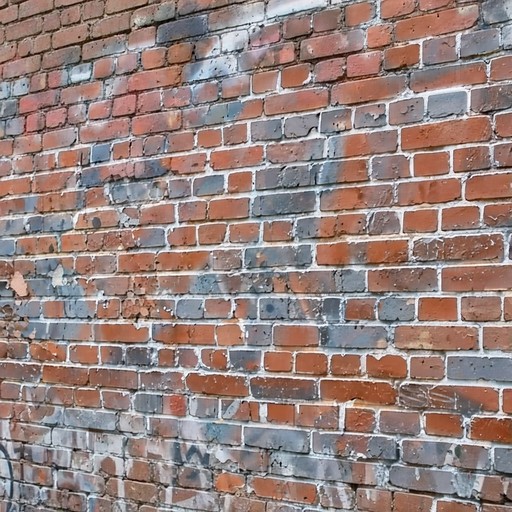} \\

    & \multicolumn{6}{l}{\scriptsize{Edit Intensity} $\xrightarrow{\hspace{120pt}}$} \\

\end{tabular}
}

\vspace{-5pt}
\caption{\footnotesize
\textbf{Qualitative alignment ablation study.}
Direct interpolation produces abrupt transitions. Text-level alignment improves continuity but lacks fine-grained detail, while embedding alignment alone leads to unstable transformations. Combining both yields smooth and coherent transitions.
}

\vspace{-7pt}
\label{fig:ablation_graffiti}

\end{minipage}

\vspace{-8pt}
\end{figure*}
Following the evaluation protocol of Kontinuous Kontext~\cite{parihar2025kontinuous}, 
we evaluate continuous editing on a subset of PIE-Bench~\cite{ju2024pnp} containing 540 image-instruction pairs with 6 
edit strengths.
We measure \textbf{smoothness} using $\delta_{\text{smooth}}$~\cite{parihar2025kontinuous}, \textbf{semantic alignment} using normalized CLIP direction (Norm CLIP-Dir)~\cite{wolf2026continuous}, and \textbf{continuity} in CLIP space~\cite{zarei2025slideredit}, which evaluates whether semantic changes evolve uniformly across edit strengths (see supplementary material for details).

\begin{wraptable}{r}{0.50\textwidth}

\vspace{-10pt}
\centering

\caption{\footnotesize Quantitative alignment ablation study.}
\label{tab:edit_ablation}

\scriptsize
\setlength{\tabcolsep}{3pt}

\begin{tabular}{@{}lccc@{}}
\toprule
& \multicolumn{3}{c}{\textbf{PIE-Bench}} \\
\cmidrule(lr){2-4}
\textbf{Method}
& $\boldsymbol{\delta_{\textbf{smooth}}}\downarrow$
& \textbf{Continuity$\uparrow$}
& \textbf{Norm CLIP-Dir$\uparrow$} \\
\midrule

Direct
& 1.157 & 1.232 & 0.189 \\

Text align
& 0.713 & 1.294 & 0.198 \\

Embedding align
& 0.749 & 1.207 & 0.051 \\

Full w/o locality bias
& \underline{0.678} & \underline{1.417} & 0.263 \\

Full w/o temperature
& 9.498 & 0.751 & \underline{0.302} \\

\midrule

Full method
& \textbf{0.530} & \textbf{1.461} & \textbf{0.393} \\

\bottomrule
\end{tabular}

\vspace{-10pt}

\end{wraptable}

As shown in Table~\ref{tab:edit}, Kontinuous Kontext achieves the strongest smoothness, 
reflecting its design for perceptually consistent transitions, 
but produces trajectories that are less aligned with the intended 
semantic edit, 
as reflected by lower CLIP-Dir.
SliderEdit and GRAG achieve stronger semantic alignment, but exhibit reduced continuity and less stable editing trajectories across edit strengths. In contrast, our method achieves strong semantic alignment together with improved continuity, producing trajectories that follow a coherent semantic direction while maintaining competitive smoothness.

\vspace{-8pt}

\paragraph{User Study.}
We conduct a pairwise user study comparing our method against continuous blending and editing baselines. Participants evaluate transition sequences randomly sampled from the evaluation datasets based on semantic smoothness and overall preference for both tasks, visual quality for blending, and instruction faithfulness for editing (see Fig.~\ref{fig:user_study_ui}). In total, we collect 140 assessments from 20 participants. As shown in Fig.~\ref{fig:user_study}, our method achieves higher win rates across most criteria, particularly in visual quality and overall preference for blending, while also improving semantic smoothness and overall preference for editing.


\vspace{-8pt}

\subsection{Ablation Studies}
\vspace{-6pt}
\label{sec:ablation}
We evaluate the contribution of each alignment stage through qualitative (Fig.~\ref{fig:ablation_graffiti}) and quantitative (Table~\ref{tab:edit_ablation}) ablations. As shown in Fig.~\ref{fig:ablation_graffiti}, direct interpolation produces abrupt transitions, text-level alignment improves continuity but lacks fine-grained detail, and embedding-level alignment alone remains unstable without the consistent semantic structure established by text-level alignment. Combining both stages yields smooth and semantically coherent transitions. Table~\ref{tab:edit_ablation} further supports these observations quantitatively, with the full method achieving the strongest performance across all metrics. Removing the locality bias degrades performance, confirming the benefit of exploiting the positional structure established by text-level alignment, while removing temperature scaling causes the correspondence weights to become overly diffuse, substantially degrading interpolation quality.

\vspace{-6pt}
\section{Conclusion}
\vspace{-6pt}
We introduced Token-to-Token alignment, a framework for enabling coherent semantic interpolation directly in the text embedding space. At a higher level, we suggest a perspective on semantic control in text-to-image generation as a problem of representation rather than manipulation. While the text embedding space implicitly encodes a meaningful and continuous semantic structure, this structure is not directly accessible due to misalignment between prompts.

By enforcing explicit token-level correspondence, we effectively linearize this space, transforming interpolation from an unreliable operation into a principled semantic operation. In this view, the generative model plays a largely passive role, acting as a renderer of trajectories defined in the aligned embedding space. This perspective shifts the emphasis away from model modification and toward organizing existing representations as the primary mechanism for control.

At the same time, our approach relies on the quality of the structural alignment between prompts. While embedding-level alignment can resolve local mismatches, it may fail when the initial descriptions do not establish the intended semantic correspondence. In addition, while linear interpolation becomes meaningful under alignment, it may not fully capture more complex or highly non-linear semantic transformations. These limitations point to several directions for future work, including improving the robustness of alignment, exploring richer and more flexible trajectories in the embedding space, and extending this formulation to temporally coherent settings such as video, where aligned semantic paths could provide a natural mechanism for controlling continuous evolution over time.
\section*{Acknowledgments}
\addcontentsline{toc}{section}{Acknowledgments}
This research was supported in part by the Israel Science Foundation (grants no. 3441/21 and 1473/24), Len Blavatnik and the Blavatnik Family Foundation, and the Bronfman Family Foundation.
\medskip

{
\small
\bibliographystyle{plainnat}
\bibliography{mybibliography}
}


\appendix


\section{Additional details}

\subsection{Benchmarks}
\label{sec:benchmarks}

\paragraph{Morph4data}
Following the protocol of FreeMorph~\cite{cao2025freemorph}, we evaluate semantic interpolation on Morph4Data, a curated dataset of 76 image pairs designed for morphing tasks. The dataset is organized into four categories: (i) similar layout but different semantics, (ii) similar layout and semantics (including face images from CelebA-HQ and object categories such as cars), (iii) dissimilar layout and semantics sampled from ImageNet, and (iv) real-world animal images (e.g., cats and dogs) collected from the internet. Each pair defines a source and target image, and we generate intermediate samples using uniformly spaced interpolation coefficients. Following the original setup, we use 5 interpolation steps between endpoints.

\paragraph{BlendBench}
We introduce BlendBench, a benchmark for evaluating continuous semantic interpolation under correspondence defined by object interactions. The dataset consists of 100 image pairs constructed by first generating pairs of prompts that describe distinct scenes involving multiple objects participating in an action. For each pair, prompts are designed such that objects occupy analogous semantic roles within the interaction, while varying object identity, attributes, background, and spatial configuration (e.g., “a samurai standing and holding a sword in the middle of a forest” vs. “a tennis player running and hitting a ball with a racket”). This construction ensures that semantic correspondence is defined by the roles of objects within the interaction, rather than by shared appearance or spatial alignment. Each prompt is then independently used to generate an image using Gemini 2.5 \cite{comanici2025gemini}, and the resulting images form the source and target pair.

For each pair, methods are evaluated on 9 uniformly spaced intermediate samples capturing the semantic transition.

\paragraph{Quantifying Differences Between Blending Benchmarks.}
To quantify the distinction between the two blending benchmarks, we compare their scene complexity and spatial correspondence in Table~\ref{tab:benchmark_analysis}. Using an off-the-shelf DETR-ResNet-50~\cite{carion2020end} detector with a confidence threshold of 0.7, we measure the number of detected entities per image and the spatial consistency of the primary subject across endpoints.

\begin{table}[h]
\centering
\caption{Quantitative comparison of Morph4Data and BlendBench.}
\label{tab:benchmark_analysis}
\begin{tabular}{lcc}
\toprule
\textbf{Metric} & \textbf{Morph4Data} & \textbf{BlendBench} \\
\midrule
Entities per image (mean) & 1.9 & 7.4 \\
Single-entity images & 62\% & 21\% \\
Primary-subject box IoU (median) & 0.72 & 0.38 \\
Primary-subject centroid shift (median, fraction of frame) & 0.06 & 0.14 \\
\bottomrule
\end{tabular}
\end{table}

As shown in Table~\ref{tab:benchmark_analysis}, BlendBench contains substantially more entities per image and weaker spatial correspondence between endpoints than Morph4Data. These results show that Morph4Data primarily evaluates transitions under partial spatial alignment, whereas BlendBench emphasizes semantic correspondence across more complex, spatially misaligned scenes.

\paragraph{Simplified PIE-Bench.}
We follow the evaluation protocol of Kontinuous Kontext~\cite{parihar2025kontinuous} and construct a simplified subset of PIE-Bench~\cite{ju2024pnp} for evaluating continuous semantic interpolation. For consistency with prior work and to enable fair comparison, we exclude the roughness, transparency, and style categories, resulting in 540 image–instruction pairs.

Because many original instructions involve multiple simultaneous modifications, which are not well suited for evaluating gradual transitions, we simplify the instructions using an LLM. The model is prompted to rewrite each instruction so that it involves at most two attribute changes.

\subsection{Metrics}

\paragraph{Continuous Blending metrics.}
We evaluate continuous blending quality using PPL, FID and  MUSIQ.


\textbf{PPL.}
We compute the perceptual path length (PPL) following~\cite{karras2020analyzing}, using LPIPS distances between consecutive frames with a VGG backbone. Formally,
\[
\text{PPL} = (N-1)^2 \sum_{i=1}^{N-1} \text{LPIPS}(I_{i+1}, I_i)^2.
\]
This metric captures the consistency of changes along the trajectory, where lower values indicate more uniform transitions.

\textbf{MUSIQ.}
We evaluate perceptual image quality using MUSIQ~\cite{ke2021musiq}. Scores are computed per frame and averaged across all intermediate steps and examples. Higher values indicate better visual quality.

\textbf{FID.}
We evaluate distributional similarity using Fréchet Inception Distance (FID)~\cite{heusel2017gans}. Real images correspond to all source and target images across the dataset, while generated images correspond to all intermediate frames (excluding endpoints) pooled across all examples. Lower values indicate better alignment with the real image distribution.

\paragraph{Continuous editing metrics.}
We evaluate continuous editing behavior using $\delta_{\text{smooth}}$, continuity, and normalized CLIP direction (Norm CLIP-Dir).

\textbf{$\delta_{\text{smooth}}$.}
We evaluate smoothness using the $\delta_{\text{smooth}}$ metric~\cite{parihar2025kontinuous}, which measures second-order consistency along the trajectory. It is defined based on the triangle inequality between consecutive triplets of frames, capturing deviations from linear transitions (see~\cite{parihar2025kontinuous} for full details). Lower values indicate smoother local transitions.

\textbf{Continuity.}
We measure continuity following~\cite{zarei2025slideredit} by evaluating the uniformity of similarity scores computed in CLIP embedding space along the editing trajectory. A chi-squared statistic is applied over binned similarity values to quantify deviations from a uniform progression (see~\cite{zarei2025slideredit} for full details), where higher values indicate smoother and more consistent transitions.

\textbf{Norm CLIP-Dir.}
We measure text alignment consistency using the normalized CLIP direction metric~\cite{wolf2026continuous}, which evaluates whether each step follows the intended semantic edit direction. Specifically, we compute the cosine similarity between the stepwise image direction and the text direction in CLIP embedding space, normalized by the global image-text alignment:
\[
\frac{1}{N-1} \sum_{i=1}^{N-1} \frac{
    \cos(\Delta \mathbf{v}_{img}^{(i)}, \Delta \mathbf{v}_{text})
}{
    \cos(\Delta \mathbf{v}_{img}^{(global)}, \Delta \mathbf{v}_{text})
},
\]
where $\Delta \mathbf{v}_{img}^{(i)} = \text{CLIP}(I_{i+1}) - \text{CLIP}(I_i)$ is the local edit direction, $\Delta \mathbf{v}_{img}^{(global)} = \text{CLIP}(I_{N-1}) - \text{CLIP}(I_0)$ is the global edit direction, and $\Delta \mathbf{v}_{text} = \text{CLIP}(c_{edit}) - \text{CLIP}(c_{src})$ is the text direction. Higher values indicate better alignment with the intended edit.

\subsection{User Study Details}

We conduct separate pairwise comparison studies for continuous blending and continuous editing.
Participants are presented with two transition sequences generated by our method and a baseline method, shown in randomized order.

For \textbf{continuous blending}, participants evaluate:
(i) visual quality,
(ii) smoothness of the transition,
and (iii) overall preference.

For \textbf{continuous editing}, participants evaluate:
(i) instruction faithfulness,
(ii) smoothness,
and (iii) overall preference.

\begin{figure}[t]
\vspace{-4pt}
\begin{center}
	\includegraphics[width=1\linewidth]{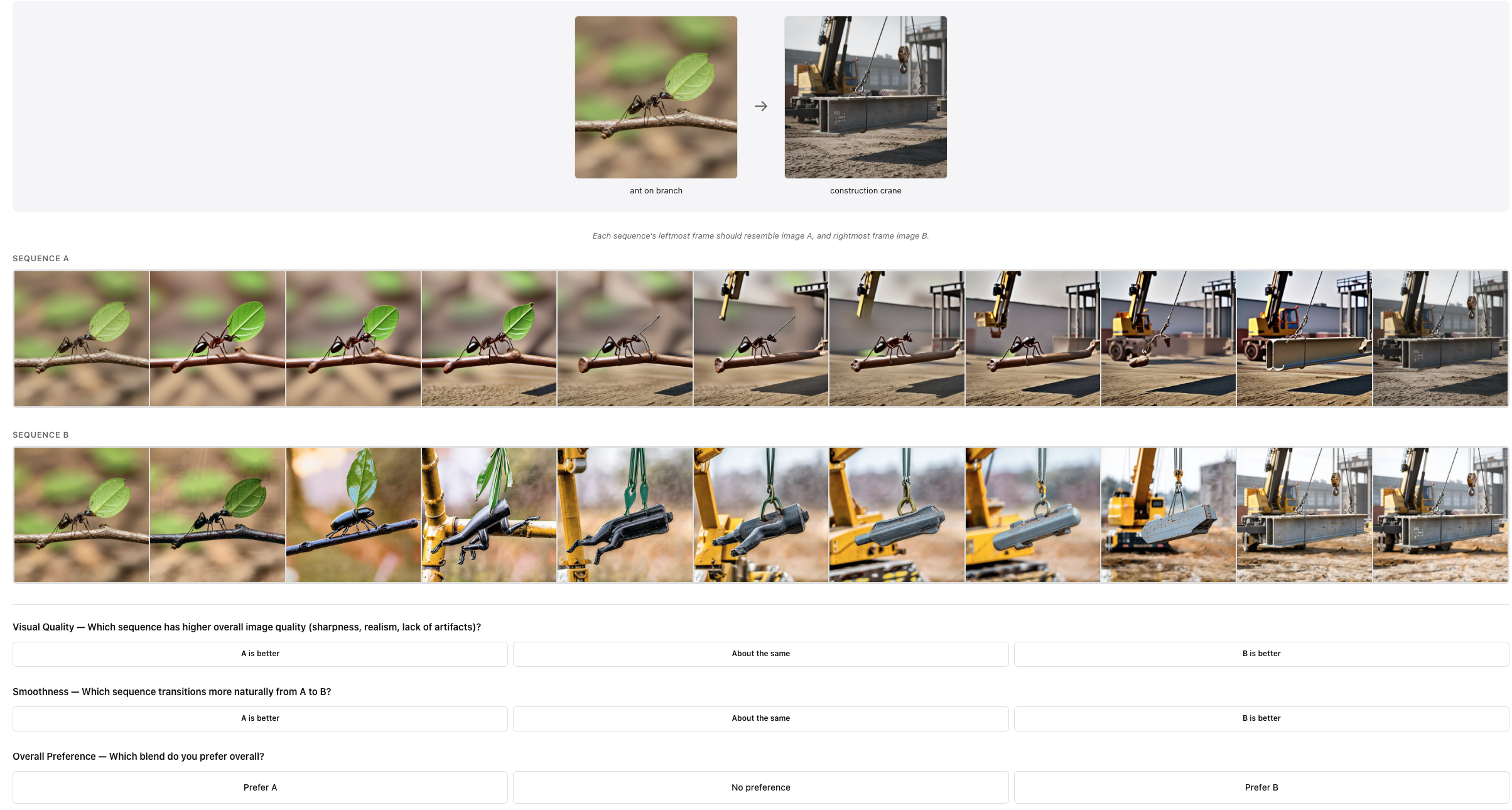}
\end{center}
\vspace{-7pt}
\caption{
User study interface used for pairwise evaluation of continuous blending.
Participants compare two transition sequences and select preferences based on visual quality, smoothness, and overall preference.
}

\vspace{-2pt}
\label{fig:user_study_ui}
\end{figure}

Figure~\ref{fig:user_study_ui} shows the interface used for the continuous blending evaluation.

\subsection{Additional Implementation Details}
\label{sec:implementation_details}

\paragraph{\textbf{Structural Prompt Alignment.}}
Figure~\ref{fig:paired_alignment} illustrates the two-stage LLM alignment pipeline used for continuous blending. We first jointly generate aligned source/edit JSON descriptions from the two input images while enforcing shared structure, field ordering, and consistent wording for unaffected content. We then perform image-aware refinement to localize textual differences only to semantically modified regions, while unchanged fields are force-copied to preserve exact correspondence. This stage produces consistent text-level alignment.


\paragraph{\textbf{Runtime Overhead.}}
The main preprocessing overhead comes from two Gemini calls for structural prompt generation and refinement, requiring 23.5\,s per prompt pair on average on PIE-Bench. The two FIBO-edit (SmolLM) text-encoder passes together with the embedding alignment require approximately 84\,ms, of which the alignment itself accounts for about 20\,ms. These operations are performed only once per prompt pair and the resulting aligned embeddings are reused across all interpolation points, amortizing the preprocessing cost over the entire trajectory.

\begin{figure}[t]
\centering
\footnotesize

\begin{tcblisting}{
    enhanced,
    colback=blue!3,
    colframe=blue!55!black,
    title={\textbf{Stage 1.} \texttt{get\_paired\_prompts\_blend}: joint generation from $(I_A, I_B)$},
    fonttitle=\bfseries,
    left=3pt,
    right=3pt,
    top=2pt,
    bottom=2pt,
    boxsep=2pt,
    listing only,
    listing options={
        basicstyle=\ttfamily\footnotesize,
        breaklines=true,
        breakatwhitespace=true,
        columns=fullflexible,
        keepspaces=true,
        showstringspaces=false
    }
}
You receive TWO images and produce TWO aligned JSONs (source, edit).

- Same keys, same #objects, same field order.
- Identical wording for fields not affected by the visual diff.
- Affected fields keep the same sentence structure; substitute only
  the minimum words required.
- edit.edit_instruction: imperative description of the change.
- source.edit_instruction: counter-instruction preserving the
  original (e.g., edit "Change hair to red" -> source "Keep hair brown").
\end{tcblisting}

\vspace{-2pt}

\begin{tcblisting}{
    enhanced,
    colback=orange!4,
    colframe=orange!70!black,
    title={\textbf{Stage 2.} \texttt{refine\_alignment}: image-aware bidirectional refinement},
    fonttitle=\bfseries,
    left=3pt,
    right=3pt,
    top=2pt,
    bottom=2pt,
    boxsep=2pt,
    listing only,
    listing options={
        basicstyle=\ttfamily\footnotesize,
        breaklines=true,
        breakatwhitespace=true,
        columns=fullflexible,
        keepspaces=true,
        showstringspaces=false
    }
}
Classify each field by word-diff ratio:

unaffected (<10%, no overlap w/ edit_instruction) -> force-copy
source -> edit

affected -> send to LLM with image I_A:

"Rewrite BOTH source and edit with parallel sentence structure
and minimal word differences. Source must match the image;
edit must reflect the edit instruction. Same #sentences, same
#clauses, same rhythm -- swap only the minimum words that must
differ."
\end{tcblisting}

\caption{
\textbf{Structural text alignment pipeline for continuous blending.}
Stage 1 prompts the LLM to jointly generate aligned source/edit JSONs from two images while enforcing shared structure, object count, field order, and consistent wording for unaffected regions. Stage 2 performs image-aware bidirectional refinement: unchanged fields are force-copied, while affected fields are rewritten in parallel with minimal word substitutions to preserve structural alignment. This concentrates textual differences only on semantically modified content.
}
\label{fig:paired_alignment}

\end{figure}






\section{Additional results}

We include in the following figures additional qualitative results complementing the main paper for the two applications evaluated in this work: continuous editing and continuous blending. These supplementary results include comparisons and transition sequences generated using both FIBO-edit and FLUX2-Klein across diverse semantic transformations.

\newcommand{\imgwidthablate}{0.136\linewidth}

\begin{figure*}
    \centering
    \setlength{\tabcolsep}{0pt}
    \begin{tabular}{cccc cc cccccc}

        \raisebox{20pt}{\rotatebox[origin=t]{90}{{DiffMorpher}}} & { } &
        \includegraphics[width=\imgwidthablate]{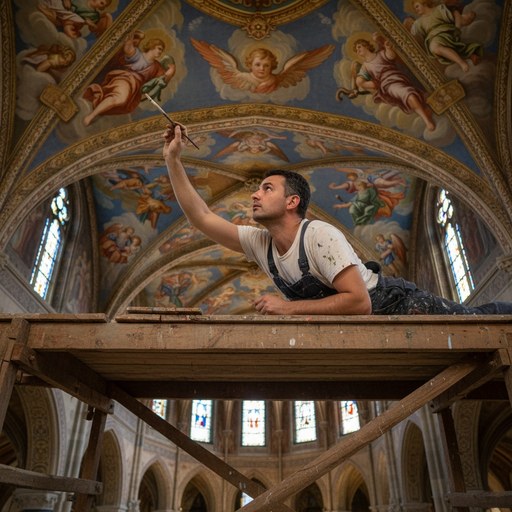} &
        \includegraphics[width=\imgwidthablate]{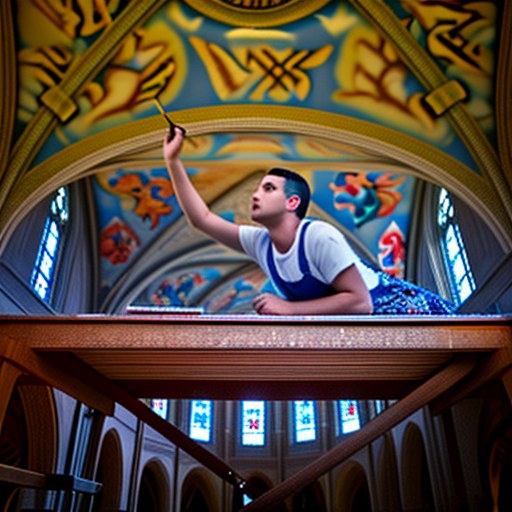} &
        \includegraphics[width=\imgwidthablate]{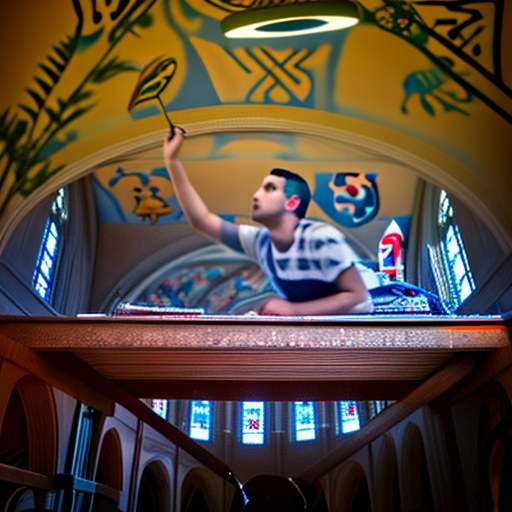} &
        \includegraphics[width=\imgwidthablate]{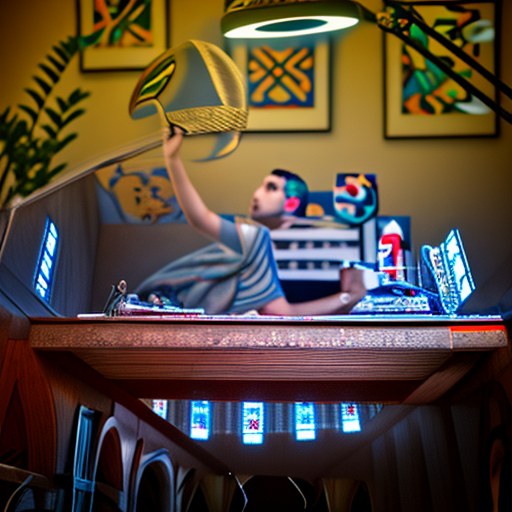} &
        \includegraphics[width=\imgwidthablate]{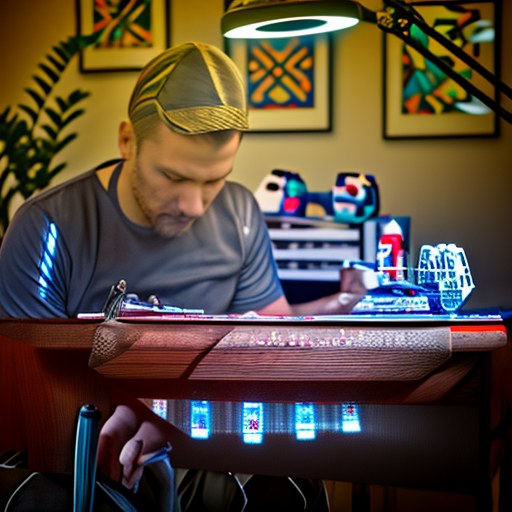} &
        \includegraphics[width=\imgwidthablate]{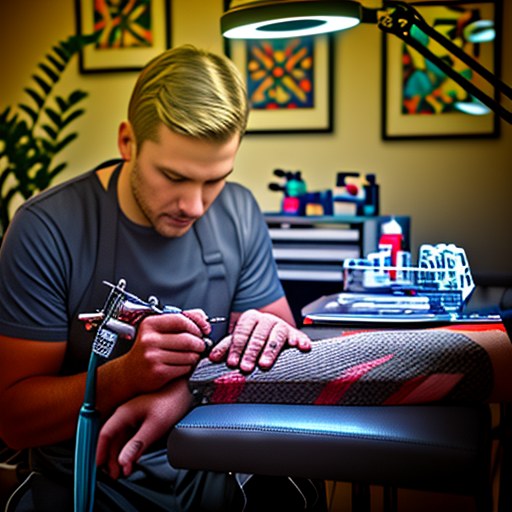} &
        \includegraphics[width=\imgwidthablate]{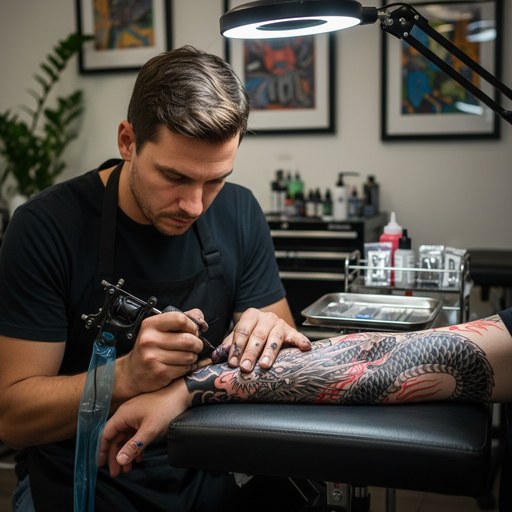} \\
        \raisebox{20pt}{\rotatebox[origin=t]{90}{{FreeMorph}}} & { } &
        \includegraphics[width=\imgwidthablate]{images/blend_comparison/tatoo/reference/image1.jpg} &
        \includegraphics[width=\imgwidthablate]{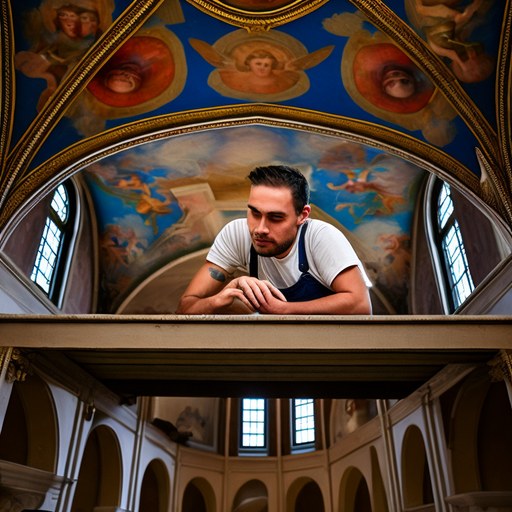} &
        \includegraphics[width=\imgwidthablate]{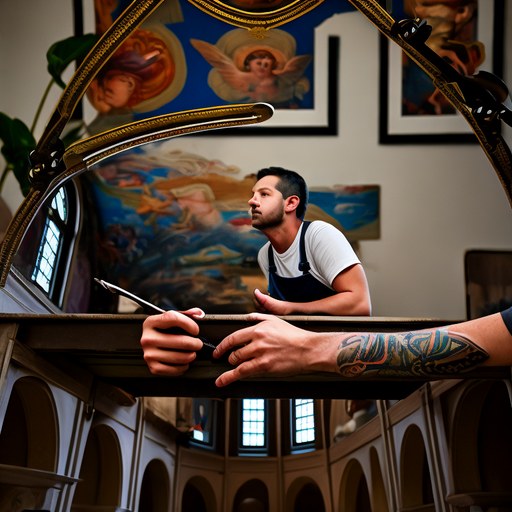} &
        \includegraphics[width=\imgwidthablate]{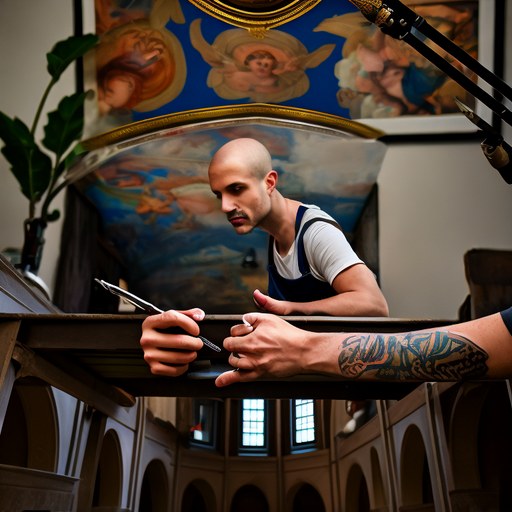} &
        \includegraphics[width=\imgwidthablate]{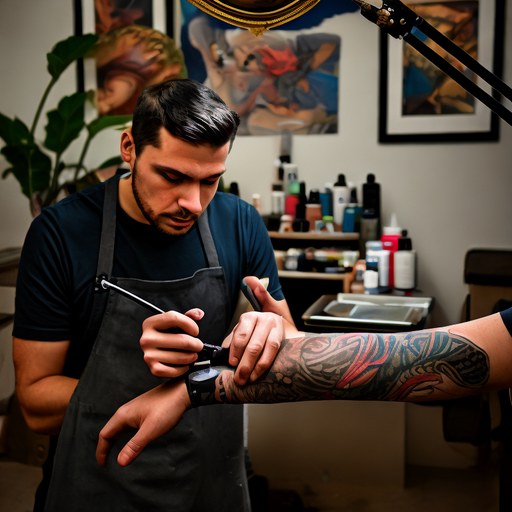} &
        \includegraphics[width=\imgwidthablate]{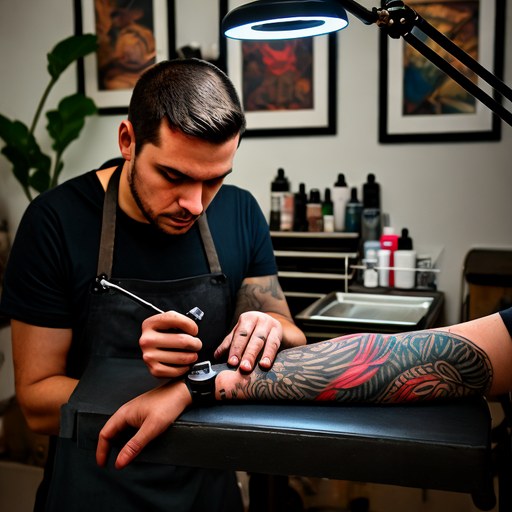} &
        \includegraphics[width=\imgwidthablate]{images/blend_comparison/tatoo/reference/image2.jpg} \\
        \raisebox{20pt}{\rotatebox[origin=t]{90}{Vibe Space}} & { } &
        \includegraphics[width=\imgwidthablate]{images/blend_comparison/tatoo/reference/image1.jpg} &
        \includegraphics[width=\imgwidthablate]{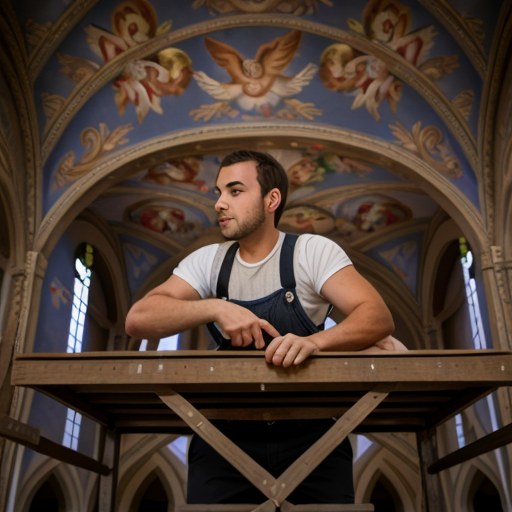} &
        \includegraphics[width=\imgwidthablate]{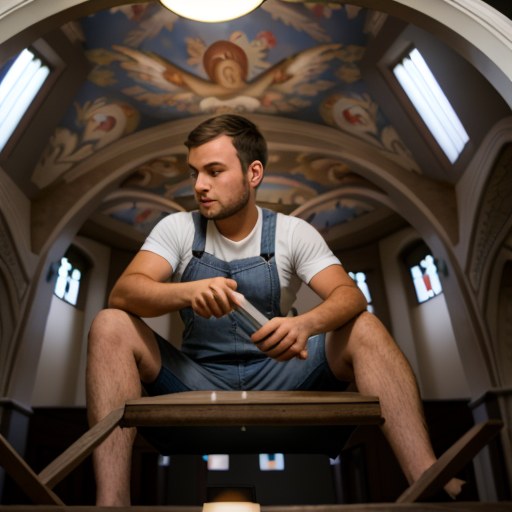} &
        \includegraphics[width=\imgwidthablate]{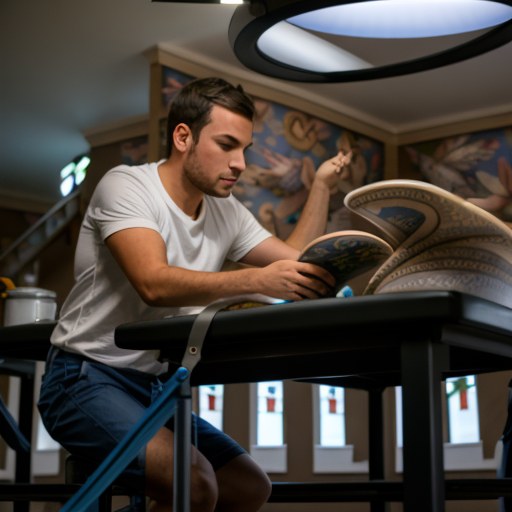} &
        \includegraphics[width=\imgwidthablate]{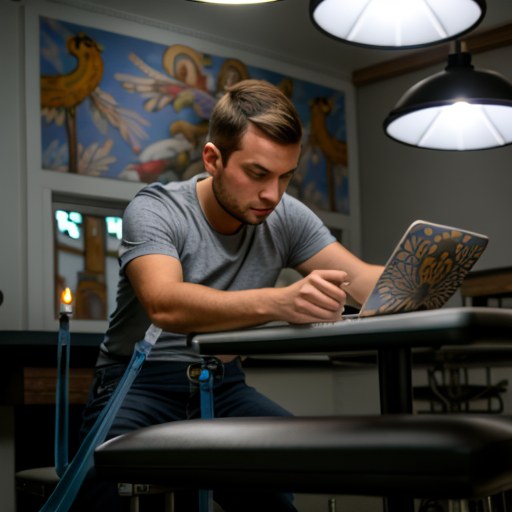} &
        \includegraphics[width=\imgwidthablate]{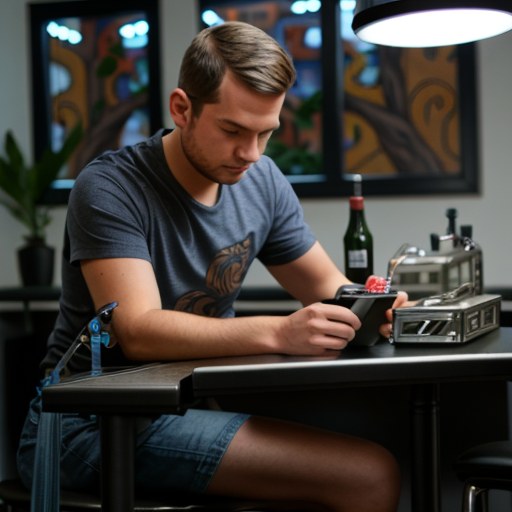} &
        \includegraphics[width=\imgwidthablate]{images/blend_comparison/tatoo/reference/image2.jpg} \\
        \raisebox{20pt}{\rotatebox[origin=t]{90}{T2T (Ours)}} & { } &
        \includegraphics[width=\imgwidthablate]{images/blend_comparison/tatoo/reference/image1.jpg} &
        \includegraphics[width=\imgwidthablate]{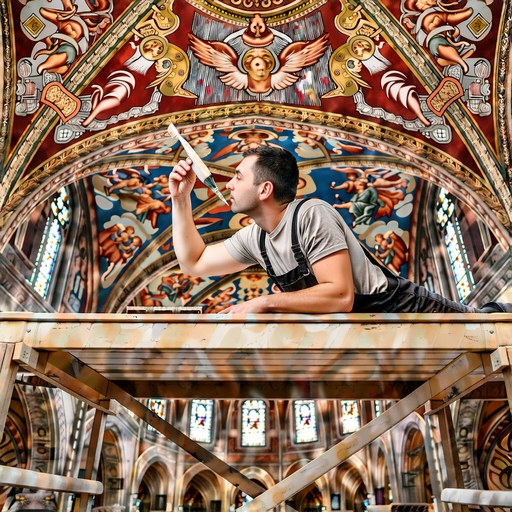} &
        \includegraphics[width=\imgwidthablate]{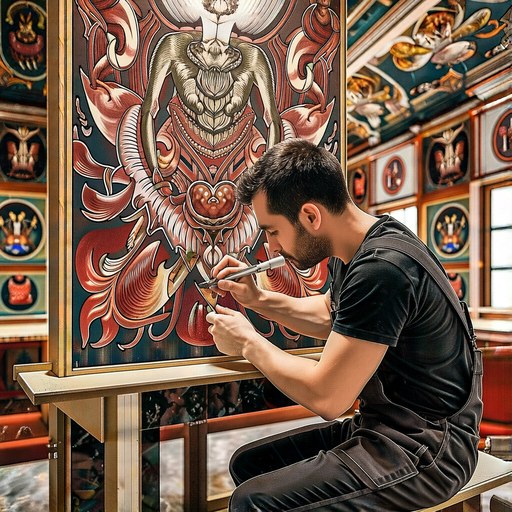} &
        \includegraphics[width=\imgwidthablate]{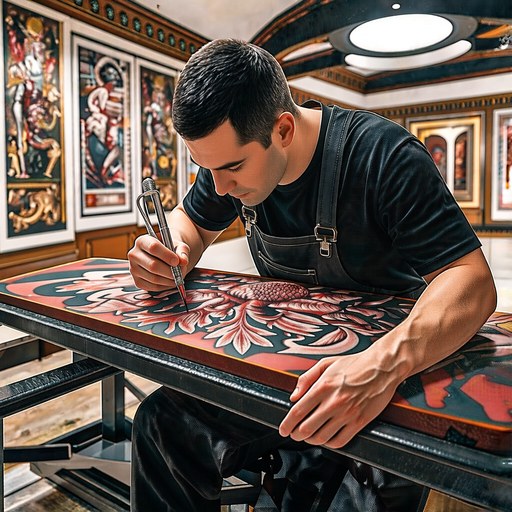} &
        \includegraphics[width=\imgwidthablate]{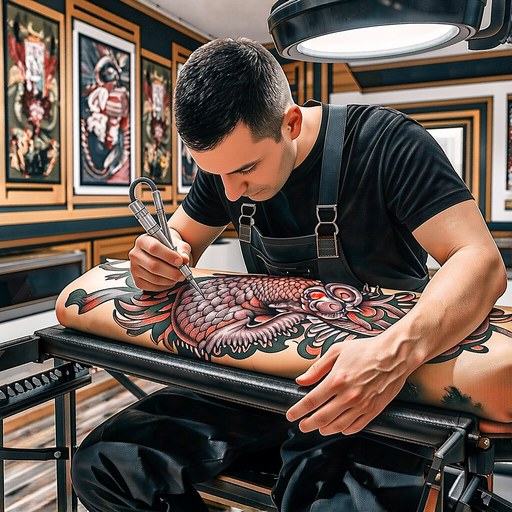} &
        \includegraphics[width=\imgwidthablate]{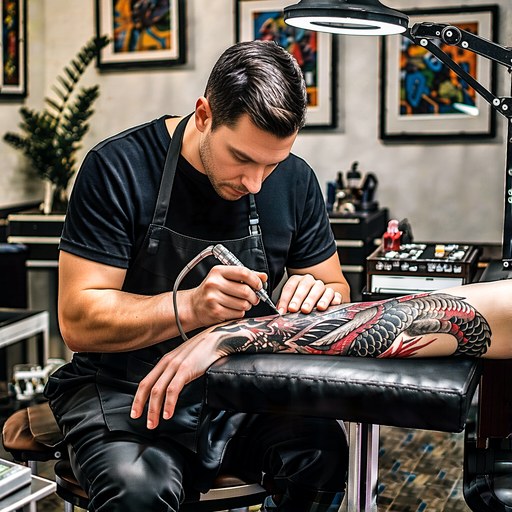} &
        \includegraphics[width=\imgwidthablate]{images/blend_comparison/tatoo/reference/image2.jpg} \\
        & & Input A & &  & &  & & Input B \\
        \\
        
        \raisebox{20pt}{\rotatebox[origin=t]{90}{{DiffMorpher}}} & { } &
        \includegraphics[width=\imgwidthablate]{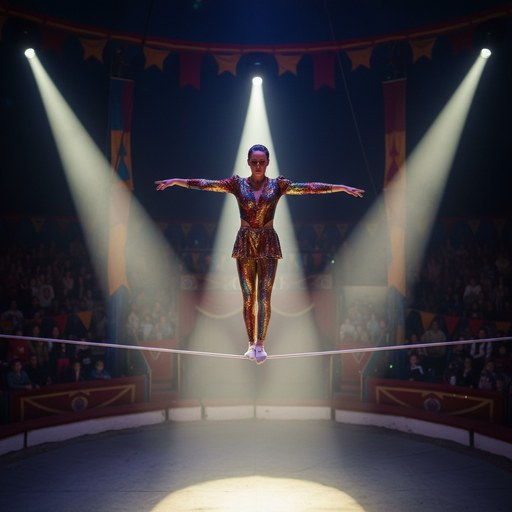} &
        \includegraphics[width=\imgwidthablate]{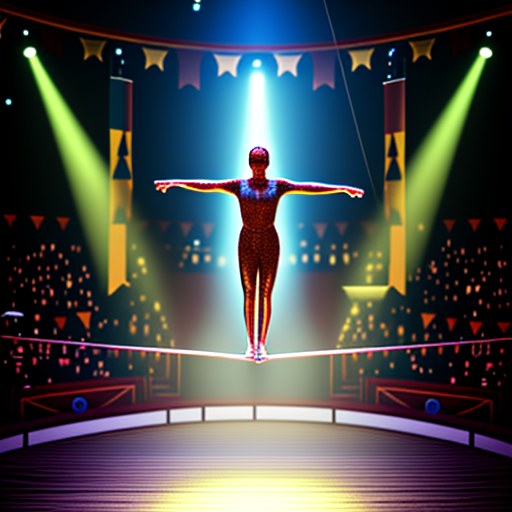} &
        \includegraphics[width=\imgwidthablate]{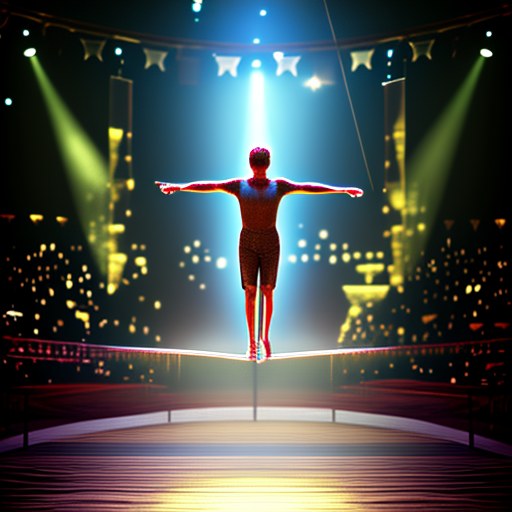} &
        \includegraphics[width=\imgwidthablate]{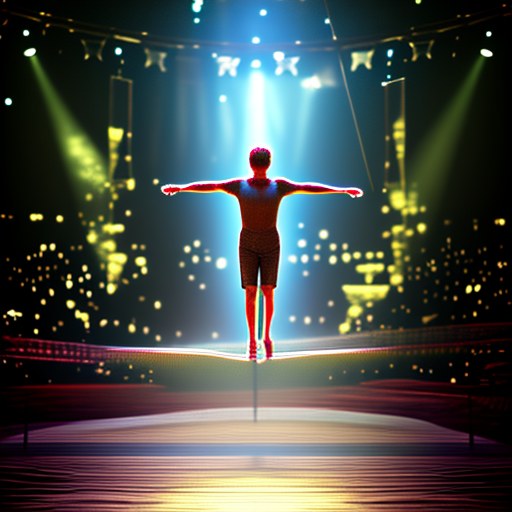} &
        \includegraphics[width=\imgwidthablate]{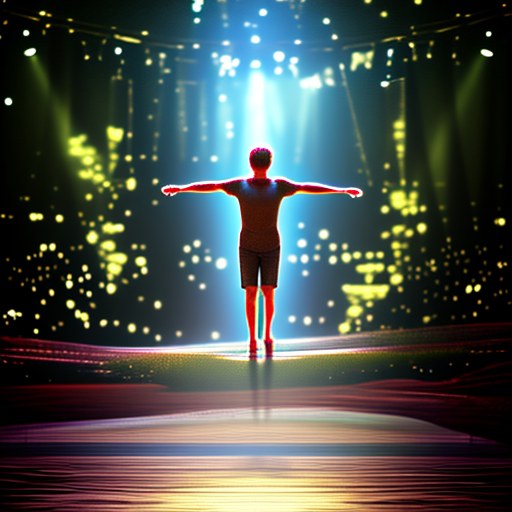} &
        \includegraphics[width=\imgwidthablate]{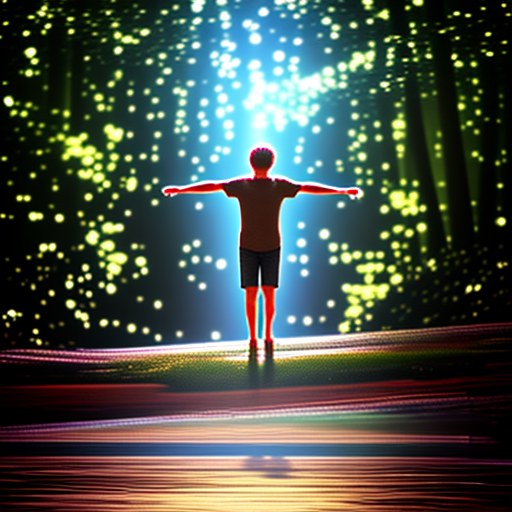} &
        \includegraphics[width=\imgwidthablate]{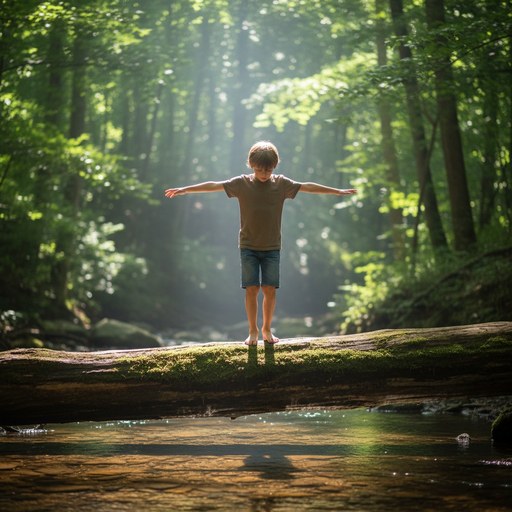} \\
        \raisebox{20pt}{\rotatebox[origin=t]{90}{{FreeMorph}}} & { } &
        \includegraphics[width=\imgwidthablate]{images/blend_comparison/circus_log/reference/image1.jpg} &
        \includegraphics[width=\imgwidthablate]{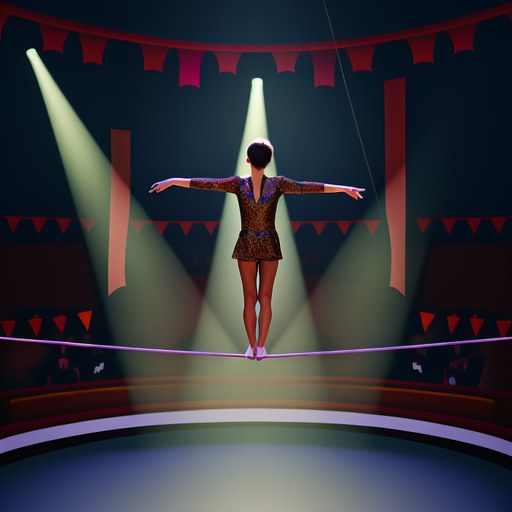} &
        \includegraphics[width=\imgwidthablate]{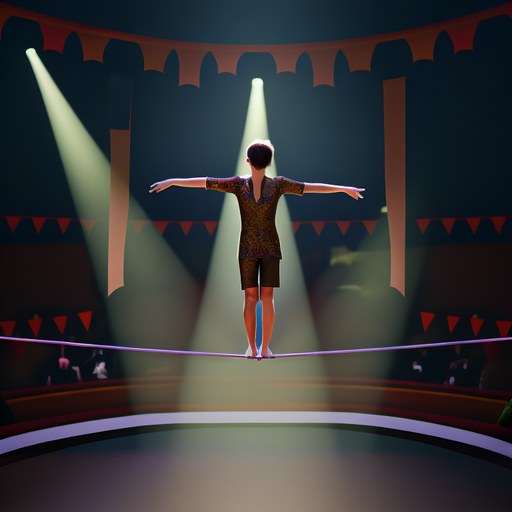} &
        \includegraphics[width=\imgwidthablate]{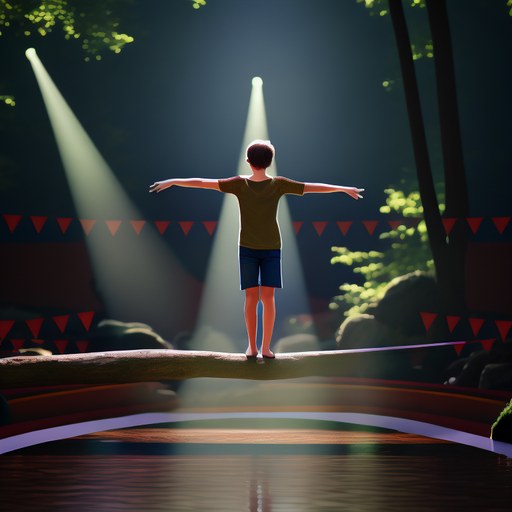} &
        \includegraphics[width=\imgwidthablate]{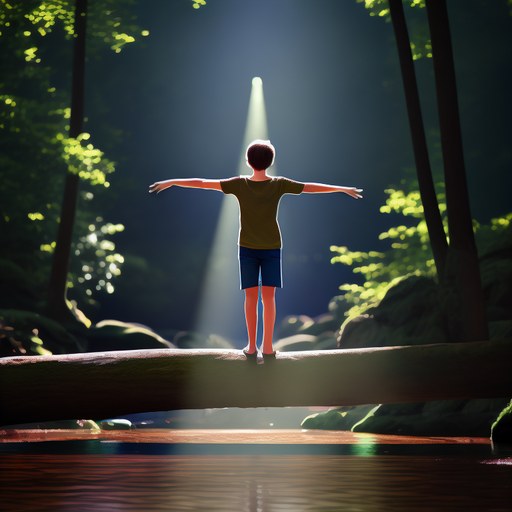} &
        \includegraphics[width=\imgwidthablate]{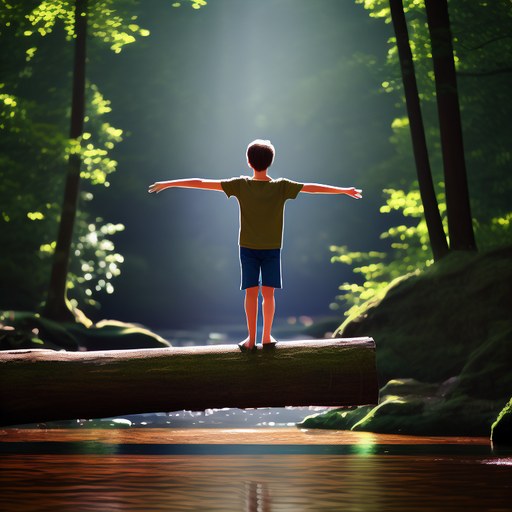} &
        \includegraphics[width=\imgwidthablate]{images/blend_comparison/circus_log/reference/image2.jpg} \\
        \raisebox{20pt}{\rotatebox[origin=t]{90}{Vibe Space}} & { } &
        \includegraphics[width=\imgwidthablate]{images/blend_comparison/circus_log/reference/image1.jpg} &
        \includegraphics[width=\imgwidthablate]{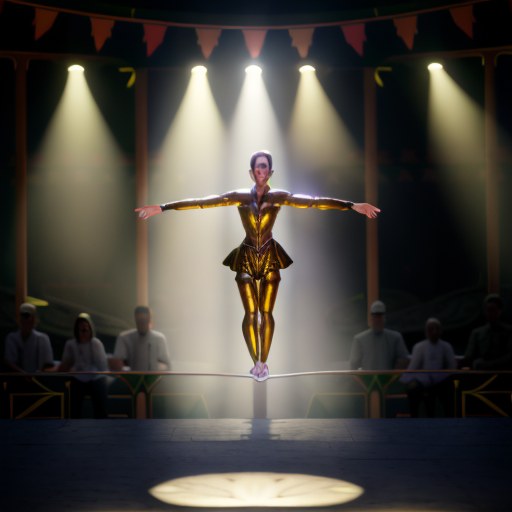} &
        \includegraphics[width=\imgwidthablate]{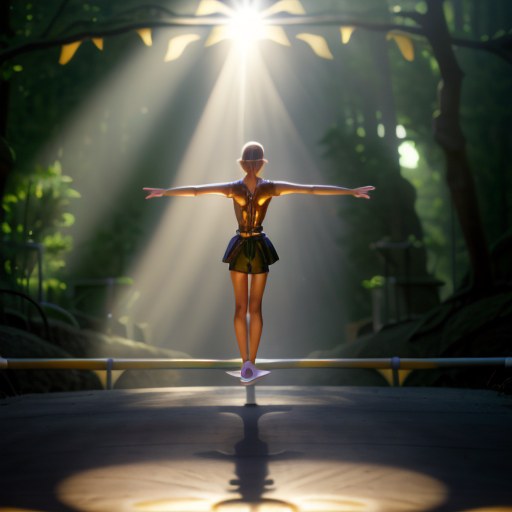} &
        \includegraphics[width=\imgwidthablate]{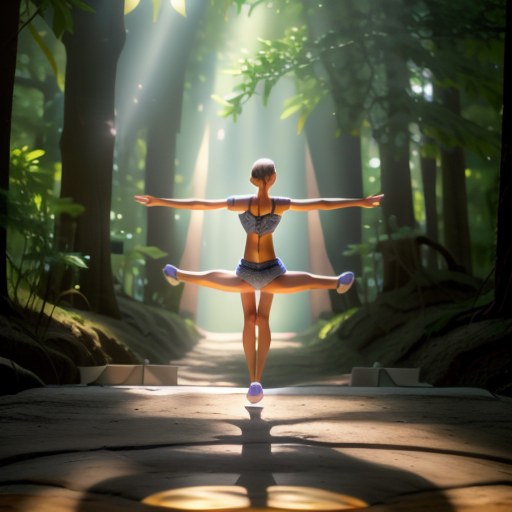} &
        \includegraphics[width=\imgwidthablate]{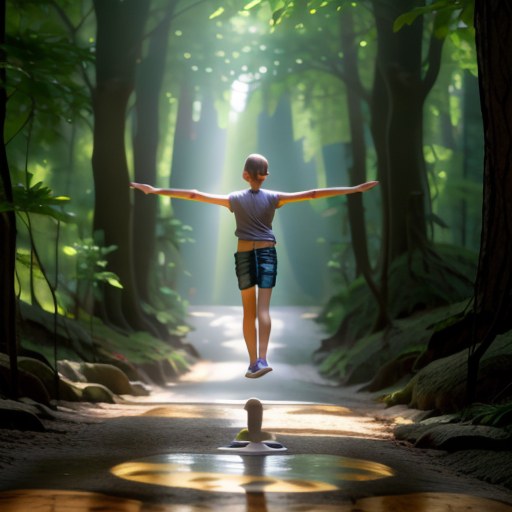} &
        \includegraphics[width=\imgwidthablate]{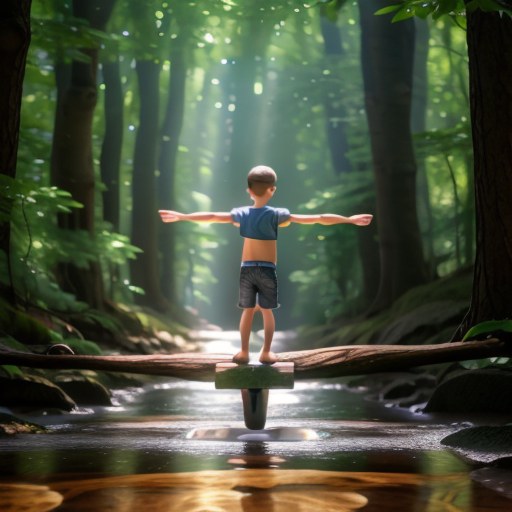} &
        \includegraphics[width=\imgwidthablate]{images/blend_comparison/circus_log/reference/image2.jpg} \\
        \raisebox{20pt}{\rotatebox[origin=t]{90}{T2T (Ours)}} & { } &
        \includegraphics[width=\imgwidthablate]{images/blend_comparison/circus_log/reference/image1.jpg} &
        \includegraphics[width=\imgwidthablate]{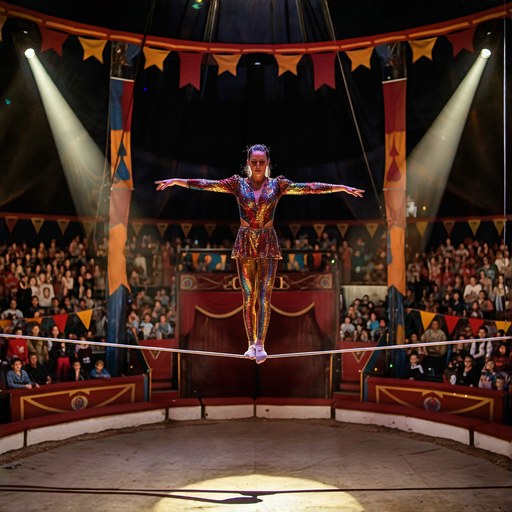} &
        \includegraphics[width=\imgwidthablate]{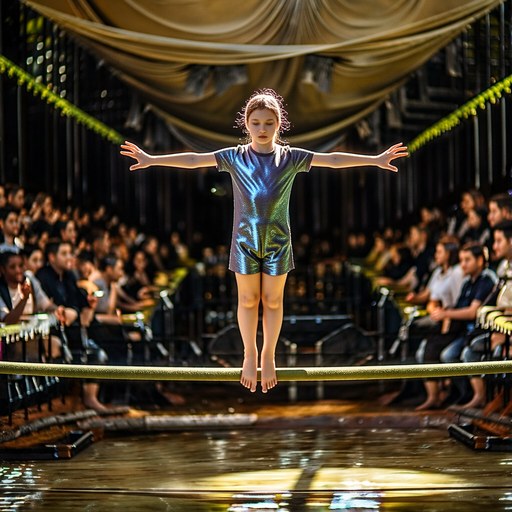} &
        \includegraphics[width=\imgwidthablate]{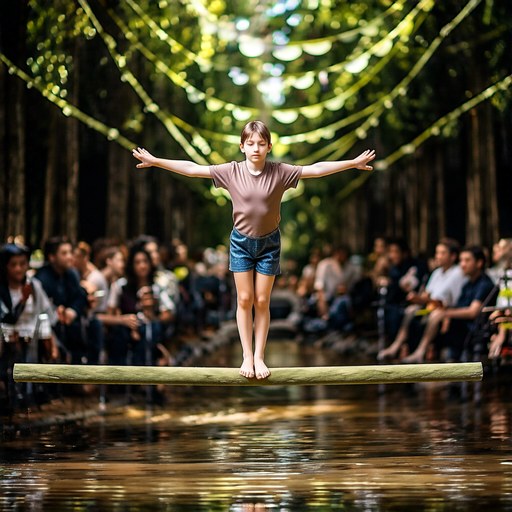} &
        \includegraphics[width=\imgwidthablate]{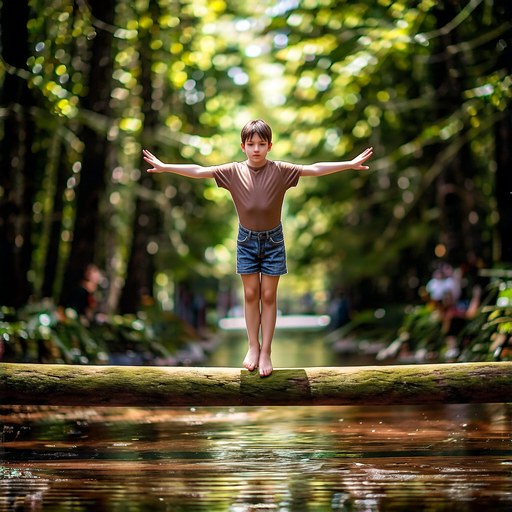} &
        \includegraphics[width=\imgwidthablate]{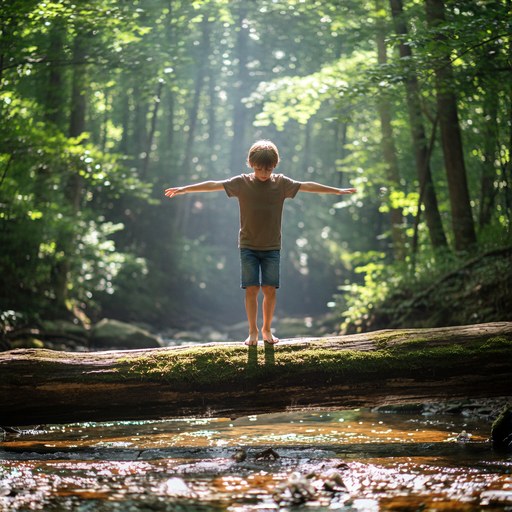} &
        \includegraphics[width=\imgwidthablate]{images/blend_comparison/circus_log/reference/image2.jpg} \\
        & & Input A & &  & &  & & Input B

    \end{tabular}
    \caption{Qualitative comparison with continuous blending methods. Results generated using FIBO-edit.}
\vspace{-10pt}
    \label{fig:supp_blend_continous}
\end{figure*}
\renewcommand{\imgwidthablate}{0.136\linewidth}

\begin{figure*}
    \centering
    \setlength{\tabcolsep}{0pt}
    \begin{tabular}{cccc cc cccccc}

        \raisebox{20pt}{\rotatebox[origin=t]{90}{{DiffMorpher}}} & { } &
        \includegraphics[width=\imgwidthablate]{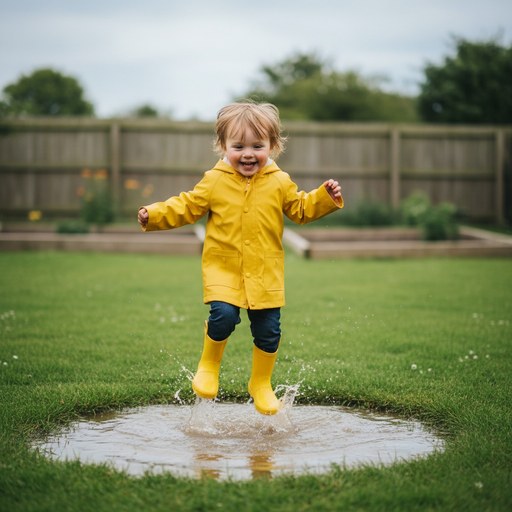} &
        \includegraphics[width=\imgwidthablate]{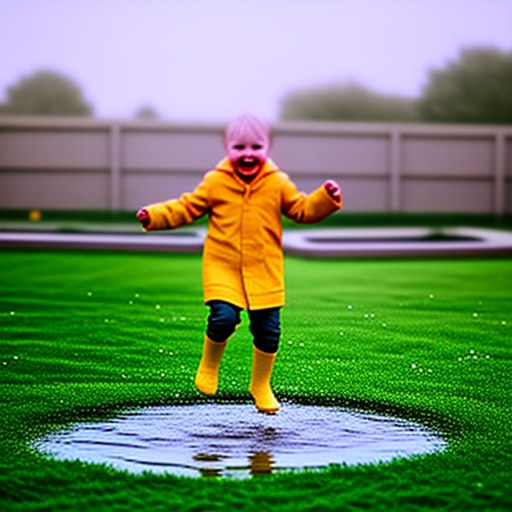} &
        \includegraphics[width=\imgwidthablate]{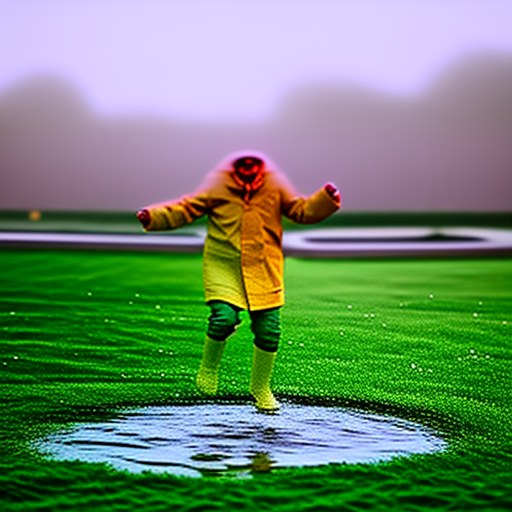} &
        \includegraphics[width=\imgwidthablate]{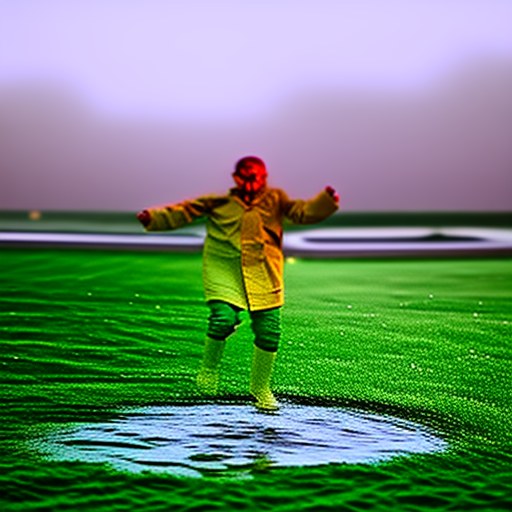} &
        \includegraphics[width=\imgwidthablate]{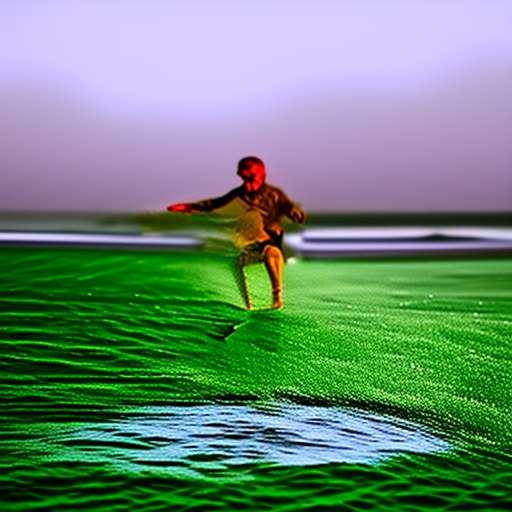} &
        \includegraphics[width=\imgwidthablate]{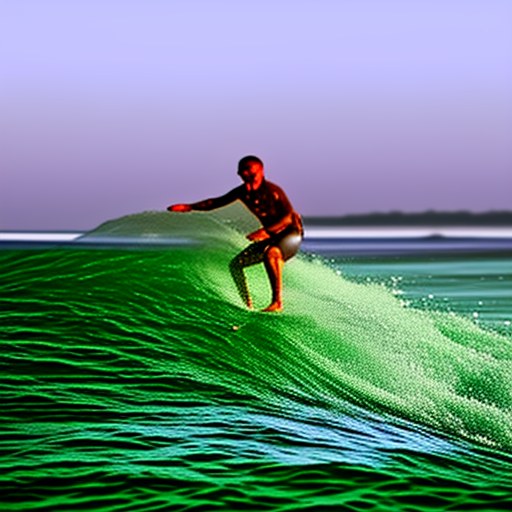} &
        \includegraphics[width=\imgwidthablate]{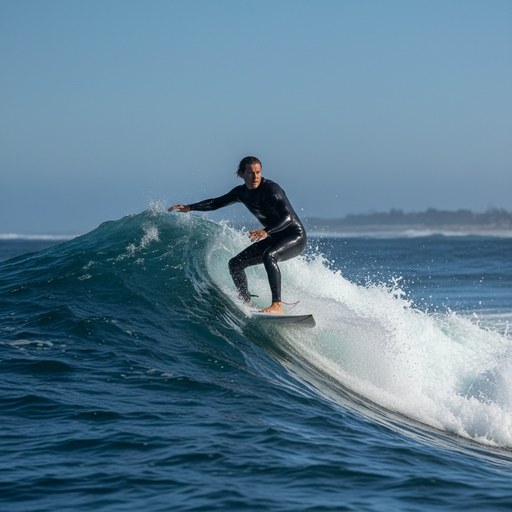} \\
        \raisebox{20pt}{\rotatebox[origin=t]{90}{{FreeMorph}}} & { } &
        \includegraphics[width=\imgwidthablate]{images/blend_comparison/surf_puddle/reference/image1.jpg} &
        \includegraphics[width=\imgwidthablate]{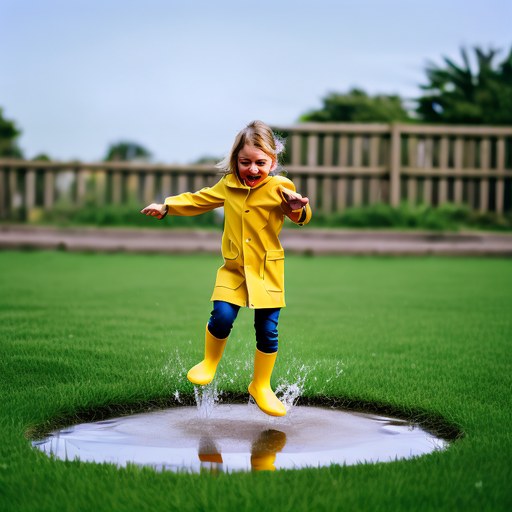} &
        \includegraphics[width=\imgwidthablate]{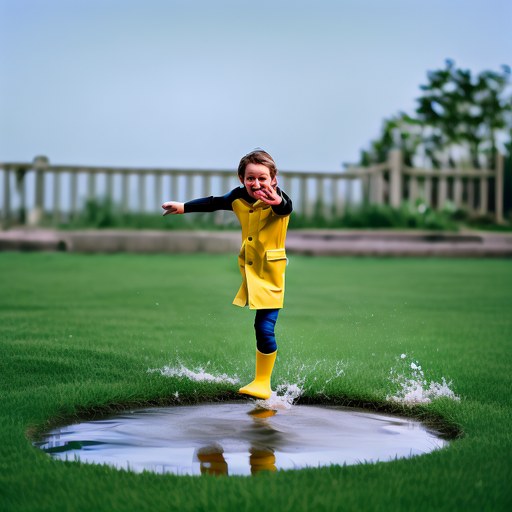} &
        \includegraphics[width=\imgwidthablate]{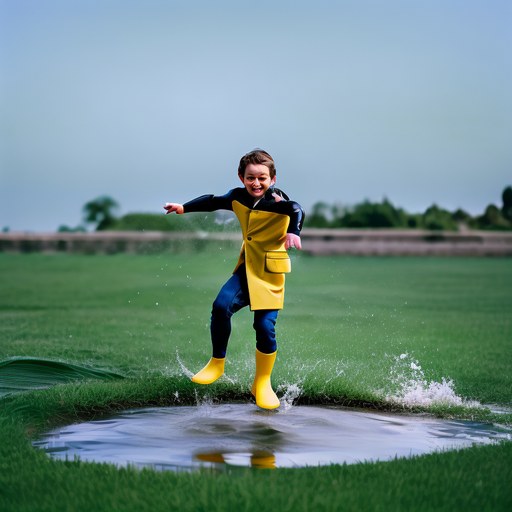} &
        \includegraphics[width=\imgwidthablate]{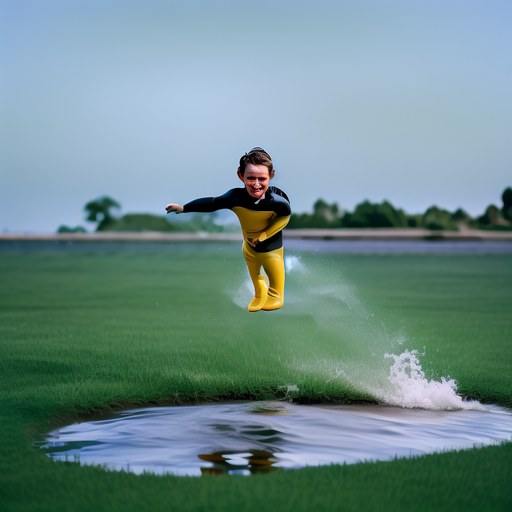} &
        \includegraphics[width=\imgwidthablate]{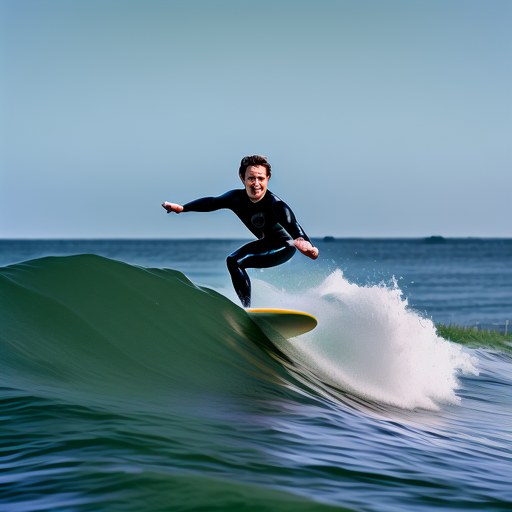} &
        \includegraphics[width=\imgwidthablate]{images/blend_comparison/surf_puddle/reference/image2.jpg} \\
        \raisebox{20pt}{\rotatebox[origin=t]{90}{Vibe Space}} & { } &
        \includegraphics[width=\imgwidthablate]{images/blend_comparison/surf_puddle/reference/image1.jpg} &
        \includegraphics[width=\imgwidthablate]{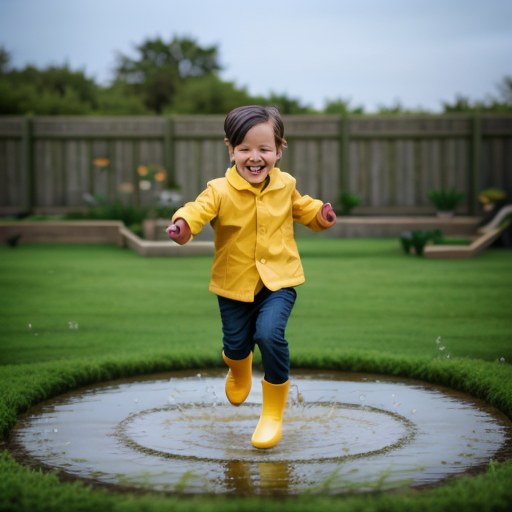} &
        \includegraphics[width=\imgwidthablate]{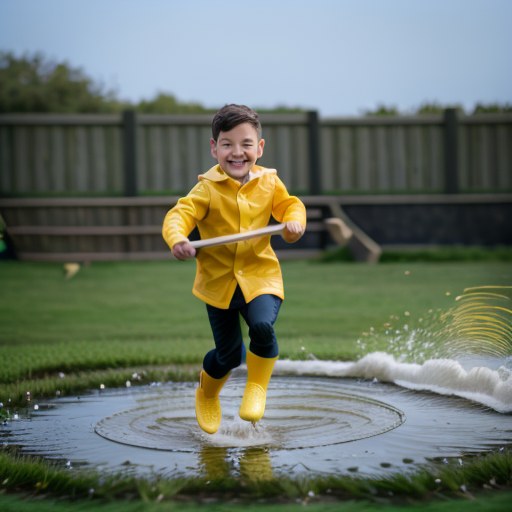} &
        \includegraphics[width=\imgwidthablate]{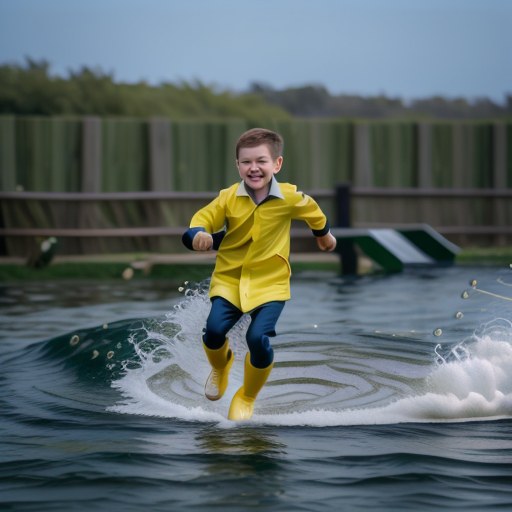} &
        \includegraphics[width=\imgwidthablate]{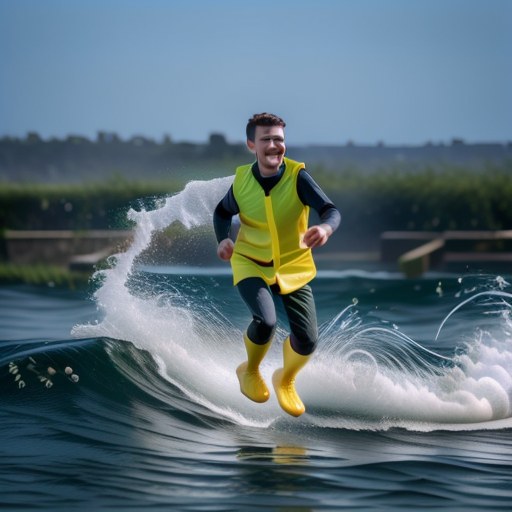} &
        \includegraphics[width=\imgwidthablate]{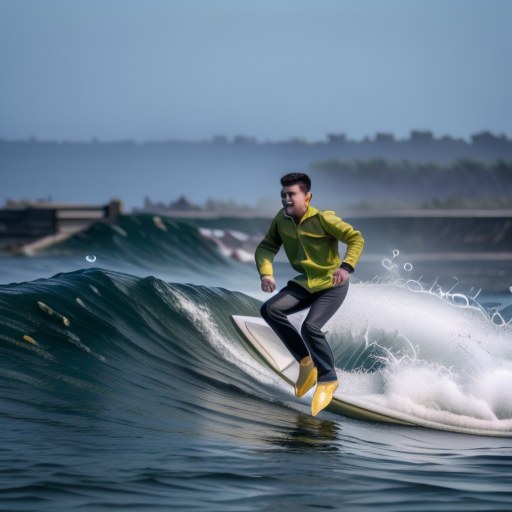} &
        \includegraphics[width=\imgwidthablate]{images/blend_comparison/surf_puddle/reference/image2.jpg} \\
        \raisebox{20pt}{\rotatebox[origin=t]{90}{T2T (Ours)}} & { } &
        \includegraphics[width=\imgwidthablate]{images/blend_comparison/surf_puddle/reference/image1.jpg} &
        \includegraphics[width=\imgwidthablate]{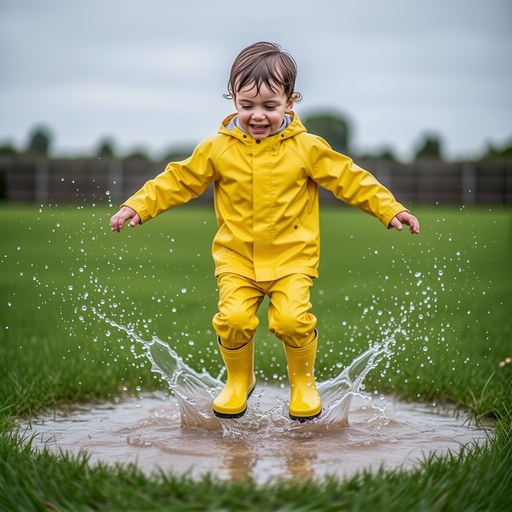} &
        \includegraphics[width=\imgwidthablate]{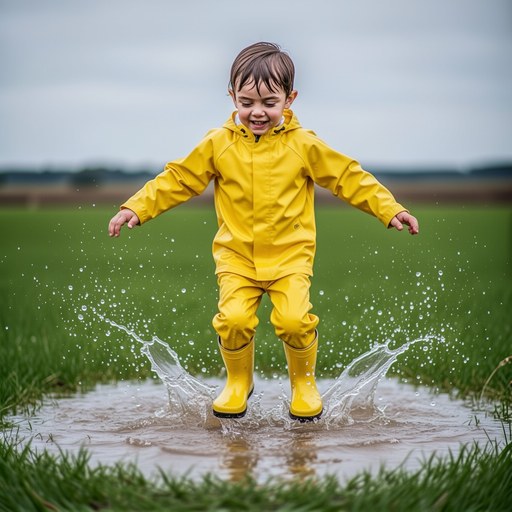} &
        \includegraphics[width=\imgwidthablate]{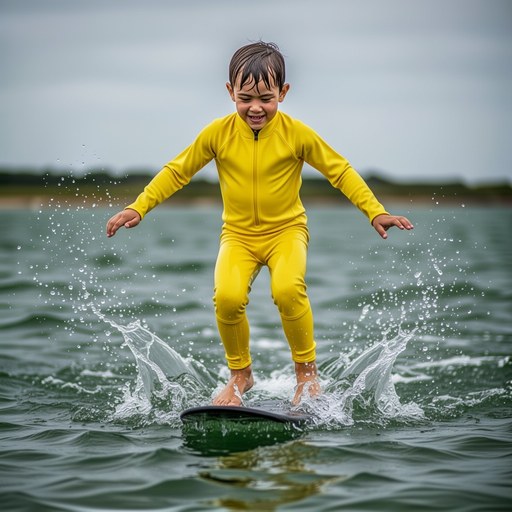} &
        \includegraphics[width=\imgwidthablate]{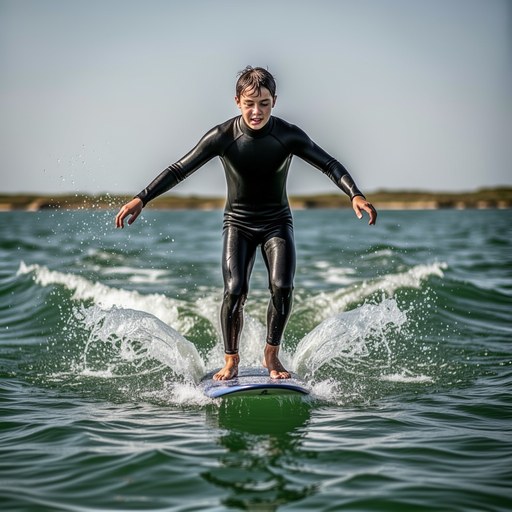} &
        \includegraphics[width=\imgwidthablate]{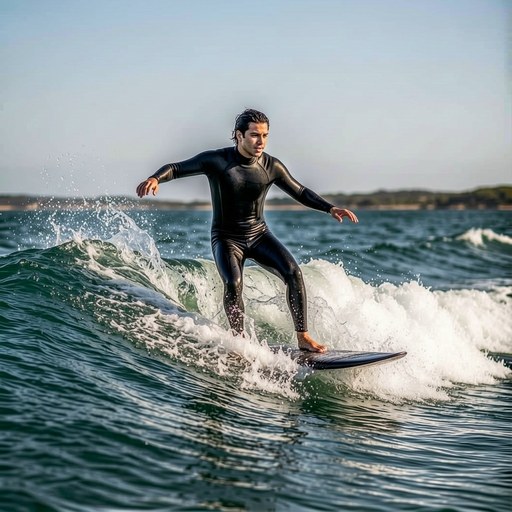} &
        \includegraphics[width=\imgwidthablate]{images/blend_comparison/surf_puddle/reference/image2.jpg} \\
        & & Input A & &  & &  & & Input B \\
        \\
        
        \raisebox{20pt}{\rotatebox[origin=t]{90}{{DiffMorpher}}} & { } &
        \includegraphics[width=\imgwidthablate]{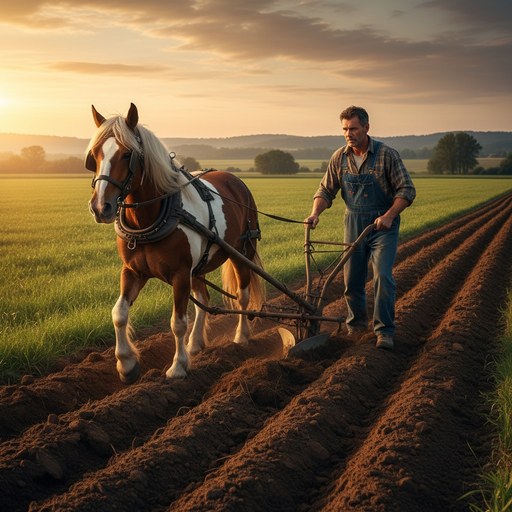} &
        \includegraphics[width=\imgwidthablate]{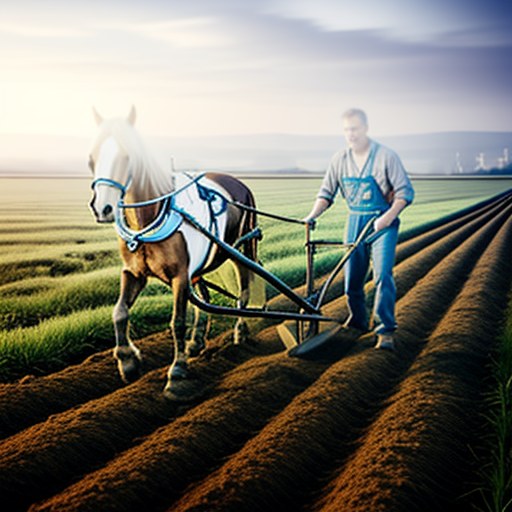} &
        \includegraphics[width=\imgwidthablate]{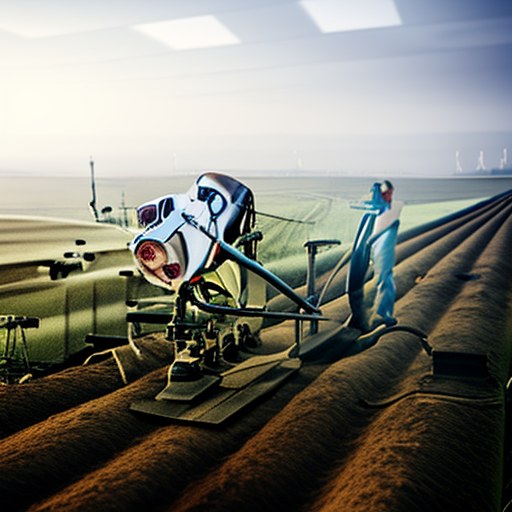} &
        \includegraphics[width=\imgwidthablate]{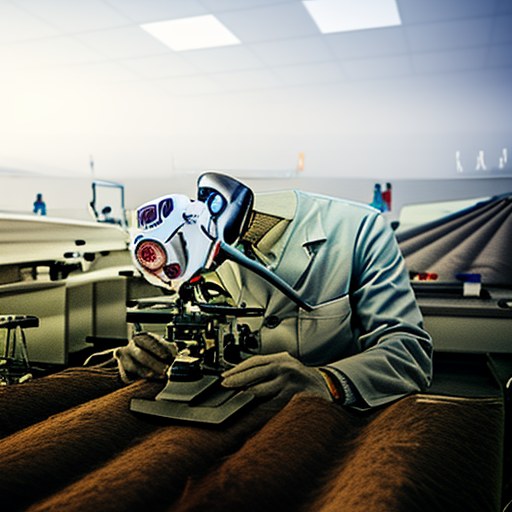} &
        \includegraphics[width=\imgwidthablate]{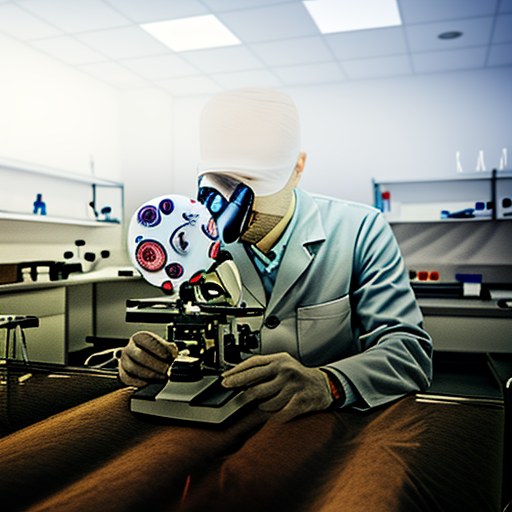} &
        \includegraphics[width=\imgwidthablate]{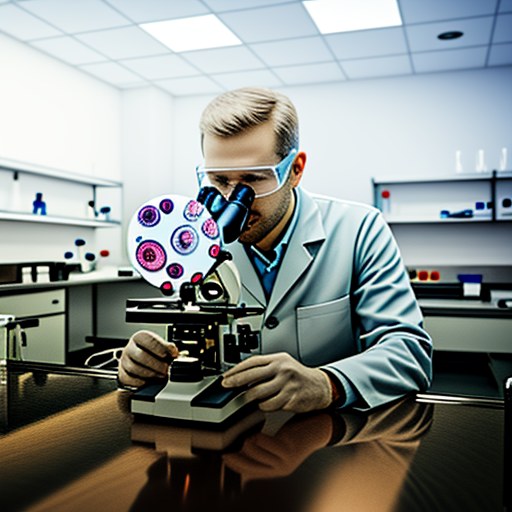} &
        \includegraphics[width=\imgwidthablate]{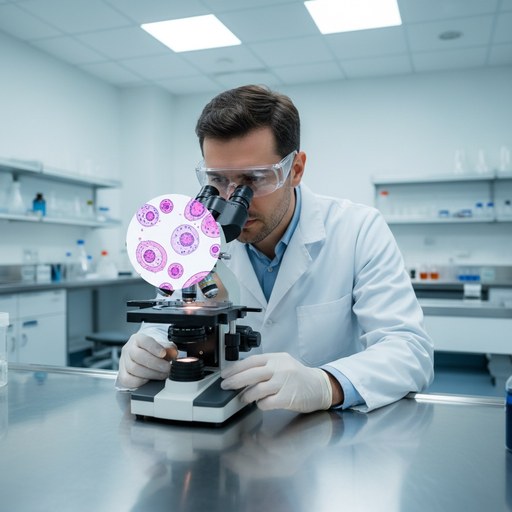} \\
        \raisebox{20pt}{\rotatebox[origin=t]{90}{{FreeMorph}}} & { } &
        \includegraphics[width=\imgwidthablate]{images/blend_comparison/farmer_scienctist/reference/image1.jpg} &
        \includegraphics[width=\imgwidthablate]{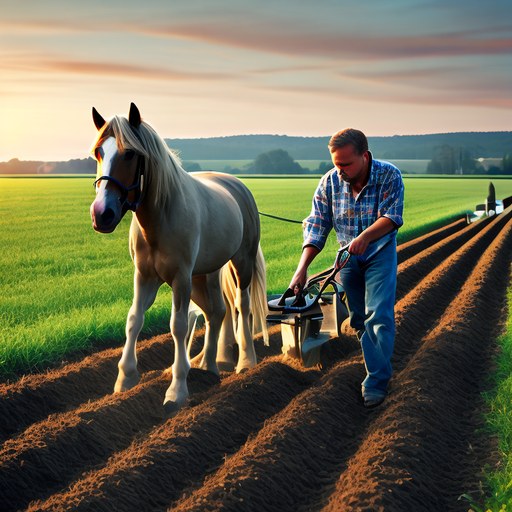} &
        \includegraphics[width=\imgwidthablate]{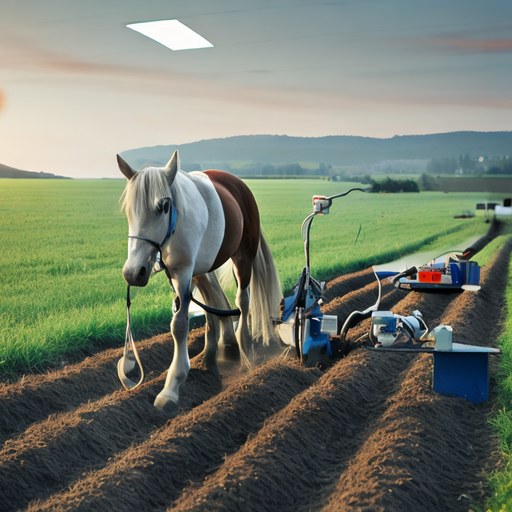} &
        \includegraphics[width=\imgwidthablate]{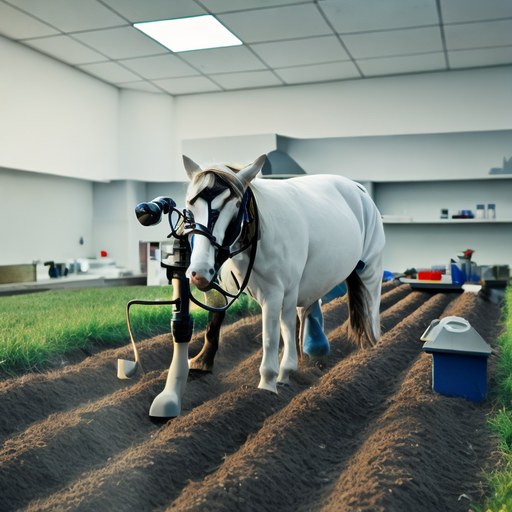} &
        \includegraphics[width=\imgwidthablate]{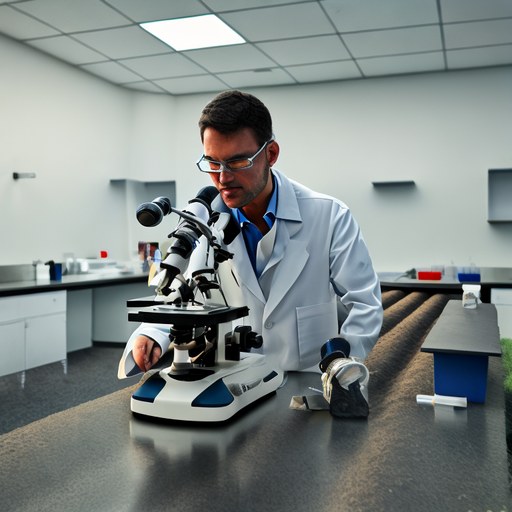} &
        \includegraphics[width=\imgwidthablate]{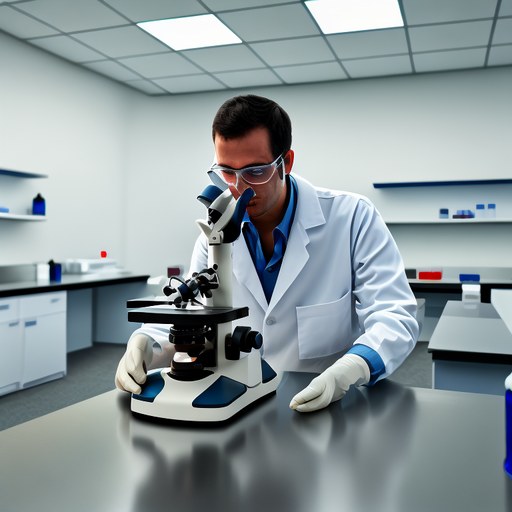} &
        \includegraphics[width=\imgwidthablate]{images/blend_comparison/farmer_scienctist/reference/image2.jpg} \\
        \raisebox{20pt}{\rotatebox[origin=t]{90}{Vibe Space}} & { } &
        \includegraphics[width=\imgwidthablate]{images/blend_comparison/farmer_scienctist/reference/image1.jpg} &
        \includegraphics[width=\imgwidthablate]{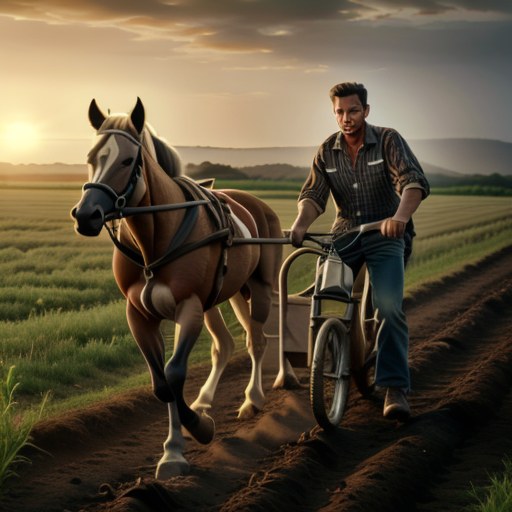} &
        \includegraphics[width=\imgwidthablate]{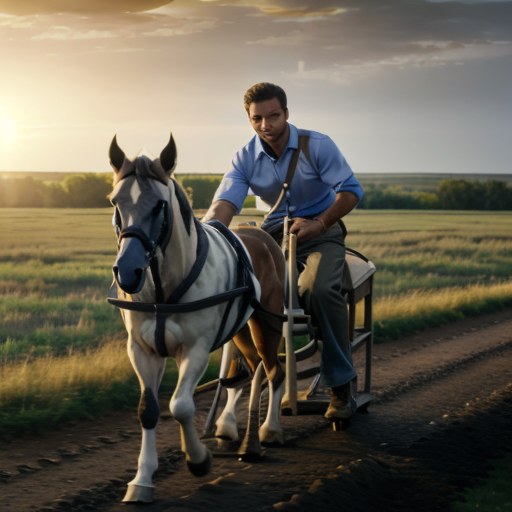} &
        \includegraphics[width=\imgwidthablate]{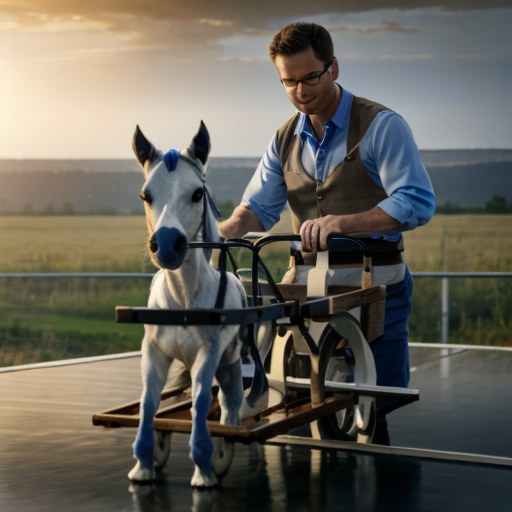} &
        \includegraphics[width=\imgwidthablate]{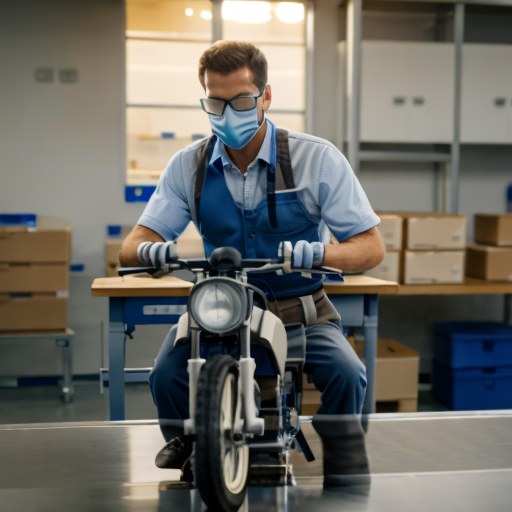} &
        \includegraphics[width=\imgwidthablate]{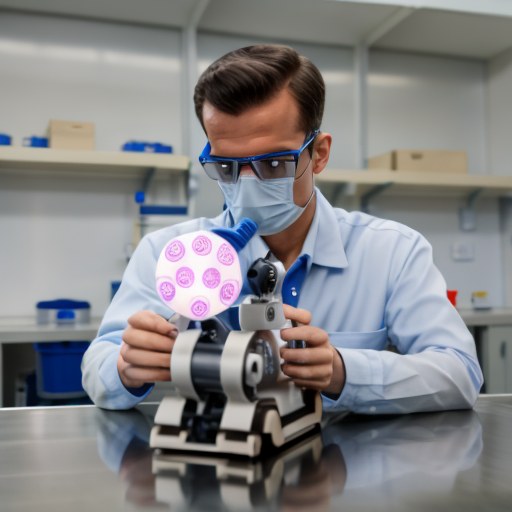} &
        \includegraphics[width=\imgwidthablate]{images/blend_comparison/farmer_scienctist/reference/image2.jpg} \\
        \raisebox{20pt}{\rotatebox[origin=t]{90}{T2T (Ours)}} & { } &
        \includegraphics[width=\imgwidthablate]{images/blend_comparison/farmer_scienctist/reference/image1.jpg} &
        \includegraphics[width=\imgwidthablate]{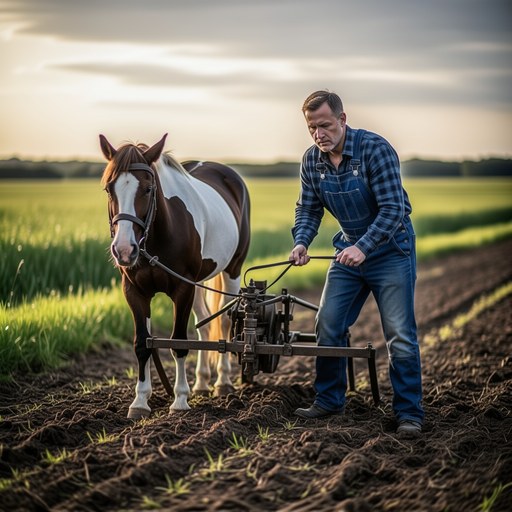} &
        \includegraphics[width=\imgwidthablate]{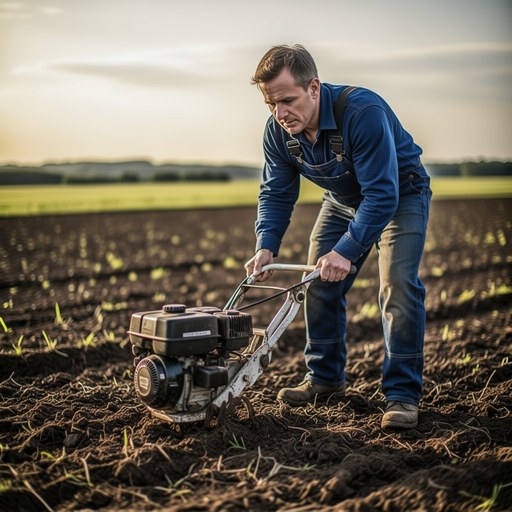} &
        \includegraphics[width=\imgwidthablate]{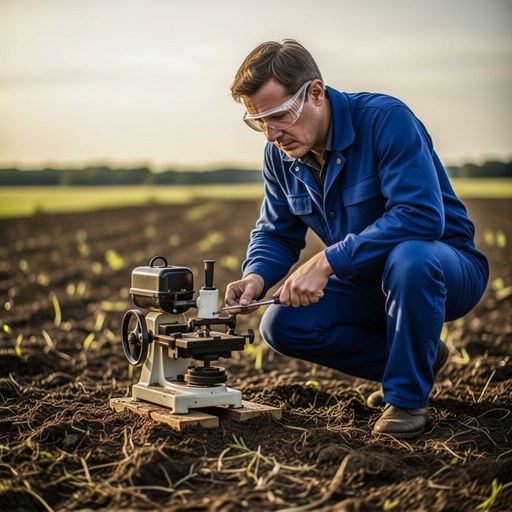} &
        \includegraphics[width=\imgwidthablate]{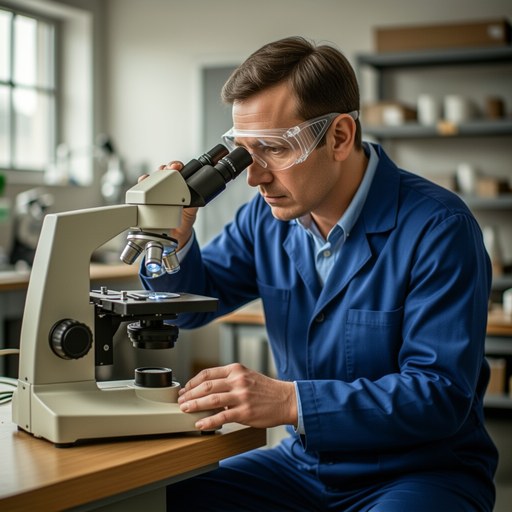} &
        \includegraphics[width=\imgwidthablate]{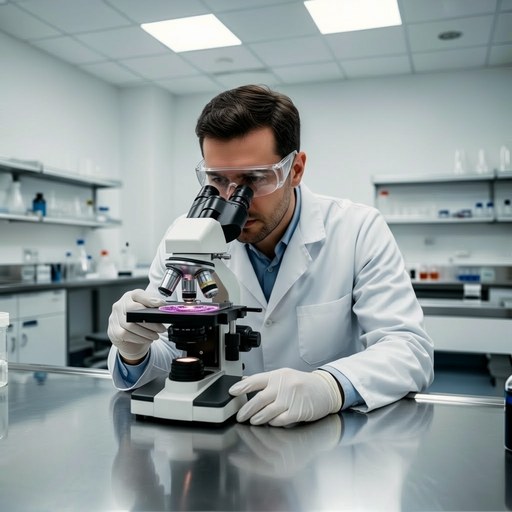} &
        \includegraphics[width=\imgwidthablate]{images/blend_comparison/farmer_scienctist/reference/image2.jpg} \\
        & & Input A & &  & &  & & Input B

    \end{tabular}
    \caption{Qualitative comparison with continuous blending methods. Results generated using FLUX2-Klein.}
\vspace{-10pt}
    \label{fig:supp_blend_continous2}
\end{figure*}
\renewcommand{\imgwidthablate}{0.14\linewidth}

\begin{figure*}
    \centering
    \setlength{\tabcolsep}{1pt}
    \begin{tabular}{ccccccc}
        \includegraphics[width=\imgwidthablate]{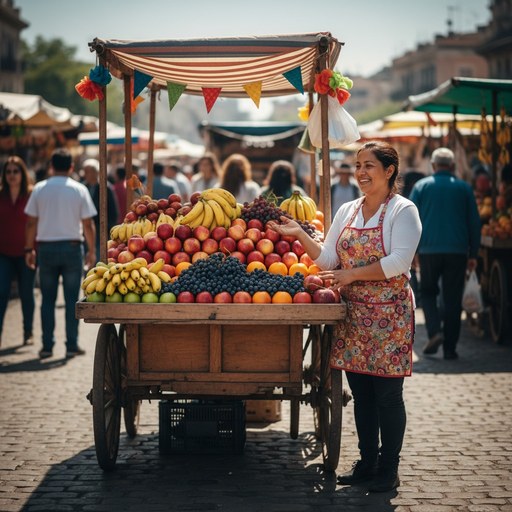} &
        \includegraphics[width=\imgwidthablate]{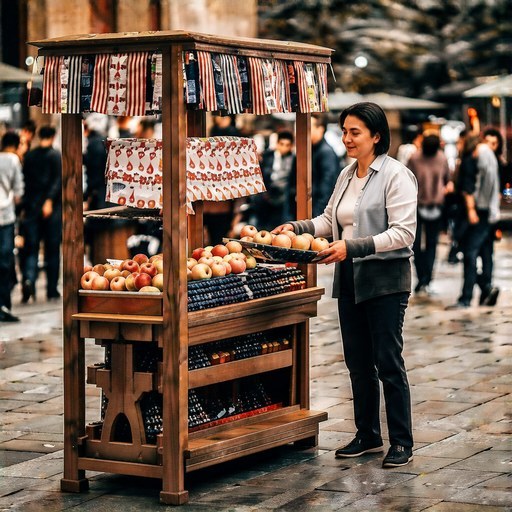} &
        \includegraphics[width=\imgwidthablate]{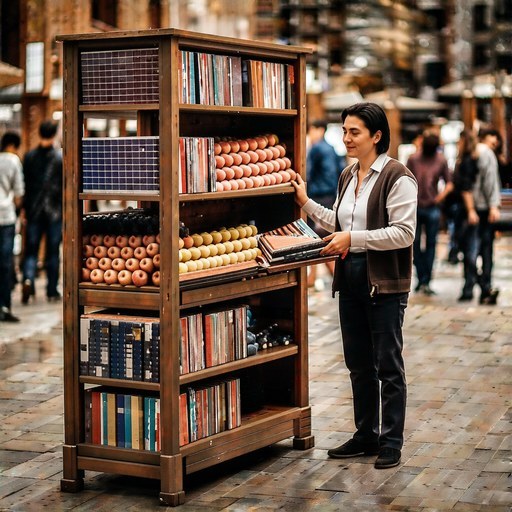} &
        \includegraphics[width=\imgwidthablate]{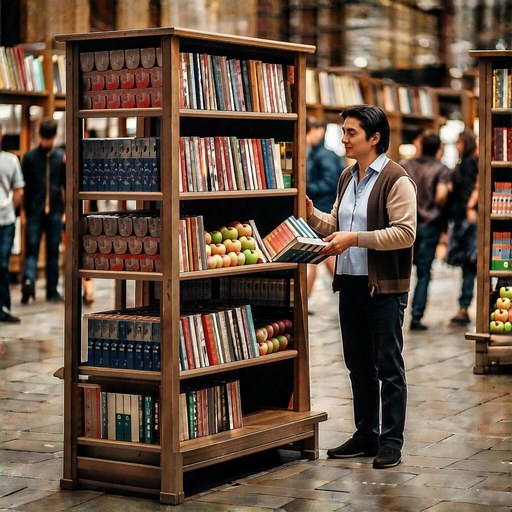} &
        \includegraphics[width=\imgwidthablate]{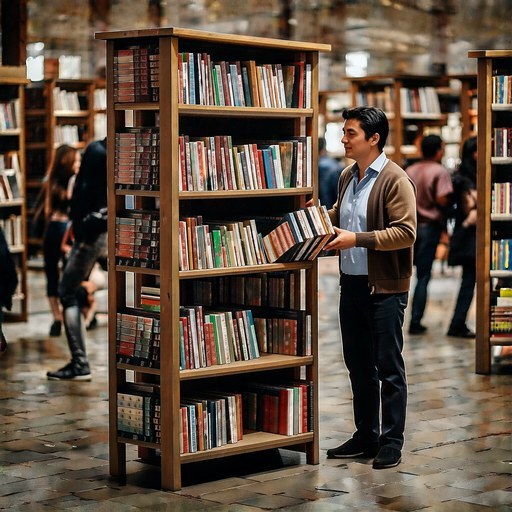} &
        \includegraphics[width=\imgwidthablate]{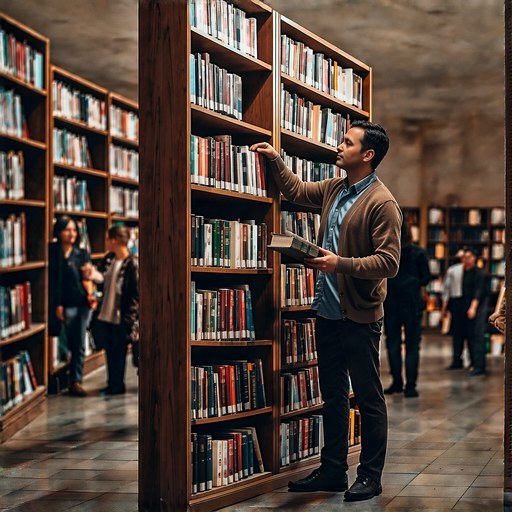} &
        \includegraphics[width=\imgwidthablate]{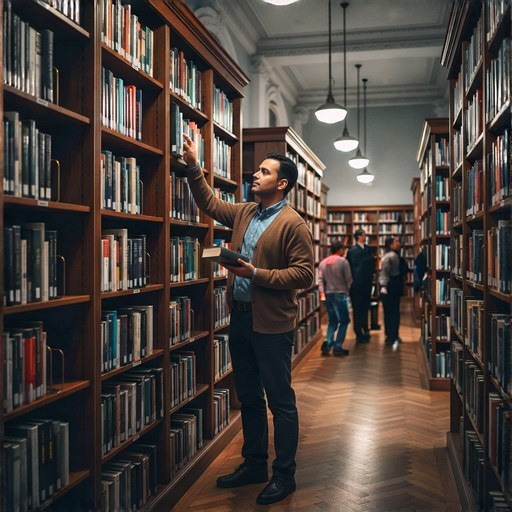} 
        \\
        \includegraphics[width=\imgwidthablate]{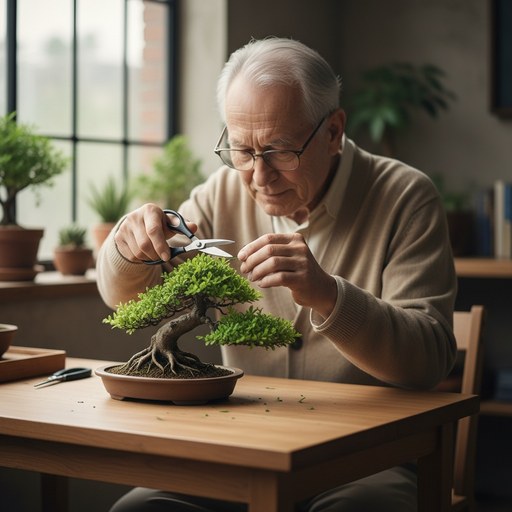} &
        \includegraphics[width=\imgwidthablate]{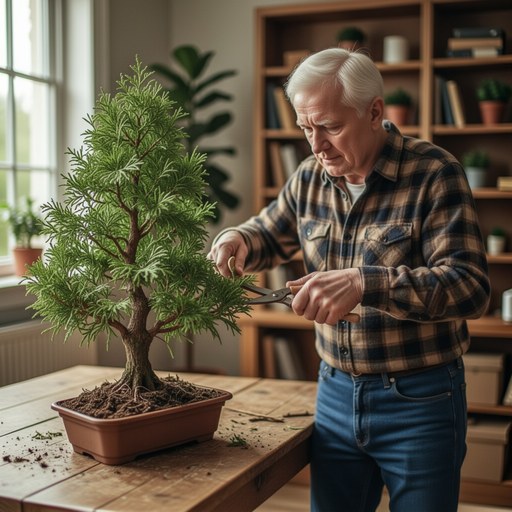} &
        \includegraphics[width=\imgwidthablate]{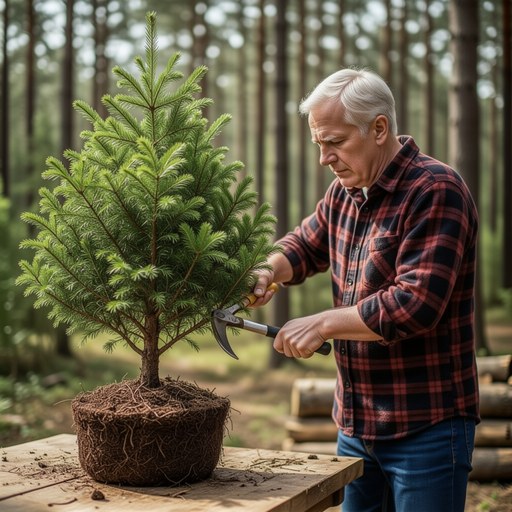} &
        \includegraphics[width=\imgwidthablate]{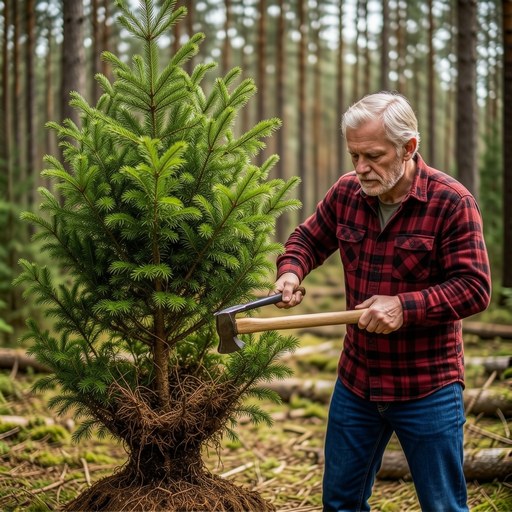} &
        \includegraphics[width=\imgwidthablate]{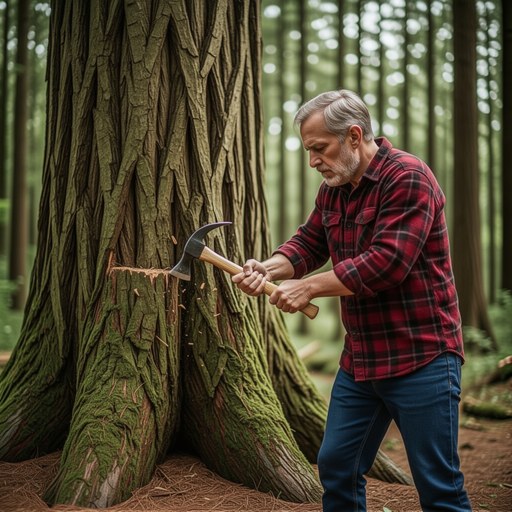} &
        \includegraphics[width=\imgwidthablate]{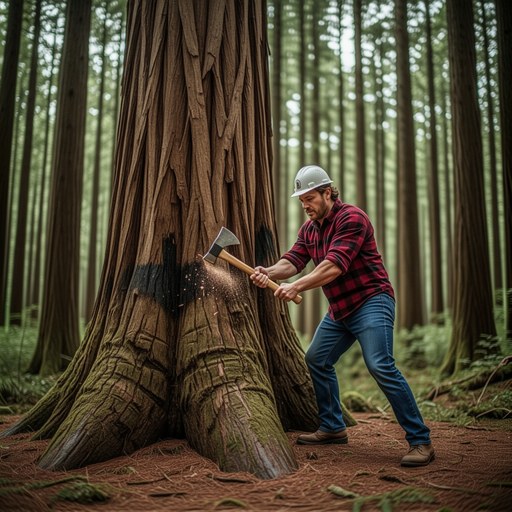} &
        \includegraphics[width=\imgwidthablate]{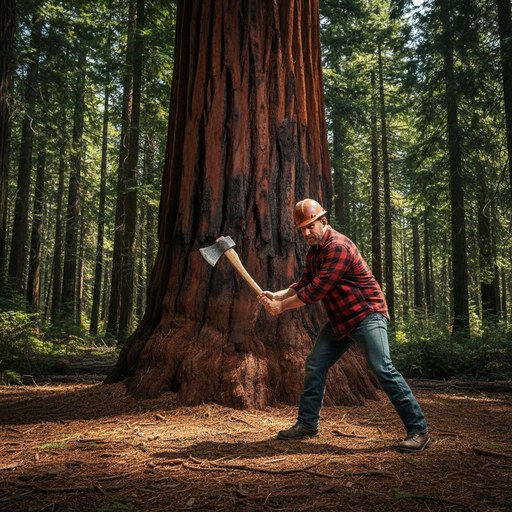}
        \\
        \includegraphics[width=\imgwidthablate]{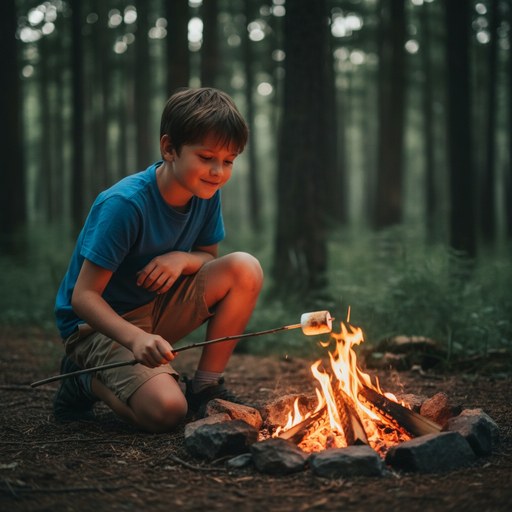} &
        \includegraphics[width=\imgwidthablate]{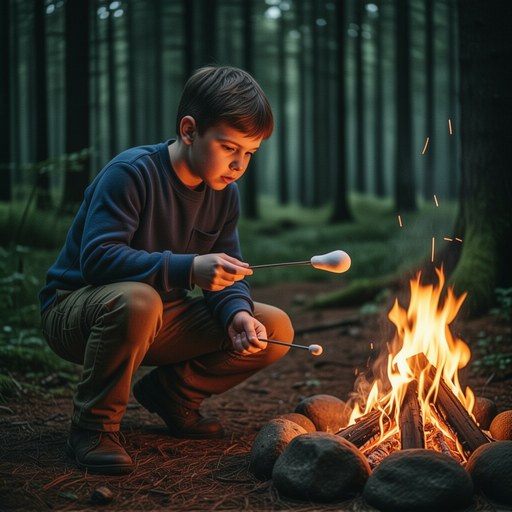} &
        \includegraphics[width=\imgwidthablate]{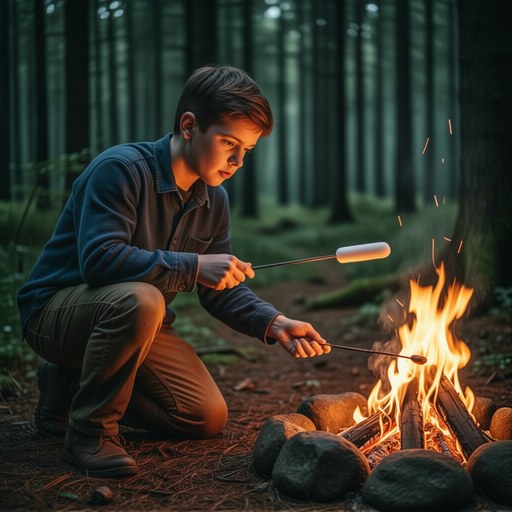} &
        \includegraphics[width=\imgwidthablate]{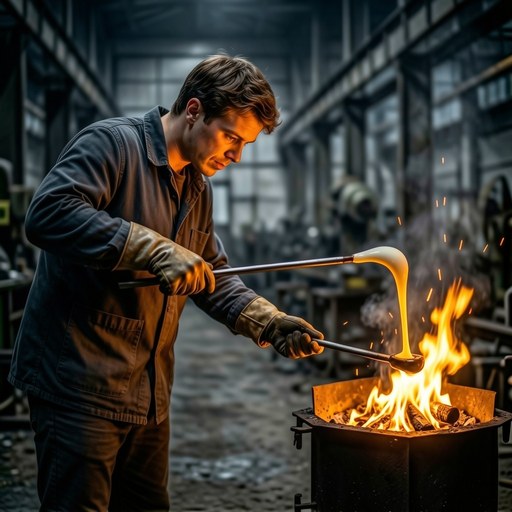} &
        \includegraphics[width=\imgwidthablate]{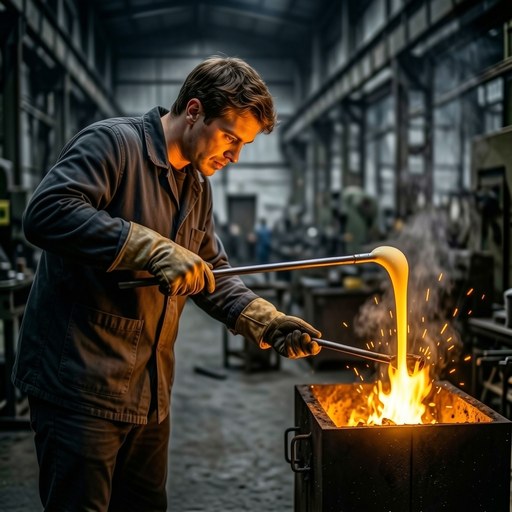} &
        \includegraphics[width=\imgwidthablate]{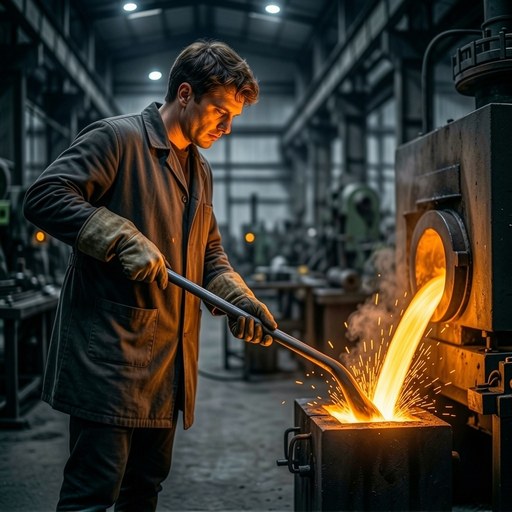} &
        \includegraphics[width=\imgwidthablate]{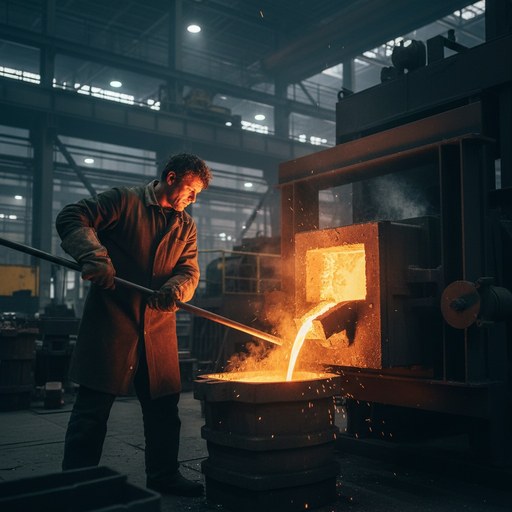} \\

        Input A & &  & &  & & Input B
    \end{tabular}
    \caption{Additional continuous blending results. Each row shows a gradual transition between two input images. Top row: results generated using FIBO-edit. Bottom two rows: results generated using FLUX2-Klein.}
\vspace{-10pt}
    \label{fig:supp_blend_continous}
\end{figure*}
\newcommand{\imgwidthNight}{0.133\linewidth}

\renewcommand{\arraystretch}{0.2}
\begin{figure*}
    \centering
    \setlength{\tabcolsep}{0pt}
    \begin{tabular}{cccc cc cccccc}
        \multicolumn{12}{c}{\textit{``Change the scene to nighttime''}} \\
        \raisebox{20pt}{\rotatebox[origin=t]{90}{{Kontinuous}}} & { } &
        \raisebox{20pt}{\rotatebox[origin=t]{90}{{Kontext}}} & { } &
        \includegraphics[width=\imgwidthNight]{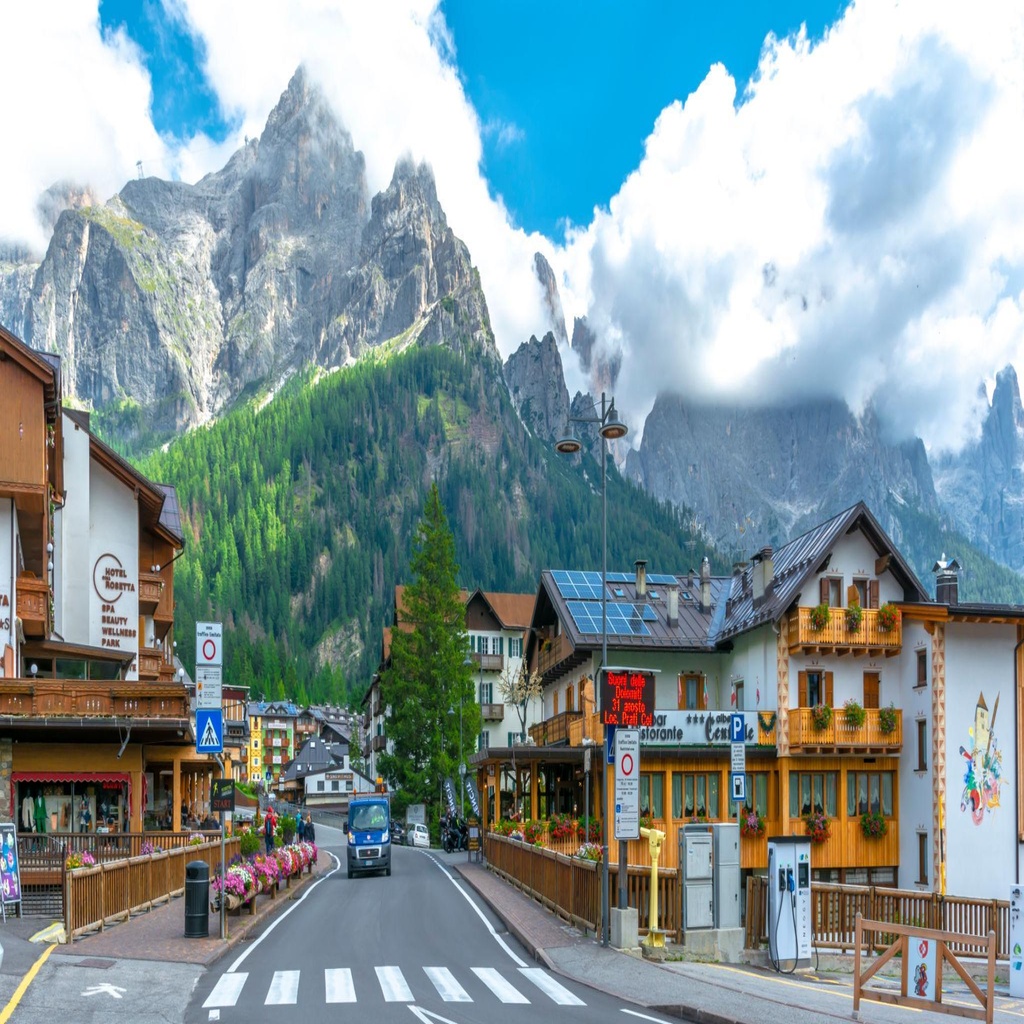} & { } &
        \includegraphics[width=\imgwidthNight]{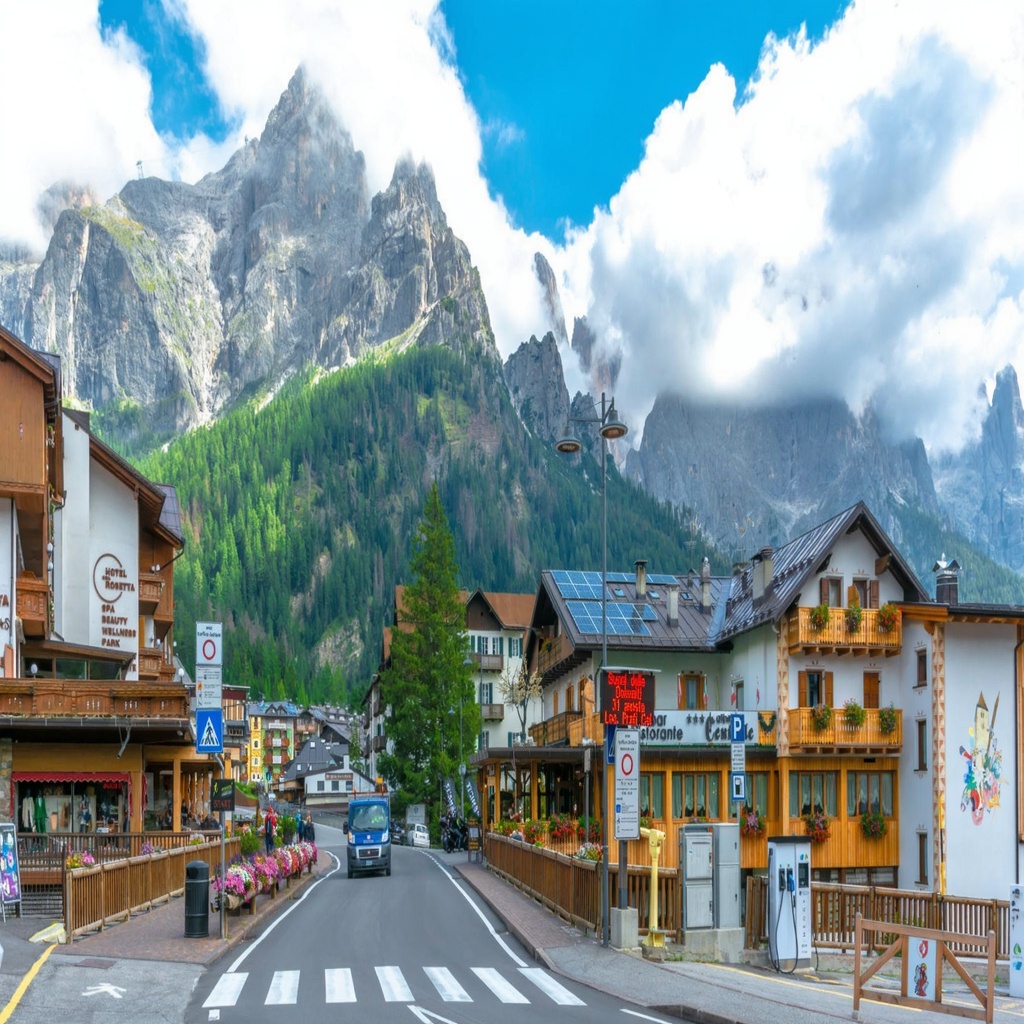} &
        \includegraphics[width=\imgwidthNight]{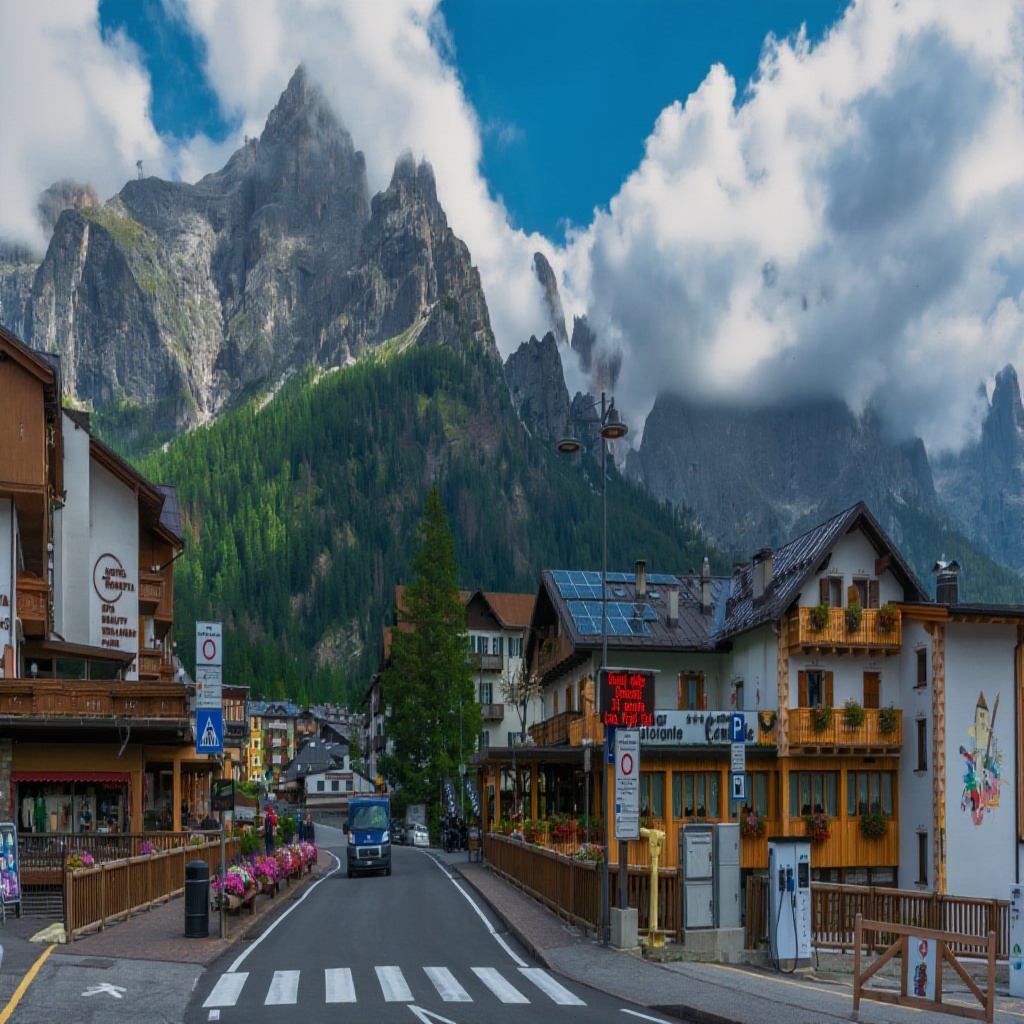} &
        \includegraphics[width=\imgwidthNight]{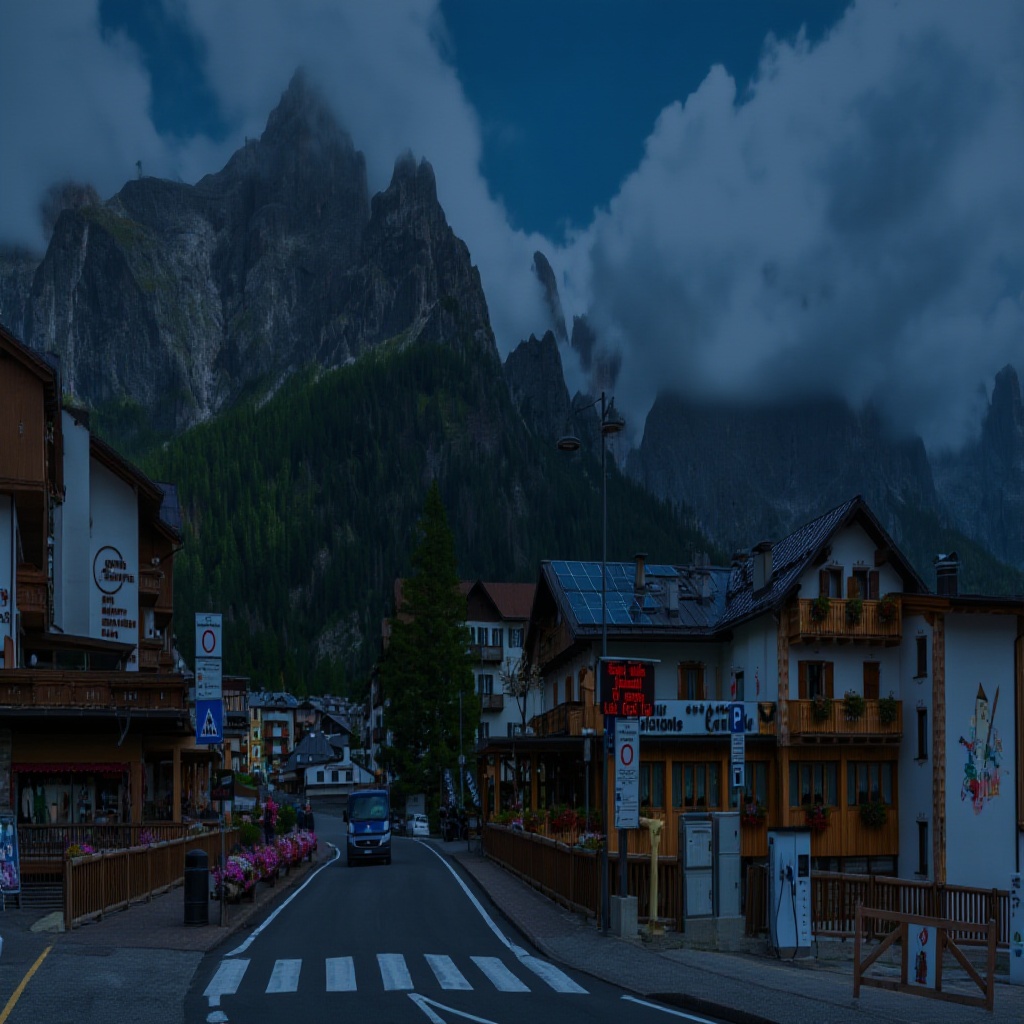} &
        \includegraphics[width=\imgwidthNight]{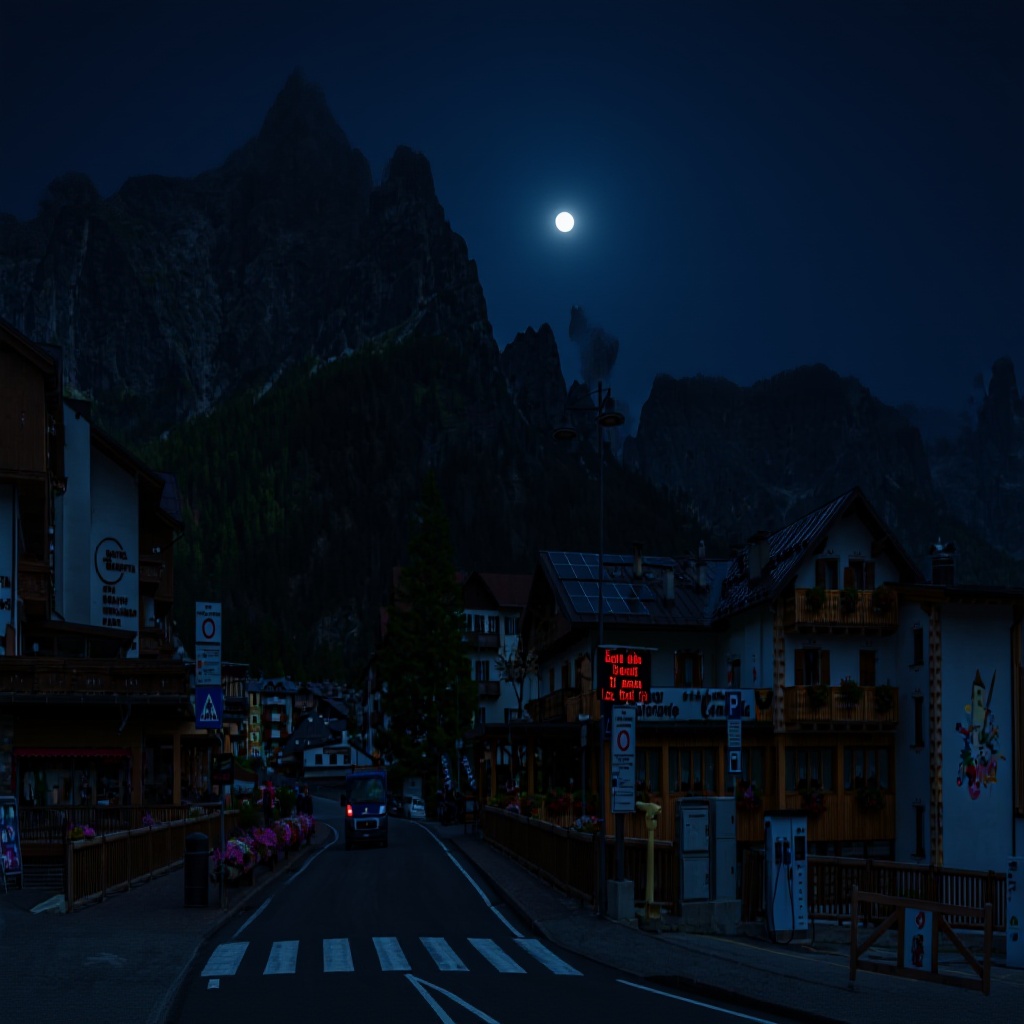} &
        \includegraphics[width=\imgwidthNight]{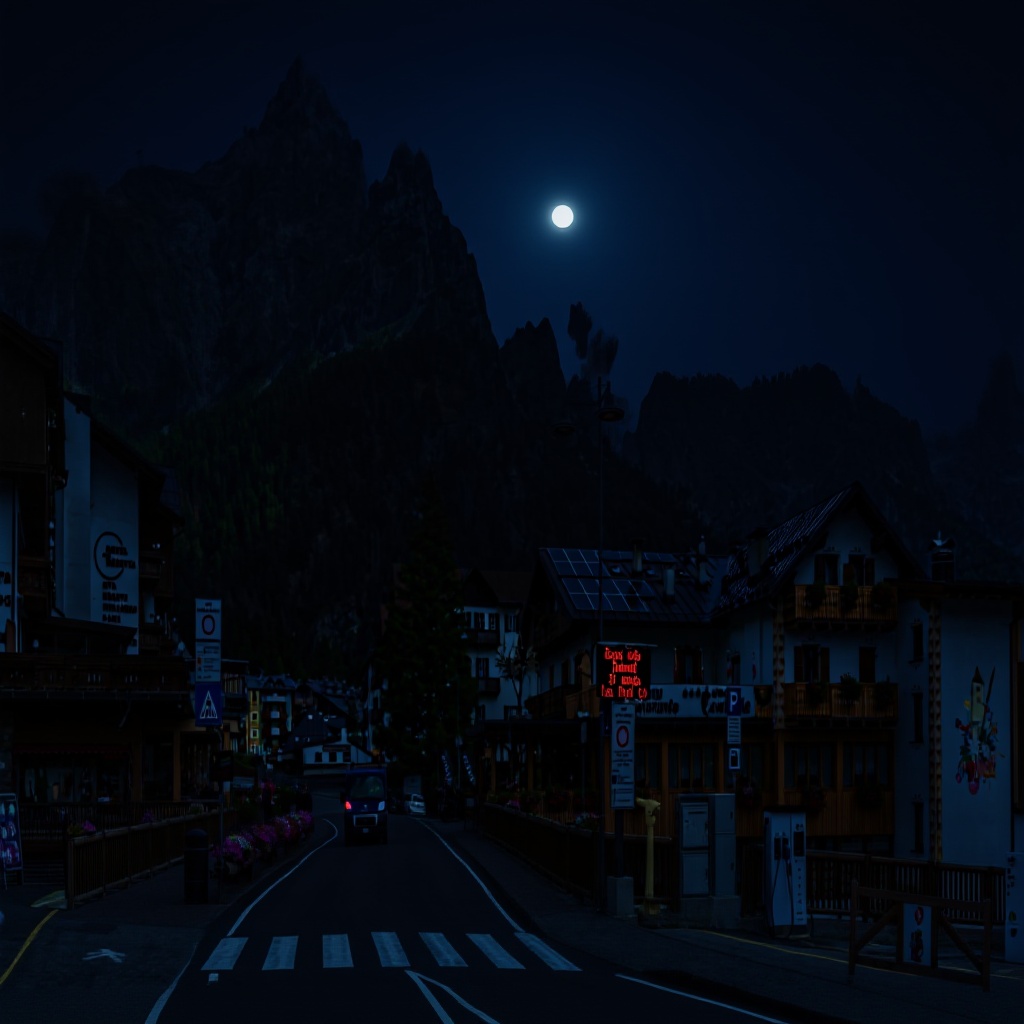} &
        \includegraphics[width=\imgwidthNight]{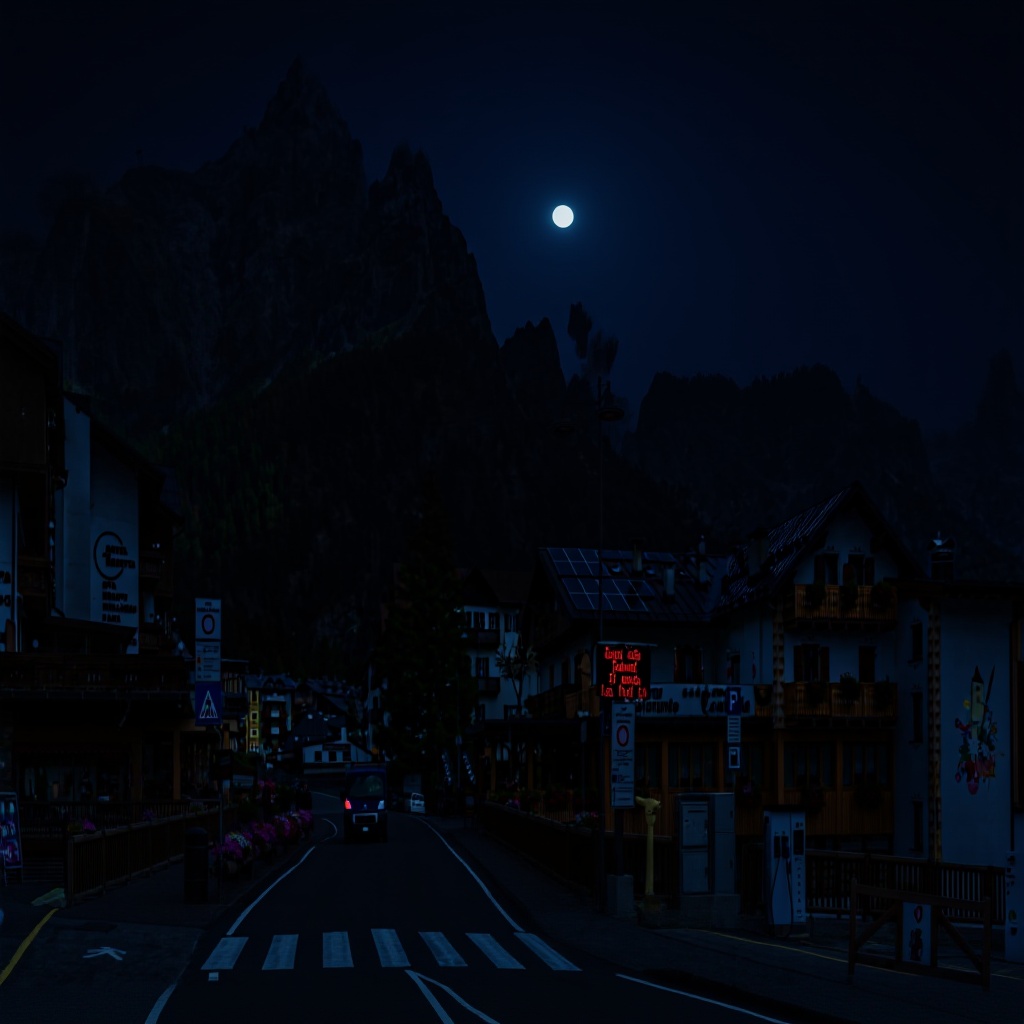} \\
        && \raisebox{20pt}{\rotatebox[origin=t]{90}{{SliderEdit}}} & { } &
        \includegraphics[width=\imgwidthNight]{images/edit_comparison/day_to_night/src.jpg} & { } &
        \includegraphics[width=\imgwidthNight]{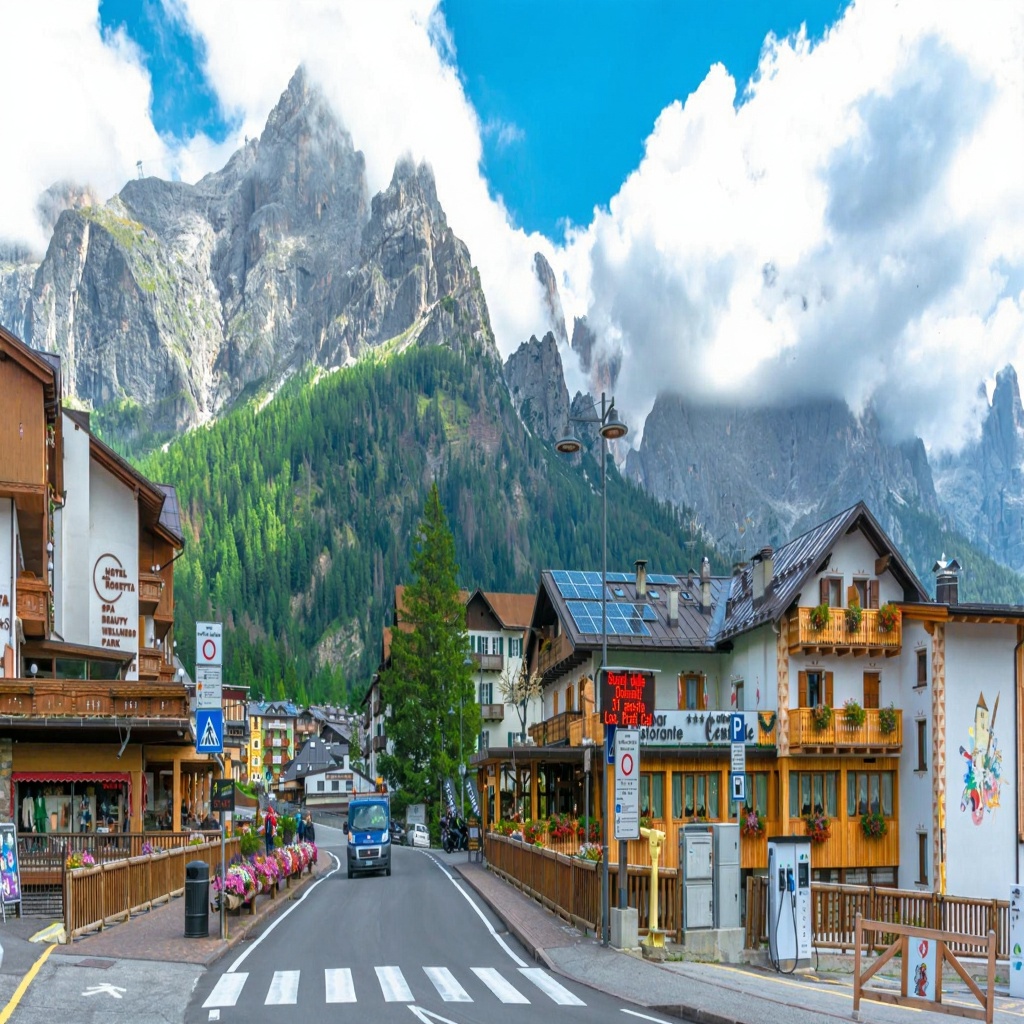} &
        \includegraphics[width=\imgwidthNight]{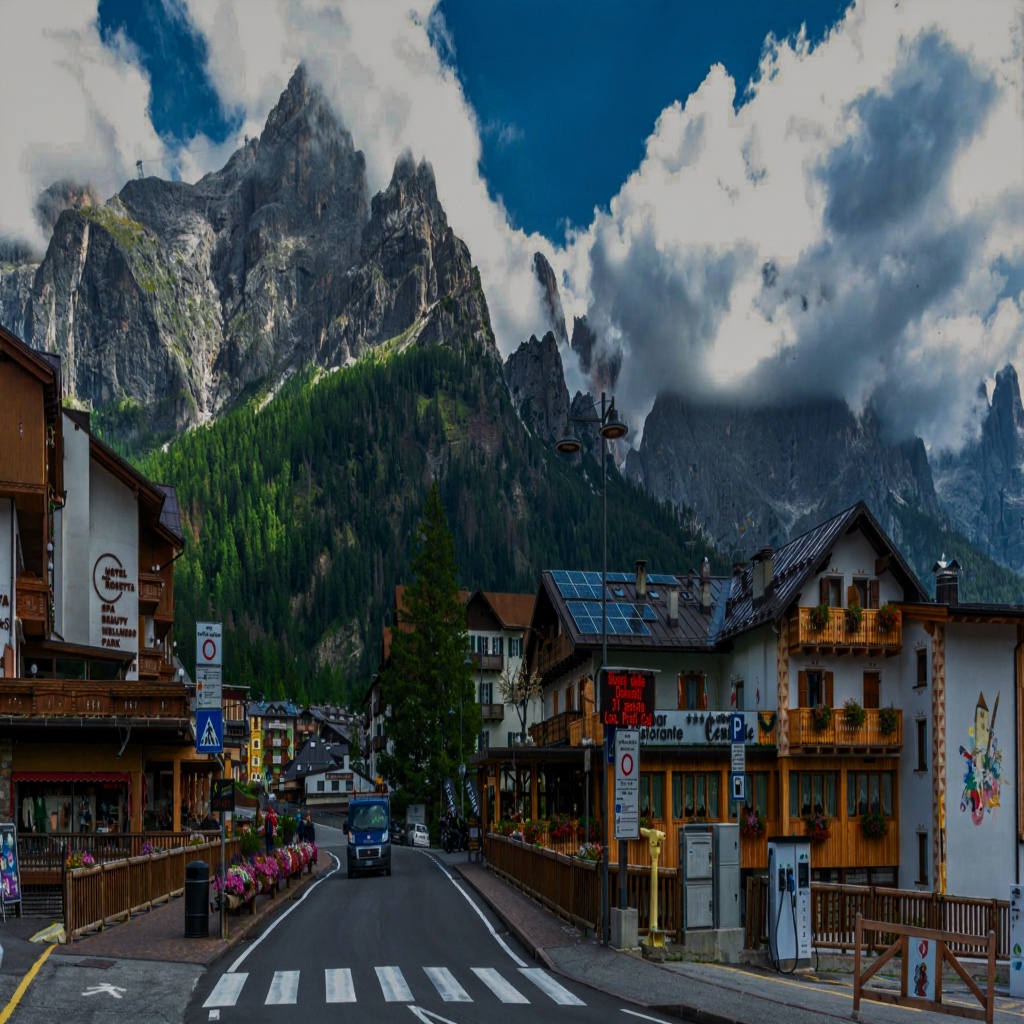} &
        \includegraphics[width=\imgwidthNight]{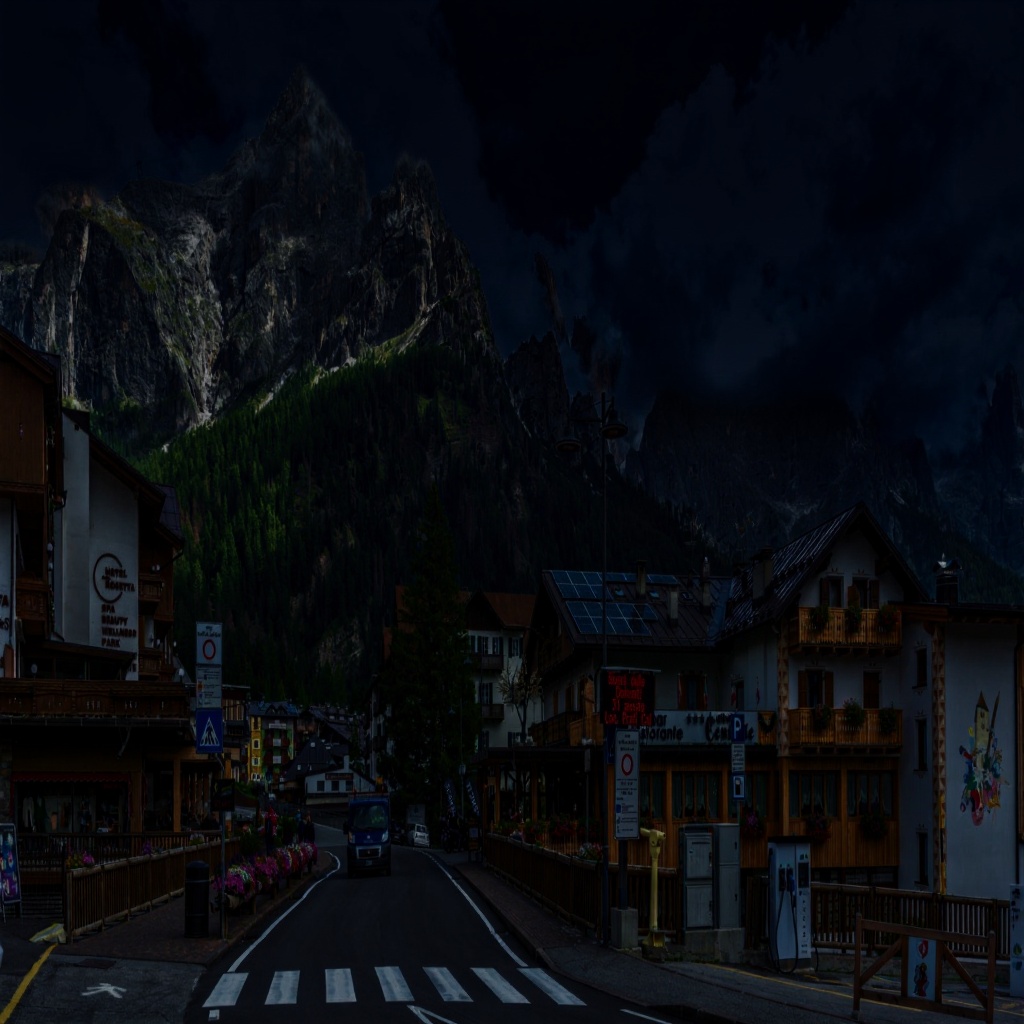} &
        \includegraphics[width=\imgwidthNight]{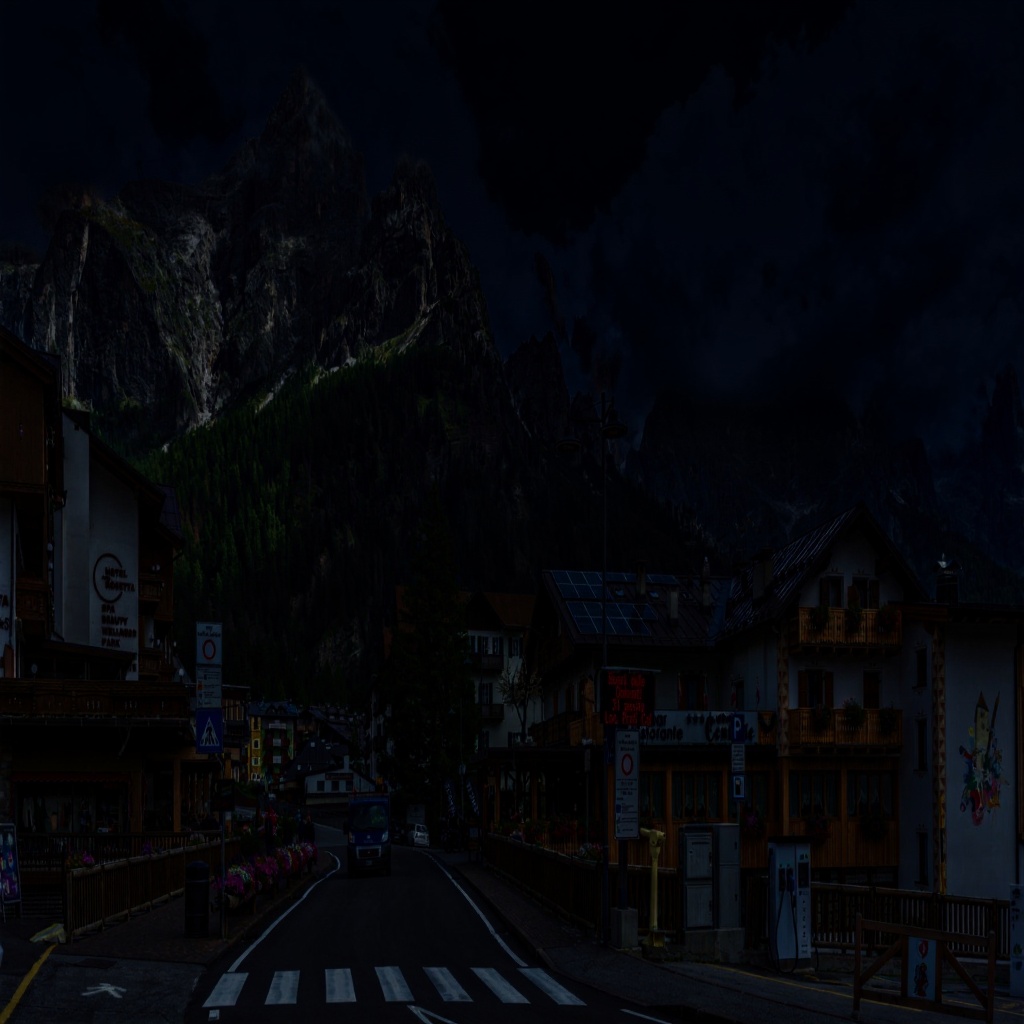} &
        \includegraphics[width=\imgwidthNight]{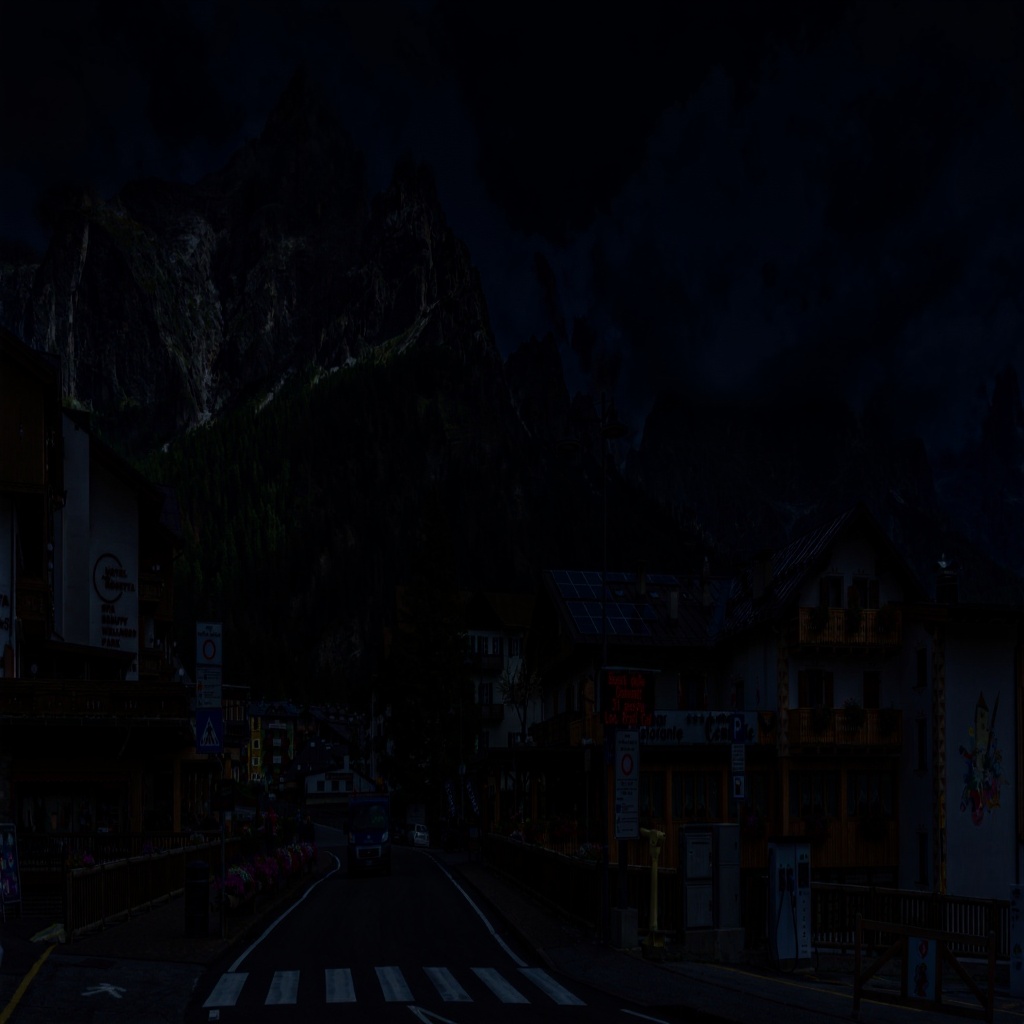} &
        \includegraphics[width=\imgwidthNight]{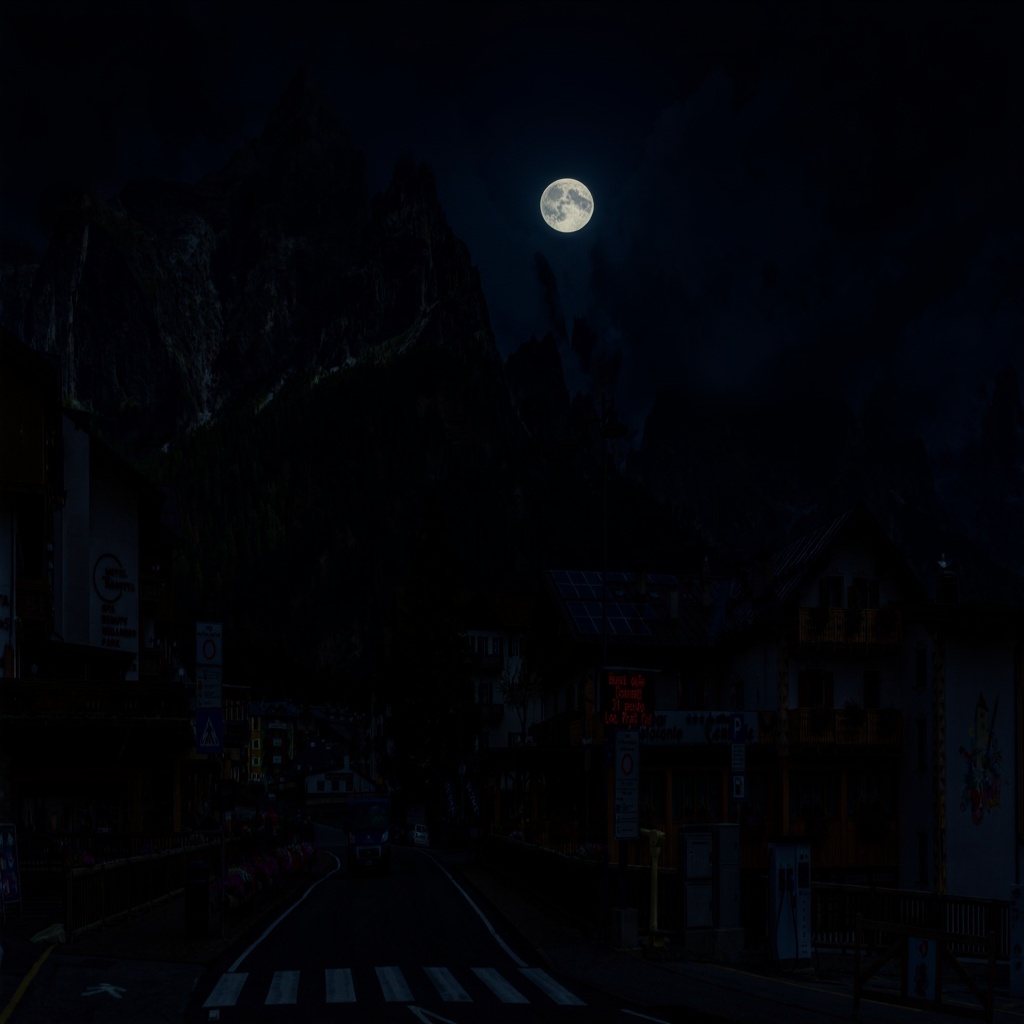} \\
        && \raisebox{20pt}{\rotatebox[origin=t]{90}{{GRAG}}} & { } &
        \includegraphics[width=\imgwidthNight]{images/edit_comparison/day_to_night/src.jpg} & { } &
        \includegraphics[width=\imgwidthNight]{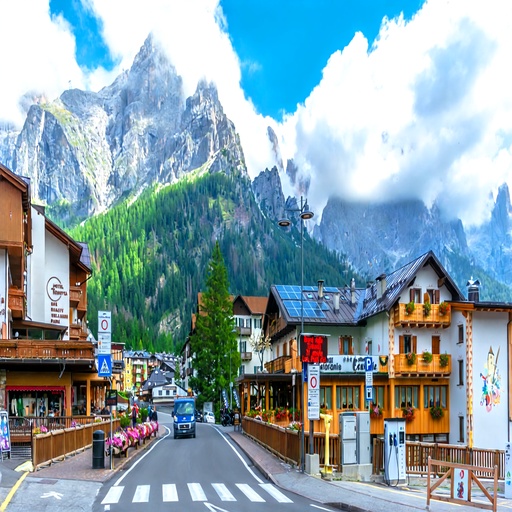} &
        \includegraphics[width=\imgwidthNight]{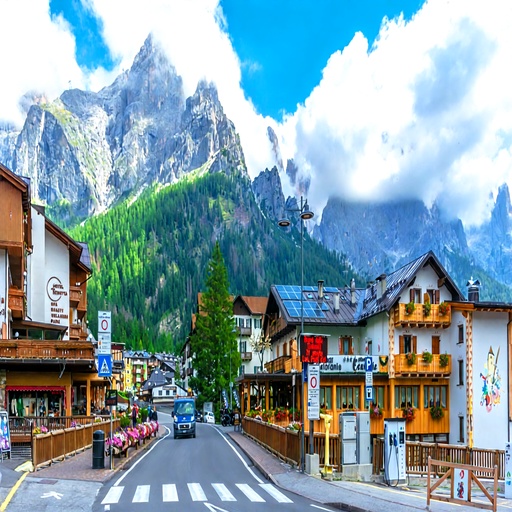} &
        \includegraphics[width=\imgwidthNight]{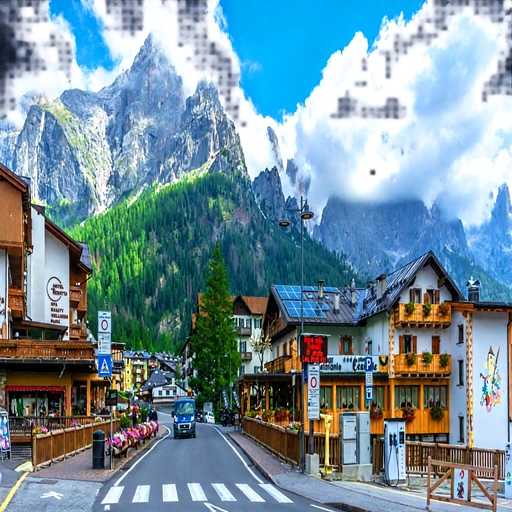} &
        \includegraphics[width=\imgwidthNight]{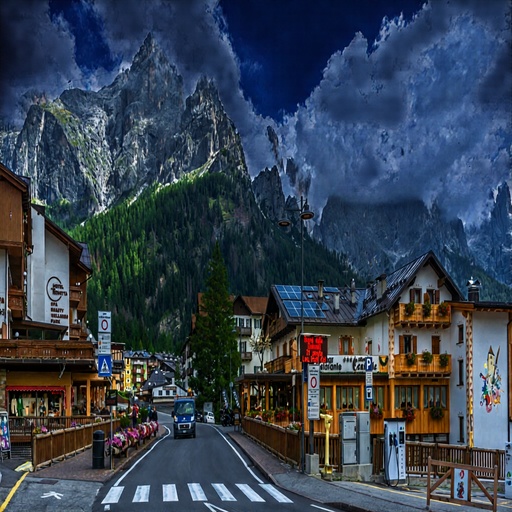} &
        \includegraphics[width=\imgwidthNight]{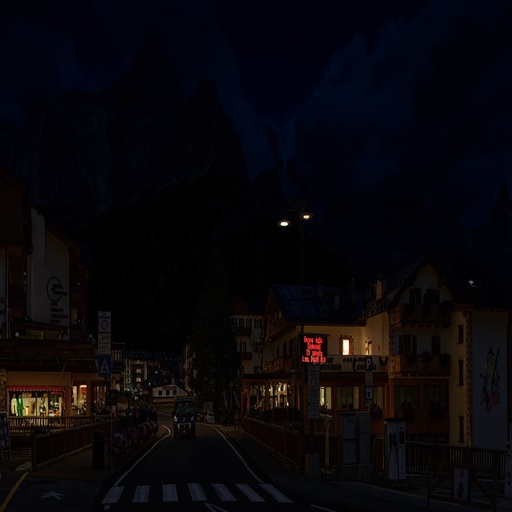} &
        \includegraphics[width=\imgwidthNight]{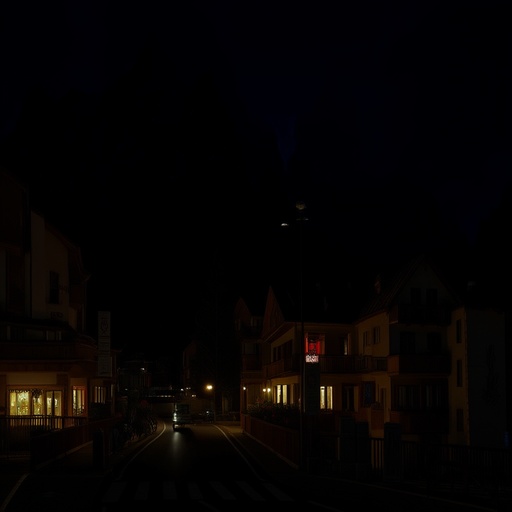} \\
        && \raisebox{20pt}{\rotatebox[origin=t]{90}{T2T (Ours)}} & { } &
        \includegraphics[width=\imgwidthNight]{images/edit_comparison/day_to_night/src.jpg} & { } &
        \includegraphics[width=\imgwidthNight]{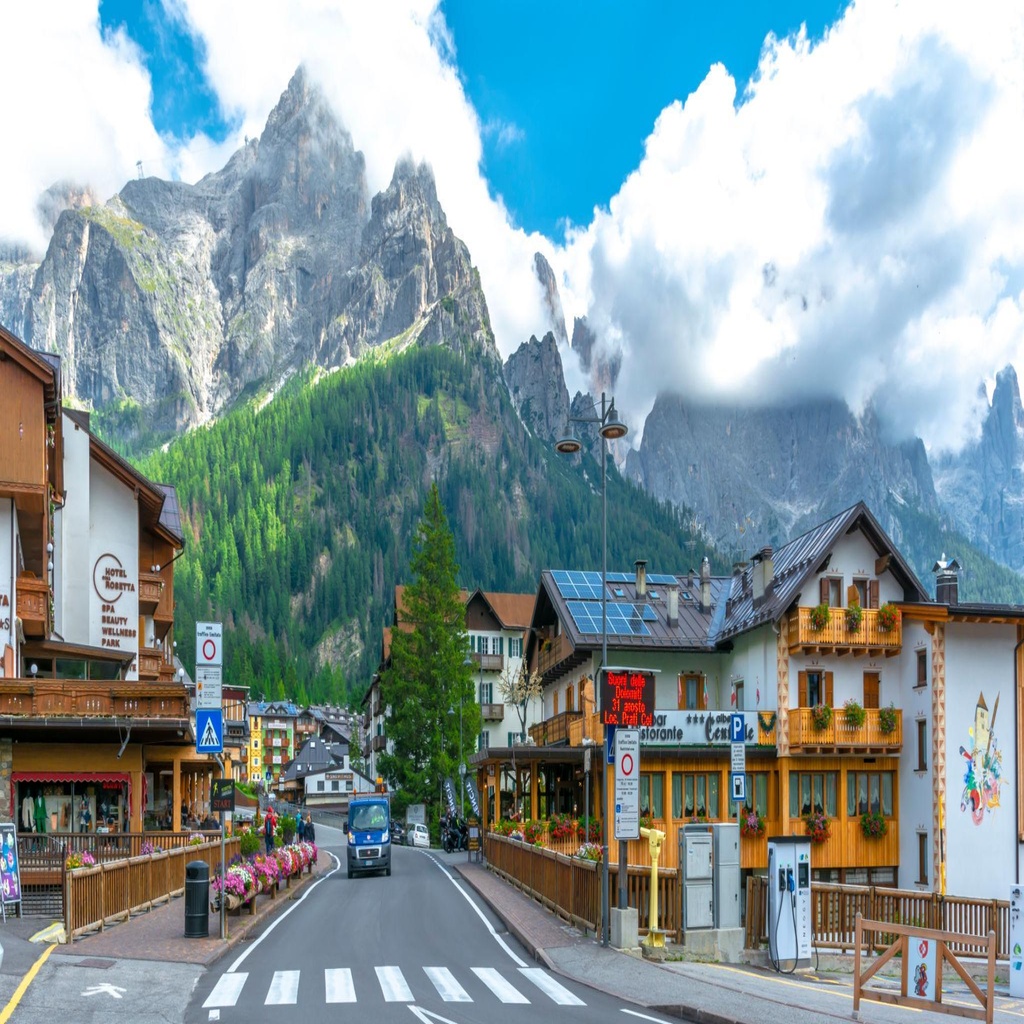} &
        \includegraphics[width=\imgwidthNight]{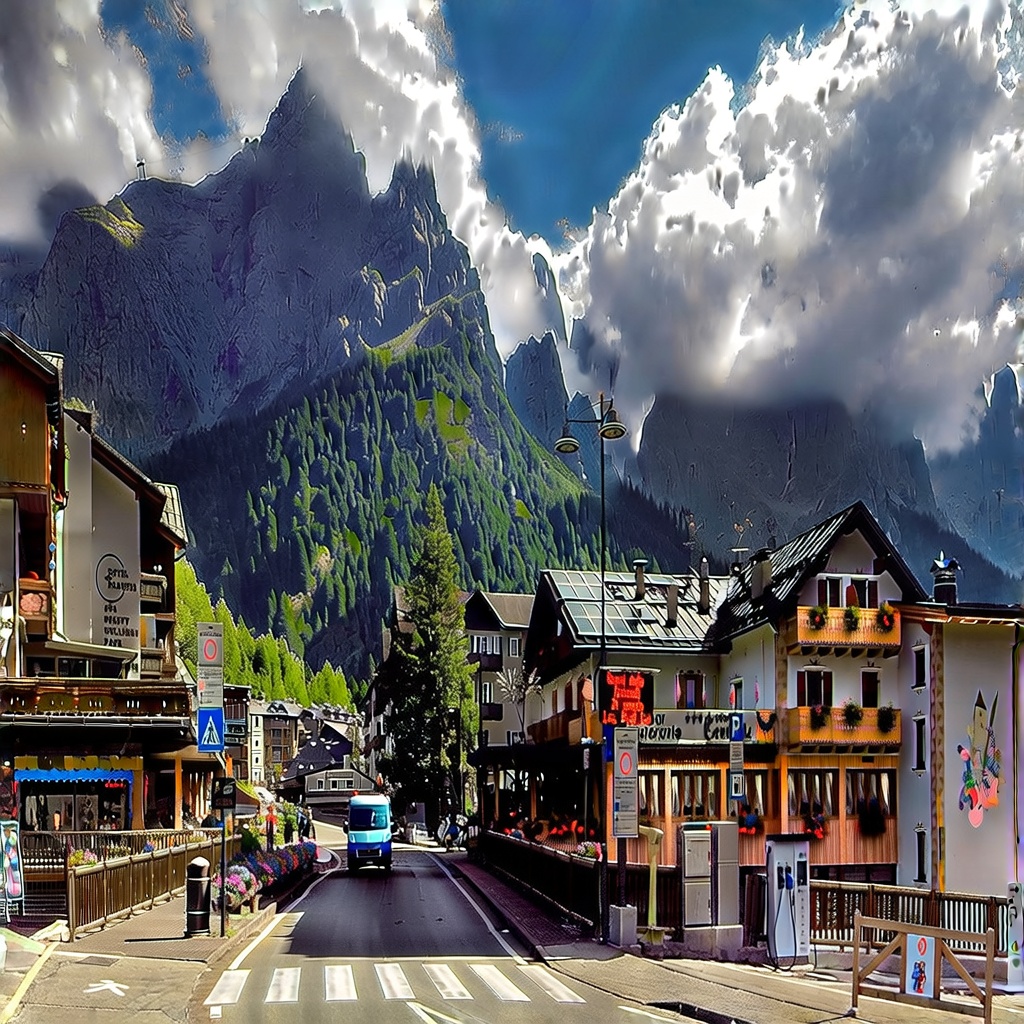} &
        \includegraphics[width=\imgwidthNight]{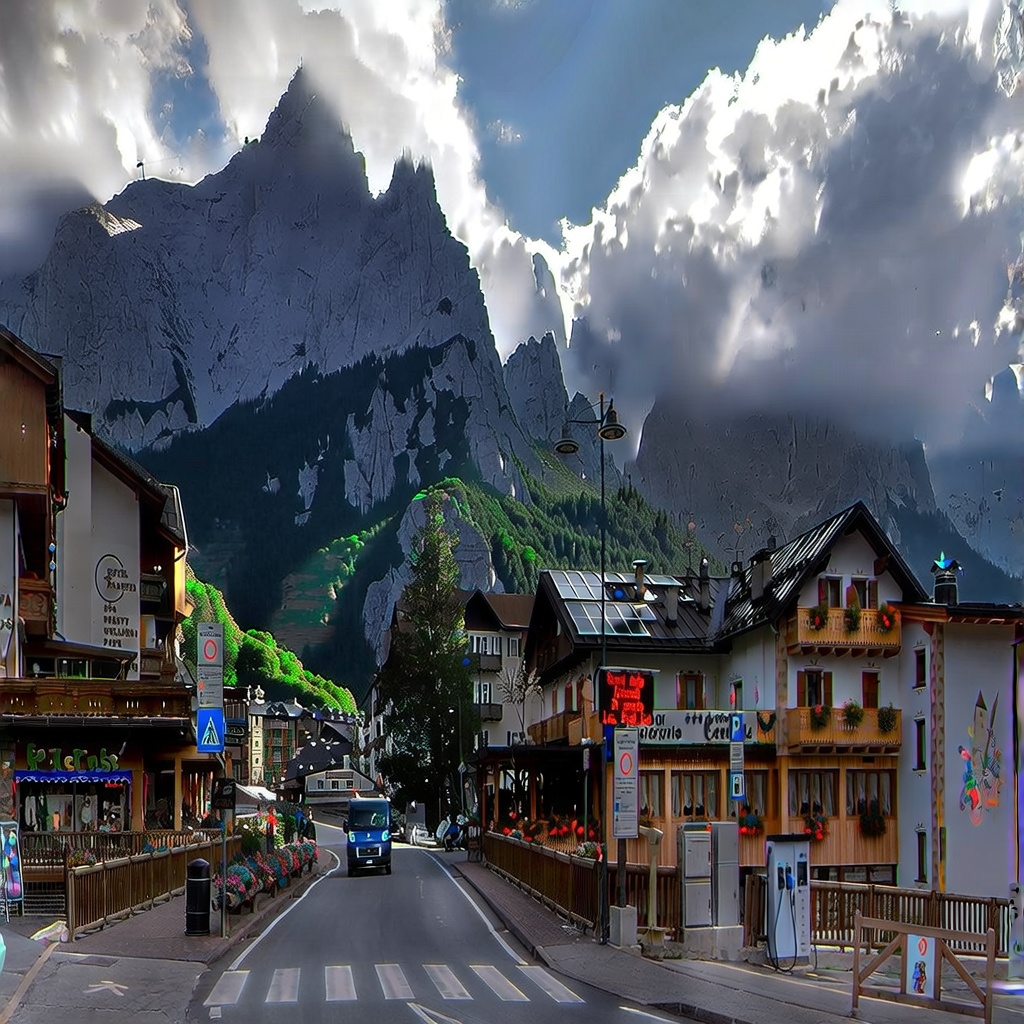} &
        \includegraphics[width=\imgwidthNight]{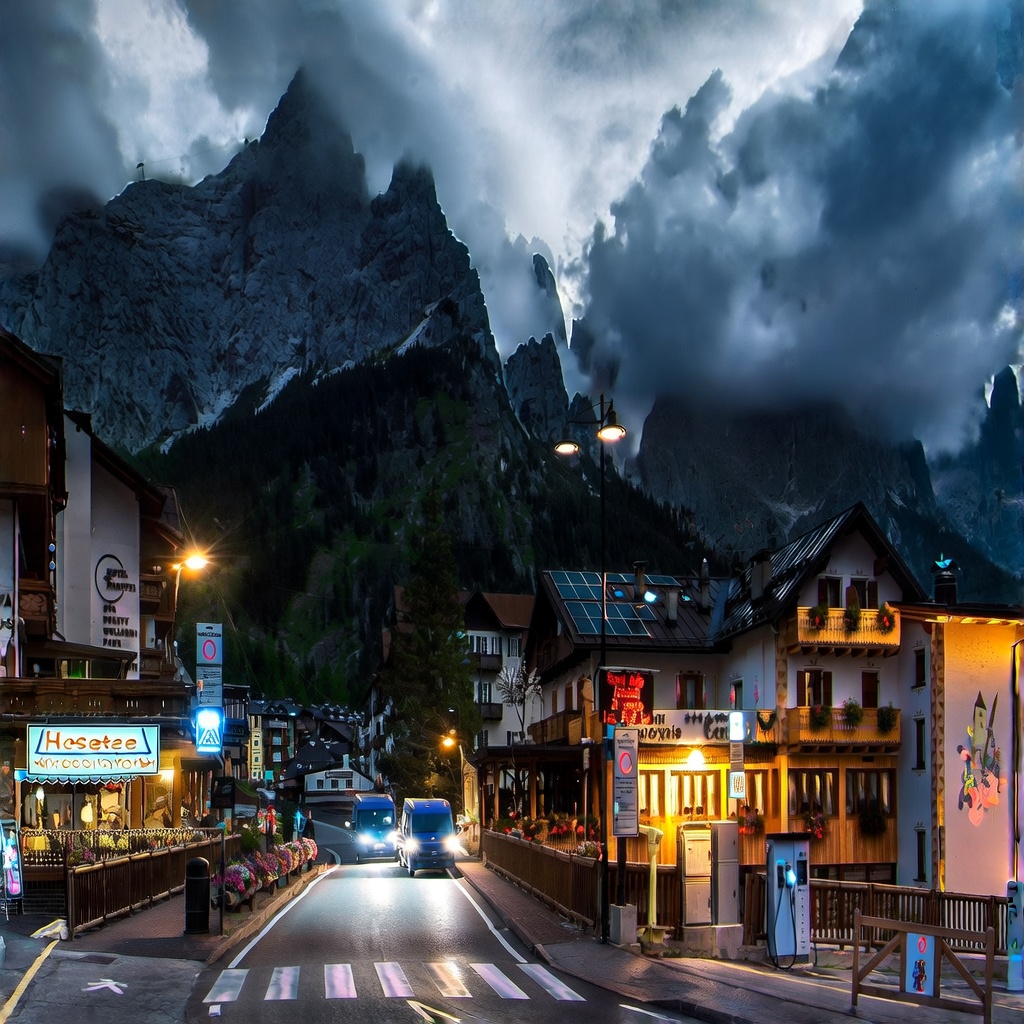} &
        \includegraphics[width=\imgwidthNight]{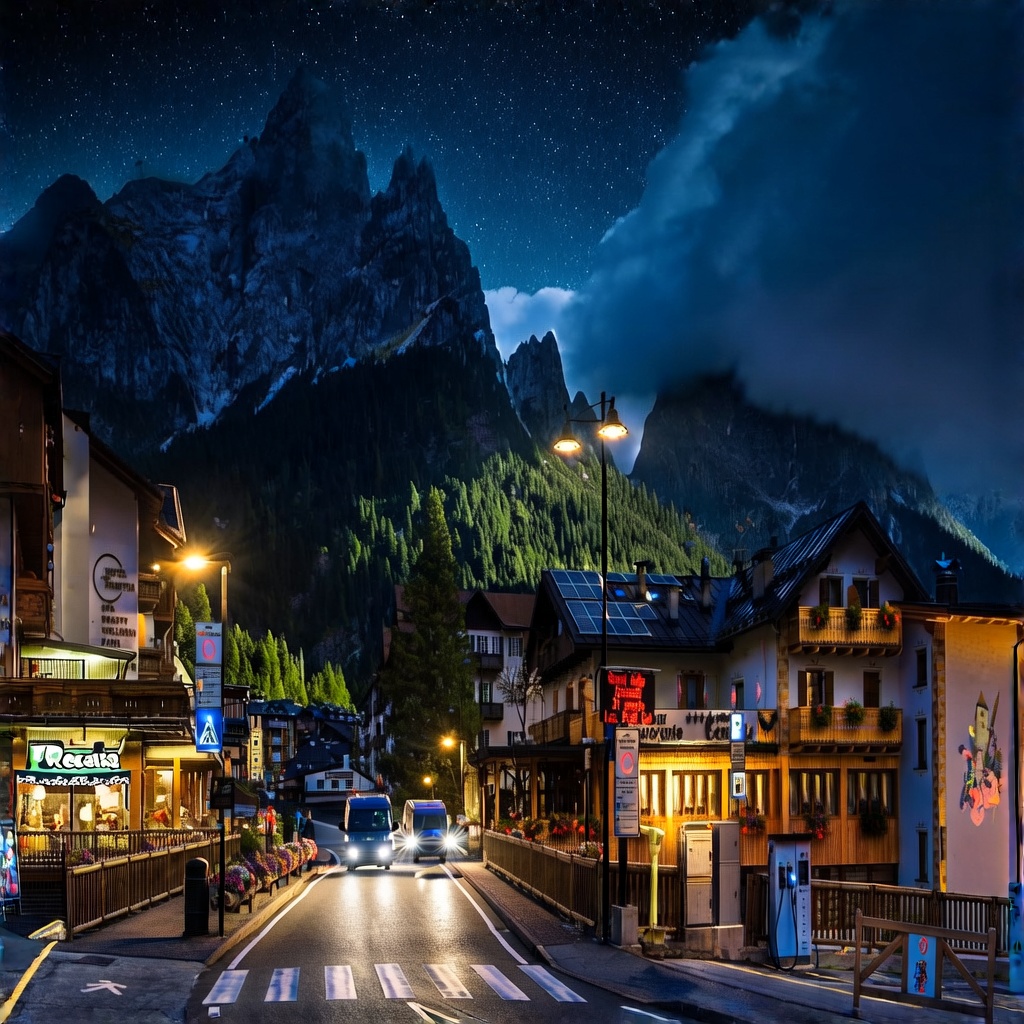} &
        \includegraphics[width=\imgwidthNight]{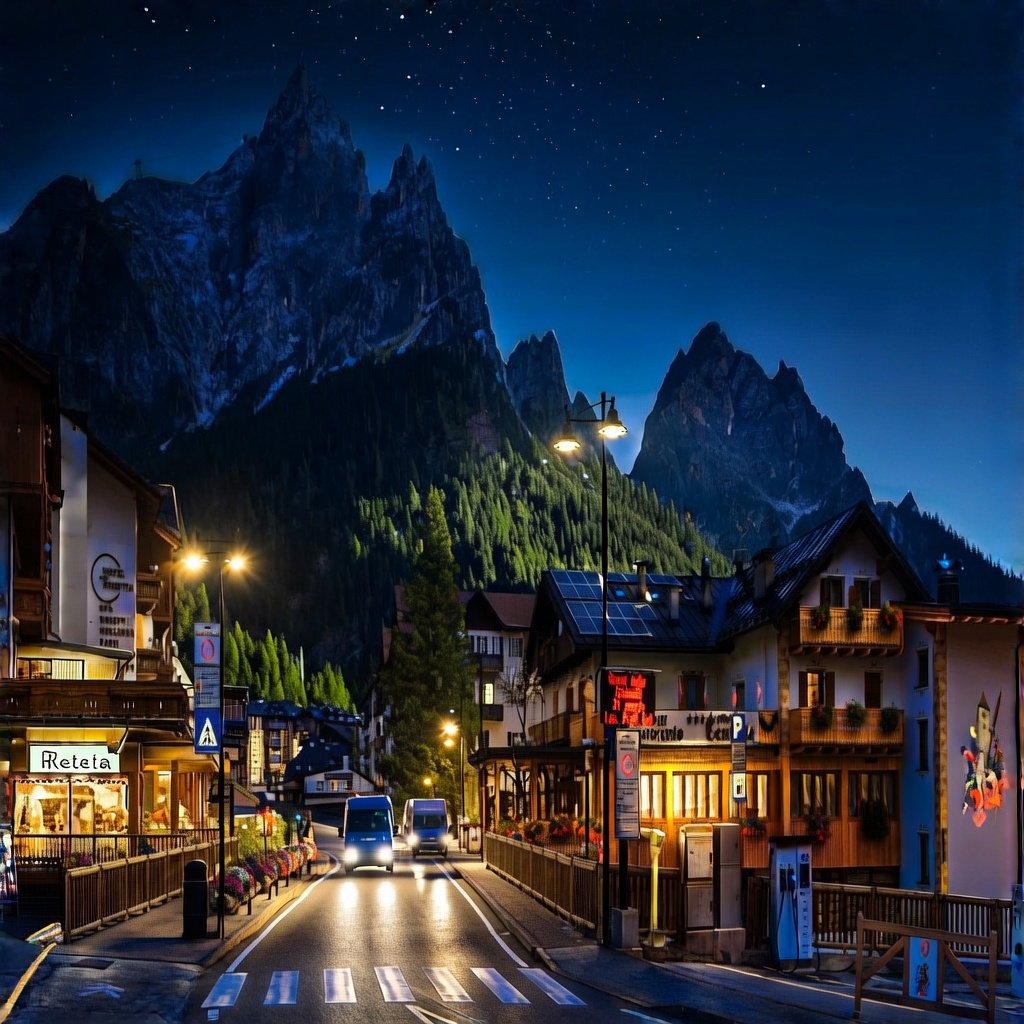} \\
        &&&& Input && 0.0 & 0.2 & 0.4 & 0.6 & 0.8 & 1.0
    \end{tabular}
    \caption{\textbf{Qualitative comparison with continuous editing methods}. For the instruction “change the scene to nighttime,” prior approaches mainly rely on appearance-based transitions such as progressive darkening, whereas our method produces a more coherent semantic transition with gradual changes in illumination and shadow placement consistent with the changing time of day.}

\vspace{-10pt}
    \label{fig:edit_comparison_supp}
\end{figure*}

\renewcommand{\arraystretch}{1}
\renewcommand{\imgwidthNight}{0.133\linewidth}

\begin{figure*}
    \centering
    \setlength{\tabcolsep}{0pt}
    \begin{tabular}{cccc cc cccccc}

        \multicolumn{12}{c}{\textit{``Change the cat to a dog and the flower to a yellow ball''}} \\
        \raisebox{20pt}{\rotatebox[origin=t]{90}{{Kontinuous}}} & { } &
        \raisebox{20pt}{\rotatebox[origin=t]{90}{{Kontext}}} & { } &
        \includegraphics[width=\imgwidthNight]{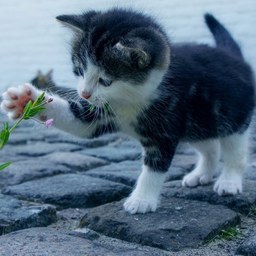} & { } &
        \includegraphics[width=\imgwidthNight]{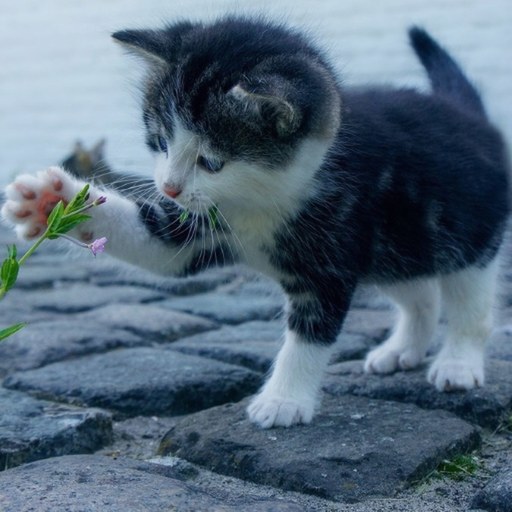} &
        \includegraphics[width=\imgwidthNight]{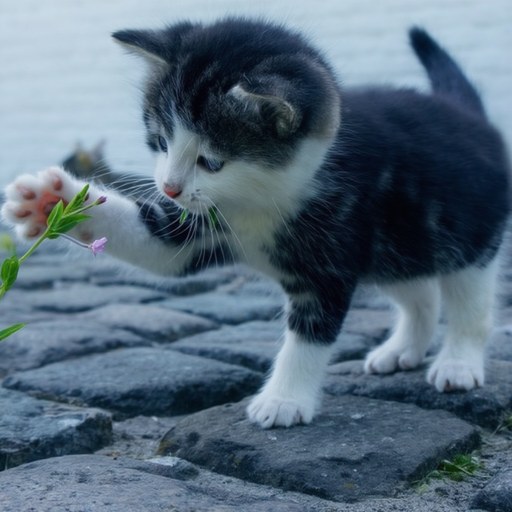} &
        \includegraphics[width=\imgwidthNight]{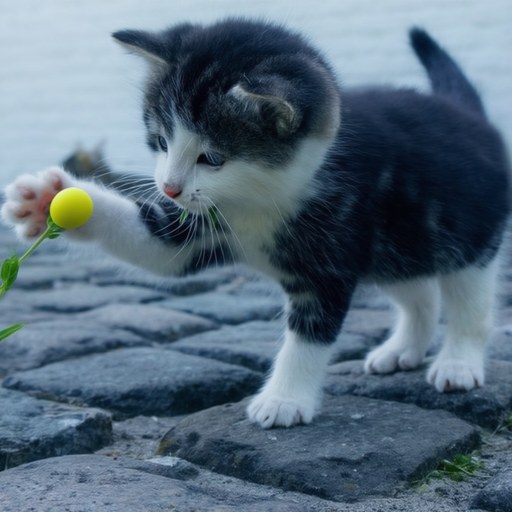} &
        \includegraphics[width=\imgwidthNight]{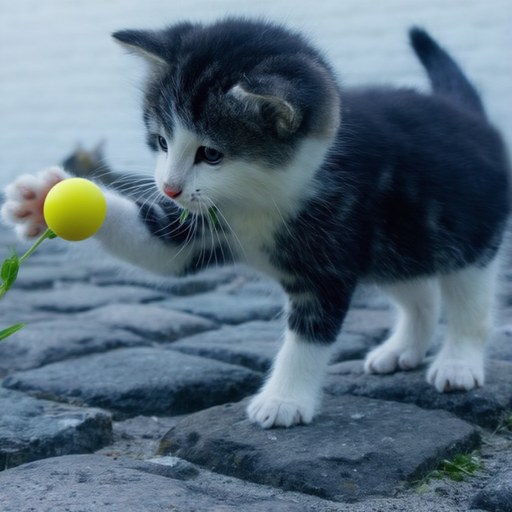} &
        \includegraphics[width=\imgwidthNight]{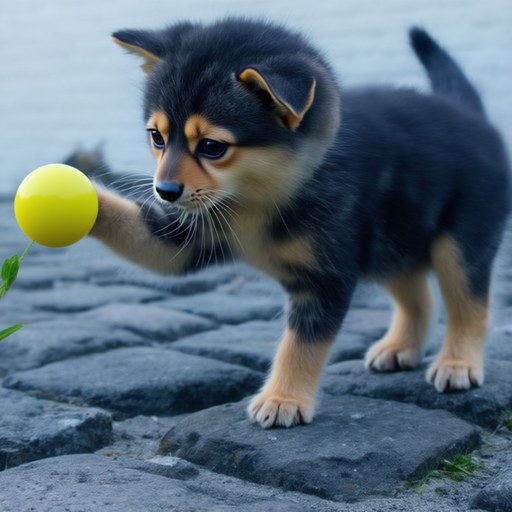} &
        \includegraphics[width=\imgwidthNight]{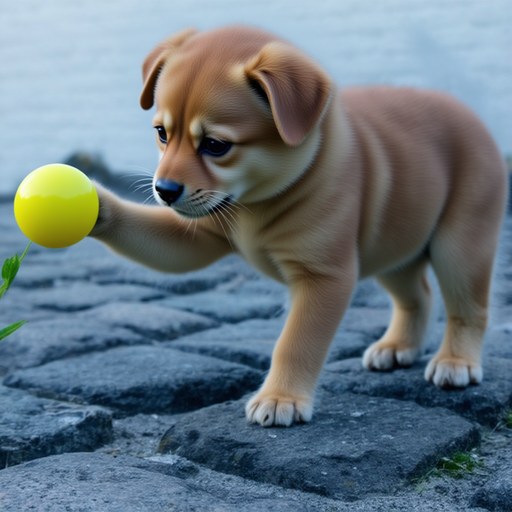} \\
        && \raisebox{20pt}{\rotatebox[origin=t]{90}{{SliderEdit}}} & { } &
        \includegraphics[width=\imgwidthNight]{images/edit_comparison/flower_ball_cat_dog/src.jpg} & { } &
        \includegraphics[width=\imgwidthNight]{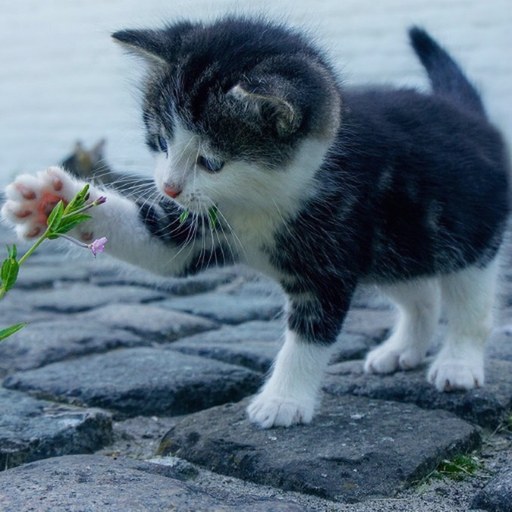} &
        \includegraphics[width=\imgwidthNight]{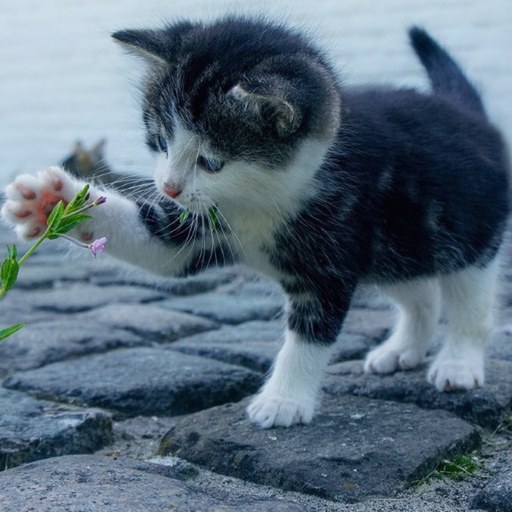} &
        \includegraphics[width=\imgwidthNight]{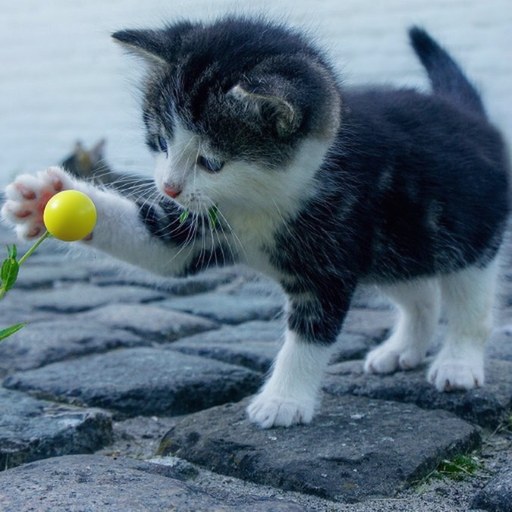} &
        \includegraphics[width=\imgwidthNight]{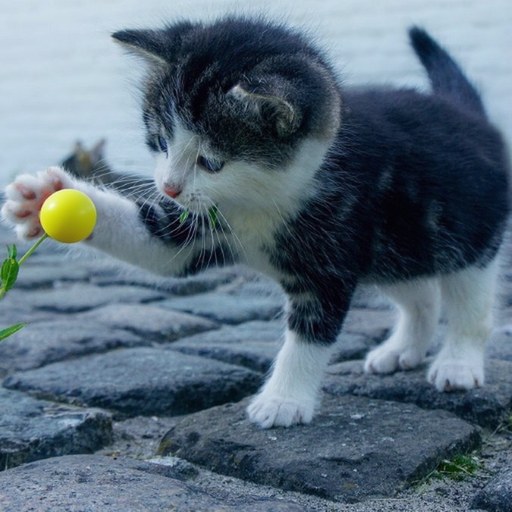} &
        \includegraphics[width=\imgwidthNight]{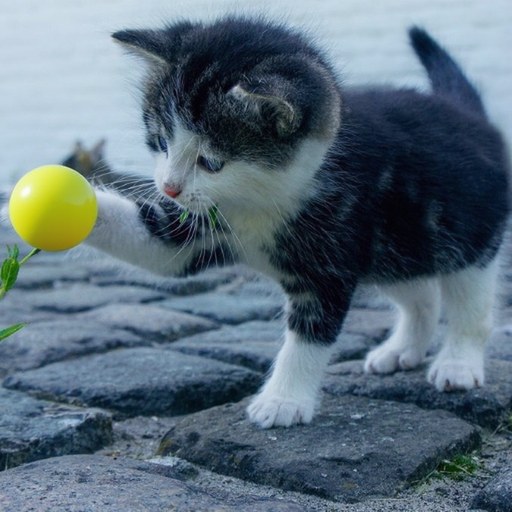} &
        \includegraphics[width=\imgwidthNight]{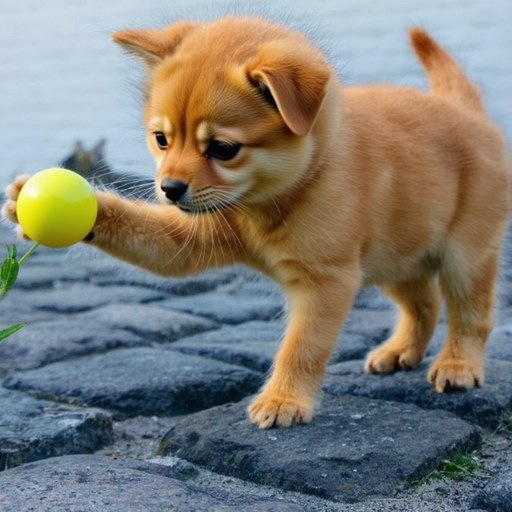} \\
                && \raisebox{20pt}{\rotatebox[origin=t]{90}{{GRAG}}} & { } &
        \includegraphics[width=\imgwidthNight]{images/edit_comparison/flower_ball_cat_dog/src.jpg} & { } &
        \includegraphics[width=\imgwidthNight]{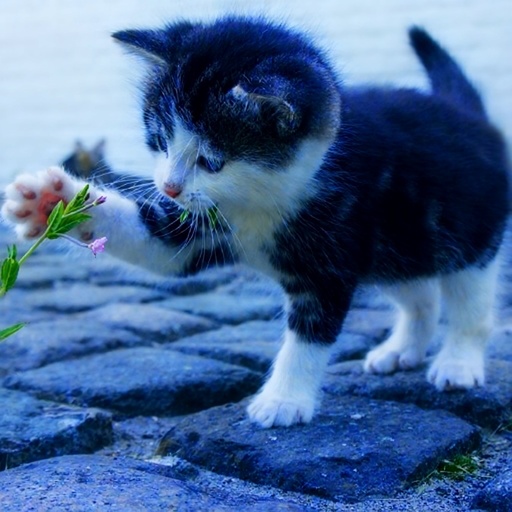} &
        \includegraphics[width=\imgwidthNight]{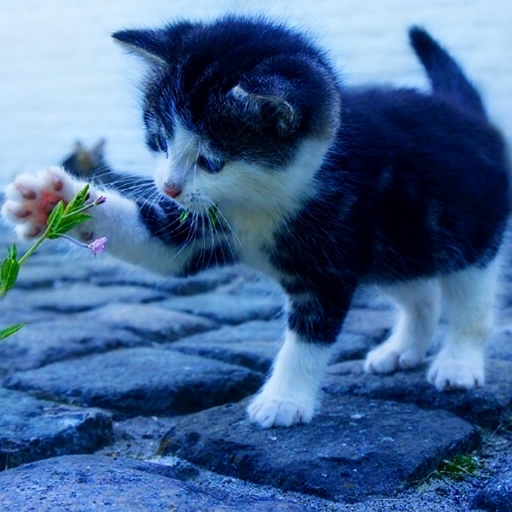} &
        \includegraphics[width=\imgwidthNight]{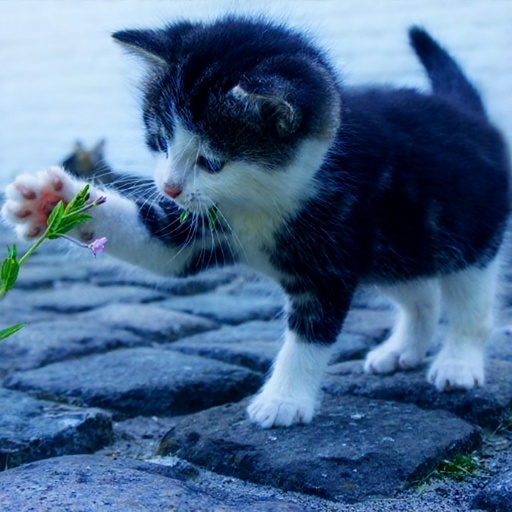} &
        \includegraphics[width=\imgwidthNight]{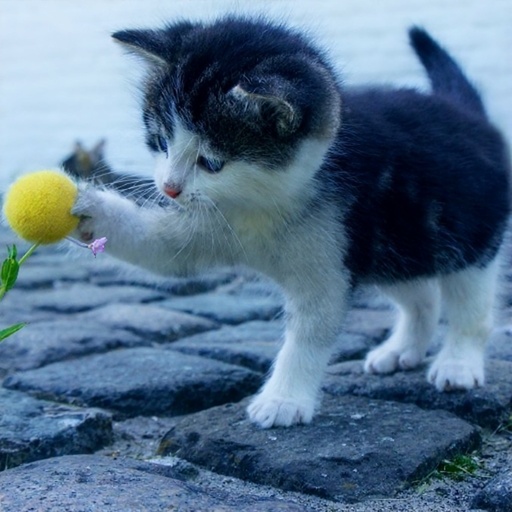} &
        \includegraphics[width=\imgwidthNight]{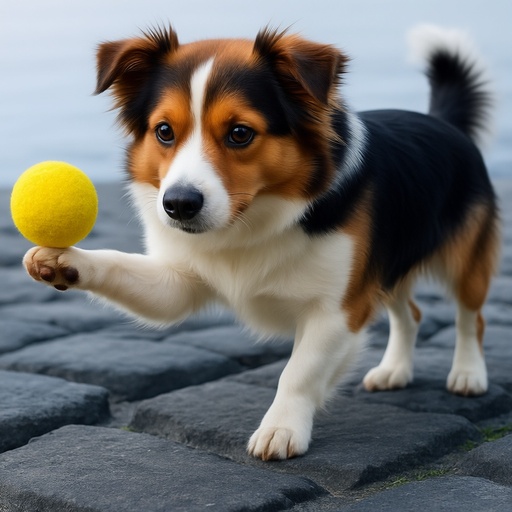} &
        \includegraphics[width=\imgwidthNight]{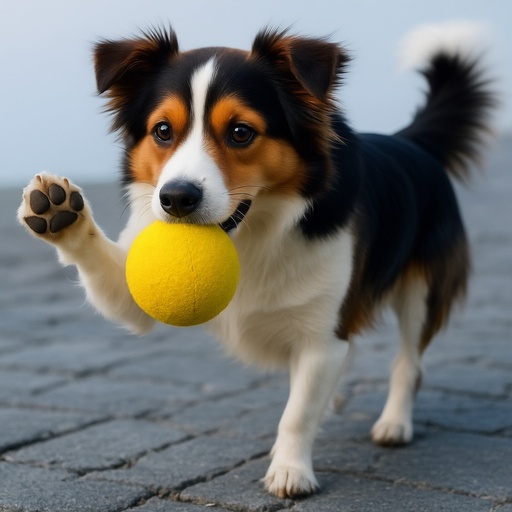} \\
        && \raisebox{20pt}{\rotatebox[origin=t]{90}{T2T (Ours)}} & { } &
        \includegraphics[width=\imgwidthNight]{images/edit_comparison/flower_ball_cat_dog/src.jpg} & { } &
        \includegraphics[width=\imgwidthNight]{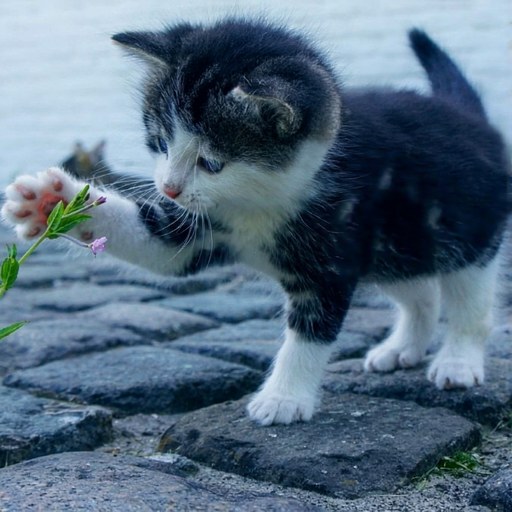} &
        \includegraphics[width=\imgwidthNight]{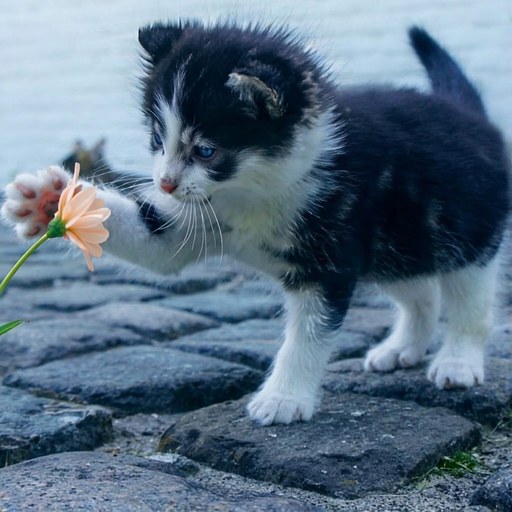} &
        \includegraphics[width=\imgwidthNight]{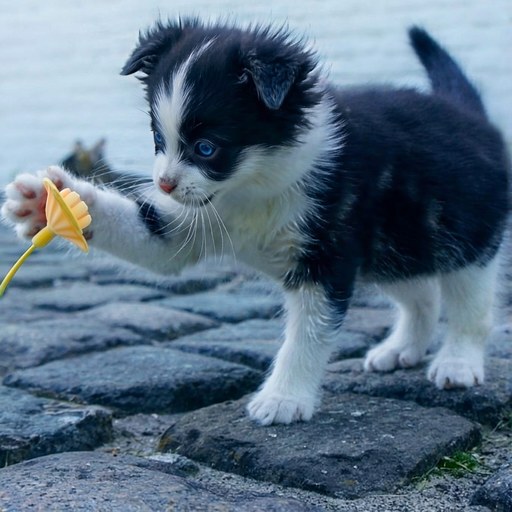} &
        \includegraphics[width=\imgwidthNight]{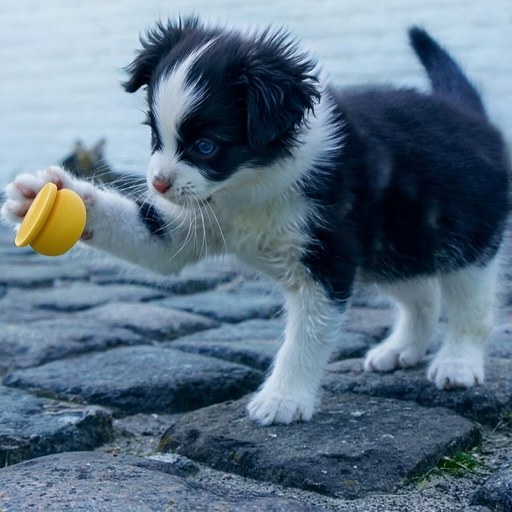} &
        \includegraphics[width=\imgwidthNight]{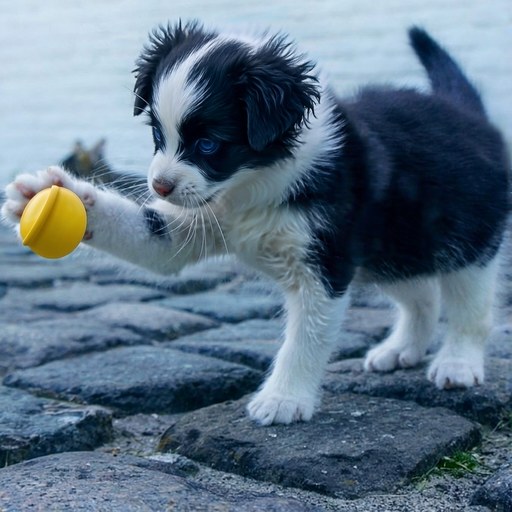} &
        \includegraphics[width=\imgwidthNight]{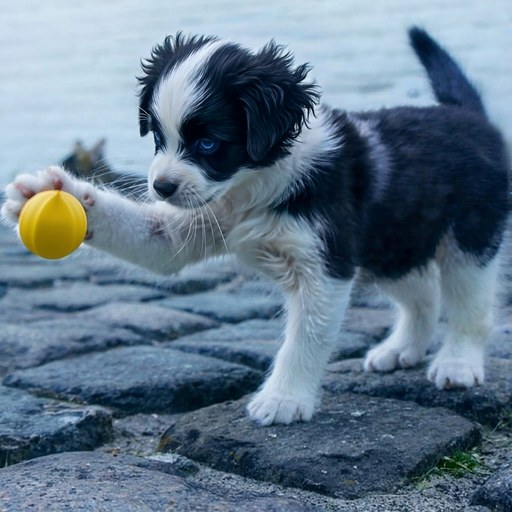} \\
        &&&& Input && 0.0 & 0.2 & 0.4 & 0.6 & 0.8 & 1.0 \\

       \multicolumn{12}{c}{\textit{``Change the room to a garden''}} \\
        \raisebox{20pt}{\rotatebox[origin=t]{90}{{Kontinuous}}} & { } &
        \raisebox{20pt}{\rotatebox[origin=t]{90}{{Kontext}}} & { } &
        \includegraphics[width=\imgwidthNight]{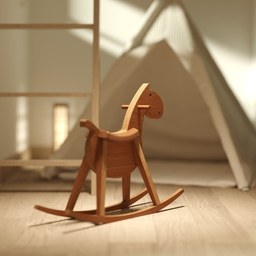} & { } &
        \includegraphics[width=\imgwidthNight]{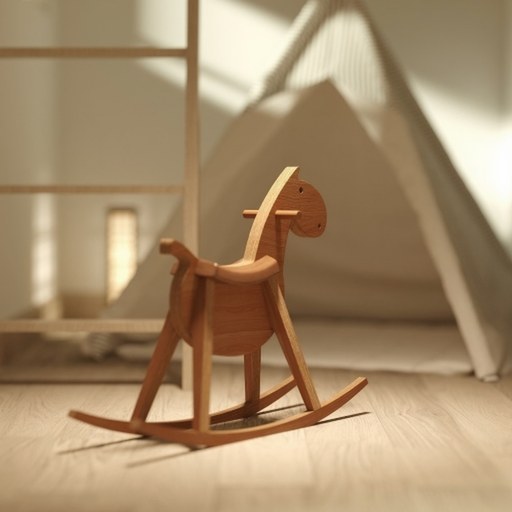} &
        \includegraphics[width=\imgwidthNight]{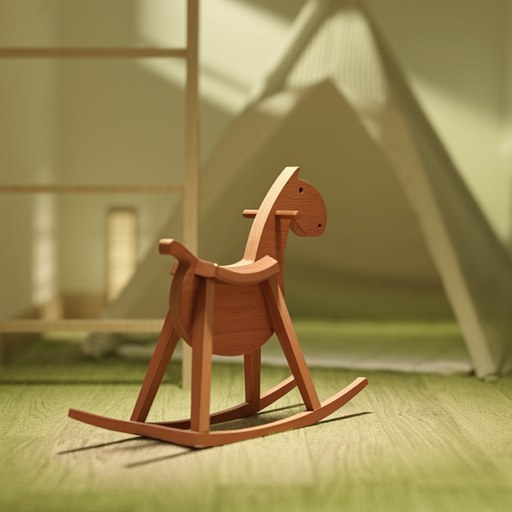} &
        \includegraphics[width=\imgwidthNight]{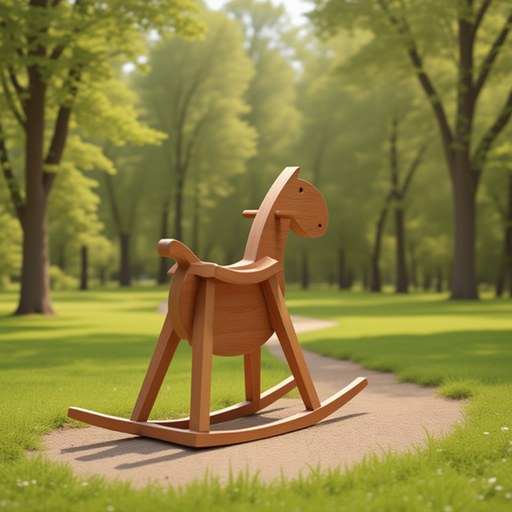} &
        \includegraphics[width=\imgwidthNight]{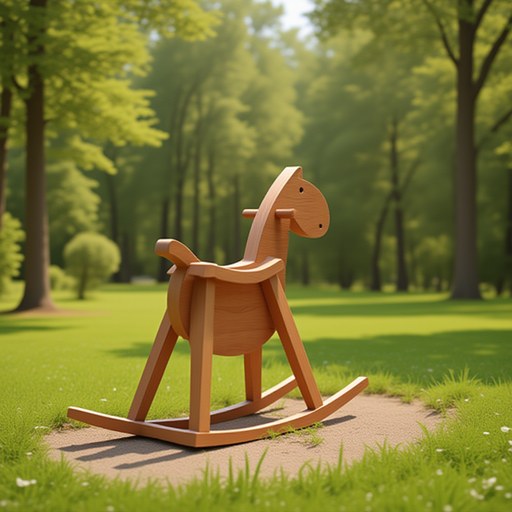} &
        \includegraphics[width=\imgwidthNight]{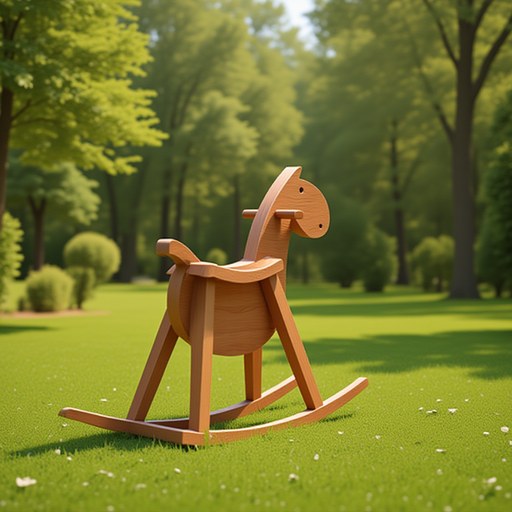} &
        \includegraphics[width=\imgwidthNight]{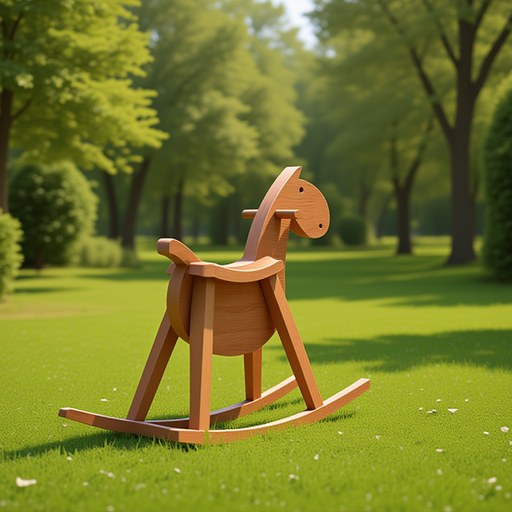} \\
        && \raisebox{20pt}{\rotatebox[origin=t]{90}{{SliderEdit}}} & { } &
        \includegraphics[width=\imgwidthNight]{images/edit_comparison/horse_swing_garden/src.jpg} & { } &
        \includegraphics[width=\imgwidthNight]{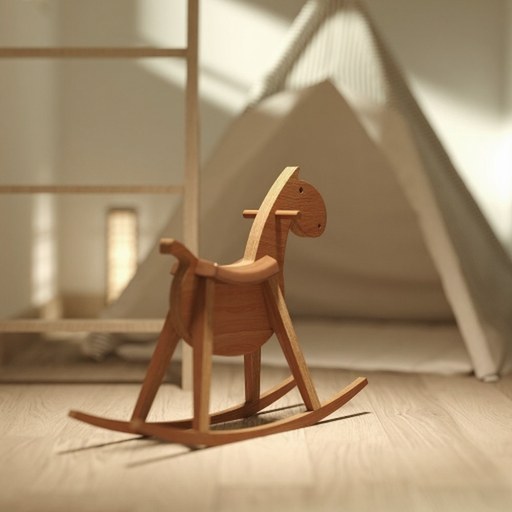} &
        \includegraphics[width=\imgwidthNight]{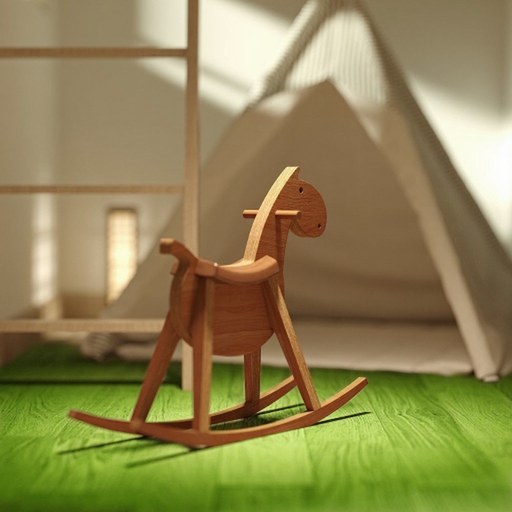} &
        \includegraphics[width=\imgwidthNight]{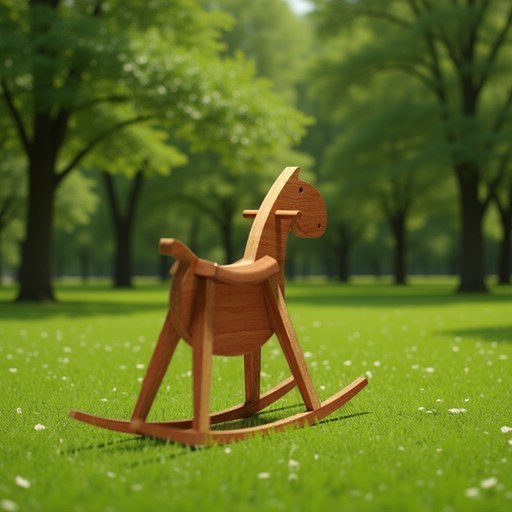} &
        \includegraphics[width=\imgwidthNight]{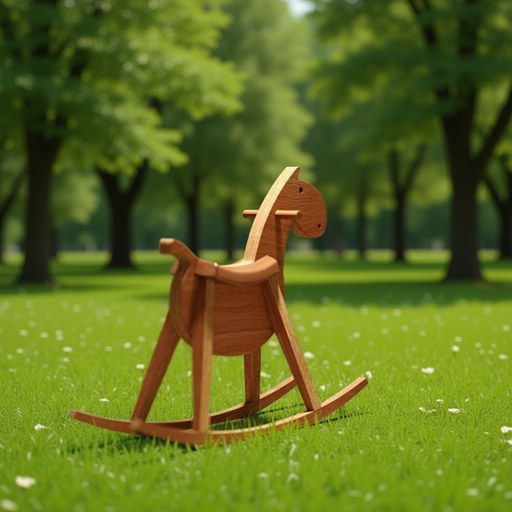} &
        \includegraphics[width=\imgwidthNight]{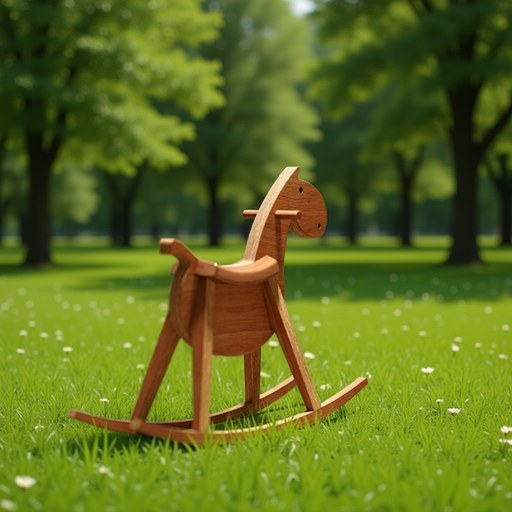} &
        \includegraphics[width=\imgwidthNight]{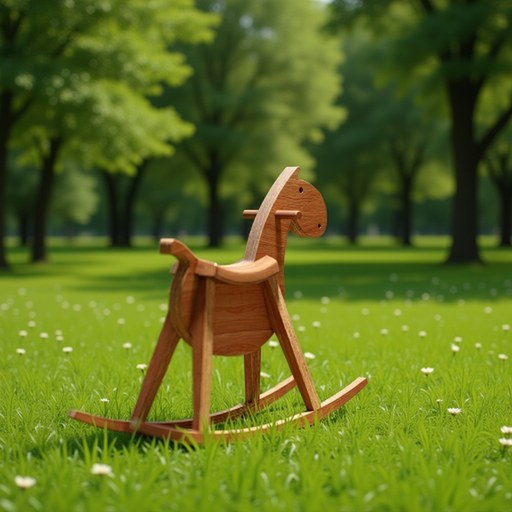} \\
        && \raisebox{20pt}{\rotatebox[origin=t]{90}{{GRAG}}} & { } &
        \includegraphics[width=\imgwidthNight]{images/edit_comparison/horse_swing_garden/src.jpg} & { } &
        \includegraphics[width=\imgwidthNight]{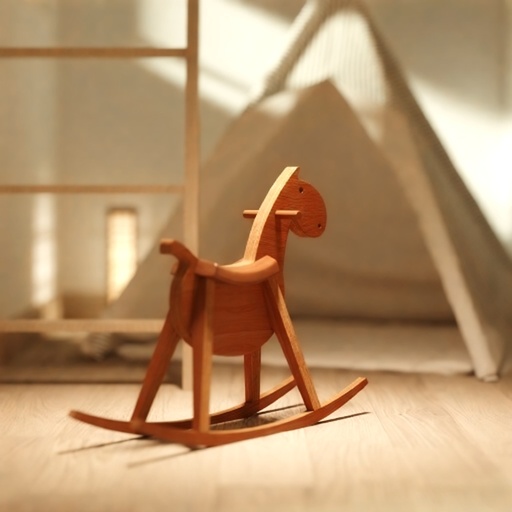} &
        \includegraphics[width=\imgwidthNight]{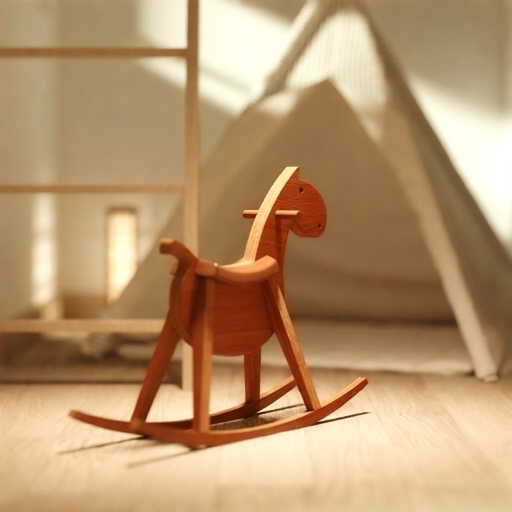} &
        \includegraphics[width=\imgwidthNight]{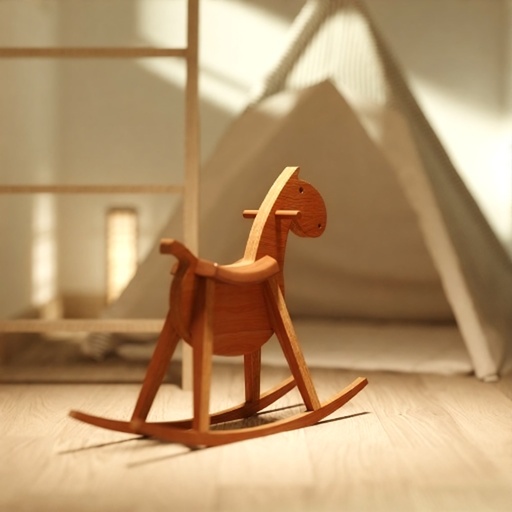} &
        \includegraphics[width=\imgwidthNight]{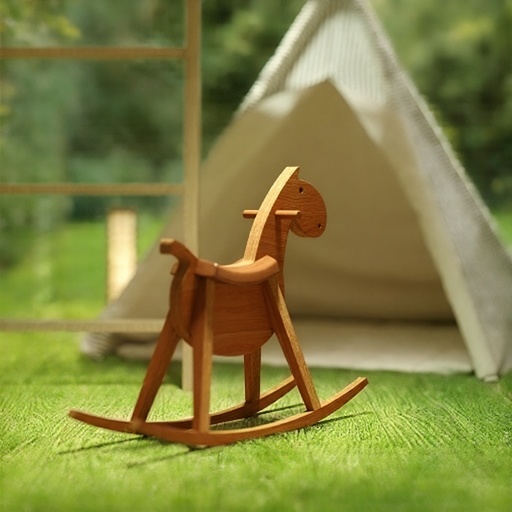} &
        \includegraphics[width=\imgwidthNight]{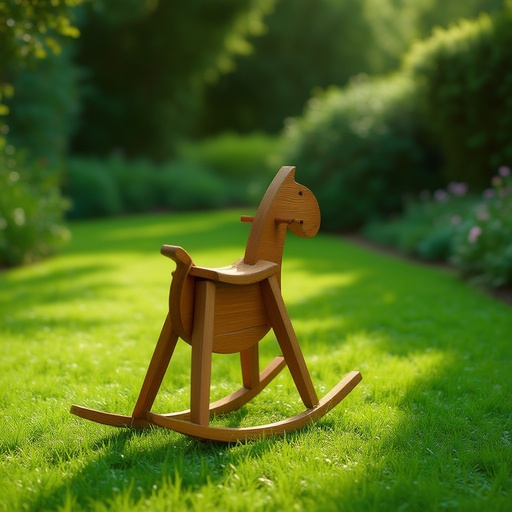} &
        \includegraphics[width=\imgwidthNight]{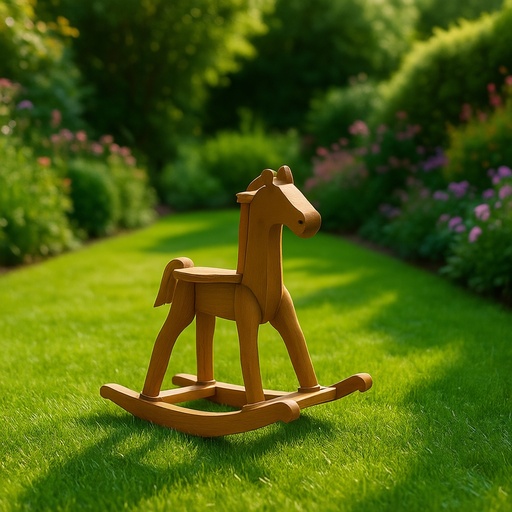} \\
        && \raisebox{20pt}{\rotatebox[origin=t]{90}{T2T (Ours)}} & { } &
        \includegraphics[width=\imgwidthNight]{images/edit_comparison/horse_swing_garden/src.jpg} & { } &
        \includegraphics[width=\imgwidthNight]{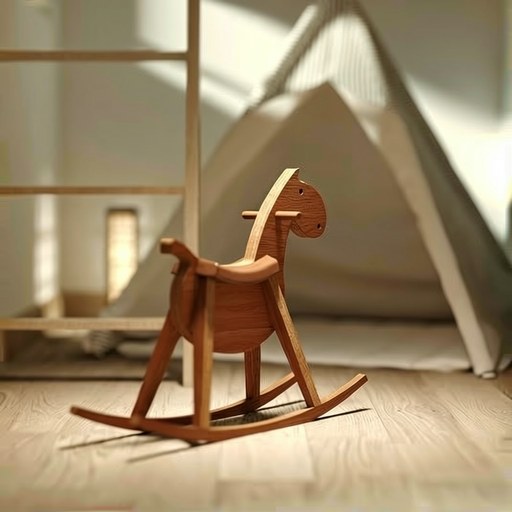} &
        \includegraphics[width=\imgwidthNight]{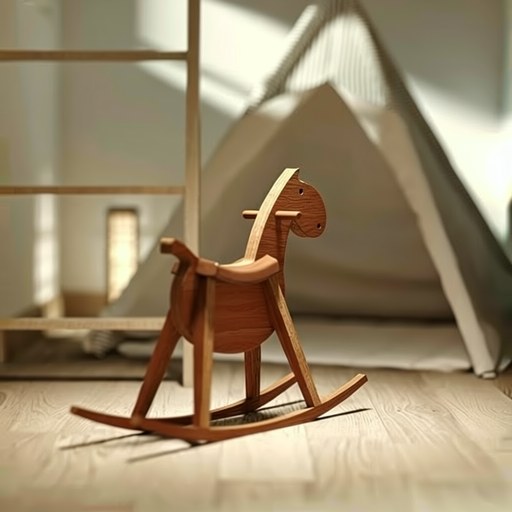} &
        \includegraphics[width=\imgwidthNight]{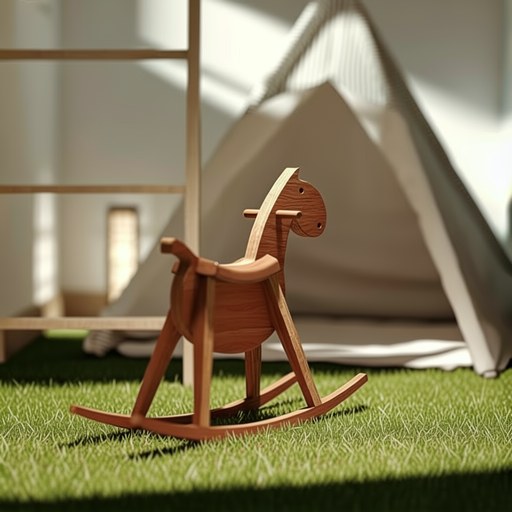} &
        \includegraphics[width=\imgwidthNight]{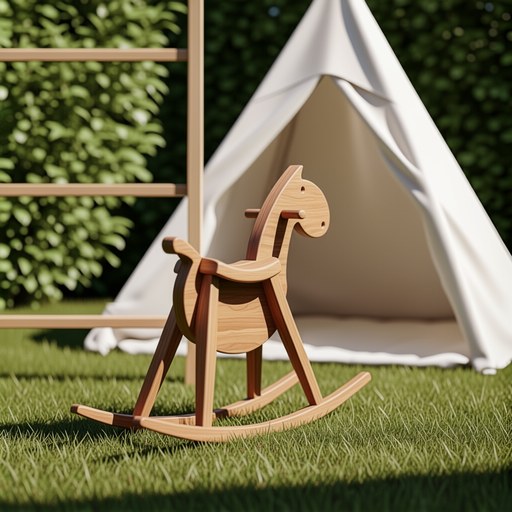} &
        \includegraphics[width=\imgwidthNight]{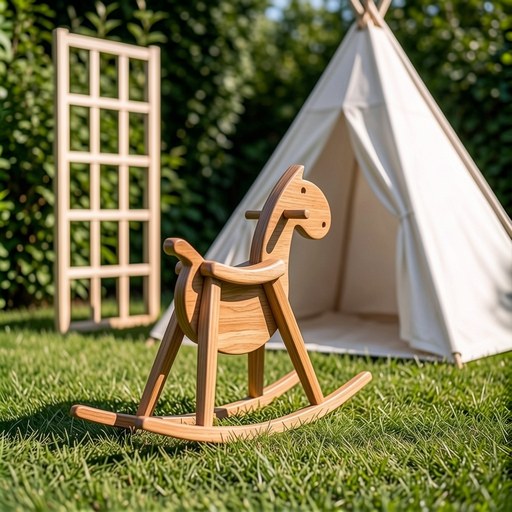} &
        \includegraphics[width=\imgwidthNight]{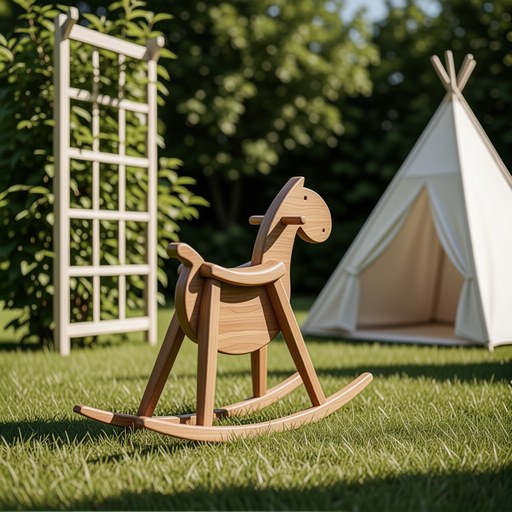} \\
        &&&& Input && 0.0 & 0.2 & 0.4 & 0.6 & 0.8 & 1.0 \\

    \end{tabular}
    \caption{Qualitative comparison with continuous editing methods. Top example: results generated using FIBO-edit. Bottom example: results generated using FLUX2-Klein.} 
\vspace{-10pt}
    \label{fig:supp_edit_comparison}
\end{figure*}
\renewcommand{\imgwidthablate}{0.14\linewidth}

\begin{figure*}
    \centering
    \setlength{\tabcolsep}{0pt}
    \begin{tabular}{ccccccc}
        \multicolumn{7}{c}{\textit{``Transform the wooden birdhouse into a white security camera.''}} \\
        \includegraphics[width=\imgwidthablate]{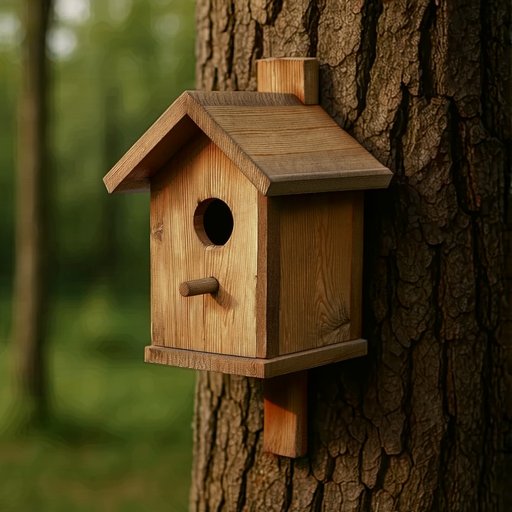} &
        \includegraphics[width=\imgwidthablate]{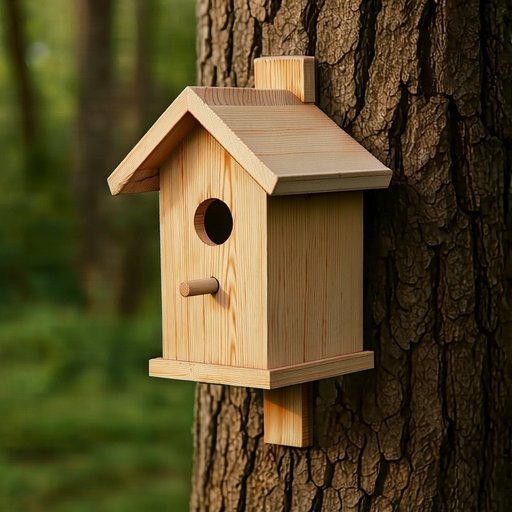} &
        \includegraphics[width=\imgwidthablate]{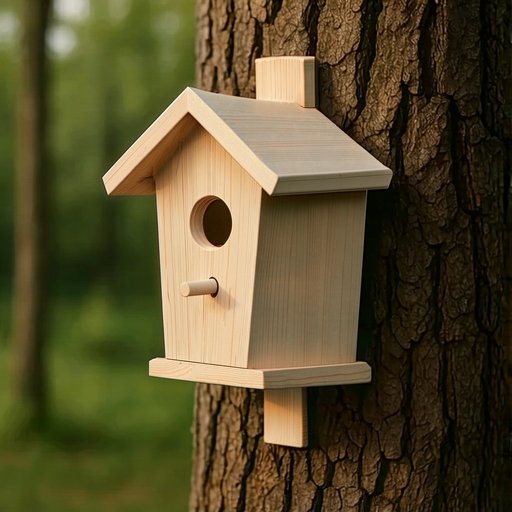} &
        \includegraphics[width=\imgwidthablate]{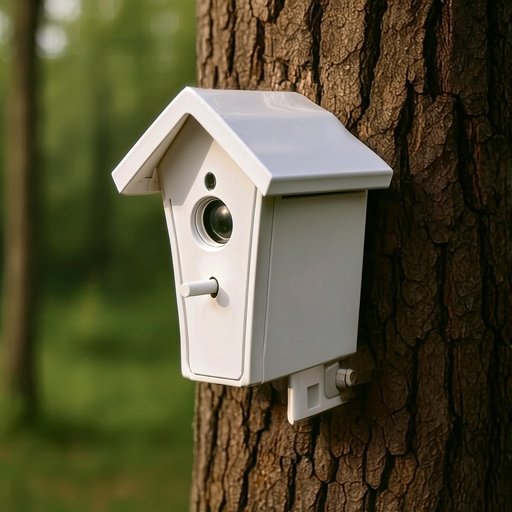} &
        \includegraphics[width=\imgwidthablate]{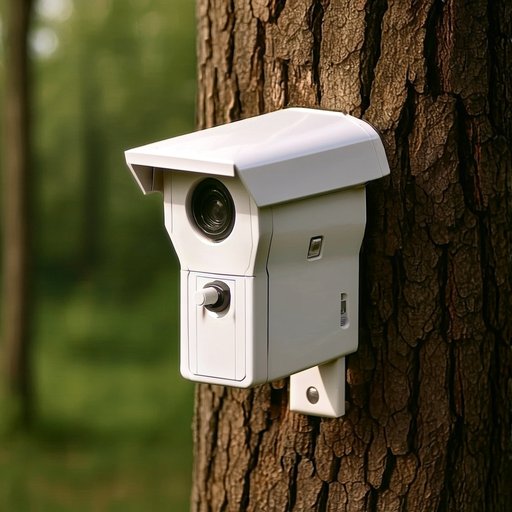} &
        \includegraphics[width=\imgwidthablate]{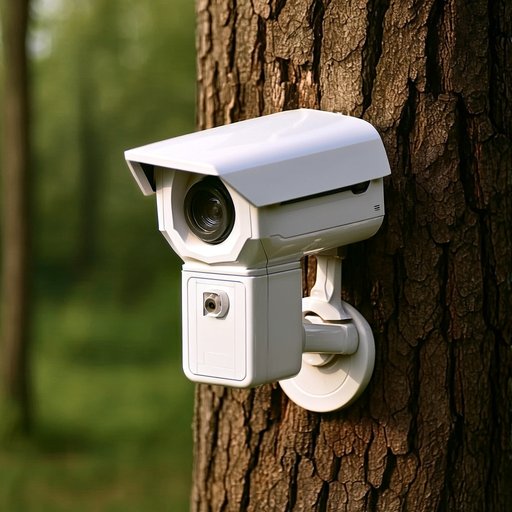} &
        \includegraphics[width=\imgwidthablate]{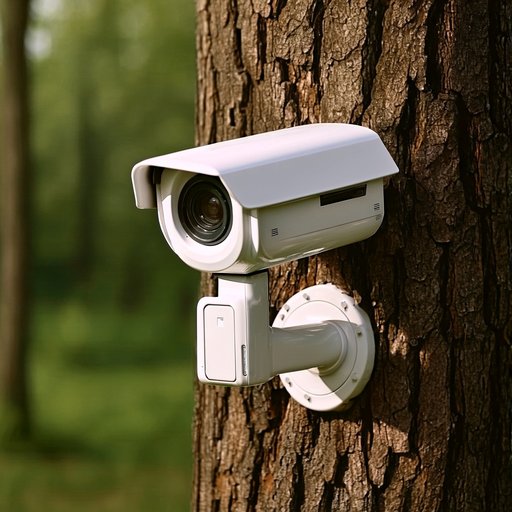} \\

        \multicolumn{7}{c}{\textit{``Change the marble wall to a flower wall.''}} \\
        \includegraphics[width=\imgwidthablate]{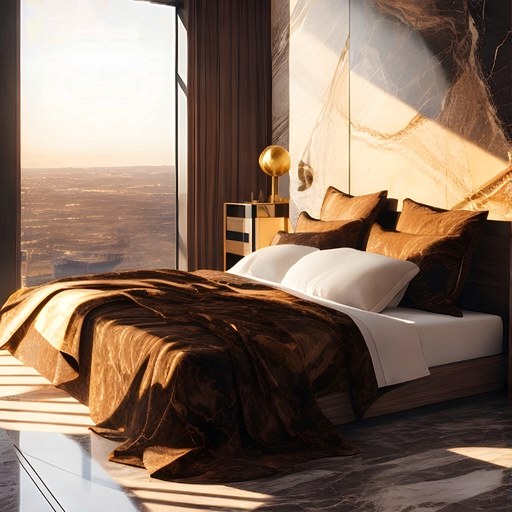} &
        \includegraphics[width=\imgwidthablate]{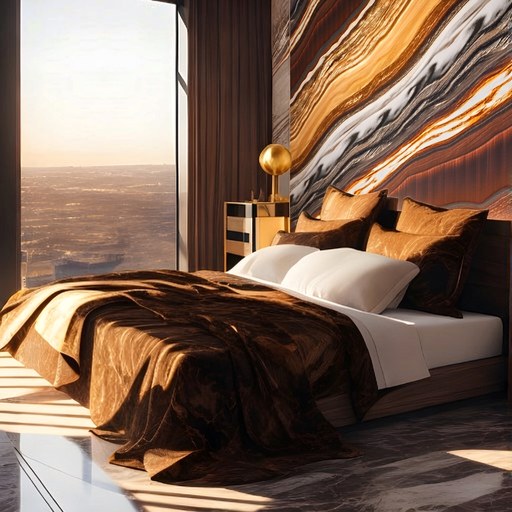} &
        \includegraphics[width=\imgwidthablate]{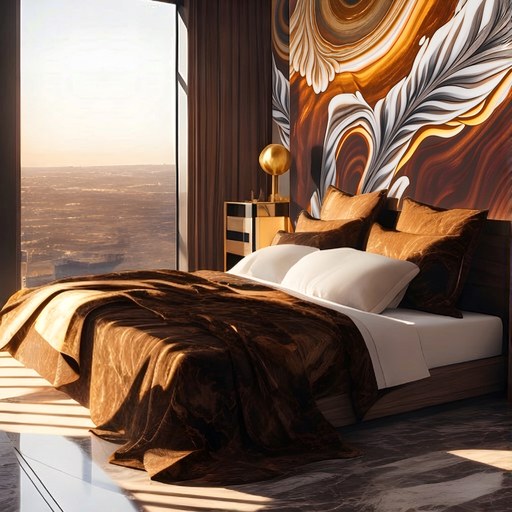} &
        \includegraphics[width=\imgwidthablate]{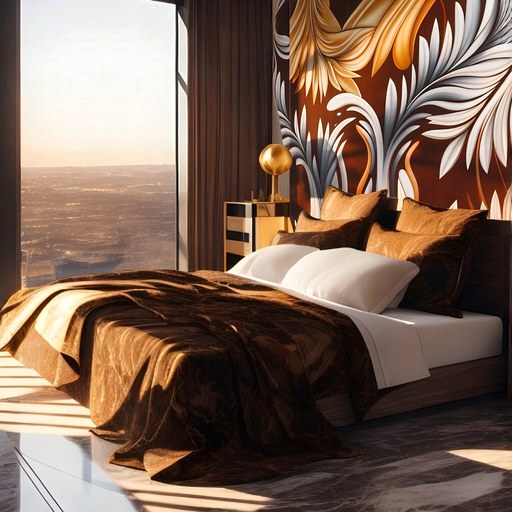} &
        \includegraphics[width=\imgwidthablate]{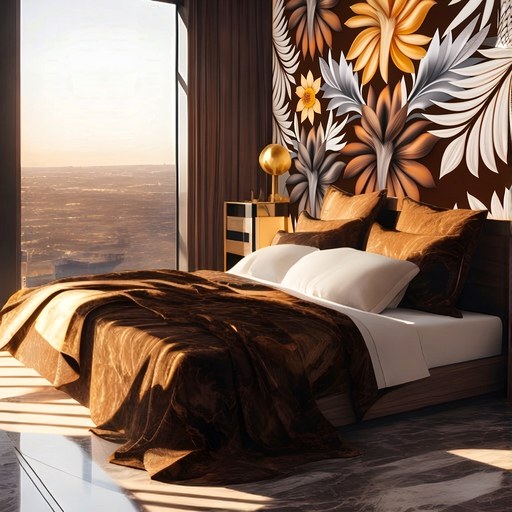} &
        \includegraphics[width=\imgwidthablate]{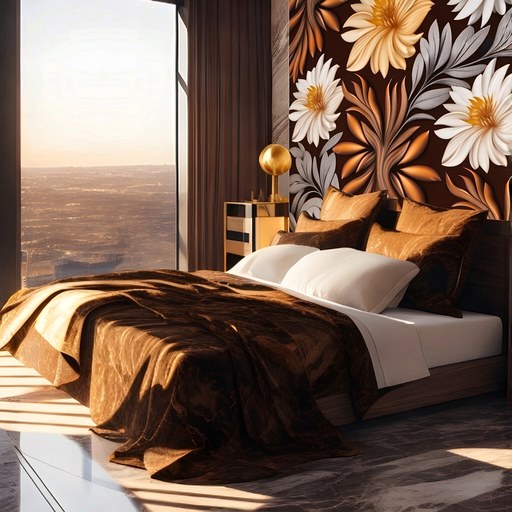} &
        \includegraphics[width=\imgwidthablate]{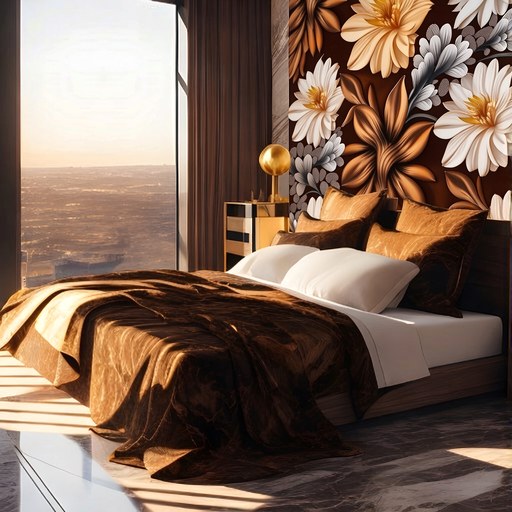} \\

        \multicolumn{7}{c}{\textit{``Make the bacon strips crisp and burnt.''}} \\
        \includegraphics[width=\imgwidthablate]{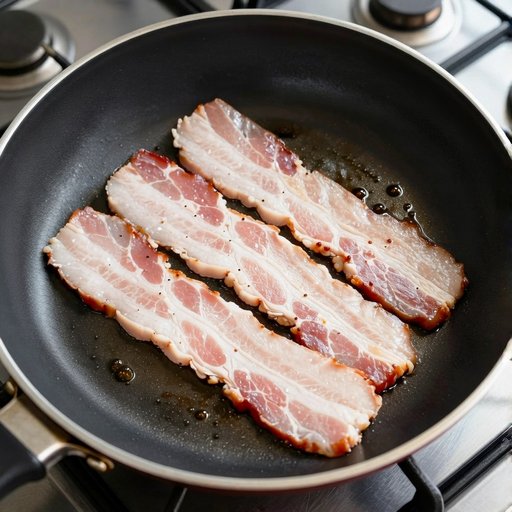} &
        \includegraphics[width=\imgwidthablate]{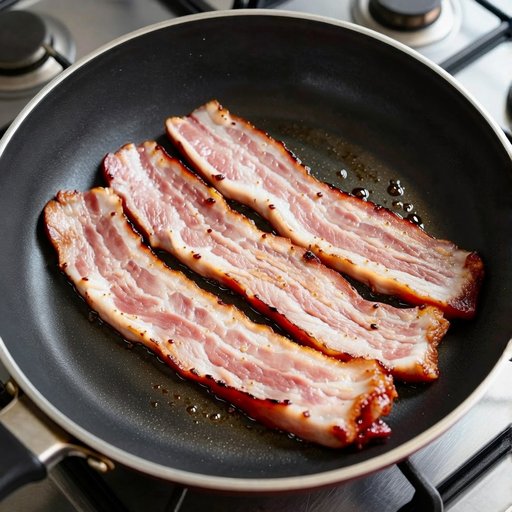} &
        \includegraphics[width=\imgwidthablate]{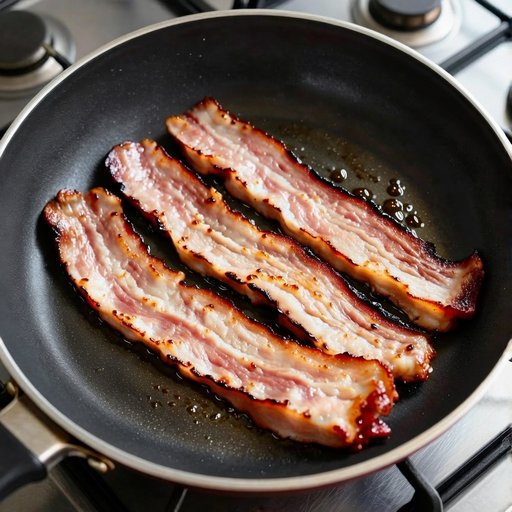} &
        \includegraphics[width=\imgwidthablate]{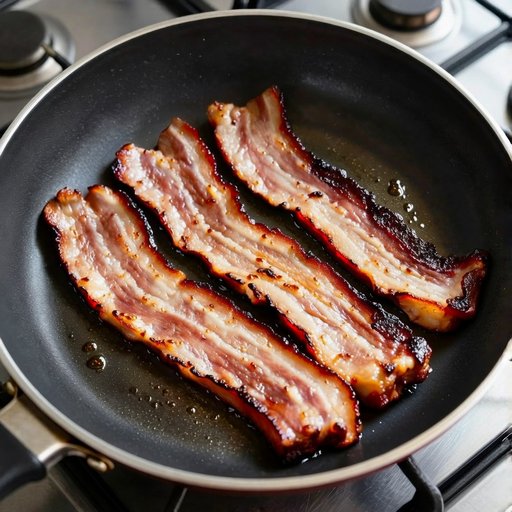} &
        \includegraphics[width=\imgwidthablate]{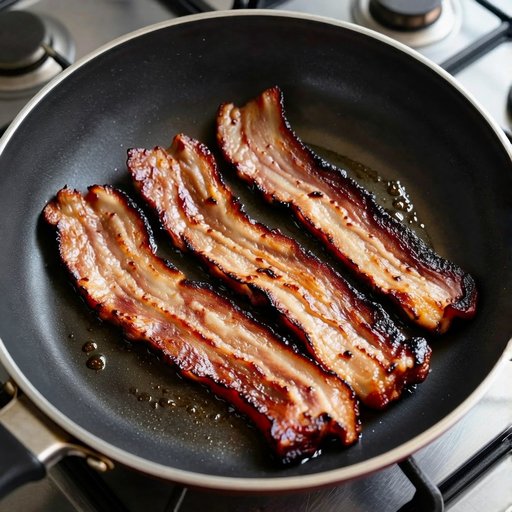} &
        \includegraphics[width=\imgwidthablate]{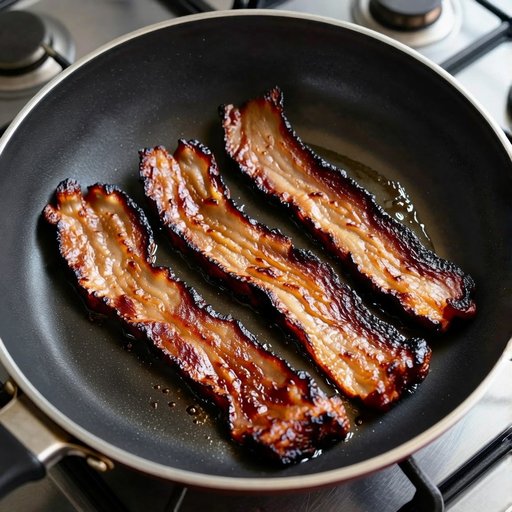} &
        \includegraphics[width=\imgwidthablate]{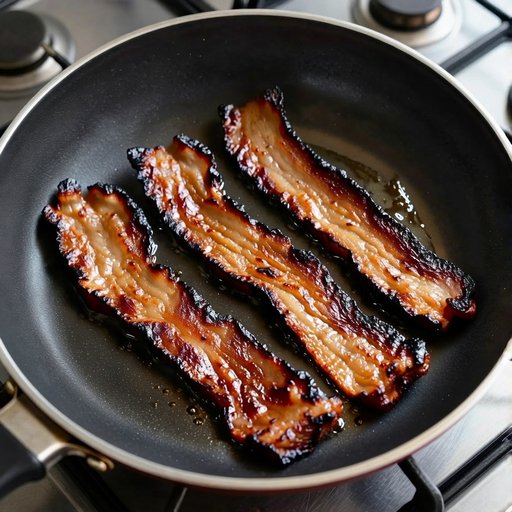} \\

        \multicolumn{7}{c}{\textit{``Replace the meat balls with tuna sushi and change the plate material from ceramic to metal.''}} \\
        \includegraphics[width=\imgwidthablate]{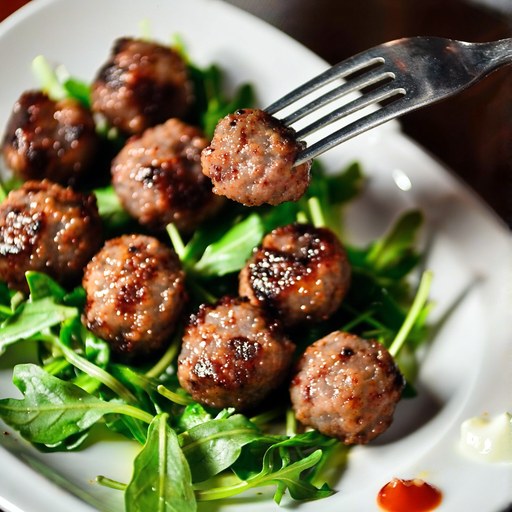} &
        \includegraphics[width=\imgwidthablate]{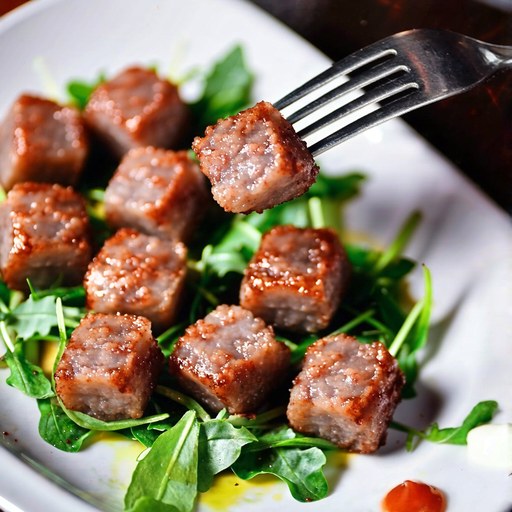} &
        \includegraphics[width=\imgwidthablate]{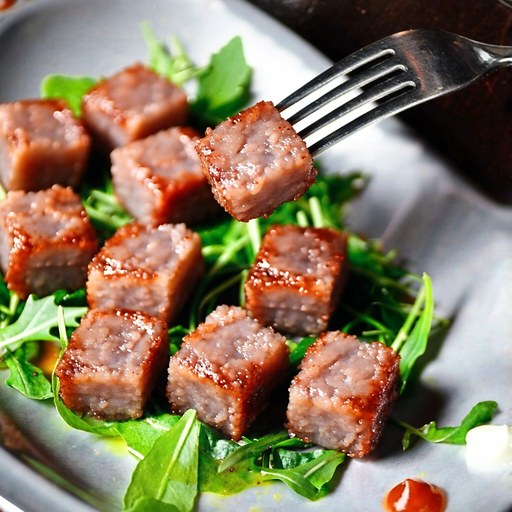} &
        \includegraphics[width=\imgwidthablate]{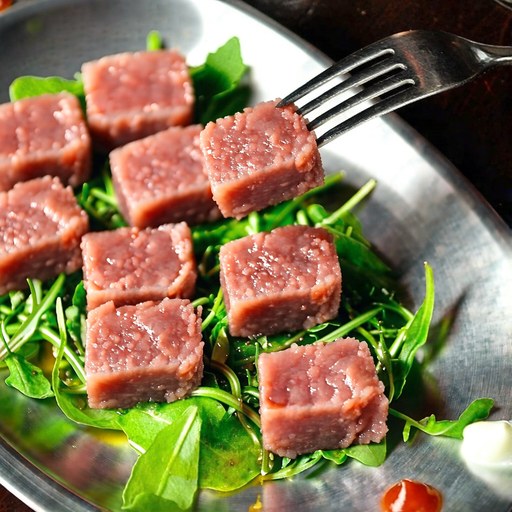} &
        \includegraphics[width=\imgwidthablate]{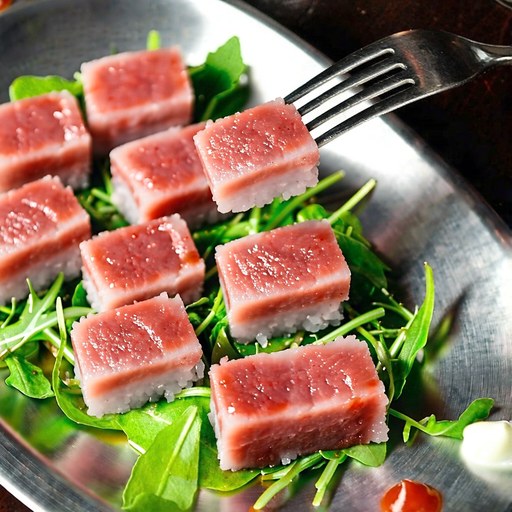} &
        \includegraphics[width=\imgwidthablate]{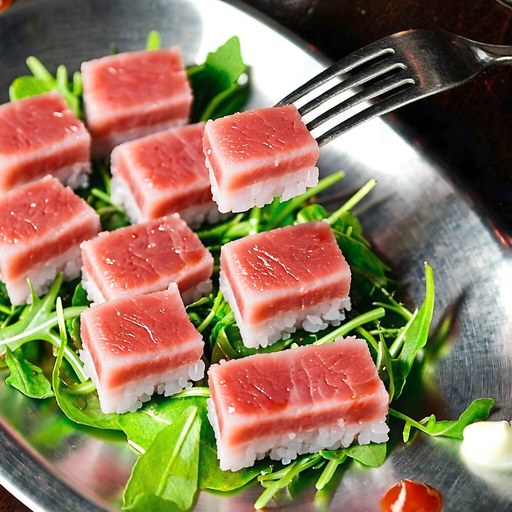} &
        \includegraphics[width=\imgwidthablate]{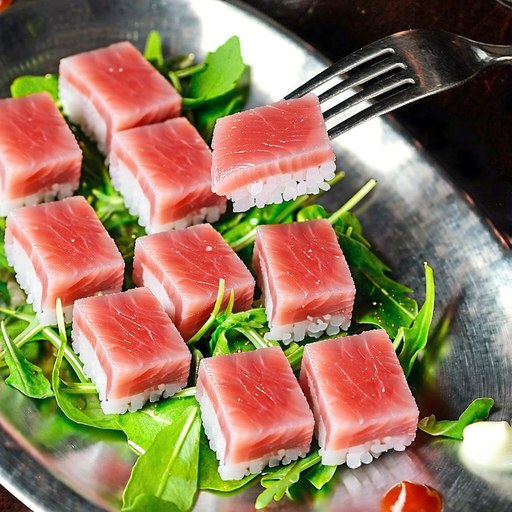} \\

        \multicolumn{7}{c}{\textit{``Make the woman laughing and change the stars in her hair to flowers.''}} \\
        \includegraphics[width=\imgwidthablate]{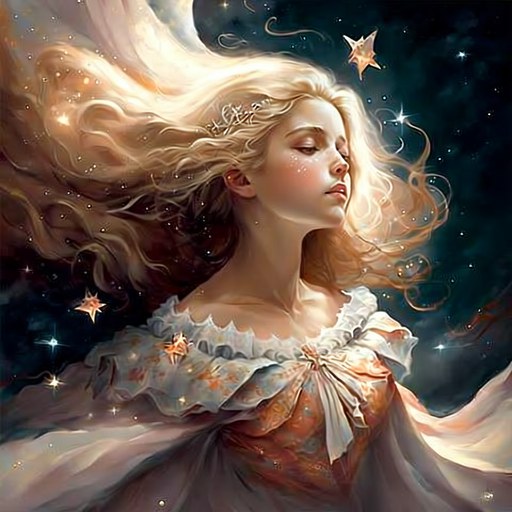} &
        \includegraphics[width=\imgwidthablate]{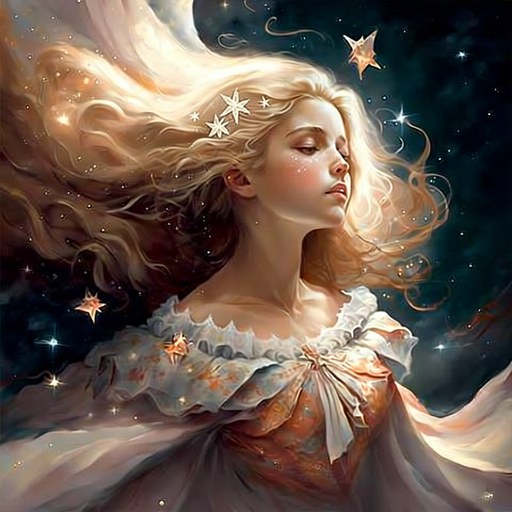} &
        \includegraphics[width=\imgwidthablate]{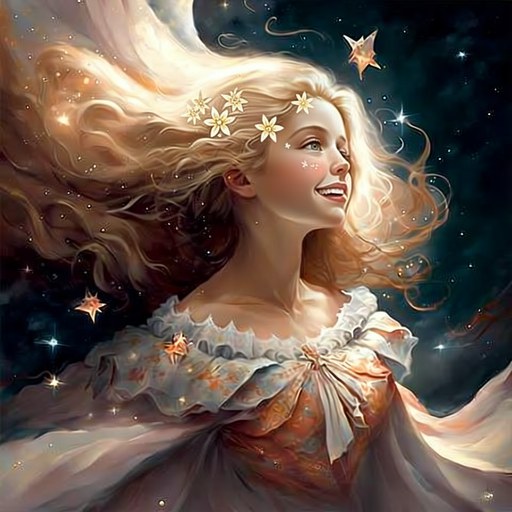} &
        \includegraphics[width=\imgwidthablate]{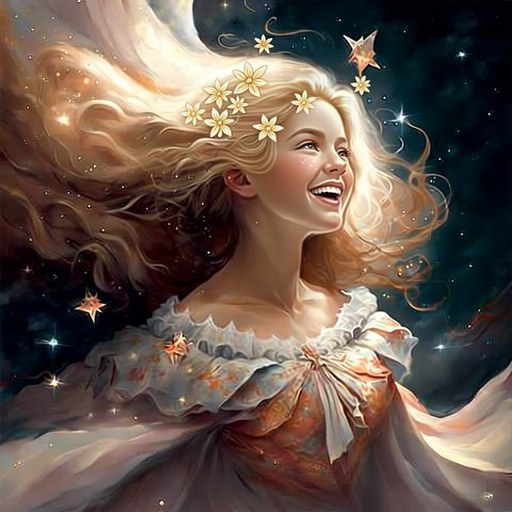} &
        \includegraphics[width=\imgwidthablate]{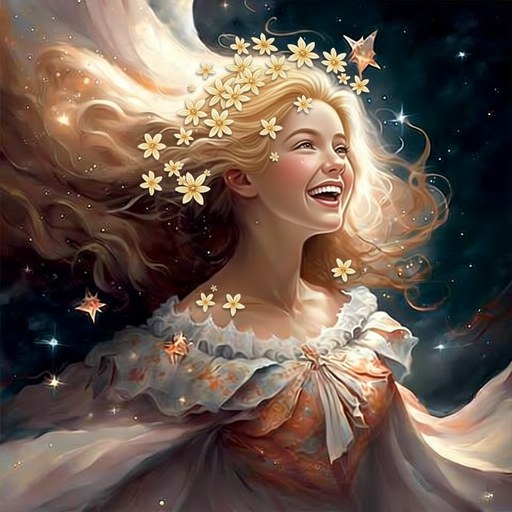} &
        \includegraphics[width=\imgwidthablate]{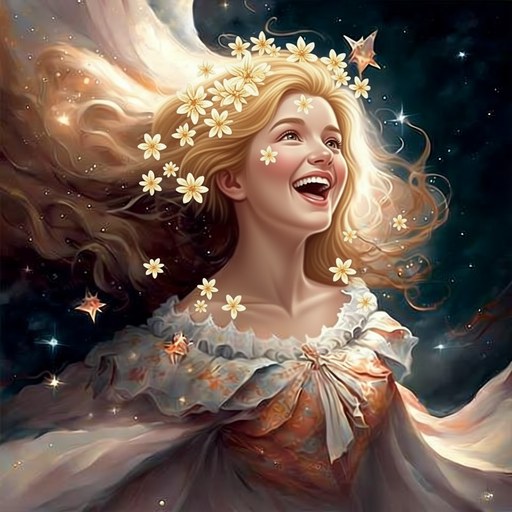} &
        \includegraphics[width=\imgwidthablate]{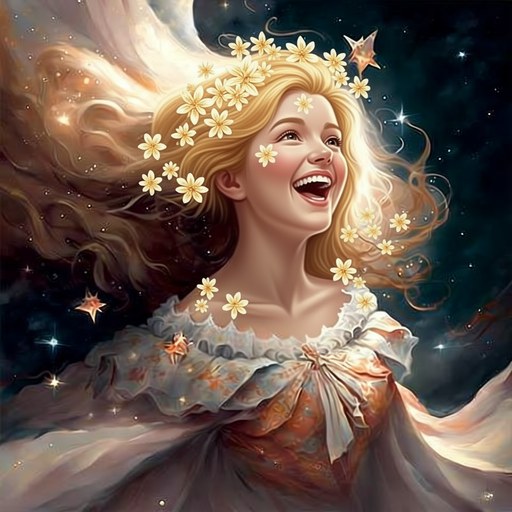} \\

        \multicolumn{7}{c}{\textit{``Change the horse to a dog and its action from running to jumping.''}} \\
        \includegraphics[width=\imgwidthablate]{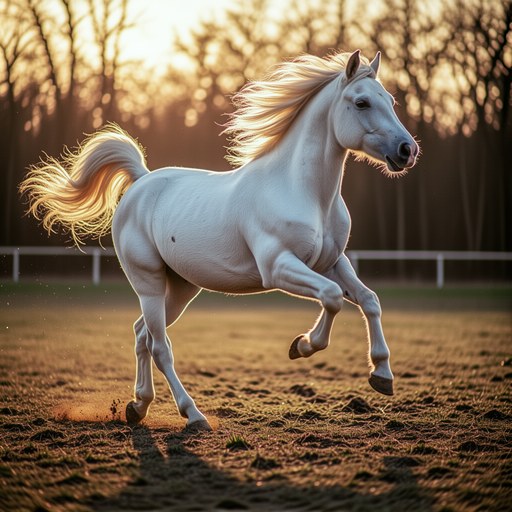} &
        \includegraphics[width=\imgwidthablate]{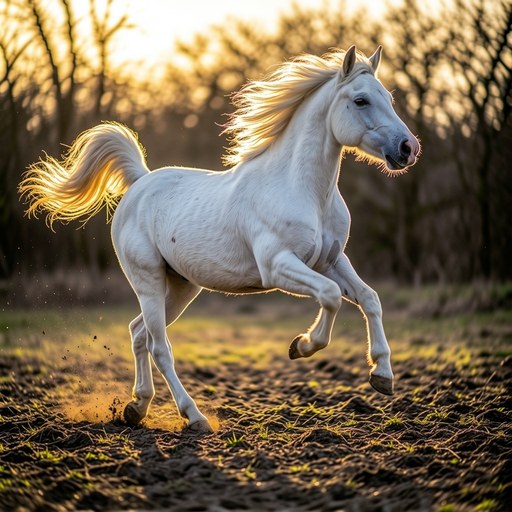} &
        \includegraphics[width=\imgwidthablate]{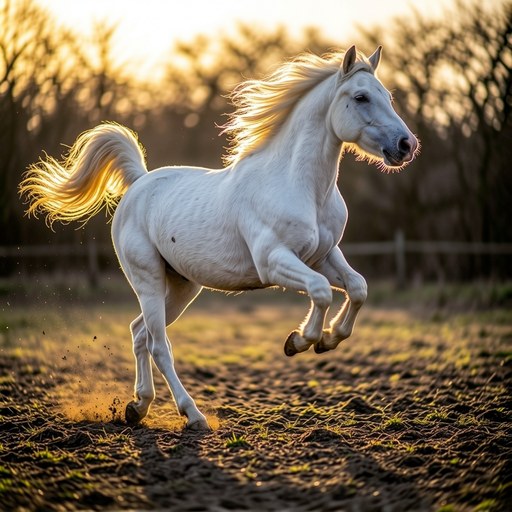} &
        \includegraphics[width=\imgwidthablate]{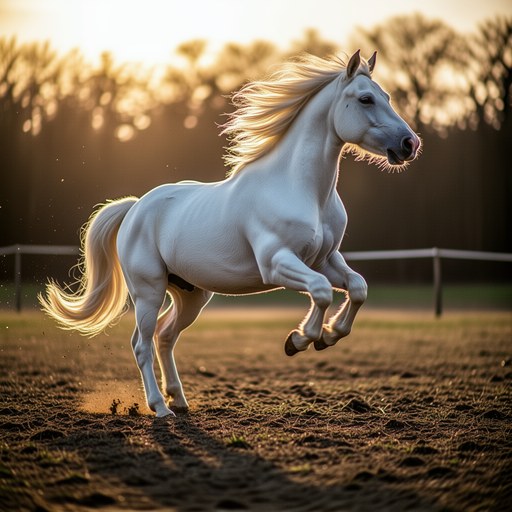} &
        \includegraphics[width=\imgwidthablate]{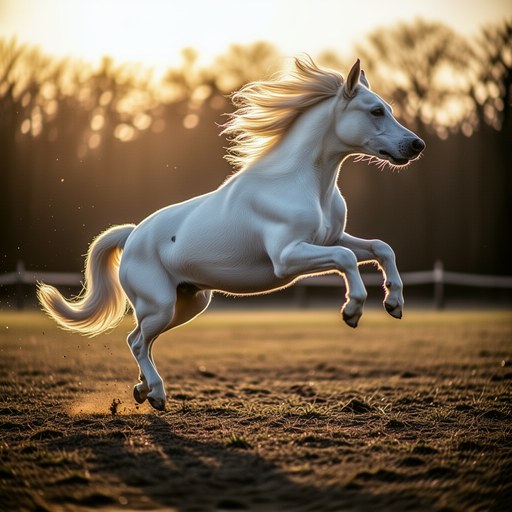} &
        \includegraphics[width=\imgwidthablate]{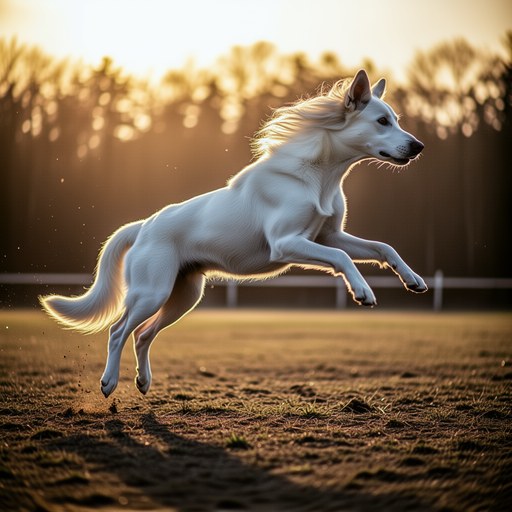} &
        \includegraphics[width=\imgwidthablate]{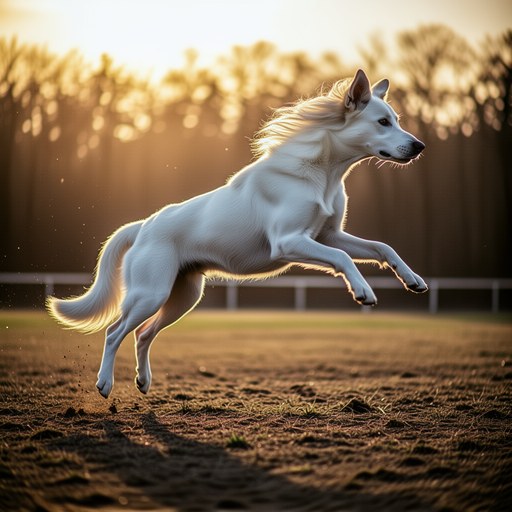} \\

        \multicolumn{7}{c}{\textit{``Change the rose from blooming to dried.''}} \\
        \includegraphics[width=\imgwidthablate]{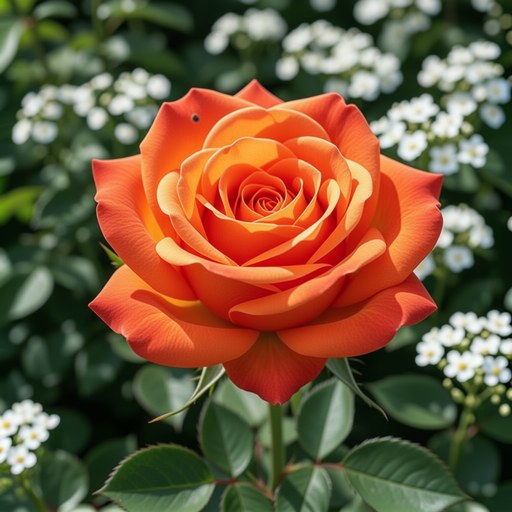} &
        \includegraphics[width=\imgwidthablate]{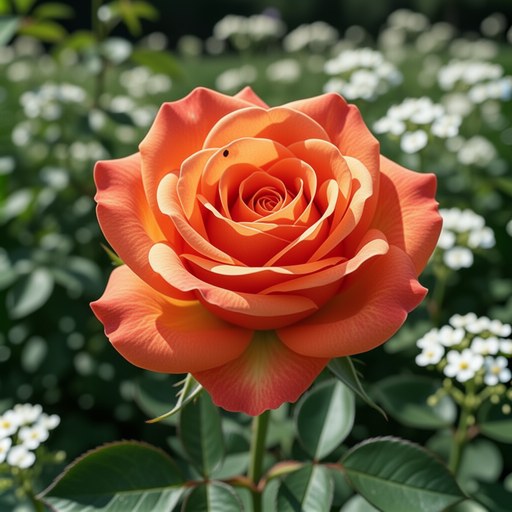} &
        \includegraphics[width=\imgwidthablate]{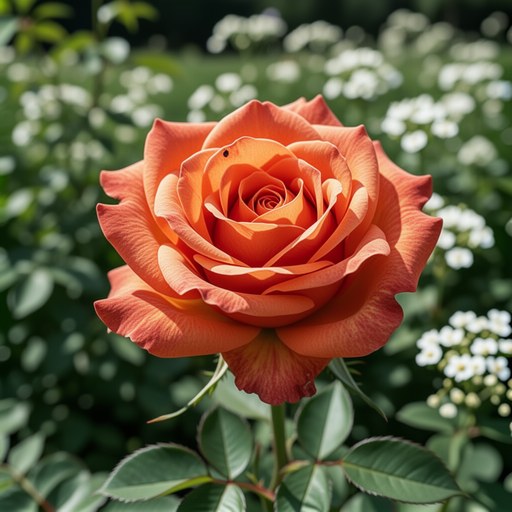} &
        \includegraphics[width=\imgwidthablate]{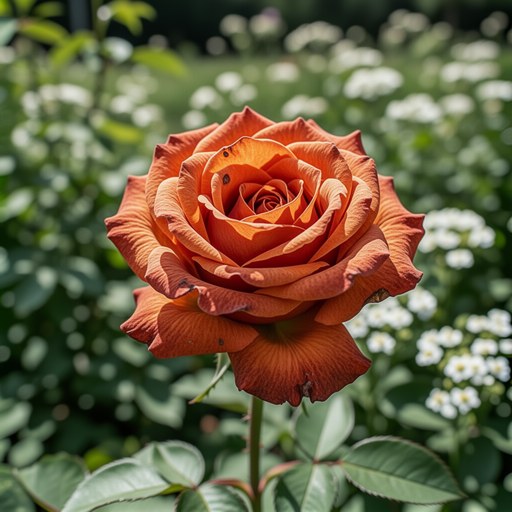} &
        \includegraphics[width=\imgwidthablate]{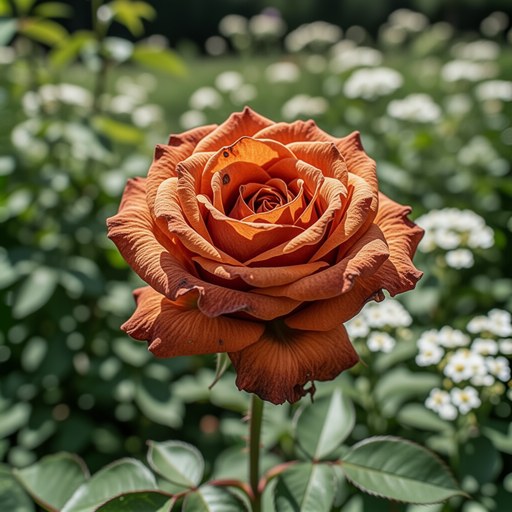} &
        \includegraphics[width=\imgwidthablate]{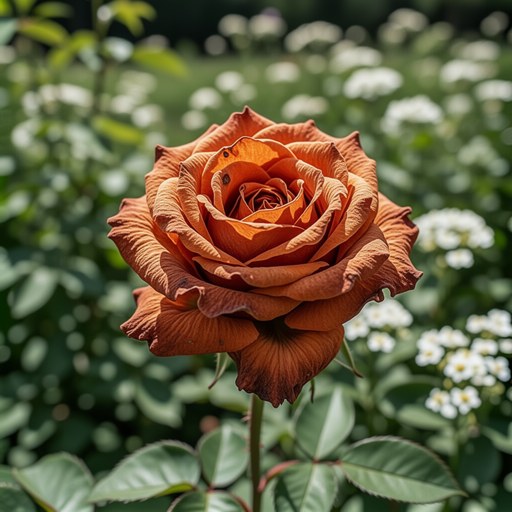} &
        \includegraphics[width=\imgwidthablate]{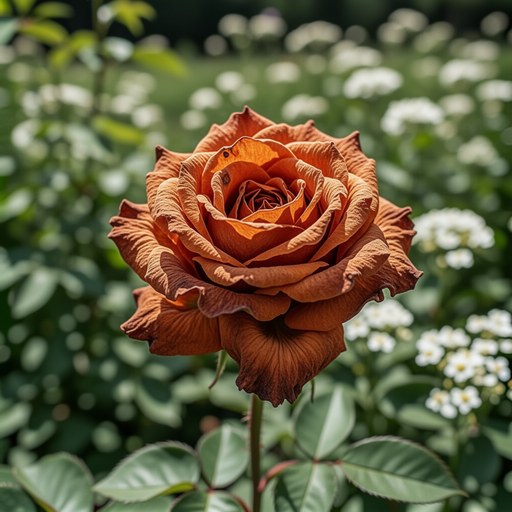} \\

        Input & \multicolumn{6}{l}{\small{Edit Intensity} $\xrightarrow{\hspace{260pt}}$} \\

    \end{tabular}
    \caption{Additional continuous editing results. Each row shows a gradual edit applied to a reference image along the specified direction. Top three rows: results generated using FIBO-edit. Bottom four rows: results generated using FLUX2-Klein.
        } 
\vspace{-10pt}
    \label{fig:supp_edit_continous}
\end{figure*}







\end{document}